\documentclass[11pt]{article}
\usepackage{acl}       

\newif\ifincludeappendix
\includeappendixtrue  

\usepackage{times}
\usepackage{latexsym}
\usepackage[T1]{fontenc}
\usepackage[utf8]{inputenc}
\usepackage{microtype}
\usepackage{inconsolata}
\usepackage{graphicx}
\usepackage{caption}
\usepackage{cuted}

\newcommand{\eg}{e.g.\xspace}
\newcommand{\ie}{i.e.\xspace}

\usepackage{booktabs}

\usepackage[accsupp]{axessibility}  

\usepackage[table]{xcolor}   

\usepackage{chngcntr}

\usepackage{hyperref}

\usepackage{orcidlink}

\usepackage{tikz}
\usepackage[most]{tcolorbox}
\tcbuselibrary{listings,skins,breakable,many,}
\usepackage{adjustbox}
\usepackage{arydshln} 

\usepackage{mathtools}

\usepackage{pifont}   
\usepackage[dvipsnames]{xcolor}
\usepackage{rotating}

\definecolor{twoturnbg}{RGB}{240, 240, 240}
\definecolor{multiturnbg}{RGB}{243, 243, 255}  

\newcommand{\subsec}[1]{\noindent\textbf{#1}~~}

\usepackage{multirow}
\usepackage{multicol}
\usepackage{xspace}
\usepackage{booktabs}
\usepackage{tabularx} 
\usepackage{tikz}

\usepackage{capt-of}

\usepackage{float}  

\usepackage[capitalize]{cleveref}
\usepackage{bm}
\usepackage{arydshln}

\global
\tcbset{
  aibox/.style={
    width=\textwidth,
    top=10pt,
    colback=white,
    colframe=black,
    colbacktitle=black,
    center,
  }
}
\newtcolorbox{AIbox}[2][]{aibox,title=#2,#1}
\tcbset{
  aiboxsmall/.style={
    width=0.62\textwidth,
    top=10pt,
    colback=white,
    colframe=black,
    colbacktitle=black,
    enhanced,
    top,
    attach boxed title to top left={yshift=-0.1in,xshift=0.15in},
    boxed title style={boxrule=0pt,colframe=white,},
  }
}
\newtcolorbox{AIboxSmall}[2][]{aiboxsmall,title=#2,#1}
\definecolor{aigold}{RGB}{244,210, 1} 
\definecolor{aired}{RGB}{255,180,181}
\newlength\savewidth

\definecolor{defaultcolor}{gray}{0.9}

\crefname{figure}{Fig.}{Figs.}
\Crefname{figure}{Fig.}{Figs.}

\crefname{table}{Tab.}{Tabs.}
\Crefname{table}{Tab.}{Tabs.}

\crefname{section}{Sec.}{Secs.}
\Crefname{section}{Sec.}{Secs.}

\newcommand{\increase}[1]{(\textcolor{ForestGreen}{+#1})}
\newcommand{\increasenoparent}[1]{\textcolor{ForestGreen}{+#1}}

\newcommand{\decrease}[1]{(\textcolor{red}{-#1})}

\newcommand{\gptfive}{GPT-5\xspace}

\newcommand{\geminiproThree}{Gemini-3-Pro-Preview\xspace}

\newcommand{\nanobanana}{Gemini-2.5-Flash-Image\xspace}
\newcommand{\nanobananapro}{Nano Banana Pro\xspace}

\newcommand{\thinkmorph}{ThinkMorph\xspace}

\newcommand{\vilasr}{ViLaSR\xspace}

\newcommand{\connectdots}{Connect-the-Dots\xspace}
\newcommand{\counting}{Counting Objects\xspace}
\newcommand{\drawingshapes}{Drawing Shapes around Objects\xspace}
\newcommand{\labeling}{Part Labeling\xspace}
\newcommand{\maze}{Maze Navigation\xspace}

\newcommand{\vpct}{VPCT\xspace}
\newcommand{\balldrop}{Ball Drop\xspace}

\newcommand{\cmark}{\textcolor{ForestGreen}{\ding{51}}} 
\newcommand{\xmark}{\textcolor{red}{\ding{55}}}         

\newcommand{\SketchVLM}{SketchVLM\xspace}
\newcommand{\SketchVLMs}{SketchVLMs\xspace}
\newcommand{\kimithree}{Kimi K3\xspace}
\newcommand{\kimitwofive}{Kimi K2.5\xspace}
\newcommand{\gemmafour}{Gemma-4-31B\xspace}
\newcommand{\qwenthreefive}{Qwen3.5-397B-A17B\xspace}

\newcommand{\geminilogo}{\raisebox{-1pt}{\includegraphics[scale=0.71]{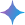}}\xspace}

\newcommand{\geminisketch}{\raisebox{-1pt}{\includegraphics[scale=0.64]{AI-Logos/Google/gemini_main.pdf}} sketchVLM \xspace}

\newcommand{\gptlogo}{\raisebox{-1pt}{\includegraphics[scale=0.013]{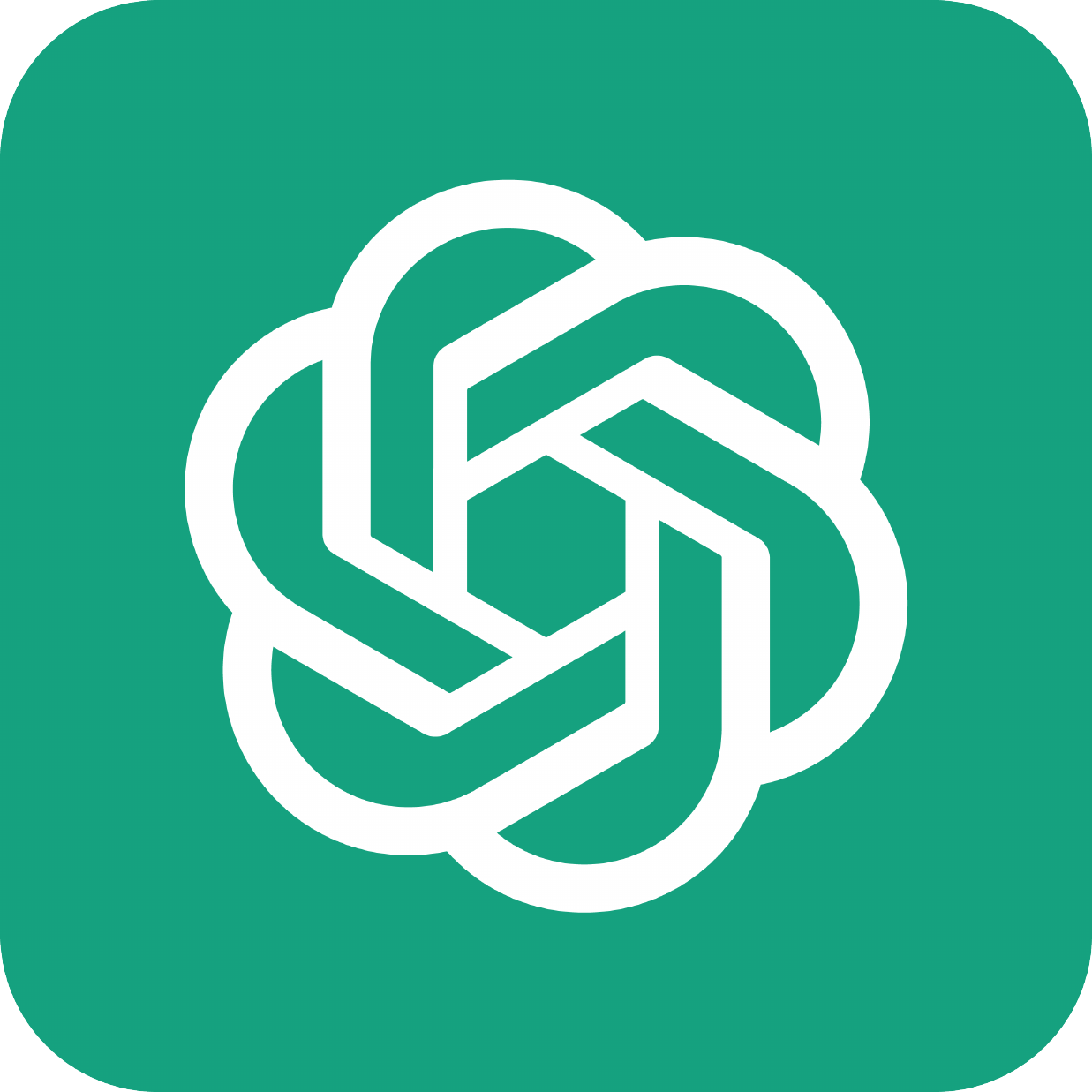}}\xspace}

\newcommand{\gptsketch}{\raisebox{-1pt}{\includegraphics[scale=0.013]{AI-Logos/OpenAI/openai_green.pdf}} sketchVLM\xspace}

\newcommand{\kimisketch}{\raisebox{-1pt}{\includegraphics[scale=0.008]{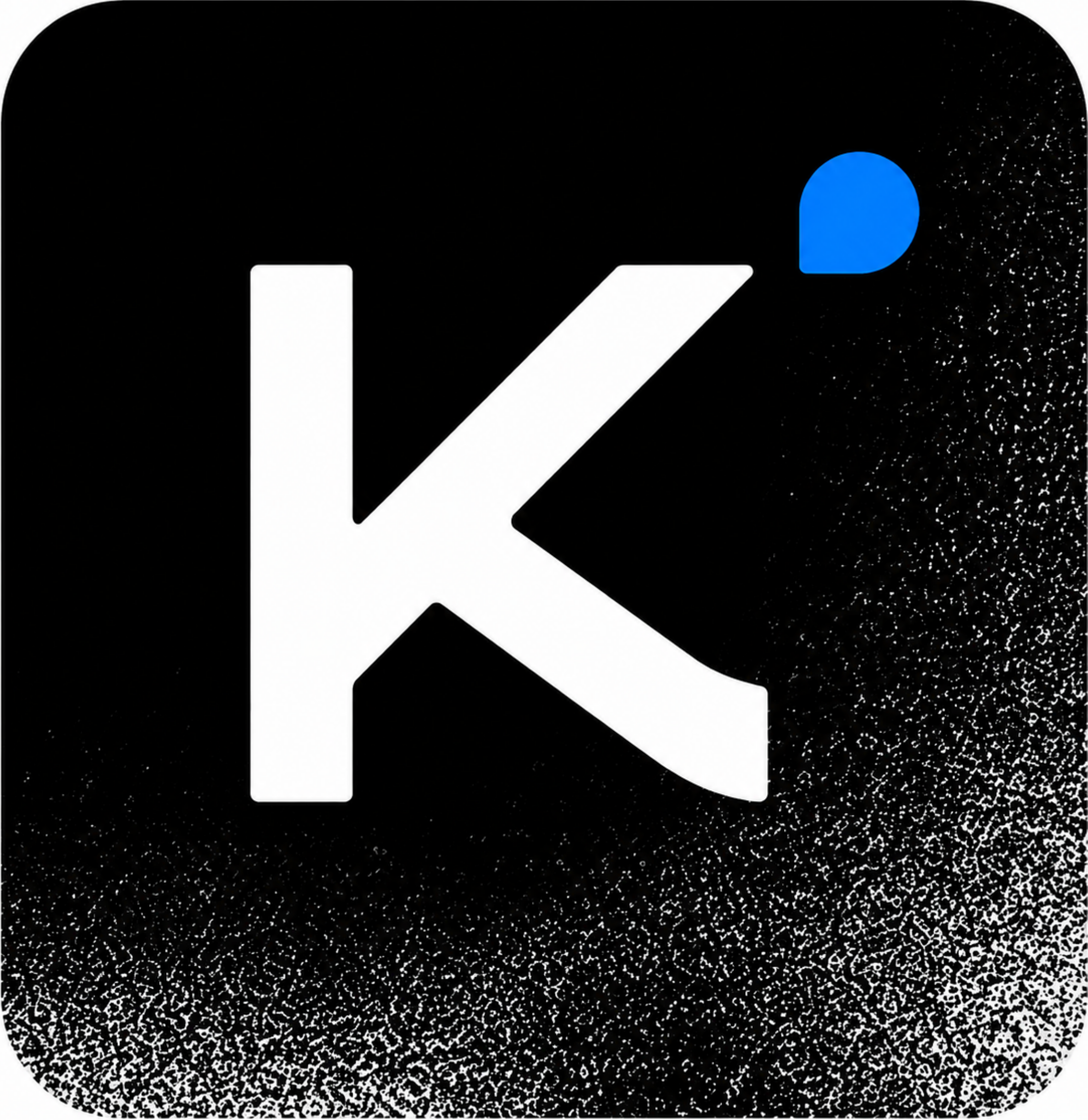}} sketchVLM\xspace}

\newcommand{\geminiplussketch}{\raisebox{-1pt}{\includegraphics[scale=0.64]{AI-Logos/Google/gemini_main.pdf}} sketchVLM \xspace}
\newcommand{\gptplussketch}{\includegraphics[scale=0.013]{AI-Logos/OpenAI/openai_green.pdf} sketchVLM\xspace}
\newcommand{\nanoplusgemini}{Nano Banana + Gemini}

\newcommand{\vilasrlogo}{{\includegraphics[scale=0.09]{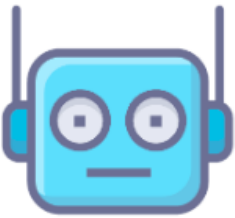}}\xspace}

\newcommand{\thinkmorphlogo}{{\includegraphics[scale=0.13]{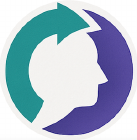}}\xspace}

\newcommand{\nanologo}{\raisebox{-1pt}{\includegraphics[scale=0.05]{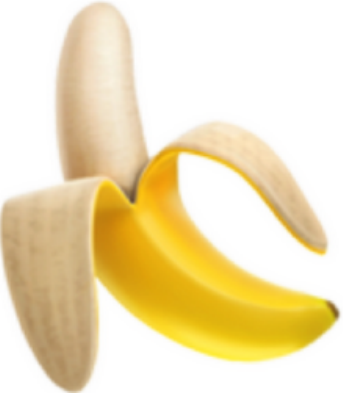}}\xspace}

\newcommand{\thickplus}{\ooalign{\hfil$\bm{+}$\hfil\cr\kern0.06ex$\bm{+}$\cr}}
\newcommand{\nanogeminilogo}{{\nanologo \hspace{-3pt}\raisebox{2pt}{\scalebox{0.8}{\thickplus}}\hspace{-3pt} \geminilogo}\xspace}

\newcommand{\correctemoji}{\raisebox{-1pt}{\includegraphics[scale=0.04]{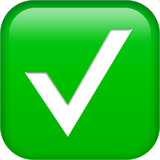}}\xspace}

\newcommand{\wrongemoji}{\raisebox{-1pt}{\includegraphics[scale=0.04]{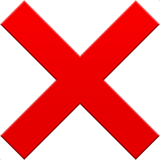}}\xspace}

\newcommand{\mazelogo}{\raisebox{-2pt}{\includegraphics[scale=0.023]{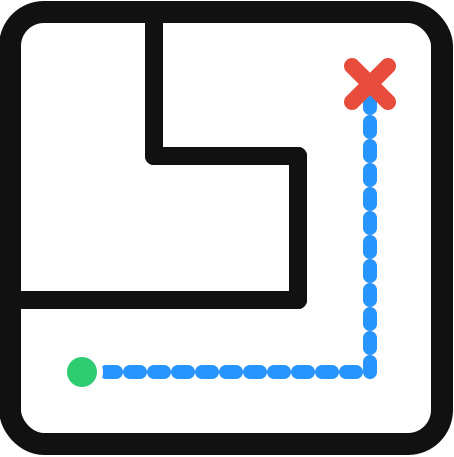}}\xspace}       
\newcommand{\connectlogo}{\raisebox{-1pt}{\includegraphics[scale=0.025]{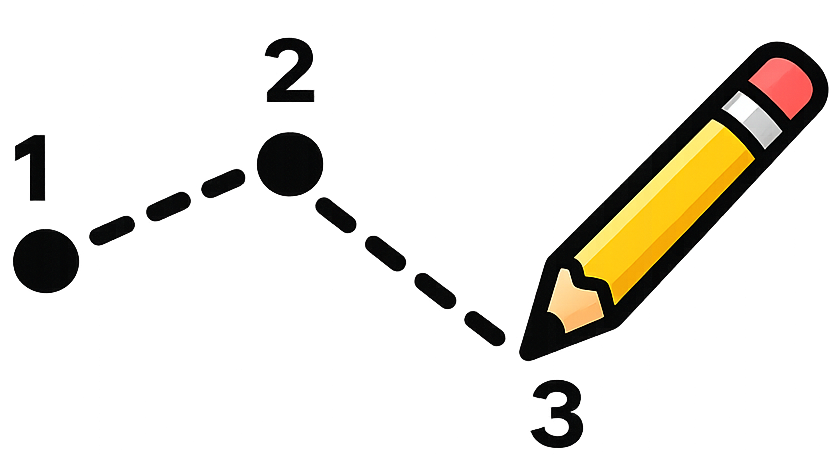}}\xspace}

\newcommand{\balllogo}{\raisebox{-2pt}{\includegraphics[scale=0.065]{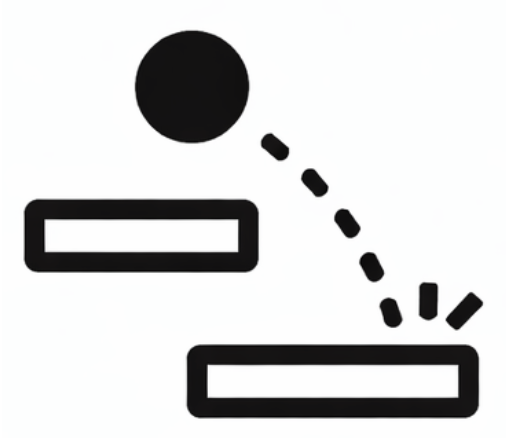}}\xspace}

\newcommand{\countlogo}{\raisebox{-2pt}{\includegraphics[scale=0.03]{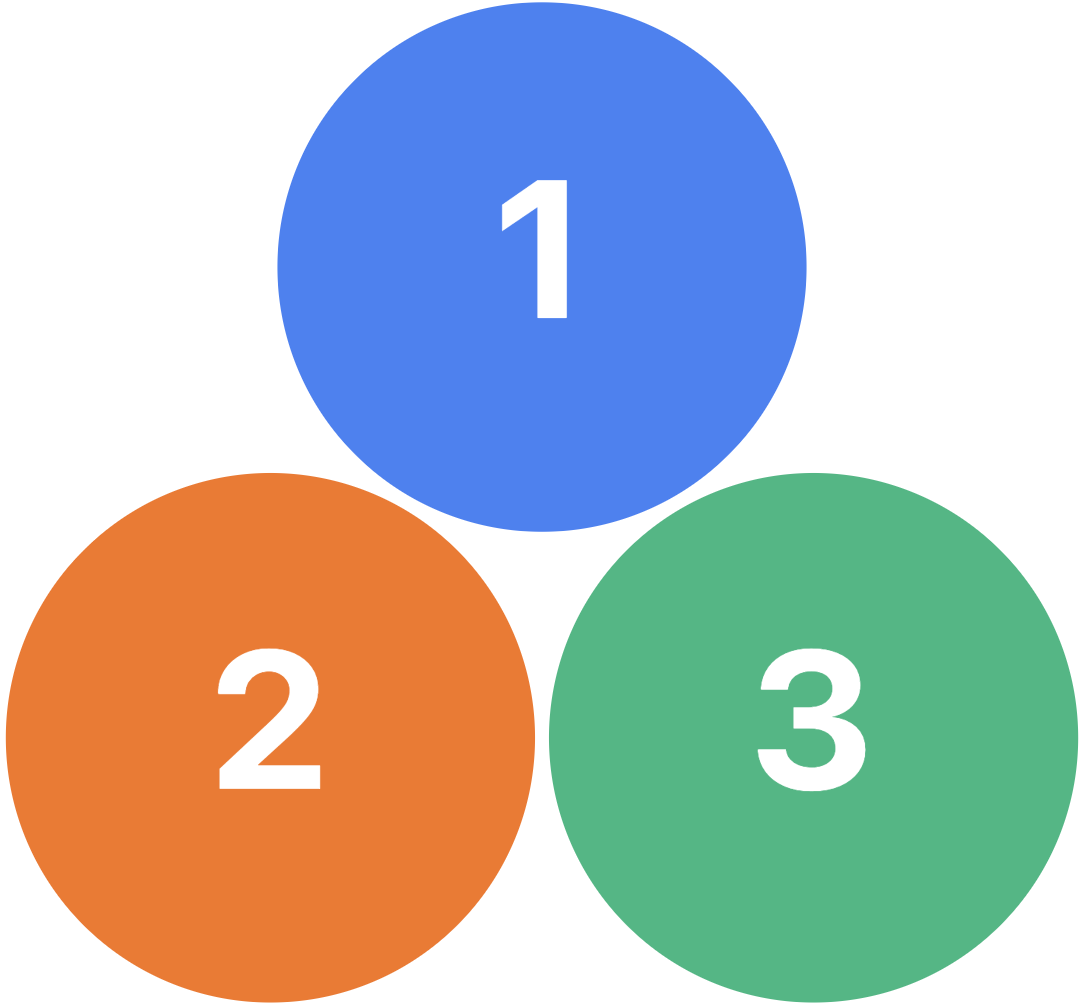}}\xspace}

\newcommand{\shapeslogo}{\raisebox{-2pt}{\includegraphics[scale=0.017]{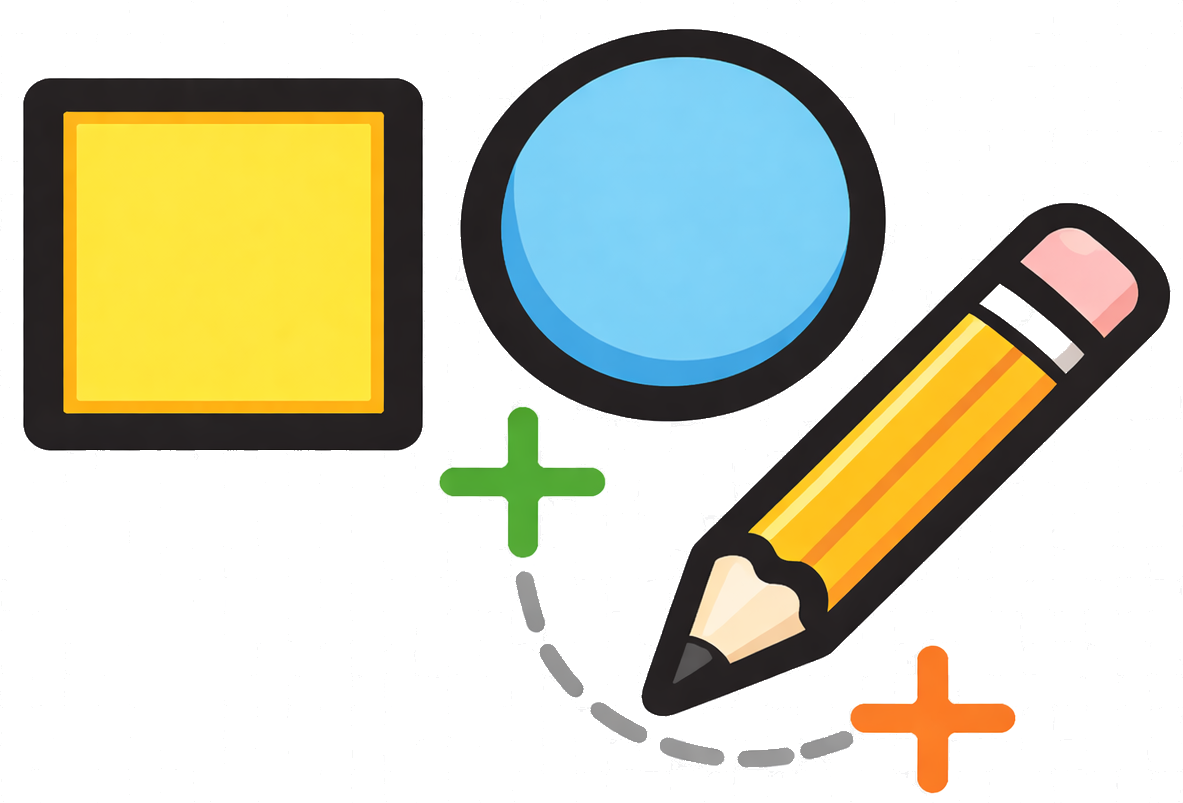}}\xspace}

\newcommand{\labellogo}{\raisebox{0pt}{\includegraphics[scale=0.015]{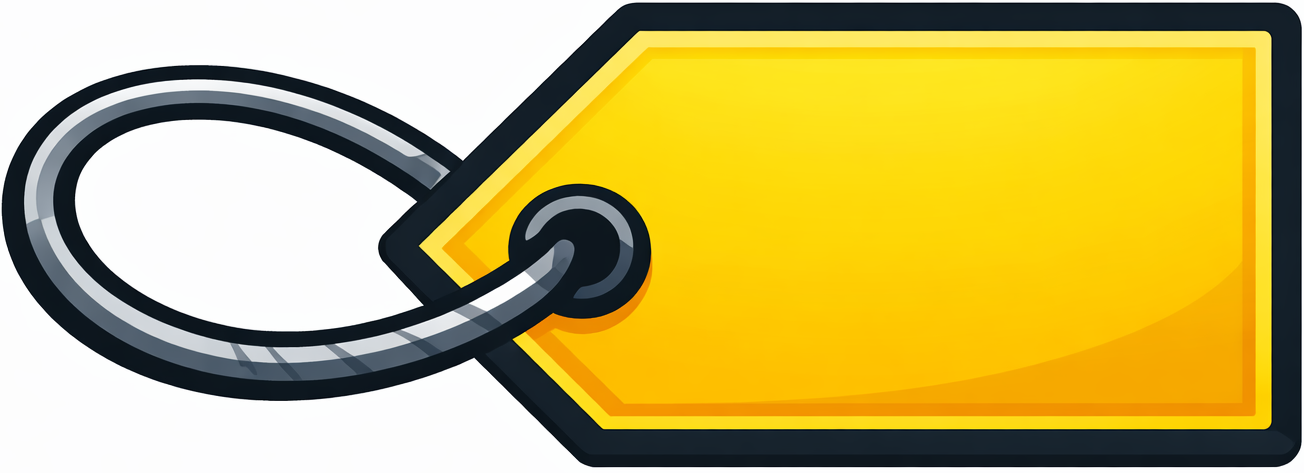}}\xspace}

\newcommand{\whiteSquareScale}{0.4}
\DeclareRobustCommand{\whitesquare}{%
  \scalebox{\whiteSquareScale}{%
    \begingroup
    \setlength{\fboxsep}{0pt}%
    \setlength{\fboxrule}{2.0pt}
    \fcolorbox{black}{white}{\makebox[6mm][c]{\rule{0pt}{6mm}}}%
    \endgroup
  }%
}

\newcommand{\whiteCircleScale}{0.5}
\DeclareRobustCommand{\whitecircle}{%
  \raisebox{-0.3ex}{%
    \scalebox{\whiteCircleScale}{%
      \tikz{
        \node[
          draw=black,
          fill=white,
          line width=2.0pt,
          circle,
          minimum size=6mm,
          inner sep=0pt
        ] {};
      }%
    }%
  }%
}

\begin{document}

\setcounter{tocdepth}{-1} 




\title{SketchVLM: Vision-Language Models Can Annotate Images to
Explain Thoughts and Guide Users}

\author{
\textbf{Brandon Collins\textsuperscript{1,*}},
\textbf{Logan Bolton\textsuperscript{1,*}},
\textbf{Hung Huy Nguyen\textsuperscript{1,*}}
\\
\textbf{Mohammad Reza Taesiri\textsuperscript{2}},
\textbf{Trung Bui\textsuperscript{3}},
\textbf{Anh Totti Nguyen\textsuperscript{1}}
\\[4pt]
\textsuperscript{1}Auburn University, Auburn, Alabama, USA
\\
\textsuperscript{2}Independent Researcher
\qquad
\textsuperscript{3}Adobe Research
\\[4pt]
\small{
\texttt{blc0063@auburn.edu},
\texttt{logan.bolton@auburn.edu},
\texttt{huyhung.dknec@gmail.com}
}
\\
\small{
\texttt{mtaesiri@gmail.com},
\texttt{bui@adobe.com},
\texttt{anh.ng8@gmail.com}
}
}


\maketitle

\begingroup
\renewcommand{\thefootnote}{\fnsymbol{footnote}}
\footnotetext[1]{All three co-first authors made major contributions to
dataset creation, experiments, and the manuscript. See the detailed
\nameref{sec:author_contribution}.}
\endgroup

\begin{abstract}
When answering questions about images, humans naturally point, label, and draw to explain their reasoning. In contrast, modern vision–language models (VLMs) such as Gemini-3-Pro and \gptfive only respond with text, which can be difficult for users to verify. We present \SketchVLM, a training-free, model-agnostic framework that enables VLMs to produce non-destructive, editable SVG overlays on the input image to visually explain their answers. Across seven benchmarks spanning visual reasoning (maze navigation, ball-drop trajectory prediction, and object counting) and drawing (part labeling, connecting-the-dots, and drawing shapes around objects), \SketchVLM improves visual reasoning task accuracy by up to \increasenoparent{28.5} percentage points and annotation quality by up to 1.48$\times$ relative to image-editing and fine-tuned sketching baselines, while also producing annotations that are more faithful to the model's stated answer. We find that single-turn generation already achieves strong accuracy and annotation quality, and multi-turn generation opens up further opportunities for human-AI collaboration. 
An interactive demo and code can be found at \href{https://sketchvlm.github.io/}{sketchVLM.github.io}.
\end{abstract}

\section{Introduction}

From browsers (Atlas, MS Edge \cite{aibrowsers}) to office products (MS Copilot, Gemini \cite{team2023gemini}), the market for LLM-powered chatbots for general end users is estimated to grow from \$1.1B in 2023 to \$83B in 2032 \cite{generative_ai_outlook}. As these assistants increasingly answer image-based questions in high-frequency consumer and productivity workflows~\cite{aibrowsers,chatapps,team2023gemini,generative_ai_outlook}, users need responses they can quickly understand and verify \cite{nguyen2025hot}.

\begin{figure*}[t]
  \centering
  \includegraphics[width=\linewidth]{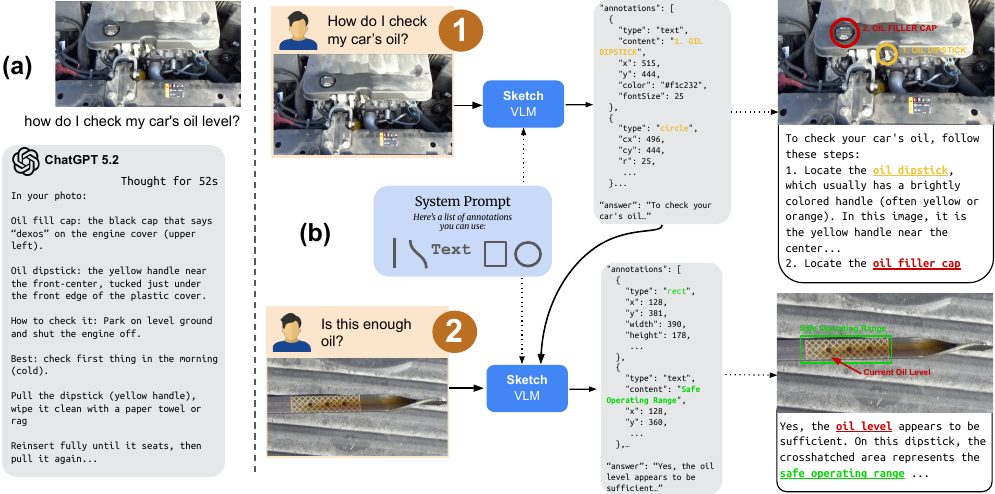}
  \vspace{-17pt}
  \caption{
  For complex questions, modern chatbots like ChatGPT often return long text responses \textbf{(a)} that are hard for users to understand, verify, and follow. In contrast, \SketchVLM guides users \textbf{(b)} step-by-step by annotating the input image and grounding answers to relevant image regions---here, guiding a user how to check their car's oil level (source: \url{https://www.youtube.com/watch?v=tNNyu9S65E4}).
  }
  \label{fig:teaser}
\end{figure*}

However, modern VLMs such as Gemini-3-Pro and \gptfive typically respond with a block of text, which can be difficult to verify \cite{zhou2025improving} (\cref{fig:image_chat_pairs}). 
To explain their thoughts on an image problem, humans often point, circle, underline, label, and annotate directly on the image. 
For example, when a user asks to check a car's oil level, a visual annotation is easier to verify than a paragraph of text (\cref{fig:teaser}) \cite{caradvertisementopenai, openai2026fixwithchatgpt}.

Some VLMs emit point coordinates for referencing objects (\eg, MoonDream \cite{vikhyat_moondream_github}, Molmo \cite{allenai_molmo_2024}), but they do not support free-form visual annotation. 
Other models are trained to generate visual reasoning traces but do not generalize well beyond their training domains \cite{wu2025reinforcingspatialreasoningvisionlanguage,gu2025thinkmorph}. Image-editing models can visualize intermediate reasoning steps for multimodal questions \cite{zou2025unimmmumassivemultidisciplinemultimodal}, but risk altering the source image in unintended ways, undermining user trust \cite{taesiri2025understanding_genai_image_editing}.

To address these issues, we propose \SketchVLM, a state-of-the-art (SotA) system that draws SVG annotations in a separate layer overlaid on top of the input image to explain its reasoning.
\SketchVLM grounds its annotations directly on the original image without modifying it and without requiring task-specific training. 
We test \SketchVLM with multiple VLM backbones, including \geminiproThree~(\geminilogo) \cite{pichai2025gemini3} and \gptfive~(\gptlogo) \cite{singh2025openaigpt5card}. We collectively refer to these models harnessed with \SketchVLM as \SketchVLMs and individually as \geminisketch and \gptsketch. We compare against the leading alternative approaches for generating visual annotations such as the SotA image-editing model \nanobananapro~\cite{Google025NanoBananaPro} and specialized VLMs fine-tuned to produce image annotations, \vilasr~\cite{wu2025reinforcingspatialreasoningvisionlanguage} and \thinkmorph~\cite{gu2025thinkmorph}. 

Across \textbf{seven tasks}: (a) \textbf{three drawing} tasks (connecting-the-dots, labeling parts of an object, and drawing shapes around objects) and (b) \textbf{four visual reasoning} tasks (two physics understanding tasks, one counting task and one maze navigation task), our main findings are:

\begin{figure*}[t]
\centering

\newlength{\labw}
\newlength{\imgwidth}
\newlength{\imgh}
\setlength{\labw}{0.04\linewidth}
\setlength{\imgwidth}{0.225\linewidth}
\setlength{\imgh}{0.16\textheight}
\setlength{\tabcolsep}{1pt}

\begin{tabular}{@{}c c c c c@{}}

& {Input} & {\geminisketch} & {\nanologo \nanobananapro} & {\vilasrlogo \vilasr} \\[-15pt]
\parbox[c][\imgh][c]{\labw}{\raggedleft
  \makebox[0pt][l]{\hspace{-1.75em}\textbf{(a)}\hspace{1.5em}}
  \rotatebox[origin=c]{90}{\tiny\textbf{\balldrop}}
} &
\parbox[c][\imgh][c]{\imgwidth}{\centering
  \includegraphics[width=\linewidth,height=\imgh,keepaspectratio]{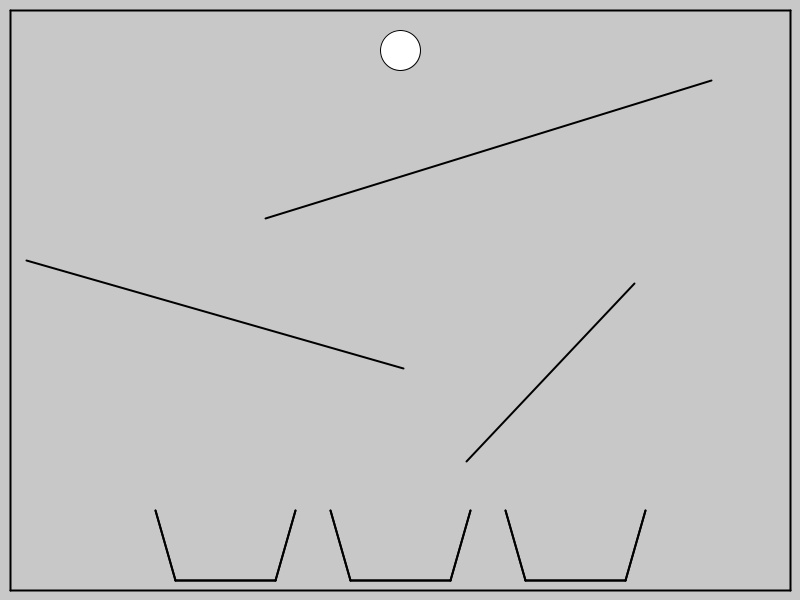}
} &
\parbox[c][\imgh][c]{\imgwidth}{\centering
  \includegraphics[width=\linewidth,height=\imgh,keepaspectratio]{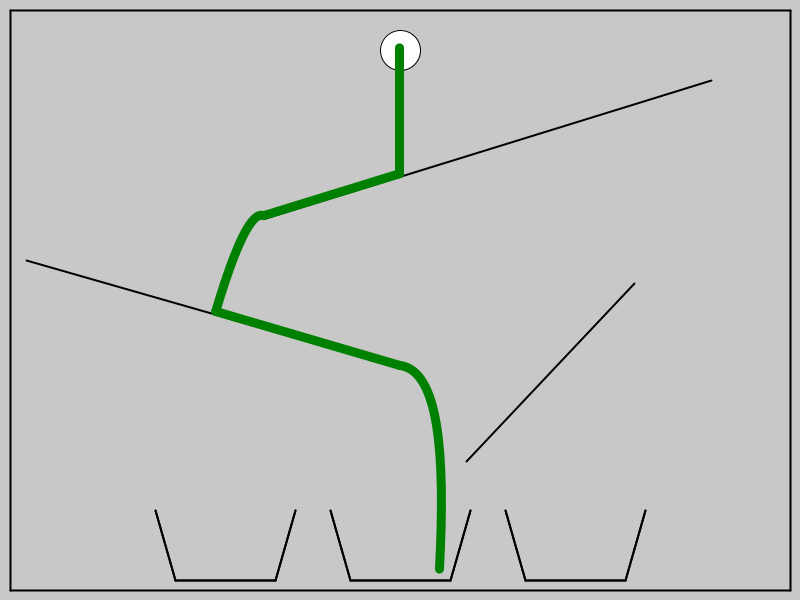}
} &
\parbox[c][\imgh][c]{\imgwidth}{\centering
  \includegraphics[width=\linewidth,height=\imgh,keepaspectratio]{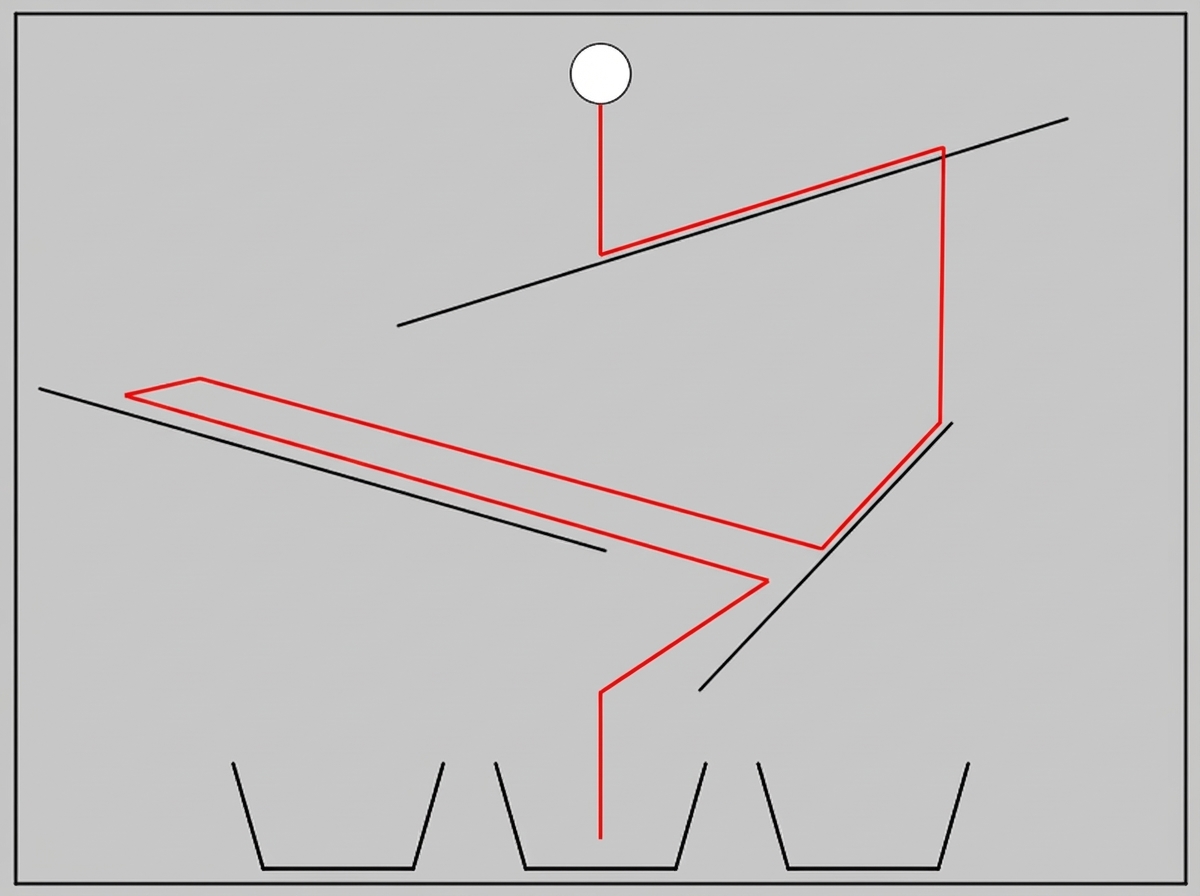}
} &
\parbox[c][\imgh][c]{\imgwidth}{\centering
  \includegraphics[width=\linewidth,height=\imgh,keepaspectratio]{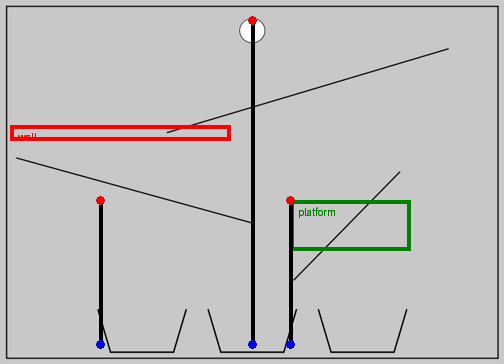}
} \\[-15pt]
\parbox[c]{\labw}{} &
\parbox[t]{\imgwidth}{\centering\scriptsize\textit{Which bucket will the ball fall into?}} &
\parbox[t]{\imgwidth}{\centering\scriptsize \texttt{Answer: 2} \correctemoji} &
\parbox[t]{\imgwidth}{\centering\scriptsize \texttt{Answer: 2} \correctemoji \\
\textit{(Ball goes through platform} \wrongemoji)} &
\parbox[t]{\imgwidth}{\centering\scriptsize \texttt{Answer: 3} \wrongemoji \\
\textit{(Ball goes through platform} \wrongemoji)} \\[11pt]

& {Input} & {\geminisketch} & {\thinkmorphlogo ThinkMorph} & {\vilasrlogo ViLaSR} \\[-5pt]
\parbox[c][\imgh][c]{\labw}{\raggedleft
  \makebox[0pt][l]{\hspace{-1.75em}\textbf{(b)}\hspace{1.5em}}
  \rotatebox[origin=c]{90}{\tiny\textbf{\connectdots}}
} &
\parbox[c][\imgh][c]{\imgwidth}{\centering
  \includegraphics[width=\linewidth,height=\imgh,keepaspectratio]{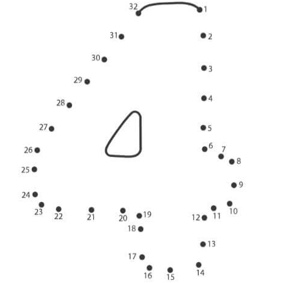}
} &
\parbox[c][\imgh][c]{\imgwidth}{\centering
  \includegraphics[width=\linewidth,height=\imgh,keepaspectratio]{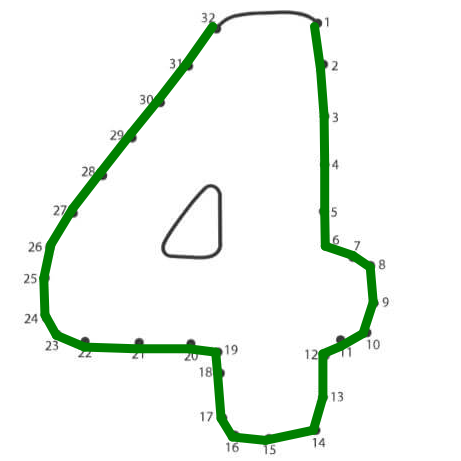}
} &
\parbox[c][\imgh][c]{\imgwidth}{\centering
  \includegraphics[width=\linewidth,height=\imgh,keepaspectratio]{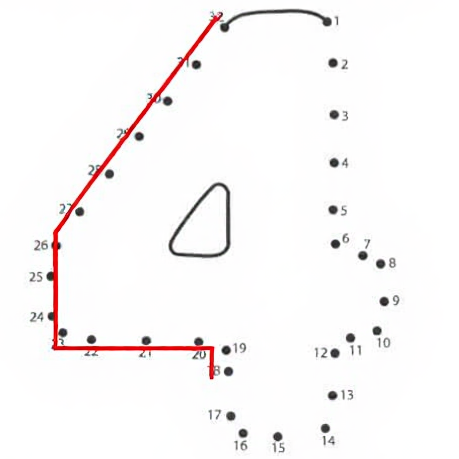}
} &
\parbox[c][\imgh][c]{\imgwidth}{\centering
  \includegraphics[width=\linewidth,height=\imgh,keepaspectratio]{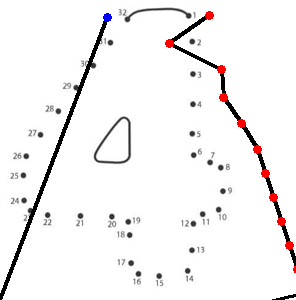}
} \\[-5pt]
\parbox[c]{\labw}{} &
\parbox[t]{\imgwidth}{\centering\scriptsize\textit{Connect the dots in order.}} &
\parbox[t]{\imgwidth}{\centering\scriptsize \textit{Correct} \correctemoji} &
\parbox[t]{\imgwidth}{\centering\scriptsize \textit{Improper Lines} \wrongemoji} &
\parbox[t]{\imgwidth}{\centering\scriptsize \textit{Improper Lines} \wrongemoji} \\[11pt]

& {Input} & {\geminisketch}
& {\nanologo \nanobananapro}
& {\thinkmorphlogo \thinkmorph} \\[-5pt]

\parbox[c][\imgh][c]{\labw}{\raggedleft
  \makebox[0pt][l]{\hspace{-1.75em}\textbf{(c)}\hspace{1.5em}}
  \rotatebox[origin=c]{90}{\tiny\textbf{\maze}}
} &
\parbox[c][\imgh][c]{\imgwidth}{\centering
  \includegraphics[width=\linewidth,height=\imgh,keepaspectratio]{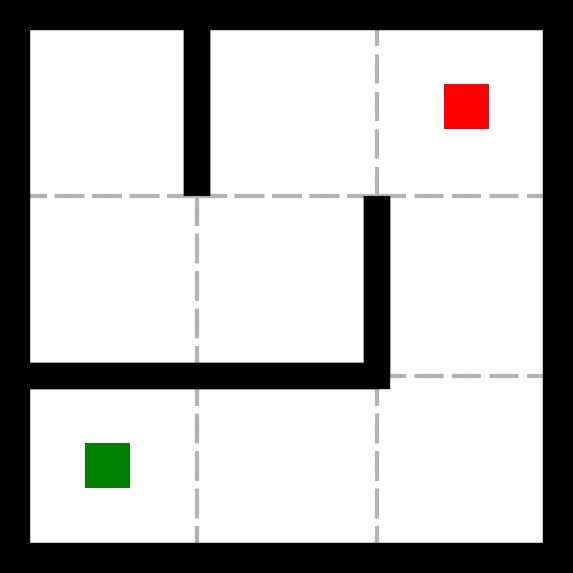}
} &
\parbox[c][\imgh][c]{\imgwidth}{\centering
  \includegraphics[width=\linewidth,height=\imgh,keepaspectratio]{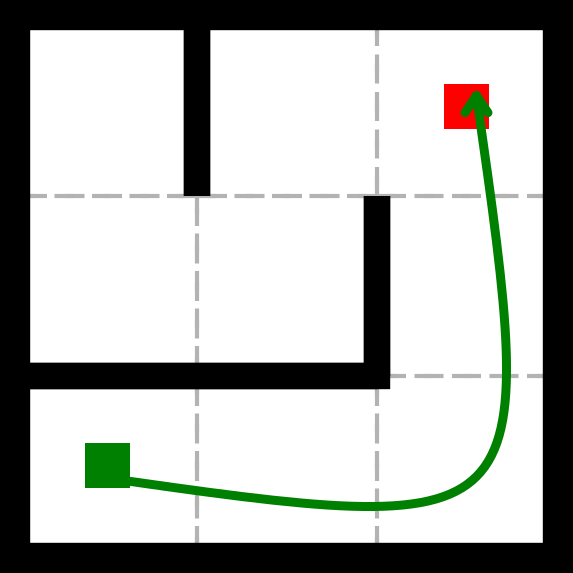}
} &
\parbox[c][\imgh][c]{\imgwidth}{\centering
  \includegraphics[width=\linewidth,height=\imgh,keepaspectratio]{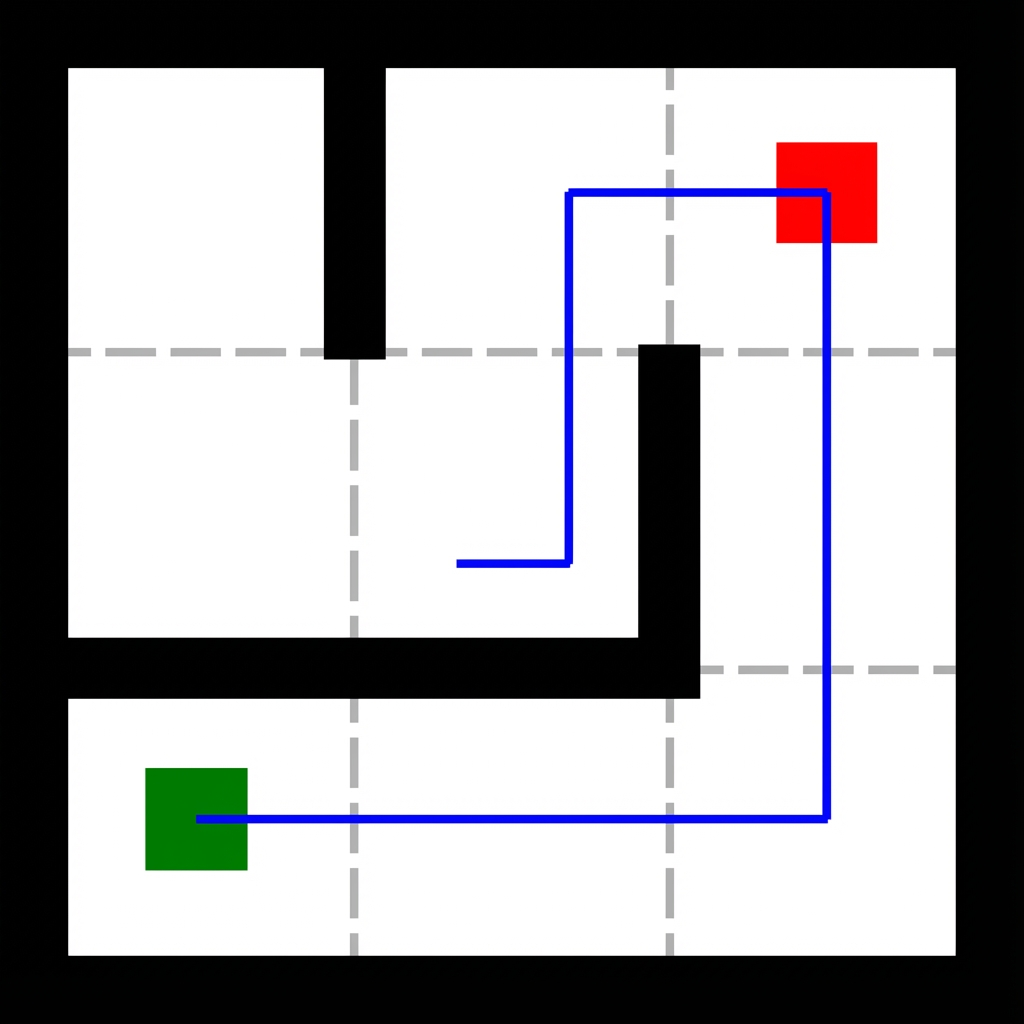}
} &
\parbox[c][\imgh][c]{\imgwidth}{\centering
  \includegraphics[width=\linewidth,height=\imgh,keepaspectratio]{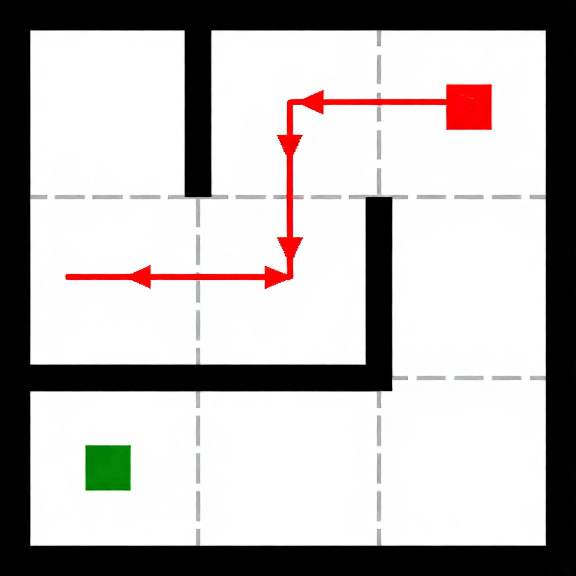}
} \\[-5pt]
\parbox[c]{\labw}{} &
\parbox[t]{\imgwidth}{\centering\scriptsize\textit{Is the path ``right, right, up, up'' from green valid?}} &
\parbox[t]{\imgwidth}{\centering\scriptsize \texttt{Answer: Yes} \correctemoji} &
\parbox[t]{\imgwidth}{\centering\scriptsize \texttt{Answer: No} \wrongemoji \quad (\textit{Wrong path} \wrongemoji)} &
\parbox[t]{\imgwidth}{\centering\scriptsize \texttt{Answer: Yes} \correctemoji \quad \textit{(Wrong path} \wrongemoji)} \\

\end{tabular}

\vspace{-5pt}
\caption{Our \geminisketch (with \geminiproThree) draws more accurate
trajectories in Ball Drop (a), connects the dots more accurately (b), and
correctly verifies the proposed maze path (c), compared with the other
baselines. \nanobananapro generates implausible trajectories while specialist fine-tuned VLMs (\thinkmorph and \vilasr) often fail to generalize to new tasks.
}
\label{fig:sketchvlm_vs_nanobanana_main}
\end{figure*}

\begin{itemize}

\item \SketchVLMs based on frontier models (\gptlogo and \geminilogo) generate annotations of superior \textbf{generalizability, accuracy, and annotation quality} compared to those by specialized fine-tuned sketching models (\vilasr and \thinkmorph) and image-editing models (\nanobananapro) (\cref{sec:results_connect_dots,sec:results_counting,sec:results_shapes,sec:results_labeling,sec:results_maze,sec:results_ball_vpct,sec:results_sketch_text_align}).

\item \SketchVLM transfers without model-specific training to four
additional open-weight VLMs, with \kimithree achieving task accuracy and
spatial precision competitive with our primary \geminiproThree backbone
(\cref{sec:results_additional_models}).

\item \SketchVLMs have similar accuracy in single-turn and multi-turn settings, but are significantly faster with single-turn generation (\cref{sec:single_vs_multi_turns}).

\item Adding an external grid of $x$--$y$ coordinates to the input image improves \gptplussketch's drawing capability and question answering accuracy but is not necessary for \geminiplussketch~(\cref{sec:best-setup-gemini-gpt}).

    
\end{itemize}

\section{Related Work}

\subsec{Native image-editing models} (\eg, GPT-Image-1.5 and \nanobananapro) can directly modify images to annotate, but they can struggle for reasoning-heavy tasks and may cause unintended alterations to the source image \cite{zou2025unimmmumassivemultidisciplinemultimodal}.
Open-source native multimodal autoregressive models (\eg, Chameleon~\cite{team2024chameleon} and Bagel~\cite{deng2025bagel}) enable interleaved text--image generation, but lack editable overlays. In contrast, \SketchVLMs generate a non-destructive SVG overlay on the input image (\cref{tab:sketch_related_work_binary}).

\subsec{Tool-calling and code generation}
Agentic systems improve visual reasoning by writing code for external tools or to manipulate the input image.
V*~\cite{wu2023vstar} uses LLM-guided visual search to localize target objects, then crops relevant regions.
Visual Sketchpad~\cite{hu2024visual} and OpenThinkIMG~\cite{su2025openthinkimg} equip VLMs with tools such as segmentation and OCR.
PyVision~\cite{zhao2025pyvision} generates Python code to draw structured overlays.
Other training-free methods leverage internal attention or gradient maps to crop and zoom~\cite{zhang2025mllms}.
These systems add additional complexity to the VLM generation process by using external tools or code rather than utilizing the VLM's natural ability to generate annotations.

\subsec{Visual prompting} \eg, drawing coordinate points or horizontal lines directly on an input image can improve the vision capabilities of VLMs~\cite{yu2024visualprompting,lei-etal-2025-scaffolding,izadi2025visual}. Similarly, SketchAgent~\cite{vinker2025sketchagent} appends a coordinate grid to the edge of the input image to allow the model to reference precise $x$--$y$ positions.


\subsec{Visual sketching}
Whiteboard-of-Thought~\cite{menon2024whiteboard} renders Matplotlib sketches before answering questions.
D2R~\cite{ou2025bridgingdynamicperceptiongap} interleaves textual reasoning with visual drafts at each reasoning step. VDLM~\cite{wang2025visually} converts images into SVG-based representations to improve visual understanding.
SketchAgent~\cite{vinker2025sketchagent} and SketchFormer~\cite{ribeiro2020sketchformertransformerbasedrepresentationsketched} focus on sketch generation on a blank canvas. In contrast, \SketchVLMs generate non-destructive, editable SVG annotations directly on input images, allowing users to inspect model reasoning without modifying the source image.

\subsec{Fine-tuned sketching models} 
MVoT~\cite{li2025imaginereasoningspacemultimodal} fine-tunes a model to generate interleaved text and image reasoning traces. LatentSketchpad~\cite{zhang2025latentsketchpad} and DeepSketcher~\cite{zhang2025deepsketcherinternalizingvisualmanipulation} both move visual reasoning into latent spaces.
ViLaSR~\cite{wu2025reinforcingspatialreasoningvisionlanguage} and \thinkmorph fine-tune VLMs to sketch on images, with \thinkmorph editing the image rather than producing overlays. Instead, we build a training-free SVG-based harness around SotA VLMs that enables them to annotate images and generalize to new domains.

\begin{table}[t]
\centering
\caption{Comparison of sketching models and methods. \textit{Annotation type} describes the visual artifact used during reasoning: \textit{Vector overlay} denotes structured, non-destructive marks (\eg, strokes/boxes/text) aligned to an image or canvas, while \textit{Image edit} denotes pixel-space image modification or synthesis that may change image content. \textit{Input image} is marked only when a provided image is the primary visual context (vs.\ blank canvas or purely generative visual thoughts). \textit{Free-form drawing} indicates support for arbitrary stroke-like annotations beyond a fixed mark set.}

\setlength{\tabcolsep}{4.5pt}
\renewcommand{\arraystretch}{1.05}

\adjustbox{max width=0.99\linewidth}{
\begin{tabular}{l c c c c c}
\toprule

Model name &
\shortstack{Training\\free} &
\shortstack{Multi\\turn} &
\shortstack{Input\\image} &
\shortstack{Free-form\\drawing} &
\shortstack{Annotation\\type} \\

\midrule

\textbf{SketchVLM (Ours)} &
\cmark & \cmark & \cmark & \cmark & Vector overlay \\

VisualSketchPad~\cite{hu2024visual} &
\cmark & \cmark & \cmark & \xmark & Vector overlay \\

PyVision~\cite{zhao2025pyvision} &
\cmark & \cmark & \cmark & \xmark & Vector overlay \\

SketchAgent~\cite{vinker2025sketchagent} &
\cmark & \cmark & \xmark & \cmark & Vector overlay \\

D2R~\cite{ou2025bridgingdynamicperceptiongap} &
\cmark & \cmark & \cmark & \xmark & Image edit \\

Whiteboard-of-Thought~\cite{menon2024whiteboard} &
\cmark & \xmark & \xmark & \xmark & Vector overlay \\

ViLaSR~\cite{wu2025reinforcingspatialreasoningvisionlanguage} &
\xmark & \cmark & \cmark & \xmark & Vector overlay \\

OpenThinkIMG~\cite{su2025openthinkimg} &
\xmark & \cmark & \cmark & \xmark & Image edit \\

ThinkMorph~\cite{gu2025thinkmorph} &
\xmark & \xmark & \cmark & \xmark & Image edit \\

MVoT~\cite{li2025imaginereasoningspacemultimodal} &
\xmark & \cmark & \cmark & \xmark & Image edit \\

LatentSketchpad~\cite{zhang2025latentsketchpad} &
\xmark & \cmark & \cmark & \xmark & Image edit \\

DeepSketcher~\cite{zhang2025deepsketcherinternalizingvisualmanipulation} &
\xmark & \cmark & \cmark & \xmark & Image edit \\

\bottomrule
\end{tabular}
}

\label{tab:sketch_related_work_binary}
\end{table}

\begin{figure*}[t]
    \centering
    \includegraphics[width=\textwidth]{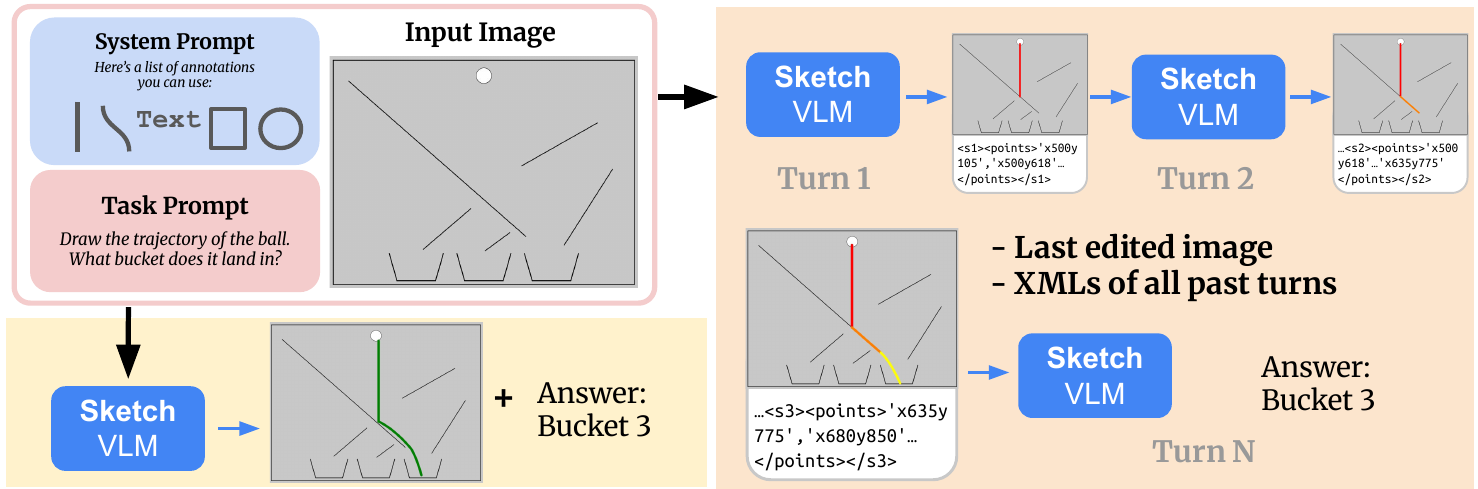}
    
    \vspace{2pt}
    \noindent
    \begin{minipage}{0.49\textwidth}
        \centering
        {(a) Single-turn}
    \end{minipage}
    \hfill
    \begin{minipage}{0.49\textwidth}
        \centering
        {(b) Multi-turn}
    \end{minipage}
    
    \caption{Single-turn and multi-turn generation on the same VPCT sample.
    In (a) \textit{single-turn}, SketchVLM receives the system prompt, the task prompt, and the input image, then outputs all annotations and the final answer in a single model call.
    In (b) \textit{multi-turn}, Turn 1 uses the same inputs and outputs one annotation.
    For later turns, the model reuses  the system prompt, the task prompt, and the previous annotations, which are provided in both the rendered image and text form.
    This process repeats until the model outputs its final text answer on the last turn.}
    \label{fig:single_multi_turn}
\end{figure*}

\section{SketchVLM}

\SketchVLM combines three components: visual prompting to aid spatial reference, a system prompt that elicits structured stroke outputs, and XML-to-SVG conversion that renders those strokes as an overlay on the source image (\cref{fig:teaser,fig:single_multi_turn}).

\subsec{\textbf{Visual prompting}} Following SketchAgent~\cite{vinker2025sketchagent}, we append a coordinate grid to the left and bottom of each image to improve spatial precision (\cref{fig:worksheet_example}).

\subsec{\textbf{Input prompt}} We introduce a system prompt (\cref{sec:sketchvlm_system_prompt}) that instructs the model to output structured annotations (\eg, XML-style \texttt{<s1>, <s2>, ..., <sN>} stroke tags). We provide instructions for drawing primitives including rectangles, arrows, text labels, straight lines, and Bézier curves. Given the task prompt (\eg, \emph{``Which bucket will the ball fall into?''}), the model annotates its reasoning before producing a final answer (\cref{fig:teaser,fig:sketchvlm_vs_nanobanana_main}).

\subsec{\textbf{SVG conversion}} We parse the model's XML outputs into SVG overlays rendered on the input image. Two-point strokes are rendered as straight lines. Otherwise, given a stroke described by $m$ ordered samples
$S_i=\{(x_j,y_j)\}_{j=1}^m$ and corresponding normalized
timestamps $T_i=\{t_j\}_{j=1}^m$ with $t_j\in[0,1]$, we fit a smooth
Bézier curve. Following
SketchAgent~\cite{vinker2025sketchagent}, a least squares solution is found for the control points in a cubic Bézier curve (see \cref{fig:appendix_annotation_example} for an example). Following SketchAgent, we run all experiments using XML format. Given that VLMs produce JSON at comparable quality and JSON is more human-readable, we use JSON for our interactive demo.

\section{Evaluation}
\subsection{7 Tasks}
\label{sec:datasets}

\subsec{1. {\connectdots~}\connectlogo} Models must locate each dot and connect them in order (\cref{app:connect_dots_dataset_creation,fig:sketchvlm_vs_nanobanana_main,app:qual-connect-dots}).

\subsec{2. {\counting~}\countlogo} Models must count the target objects and place numbered markers on each one (\cref{app:counting_dataset_creation,fig:counting_task_example_37,app:qual_counting}).

\subsec{3. {\drawingshapes~}\shapeslogo} Models must localize objects by drawing rectangles or ovals (\cref{app:shapes_dataset_creation,fig:object_detection_comparison_142238,fig:drawing_more_samples}).

\subsec{4. {\labeling~}\labellogo} Models must place the correct text labels at the corresponding part locations (\cref{app:labeling,fig:part_labeling_task_example_blender,fig:part_labelling_more_samples}).

\subsec{5. \textbf{\maze~}\mazelogo} Models must trace a proposed path and determine if it reaches the goal without crossing walls (\cref{app:maze_dataset_creation,fig:sketchvlm_vs_nanobanana_main,app:qual_maze}).

\subsec{6. {\vpct}} is the Visual Physics Comprehension Test~\cite{cbrowerVPCT2025}. Models must predict which container a dropped ball will land in (\cref{app:ball_drop_dataset_creation,fig:sketchvlm_vs_nanobanana_main,fig:vpct_qual_ex}).

\subsec{7. {\balldrop}~\balllogo} contains synthetically generated images with ground-truth ball trajectories produced using PHYRE~\cite{bakhtin2019phyre}. Models must predict the landing container and trace the ball trajectory (\cref{app:ball_drop_dataset_creation,fig:ball_drop_examples,fig:vpct_accuracy_mse,fig:ball_drop_qual_ex}).


\subsection{Setup}
\label{sec:single_turn_and_multi_turn}

\subsec{Single-turn and multi-turn} We evaluate \SketchVLMs in (1) \textit{single-turn}, where the models output all annotations and their final answer in one response, and (2) \textit{multi-turn}, producing one stroke per turn, to simulate iterative, real-world conversations. During each turn, the VLM receives the image with all previous annotations rendered, and the annotations' text representations. The model then outputs its final text answer on the last turn (\cref{fig:single_multi_turn}).

\subsec{Visual prompting} We evaluate the necessity of a coordinate grid for the model to reference specific points in an image. When omitting the grid, the model outputs coordinates on a normalized 1000$\times$1000 scale (\cref{fig:coord_1000_vs_2000}).

\subsection{Baselines}


\subsec{Image Editing Model + VLM:} \nanobananapro~is an image-editing model that can annotate images, but produces no text answer. To obtain a text response, we feed its edited image to \geminiproThree (\geminilogo), denoted as \nanoplusgemini.

\subsec{Fine-tuned Sketching Models:} \vilasr~is fine-tuned from Qwen-2.5-VL-7B \cite{bai2025qwen25vltechnicalreport} to autoregressively generate SVG annotations on the input image over multiple turns, and \thinkmorph~is fine-tuned from BAGEL-7B-MoT \cite{deng2025bagel} to directly edit the input image and produce a text answer.

\subsec{Default VLMs:} We additionally evaluate text-only  \geminiproThree (\geminilogo) and \gptfive (\gptlogo).

\subsection{Metrics}
\label{sec:metrics}

We measure three aspects to test the accuracy and consistency between text responses and annotations. \textbf{1) Accuracy} serves as the primary measure of task performance and tests the effect of sketching on answer correctness. \textbf{2) Annotation--text alignment} measures how faithful the visual traces are to the text answer.  We ask a VLM judge \cite{pmlr-v235-chen24h} to infer the answer from the annotations and report how often the inferred answer matches the final text answer. \textbf{3) Annotation quality} allows us to distinguish models that produce informative annotations from those that output low-quality or incoherent drawings. We adopt a VLM-as-a-Judge approach using a rubric scored from 1 to 5 (\cref{vlm-judge-details}) that evaluates annotation plausibility and visual clarity for each task.

\noindent\section{Results}

\subsection{Grid prompting improves  \gptfive's \gptlogo annotation precision, but hurts \geminiproThree \geminilogo}
\label{sec:best-setup-gemini-gpt}

\begin{table}[h]
\centering
\setlength{\tabcolsep}{6pt}
\renewcommand{\arraystretch}{1.0}

\caption{Ablation across inputs in single-turn mode. ``Sketch'' adds
strokes/system prompt; ``Grid'' additionally overlays the coordinate grid.
Median per-sample RMSE is reported for \connectdots while accuracy is
reported for the other tasks. Lower median RMSE is better.
\gptsketch works best with the grid while \geminisketch works best without
the grid.\\
\geminilogo{} = \geminiproThree, \gptlogo{} = \gptfive.}
\label{tab:ablation_models}

\adjustbox{max width=\columnwidth}{%
\begin{tabular}{l l c c c c}
\toprule
Model
& Input
& \vpct
& \balllogo
& \mazelogo
& \connectlogo \\
\midrule

\multirow{5}{*}{\gptlogo}
& Image
& 63.5
& 66.0
& 92.3
& N/A \\

& + Grid
& 67.0
& 67.2
& \textbf{98.0}
& N/A \\

& + Sketch
& 59.0
& 63.1
& 88.7
& 110.5 \\

& + Sketch + Grid + Multi-turn
& 68.0
& 63.1
& 82.1
& 41.7 \\

& + Sketch + Grid (\ie, \textbf{\gptsketch})
& \textbf{70.0}
& \textbf{68.5}
& 92.8
& \textbf{40.1} \\

\midrule

\multirow{5}{*}{\geminilogo}
& Image
& 89.3
& \textbf{83.8}
& 99.3
& N/A \\

& + Grid
& 90.0
& 81.3
& \textbf{99.5}
& N/A \\

& + Sketch (\ie, \textbf{\geminisketch})
& \textbf{96.0}
& 79.7
& 98.0
& \textbf{4.6} \\

& + Sketch + Multi-turn
& 80.0
& 79.8
& 98.2
& 4.8 \\

& + Sketch + Grid
& 91.0
& 82.3
& \textbf{99.5}
& 33.2 \\

\bottomrule
\end{tabular}%
}
\end{table}

\newcommand{\tighticon}[1]{\raisebox{0.2ex}[0pt][0pt]{#1}}

\begin{table*}[t]
\centering
\caption{\SketchVLMs (using \geminilogo \geminiproThree and
\gptlogo \gptfive) produce visual reasoning traces while maintaining
competitive accuracy. \nanoplusgemini~underperforms default
\geminilogo \geminiproThree, and fine-tuned sketching models
(\vilasr, \thinkmorph) perform poorly on all tasks.}
\label{tab:full_results}
\small

\adjustbox{max width=\textwidth}{%
\begin{tabular}{l |
w{c}{5.5em} w{c}{5.5em} w{c}{5.5em} w{c}{4.5em} |
w{c}{4.5em} w{c}{4.5em} w{c}{4.5em} w{c}{4.5em} w{c}{4.5em}}

\multicolumn{1}{c|}{\multirow{3}{*}{Model}}
& VPCT
& \tighticon{\textbf{\balllogo}}
& \tighticon{\textbf{\mazelogo}}
& \tighticon{\textbf{\countlogo}}
& \tighticon{\textbf{\labellogo}}
& \multicolumn{2}{c|}{
    \raisebox{1.5ex}[0pt][0pt]{
        \tighticon{\textbf{\shapeslogo}}
        \hspace{0.35em}
        \raisebox{1.0ex}[0pt][0pt]{\scriptsize draw shapes}
    }
}
& \multicolumn{2}{c}{
    \raisebox{1.5ex}[0pt][0pt]{
        \tighticon{\connectlogo}
        \hspace{0.35em}
        \raisebox{1.0ex}[0pt][0pt]{\scriptsize connect dots}
    }
} \\[-0.3em]

& \scriptsize video physics
& \scriptsize ball drop
& \scriptsize maze trace
& \scriptsize counting
& \scriptsize labeling
& \multicolumn{2}{c|}{}
& \multicolumn{2}{c}{} \\[-1.8em]

\cmidrule(lr){7-8}
\cmidrule(lr){9-10}

& & & & & &
\whitesquare
& \whitecircle
& Median RMSE
& Order Acc\% \\

\midrule

\geminisketch
& \textbf{96.0 $\pm$ 1.4}
& 79.7 $\pm$ 2.8
& 98.0 $\pm$ 1.7
& \textbf{94.5}
& 60.3
& 58.8
& 55.4
& \textbf{4.6}
& \textbf{99.0} \\

\geminilogo \geminiproThree
& 89.3 $\pm$ 2.2
& \textbf{83.8 $\pm$ 3.4}
& \textbf{99.3 $\pm$ 0.8}
& 93.0
& \textbf{64.1}
& \textbf{63.1}
& \textbf{59.8}
& $\dagger$
& $\dagger$ \\

\midrule

\gptsketch
& \textbf{70.0 $\pm$ 2.9}
& \textbf{68.5 $\pm$ 2.2}
& \textbf{92.8 $\pm$ 2.5}
& \textbf{75.4}
& \textbf{20.4}
& 18.7
& 11.2
& 40.1
& 74.0 \\

\gptlogo \gptfive
& 63.5 $\pm$ 2.5
& 66.0 $\pm$ 4.7
& 92.3 $\pm$ 2.3
& 72.0
& 19.1
& \textbf{22.4}
& \textbf{15.4}
& $\dagger$
& $\dagger$ \\

\midrule

\nanogeminilogo Nano Banana + Gemini
& 63.0
& 62.6
& 93.3 $\pm$ 3.0
& 91.7
& $\dagger$
& $\dagger$
& $\dagger$
& $\dagger$
& 39.0 \\

\vilasrlogo \vilasr
& 37.0
& 35.9
& 50.8 $\pm$ 1.5
& 48.6
& --
& --
& $\dagger$
& 175.5
& 9.0 \\

\thinkmorphlogo \thinkmorph
& 27.0
& 30.3
& 62.5 $\pm$ 2.1
& 68.1
& $\dagger$
& $\dagger$
& $\dagger$
& $\dagger$
& 0.0 \\

\bottomrule
\end{tabular}
}

\vspace{4pt}
\parbox{\textwidth}{\raggedright\scriptsize
\textit{$\dagger$ Model is \textcolor{red}{unable} to output this format.}}

\end{table*}

To understand how applying an external coordinate grid to the image affects \SketchVLMs' annotation accuracy, we run ablations with different setups.

\subsec{\textbf{Experiment}} We evaluate four input configurations in single-turn mode: the base image alone, image with grid, image with sketching prompt, and image with both. We report accuracy on \vpct, \balldrop, and \maze, and median per-sample RMSE on \connectdots (\cref{tab:ablation_models}).

\subsec{\textbf{Results}} \gptplussketch~performs best with both the grid and sketching prompt, with the grid providing a consistent boost to spatial precision (\cref{fig:worksheet_example}), consistent with prior findings \cite{izadi2025visual}. \geminiplussketch~performs better without the grid, and adding it notably degrades localization on \connectdots (median per-sample RMSE increases from 4.6 to 33.2). We therefore report all results in \cref{tab:full_results} with the input grid for \gptplussketch~but without the grid for \geminiplussketch. We treat grid usage as a model-dependent hyperparameter rather than an essential component of the method. A discretized grid limits the minimum possible RMSE, and we hypothesize that stronger models localize better than this limit while smaller models improve with the grid. We investigate this discrepancy further in \cref{fig:coord_1000_vs_2000} and \cref{tab:connectdots_categories_mse}.

\subsection{\SketchVLMs can localize points and connect them in order \connectlogo}
\label{sec:results_connect_dots}

Connecting the dots tests both spatial grounding and producing multiple coherent strokes in a row.

\subsec{\textbf{Experiment}} For each sample, we compute root mean squared error (RMSE) in pixels between the ground-truth
points and their closest generated points. We report the median
per-sample RMSE across the dataset. To evaluate whether models connect points in the correct order, we compare each predicted segment \textit{i} against all ground-truth segment pairs using MSE. If segment \textit{i} has lower MSE to a different ground-truth pair than its expected pair \textit{(i, i+1)}, it is counted as an ordering error. Because \nanobananapro and \thinkmorph only produce an image response with no x-y coordinates, we manually evaluate them.

\subsec{\textbf{Results}} \geminiplussketch~can connect up to 35 points with a low median per-sample RMSE of 4.6 (\cref{tab:full_results}). \gptplussketch~has a higher median per-sample RMSE of 40.1, but performs better than \vilasr's median per-sample 175.5 RMSE. \nanobananapro, \thinkmorph, and \vilasr frequently produce ordering errors and inadvertently alter the input image (\cref{sec:nano_connect_dots_eval}). In contrast, \gptplussketch~and \geminiplussketch~correctly order points 74\% and 99\% of the time, respectively, demonstrating that \SketchVLMs can reliably scale to many strokes while maintaining spatial accuracy and logical coherence (\cref{fig:curves_vs_lines}).

\subsection{\SketchVLM improves counting accuracy over text-only baselines \countlogo}
\label{sec:results_counting}

\begin{figure*}[t]
\centering
\begin{minipage}[t]{0.26\linewidth}
  \centering
  \vspace{0pt}
  \parbox[t][1.0em][t]{\linewidth}{\centering Input\vphantom{\geminilogo}}\\[0.1ex]
  \includegraphics[width=\linewidth]{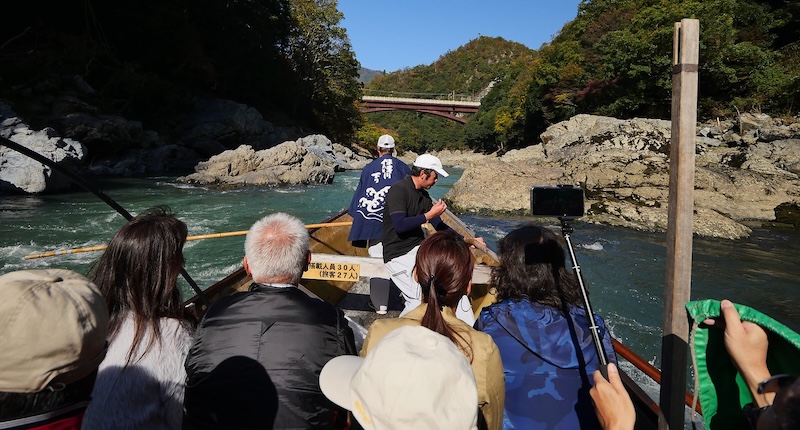}\\[0.2ex]
  \parbox[t][0.9em][t]{\linewidth}{\centering\phantom{Pred: 9\ \correctemoji}}
\end{minipage}\hfill
\begin{minipage}[t]{0.195\linewidth}
  \centering
  \vspace{0pt}
  \parbox[t][1.0em][t]{\linewidth}{\centering (a) \nanologo \nanobananapro}\\[0.1ex]
  \includegraphics[width=\linewidth]{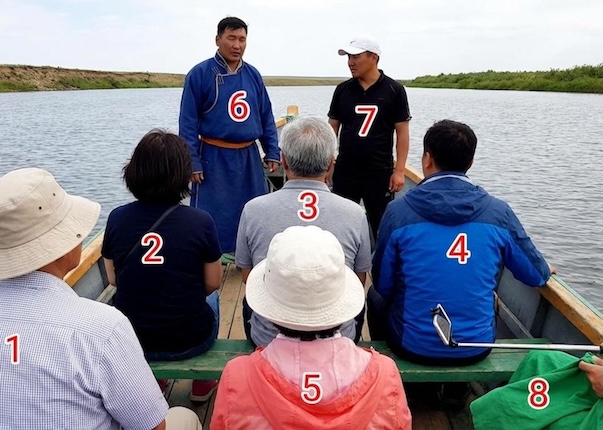}\\[0.2ex]
  \parbox[t][0.9em][t]{\linewidth}{\centering Answer: 8\ \wrongemoji}
\end{minipage}\hfill
\begin{minipage}[t]{0.26\linewidth}
  \centering
  \vspace{0pt}
  \parbox[t][1.0em][t]{\linewidth}{\centering (b) \geminisketch}\\[0.1ex]
  \includegraphics[width=\linewidth]{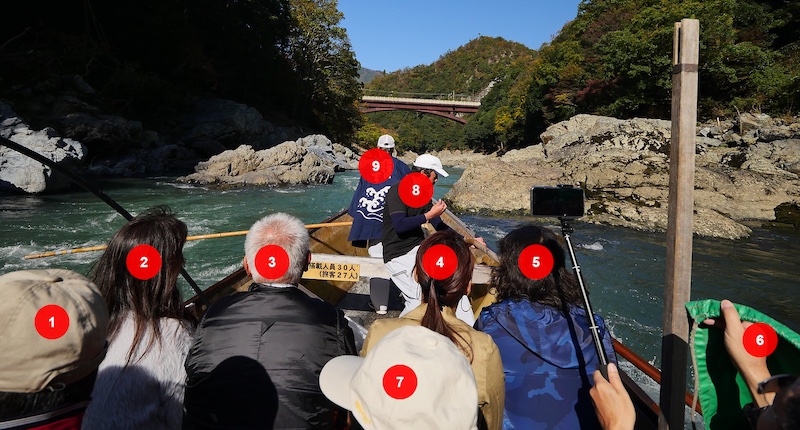}\\[0.2ex]
  \parbox[t][0.9em][t]{\linewidth}{\centering Answer: 9\ \correctemoji}
\end{minipage}\hfill
\begin{minipage}[t]{0.26\linewidth}
  \centering
  \vspace{0pt}
  \parbox[t][1.0em][t]{\linewidth}{\centering (c) \geminilogo}\\[0.1ex]
  \includegraphics[width=\linewidth]{figure/tasks/counting/37_source2.jpg}\\[0.2ex]
  \parbox[t][0.9em][t]{\linewidth}{\centering Answer: 3\ \wrongemoji}
\end{minipage}

\caption[Counting example]{%
  (a) \protect\nanologo \nanobananapro\hspace{-3pt} generates a different image and predicts an incorrect count.
  (c) \protect\geminilogo{} \geminiproThree directly outputs only a number without annotations and severely undercounts.
  In contrast, our \protect\geminisketch (b) outputs the correct answer and produces visual annotations to explain its answer.}
\label{fig:counting_task_example_37}
\end{figure*}


\noindent Existing VLMs can output point coordinates to count, but these points are unlabeled and can be tedious to verify. 
In contrast, \SketchVLMs enable direct visual verification with numeric markers.

\subsec{Experiment}
We measure the accuracy of \SketchVLMs and also how well they ground their markers on counting datasets (\cref{sec:datasets}). We consider a marker correct if it lies within the corresponding ground-truth bounding box, obtained via SAM-3~\cite{carion2025sam}, allowing at most one marker per object.

\subsec{Results}
\geminiplussketch~exhibits strong consistency between counting accuracy (94.5) and numeric-marker location accuracy (95.9), whereas \gptplussketch~achieves high counting accuracy (75.4) but substantially lower numeric-marker location accuracy (51.0) (\cref{tab:full_results,tab:text_location_accuracy_counting}). 
\vilasr attains low counting accuracy (48.6) and numeric-marker location accuracy (59.9), indicating limited performance in both aspects (\cref{tab:full_results,tab:text_location_accuracy_counting}). SketchVLM improves counting accuracy, yielding gains of \increase{1.5} points for \geminiplussketch~and \increase{3.4} points for \gptplussketch~ (\cref{tab:full_results}).
The explicit numeric markers further enable direct visual verification of model outputs (\cref{fig:counting_task_example_37,fig:counting_qualitative_all_models}). 

\begin{figure*}[t]
\centering
\newcommand{\imgH}{21mm}            
\newcommand{\wA}{0.244\linewidth}   
\setlength{\tabcolsep}{0.5mm}       

\begin{tabular}{@{} c c c c @{}}
(a) \geminisketch &
(b) \geminisketch &
(c) \nanologo \nanobananapro &
(d) \geminilogo \\[-0.2ex]
\textit{Free-form Ovals} & \textit{Free-form Rectangles} & & \textit{Oval Primitive} \\[0.6ex]
\includegraphics[height=\imgH,width=\wA,keepaspectratio]{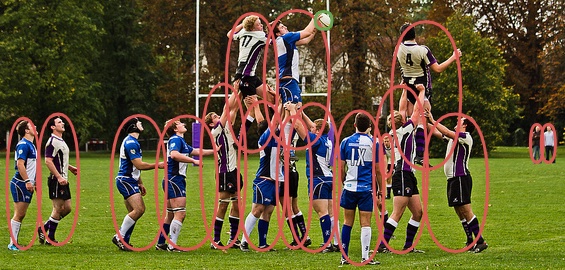} &
\includegraphics[height=\imgH,width=\wA,keepaspectratio]{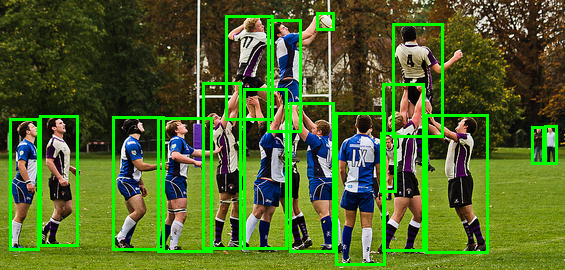} &
\includegraphics[height=\imgH,width=\wA,keepaspectratio]{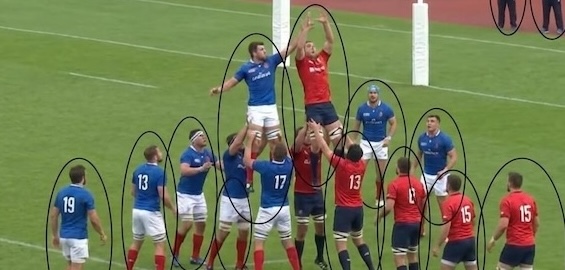} &
\includegraphics[height=\imgH,width=\wA,keepaspectratio]{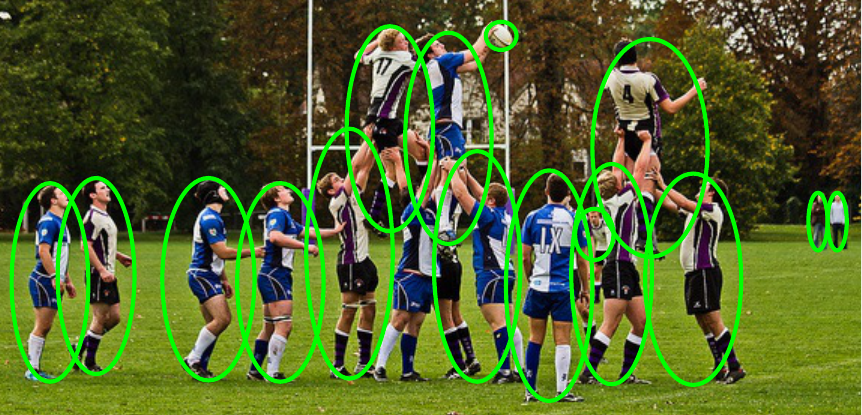}
\end{tabular}

\caption{%
  When prompted to outline the classes ``person'' and ``sports-ball'', (c) \nanologo \nanobananapro replaces the original image with a newly generated one,
  whereas \SketchVLM in (a) and (b) preserves the original image and draws shapes that align with object boundaries and locations as compared to default (d) \geminilogo \geminiproThree.}
\label{fig:object_detection_comparison_142238}
\end{figure*}

\subsection{\SketchVLMs localize objects more accurately with pre-defined shape primitives than with free-form annotations \shapeslogo}

\label{sec:results_shapes}


\noindent One design choice is whether to have \SketchVLMs generate all of their drawings through free-form strokes, or to allow them to output shape information (such as the $x$--$y$ position of the center of the circle and the radius) and then render through SVG conversion. Free-form strokes are more flexible but require point-by-point drawing. Pre-defined primitives for rectangles, ovals, etc. can be specified by a few parameters and rendered automatically.

\subsec{\textbf{Experiment}} 
We compare the accuracy of \SketchVLMs' stroke-based annotations against the baseline models that directly output parameters for shape locations and shape size (\cref{sec:datasets}). To evaluate oval annotations, we convert the rendered shape into its tight enclosing bounding box before computing metrics. Performance is measured using Average Precision (AP) at an IoU threshold of 0.5.

\subsec{\textbf{Results}} 
For \geminiplussketch~, stroke-based rectangles slightly improve performance on medium \increase{0.6} and large objects \increase{1.4}, but significantly degrade small-object detection \decrease{10.1}, reducing overall AP50 from 63.1 to 58.8 (\cref{fig:object_detection_comparison_142238,fig:drawing_shape_qualitative_all_models,tab:shape_ap50_gemini}). 

Further analysis shows that \SketchVLMs match the original model in precision but exhibit lower recall (\cref{tab:total_detection_metrics}), while prompt variations have minimal effect (\cref{tab:drawing_shape_ablation}). To combine the localization accuracy of pre-defined primitives with the expressiveness of free-form strokes, we allow the model to output both simultaneously (\cref{fig:teaser}).

\subsection{\SketchVLM improves localization accuracy for \gptfive but not for \geminiproThree when labeling parts of an object \labellogo}
\label{sec:results_labeling}


\SketchVLMs can place semantically and spatially correct labels, enabling step-by-step instructions and visual explanations (\cref{fig:aws_demo,fig:teaser}).

\subsec{\textbf{Experiment}}
Models predict both part labels and locations from a predefined label set. In \SketchVLM, the framework automatically renders the predicted text at the predicted location (with adaptive size and color). For fair comparison, we render predicted labels from the original VLMs onto the image using a Python post-processing pipeline with manually chosen text size and color.  We follow \cite{Cheng_2021_CVPR} and evaluate under boundary dilation radius $r$ to allow tolerance in spatial matching.

\subsec{\textbf{Results}}
\SketchVLM improves part labeling for \gptfive\ \increase{1.3} but slightly underperforms \geminiproThree\ at strict boundary matching \decrease{3.8} (\cref{tab:dilation_accuracy}). Under boundary dilation, the Gemini gap narrows from \decrease{3.8} at $r=0$ to \decrease{0.6} at $r=7$. Error analysis shows the original VLMs produce more missing-label errors, while \SketchVLMs produce more position errors (\cref{tab:appendix_labeling_error_breakdown}) that are typically visually negligible (\cref{fig:part_labeling_task_example_blender,fig:boundary_dilation,fig:partlabelling_qualitative_all_models}). These findings suggest that most SketchVLM errors are minor spatial shifts rather than semantic labeling failures (\cref{fig:boundary_dilation}).

\subsection{\SketchVLMs outperform models that were fine-tuned directly on path-tracing tasks \mazelogo}
\label{sec:results_maze}

\begin{figure}[h]
  \centering
  
  \setlength{\tabcolsep}{2pt}

  \begin{minipage}{0.24\linewidth}
    \centering
    {\scriptsize (a) Input} \\[1ex]
    \includegraphics[width=\linewidth]{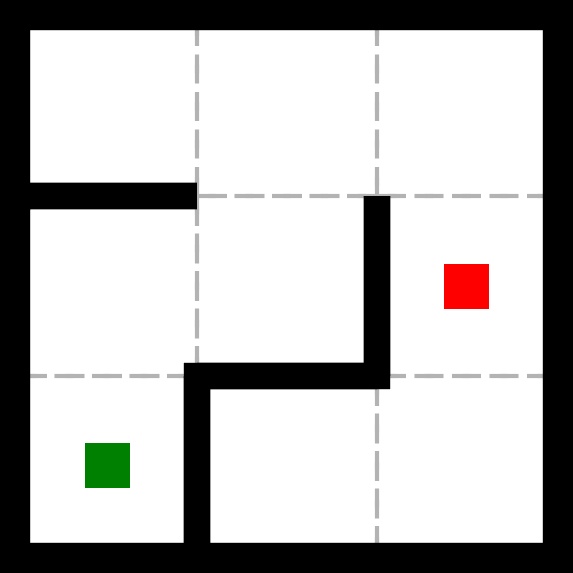}
  \end{minipage}\hfill
  \begin{minipage}{0.24\linewidth}
    \centering
    {\scriptsize (b) \geminisketch} \\[1ex]
    \includegraphics[width=\linewidth]{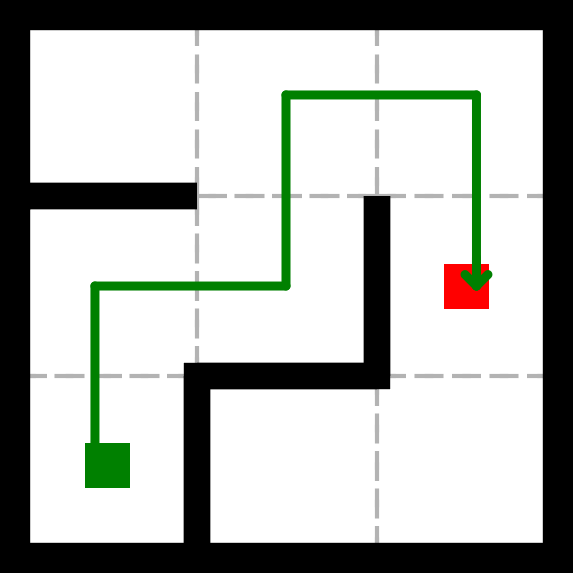}
  \end{minipage}\hfill
  \begin{minipage}{0.24\linewidth}
    \centering
    {\scriptsize (c) Input} \\[1ex]
    \includegraphics[width=\linewidth]{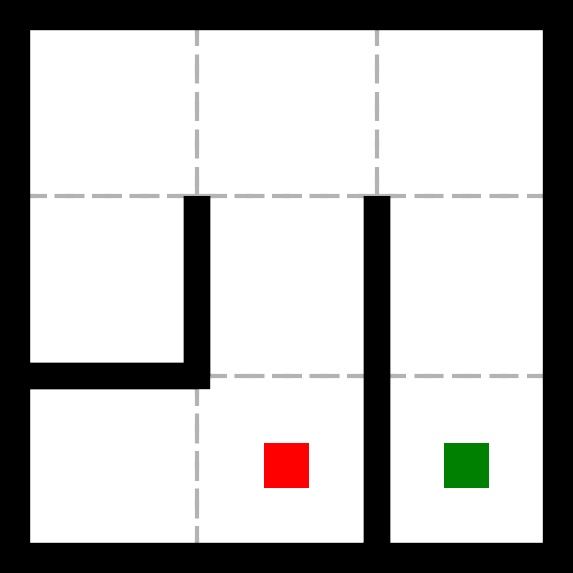}
  \end{minipage}\hfill
  \begin{minipage}{0.24\linewidth}
    \centering
    {\scriptsize (d) \geminisketch} \\[1ex]
    \includegraphics[width=\linewidth]{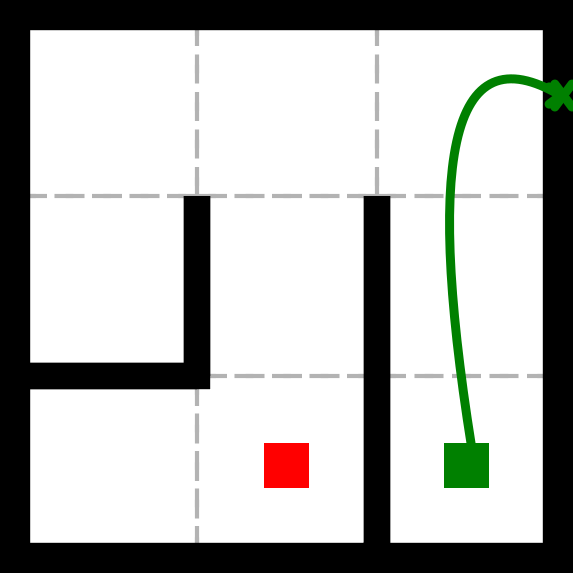}
  \end{minipage}\hfill

  \vspace{0.5em}
  \begin{tabular}{@{}c c : c c@{}}

    \multicolumn{2}{c}{\parbox[c]{0.49\columnwidth}{\centering\scriptsize
      Input Path: \textit{\texttt{\textcolor{Green}{Up, Right, Up, Right, Down}}\\ \geminilogo: ``... the path is \textbf{valid}.'' \correctemoji
    }}} &
        
    \multicolumn{2}{c}{\parbox[c]{0.49\columnwidth}{\centering\scriptsize
      Input Path: \textit{\texttt{\textcolor{Green}{Up, Up,} \textcolor{red}{\underline{Right}}, \textcolor{Green}{Down, Down}}\\\geminilogo: ``... the path is \textbf{invalid}.'' \correctemoji
    }}} \\

  \end{tabular}

  \caption{Models are presented with a blank maze such as (a) or (c) and are asked to verify whether a proposed path from the green square to the red square is feasible through annotations (b) and (d). \SketchVLMs correctly verify both valid and invalid paths by drawing the trajectory and marking where an invalid move occurs.}
  \label{fig:maze_valid_invalid_1row}
\end{figure}

\noindent \maze evaluates spatial reasoning by requiring models to follow a sequence of directions to reach a goal while avoiding obstacles. Given that \vilasr and \thinkmorph are trained on a similar maze task, we expect them to perform well.

\subsec{\textbf{Experiment}}
Given a set of directions, the model must sketch the path while also determining if the path reaches the goal without crossing any walls.

\subsec{\textbf{Results}} Surprisingly, we find that \vilasr has an accuracy of 50.8\% and \thinkmorph has an accuracy of 62.5\%, both near the random choice baseline of 50\%. \nanobananapro performs much stronger at 93.3\%; however, it sometimes alters the entire image and has other unusual outputs (\cref{fig:maze_valid_qual_ex}). Both \geminiplussketch~and \gptplussketch~perform well with accuracies of 98.0\% and 92.8\% respectively. The difference in performance between the default model baseline and the \SketchVLM mode is minor (\cref{tab:full_results}), but the annotations benefit user verification.

\begin{figure*}[t]
  \centering
  \includegraphics[width=\linewidth]{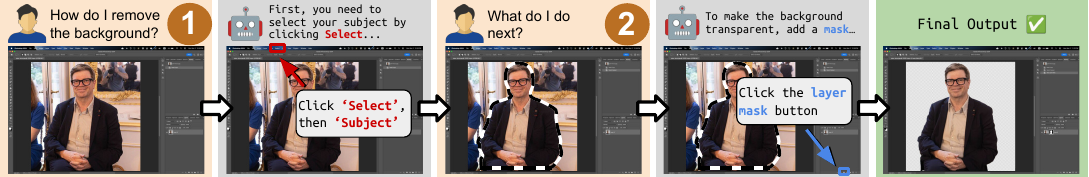}
  \caption{Multi-turn example of \SketchVLM guiding a user through how to remove an image's background. At each turn, the model receives a screenshot then annotates the screenshot with labeled arrows and highlights UI elements to indicate the next step.
  }
  \label{fig:multiturn_ex}
\end{figure*}

\subsection{Fine-tuned sketching models fail to generalize to unseen physics understanding tasks \balllogo}
\label{sec:results_ball_vpct}

\noindent In addition to spatial reasoning, we evaluate whether \SketchVLM can predict trajectories involving physical dynamics, such as a ball falling and rolling.

\subsec{\textbf{Experiment}} 
Given an image with a ball and platforms, the model must sketch the ball's trajectory and output the container number it lands in.

\subsec{\textbf{Results}} 
We find that \vilasr and \thinkmorph perform poorly on \vpct with accuracies of 37\% and 27\%, which are near the random choice baseline of 33.3\%. On our \balldrop dataset, they have accuracies of 35.9\% and 30.3\%, only slightly above the random choice baseline of 25.0\%, showing how these models fail to generalize to tasks that they are not trained on. \nanobananapro often removes ledges in the image (\cref{fig:vpct_qual_ex}) and performs significantly worse than the baseline \geminiproThree accuracy (\cref{tab:full_results}).
In contrast, \SketchVLMs output coherent annotations and have high accuracy with \geminiplussketch~ reaching 96.0\% accuracy on VPCT and 79.7\% on \balldrop (\cref{fig:ball_drop_examples}).

\subsection{\SketchVLMs produce more faithful and higher-quality annotations than image-editing and fine-tuned models}
\label{sec:results_sketch_text_align}

\begin{table}[t]
\centering
\caption{VLM-judged annotation--text alignment (higher is better) and annotation quality (1--5, higher is better). Models fine-tuned to generate annotations show lower alignment, while \geminisketch has the highest alignment and annotation quality.}
\label{tab:vlm_judge_alignment_quality}

\setlength{\tabcolsep}{4.5pt}
\renewcommand{\arraystretch}{1.05}

\adjustbox{max width=\columnwidth}{%
\begin{tabular}{l | c c c c | c c c c}
\toprule
\textbf{Model}
& \multicolumn{4}{c|}{Annotation--Text Alignment}
& \multicolumn{4}{c}{Annotation Quality} \\
\cmidrule(lr){2-5} \cmidrule(lr){6-9}

& \vpct
& \balllogo
& \mazelogo
& Mean
& \vpct
& \balllogo
& \mazelogo
& Mean \\

\midrule
\gptsketch
& 99.0 & \textbf{99.0} & \textbf{88.4} & \textbf{95.5}
& 1.83 & 1.74 & 3.20 & 2.26 \\

\geminisketch
& \textbf{100.0} & 98.5 & 84.2 & 94.2
& \textbf{3.12} & \textbf{4.28} & \textbf{3.69} & \textbf{3.70} \\

\midrule
\nanogeminilogo Nano Banana
& 93.0 & 41.4 & 80.3 & 71.6
& 1.56 & 2.56 & 3.68 & 2.60 \\

\vilasrlogo \vilasr
& 32.0 & 45.5 & 8.2 & 28.6
& 1.36 & 1.28 & 2.78 & 1.81 \\

\hspace{1.6pt}\thinkmorphlogo \thinkmorph
& 54.0 & 38.4 & 48.1 & 46.8
& 1.62 & 2.11 & 1.17 & 1.63 \\

\bottomrule
\end{tabular}%
}
\end{table}

\noindent Annotations should be visually clear, logically coherent, and aligned with the text answer. A trajectory that clips through walls, a label that floats in empty space, or uninformative annotations all undermine the user's ability to interpret the model's reasoning (\cref{fig:annotation_quality_examples}).

\subsec{\textbf{Experiment}} To measure annotation--text alignment, we show each model's annotations for \vpct, \balldrop and \maze (without its text answer) to a VLM judge (Gemini-3-Flash-Preview \cite{doshi2025gemini3flash}) and ask it to infer what the final answer should be from the annotations alone. Then to measure annotation quality, we give a VLM judge a 1--5 rubric (\cref{vlm-judge-details}) in which the judge is asked to evaluate annotation plausibility and visual clarity. We validate judge ratings in \cref{vlm-judge-details}. 

\subsec{\textbf{Results}} Despite being training-free, \SketchVLMs achieve substantially higher annotation--text alignment than all baselines (\cref{tab:vlm_judge_alignment_quality}). \gptplussketch~and \geminiplussketch reach mean alignment scores of 95.5\% and 94.2\%, respectively, compared to 71.6\% for \nanobananapro, 46.8\% for \thinkmorph, and 28.6\% for \vilasr. Based on the VLM judge's ratings,  
\geminiplussketch~achieves the highest mean annotation quality score of 3.70, which is 1.10 points higher than the next-best method, \nanobananapro, and 1.44 points higher than \gptplussketch~(\cref{tab:vlm_judge_alignment_quality}). 
Similarly, based on human ratings, \geminiplussketch~achieves the highest mean quality score of 4.14, followed by \gptplussketch~at 3.70, \nanobananapro at 3.08, \vilasr at 1.74, and \thinkmorph at 1.24 (\cref{tab:human_score_by_task}).
\SketchVLMs' high annotation--text alignment and quality indicate that users can reliably look at the annotations to verify whether the model's reasoning makes sense.

\subsection{\SketchVLM generalizes to open-weight VLMs}
\label{sec:results_additional_models}

\subsec{\textbf{Experiment}}
To evaluate whether \SketchVLM generalizes beyond our primary \gptfive and \geminiproThree backbones, we additionally test \gemmafour \cite{gemmateam2026gemma4}, \qwenthreefive \cite{qwenteam2026qwen35}, \kimitwofive \cite{kimiteam2026kimik25visualagentic}, and \kimithree \cite{kimiteam2026kimik3}. We evaluate Direct VQA and \SketchVLM on the six tasks with directly comparable answer outputs. The complete results and input ablations are provided in \cref{tab:seven_dataset_comparison,tab:ablation_other_models}.

\subsec{\textbf{Results}}
\SketchVLM successfully elicits structured visual annotations from all four additional VLMs, demonstrating that the framework transfers without model-specific training. \kimithree is the strongest additional open-weight backbone with results competitive with \geminiproThree.


Results are mixed for the other open-weight models. \gemmafour significantly improves its accuracy in some tasks (+8 points in \vpct and +18.8 points in \counting) while \kimitwofive and \qwenthreefive do not improve overall accuracy with \SketchVLM. However, all models are successfully able to generate interpretable visual annotations without task-specific training (\cref{sec:qual-samples}).

\subsection{Single-turn is as accurate as multi-turn but requires significantly fewer turns}
\label{sec:single_vs_multi_turns}

\noindent We want \SketchVLMs to visually guide users through tasks with multiple turns, such as removing the background of a photo (\cref{fig:multiturn_ex}) or setting up an EC2 instance on AWS (\cref{fig:aws_demo}). We compare accuracy between single-turn and multi-turn, and identify the best configuration for multi-turn.

\subsec{\textbf{Experiment}} We compare single-turn and multi-turn generation on \vpct, \balllogo, \mazelogo, and \connectlogo. In multi-turn, the model receives all previously drawn strokes rendered onto the input image at each turn. We also test the necessity of providing the text representation of rendered strokes across turns.

\subsec{\textbf{Results}} Single-turn achieves comparable or higher accuracy than multi-turn across all tasks (\cref{tab:ablation_models}), while requiring about 5.92x fewer turns (\cref{tab:avg_turns}). This shows that \SketchVLMs can produce high-quality annotations and accurate answers in a single pass. When we remove the text representation of prior strokes to force the model to rely solely on the rendered image, we observe notable degradation in annotation quality (\cref{fig:gemini3pro_multiturn_text_vs_no_text}). We therefore report all multi-turn results with both the rendered image and the text history provided to the model.

\section{Conclusion}

We present \SketchVLM, a training-free framework that prompts frontier VLMs to produce editable, non-destructive SVG annotations grounded on the input image. These visual explanations let users verify model reasoning at a glance, outperform sketching models by +28.5 percentage points in accuracy (\cref{tab:avg_delta_accuracy}) and +48.3\% in annotation quality (\cref{tab:avg_quality_delta}), while also generalizing well to real-world tasks (\cref{sec:real_world_applications}).

\section*{Limitations}
\SketchVLM works best with \gptfive and \geminiproThree and can transfer to strong open-weight VLMs like \kimitwofive \cite{kimiteam2026kimik25visualagentic}, \kimithree, \gemmafour, and \qwenthreefive (\cref{tab:ablation_other_models,sec:qual-samples}), but does not perform well on small VLMs that struggle with instruction following like Qwen2.5-VL-7B \cite{bai2025qwen25vltechnicalreport}. Additionally, enabling models to undo and edit strokes could be explored as a way to improve multi-turn performance.

Users may over-trust incorrect sketches, so \SketchVLM should be used as a verification aid rather than a guarantee of correctness, and annotations should be checked carefully when used for safety-critical, medical, legal, or financial tasks.

\section*{Author contribution statement}
\label{sec:author_contribution}

Brandon Collins (BC), Logan Bolton (LB), and Hung Huy Nguyen (HN) made
major contributions to benchmark creation and experimental evaluation.
\begin{itemize}
    \item HN curated the Counting, Drawing Shapes, and Part Labeling
    benchmarks from existing datasets, led their evaluation, and led the
    human-versus-VLM agreement study.
    
    \item BC created the Connect-the-Dots benchmark and led method
    development, development of the inference and evaluation code,
    ablations, and evaluation on Connect-the-Dots and VPCT.
    
    \item LB created the Ball Drop and Maze Navigation benchmarks, led
    their evaluation, and ran the open-source model experiments as well
    as the annotation-quality and annotation--text alignment evaluations.
\end{itemize}

BC and LB led the writing of the manuscript, while all authors contributed
to editing and reviewing it. BC, LB, and Anh Totti Nguyen (AN) developed
the interactive demo. BC led development of the project website, with
additional contributions from LB. AN supervised the project.

\section*{Acknowledgments}

We thank Pooyan Rahmanzadehgervi for feedback. AN was supported by NSF Grant No.~2145767 and
donations from the NaphCare Foundation and Adobe Research. LB and HN were supported
by the AU URF and PGRF programs, respectively.



%

\bibliography{main}

@article{hu2024visual,
  title={Visual sketchpad: Sketching as a visual chain of thought for multimodal language models},
  author={Hu, Yushi and Shi, Weijia and Fu, Xingyu and Roth, Dan and Ostendorf, Mari and Zettlemoyer, Luke and Smith, Noah A and Krishna, Ranjay},
  journal={Advances in Neural Information Processing Systems},
  volume={37},
  pages={139348--139379},
  year={2024}
}

@article{zhou2025improving,
  title={Improving human verification of llm reasoning through interactive explanation interfaces},
  author={Zhou, Runtao and Nguyen, Giang and Kharya, Nikita and Nguyen, Anh Totti and Agarwal, Chirag},
  journal={arXiv preprint arXiv:2510.22922},
  year={2025}
}

@article{team2024chameleon,
  title={Chameleon: Mixed-modal early-fusion foundation models},
  author={Team, Chameleon},
  journal={arXiv preprint arXiv:2405.09818},
  year={2024}
}

@article{team2023gemini,
  title={Gemini: a family of highly capable multimodal models},
  author={Team, Gemini and Anil, Rohan and Borgeaud, Sebastian and Alayrac, Jean-Baptiste and Yu, Jiahui and Soricut, Radu and Schalkwyk, Johan and Dai, Andrew M and Hauth, Anja and Millican, Katie and others},
  journal={arXiv preprint arXiv:2312.11805},
  year={2023}
}

@article{nguyen2025hot,
  title={Hot: Highlighted chain of thought for referencing supporting facts from inputs},
  author={Nguyen, Tin and Bolton, Logan and Taesiri, Mohammad Reza and Bui, Trung and Nguyen, Anh Totti},
  journal={arXiv preprint arXiv:2503.02003},
  year={2025}
}

@misc{aibrowsers,
author = {By Gyana Swain},
  title = {Keep AI browsers out of your enterprise, warns Gartner – Computerworld},
  url = "https://www.computerworld.com/article/4102569/keep-ai-browsers-out-of-your-enterprise-warns-gartner.html",
month = {},
year = {},
  note = "[Online; accessed 2026-01-28]"
}

@inproceedings{vinker2025sketchagent,
  title={Sketchagent: Language-driven sequential sketch generation},
  author={Vinker, Yael and Shaham, Tamar Rott and Zheng, Kristine and Zhao, Alex and E Fan, Judith and Torralba, Antonio},
  booktitle={Proceedings of the Computer Vision and Pattern Recognition Conference},
  pages={23355--23368},
  year={2025}
}

@misc{cbrowerVPCT2025,
  title = {VPCT Ball Drop Benchmark},
  author = {cbrower},
  year = {2025},
  howpublished = {\url{https://cbrower.dev/vpct}},
  note = {Accessed: 2025-11-09}
}

@inproceedings{bakhtin2019phyre,
  title     = {{PHYRE}: A New Benchmark for Physical Reasoning},
  author    = {Bakhtin, Anton and van der Maaten, Laurens and Johnson, Justin
               and Gustafson, Laura and Girshick, Ross},
  booktitle = {Advances in Neural Information Processing Systems},
  volume    = {32},
  year      = {2019},
  url       = {https://arxiv.org/abs/1908.05656}
}

@misc{openclipart,
  author       = {{Openclipart Contributors}},
  title        = {Openclipart Silhouette Collection},
  howpublished = {\url{https://openclipart.org/search/?query=silhouette}},
  note         = {Accessed: 2025-11-10},
  year         = {2025}
}

@article{douglas1973algorithms,
  title        = {Algorithms for the reduction of the number of points required to represent a digitized line or its caricature},
  author       = {Douglas, David H. and Peucker, Thomas K.},
  journal      = {The Canadian Cartographer},
  volume       = {10},
  number       = {2},
  pages        = {112--122},
  year         = {1973},
  publisher    = {Taylor \& Francis}
}

@misc{ribeiro2020sketchformertransformerbasedrepresentationsketched,
      title={Sketchformer: Transformer-based Representation for Sketched Structure}, 
      author={Leo Sampaio Ferraz Ribeiro and Tu Bui and John Collomosse and Moacir Ponti},
      year={2020},
      eprint={2002.10381},
      archivePrefix={arXiv},
      primaryClass={cs.CV},
      url={https://arxiv.org/abs/2002.10381}, 
}

@misc{wu2025reinforcingspatialreasoningvisionlanguage,
      title={Reinforcing Spatial Reasoning in Vision-Language Models with Interwoven Thinking and Visual Drawing}, 
      author={Junfei Wu and Jian Guan and Kaituo Feng and Qiang Liu and Shu Wu and Liang Wang and Wei Wu and Tieniu Tan},
      year={2025},
      eprint={2506.09965},
      archivePrefix={arXiv},
      primaryClass={cs.CV},
      url={https://arxiv.org/abs/2506.09965}, 
}

@article{
wang2025visually,
title={Visually Descriptive Language Model for Vector Graphics Reasoning},
author={Zhenhailong Wang and Joy Hsu and Xingyao Wang and Kuan-Hao Huang and Manling Li and Jiajun Wu and Heng Ji},
journal={Transactions on Machine Learning Research},
issn={2835-8856},
year={2025},
url={https://openreview.net/forum?id=WzS33L1iPC},
note={}
}

@article{beyer2024paligemma,
  title={Paligemma: A versatile 3b vlm for transfer},
  author={Beyer, Lucas and Steiner, Andreas and Pinto, Andr{\'e} Susano and Kolesnikov, Alexander and Wang, Xiao and Salz, Daniel and Neumann, Maxim and Alabdulmohsin, Ibrahim and Tschannen, Michael and Bugliarello, Emanuele and others},
  journal={arXiv preprint arXiv:2407.07726},
  year={2024}
}

@inproceedings{acharya2019tallyqa,
  title={Tallyqa: Answering complex counting questions},
  author={Acharya, Manoj and Kafle, Kushal and Kanan, Christopher},
  booktitle={Proceedings of the AAAI conference on artificial intelligence},
  volume={33},
  number={01},
  pages={8076--8084},
  year={2019}
}

@inproceedings{deitke2025molmo,
  title={Molmo and pixmo: Open weights and open data for state-of-the-art vision-language models},
  author={Deitke, Matt and Clark, Christopher and Lee, Sangho and Tripathi, Rohun and Yang, Yue and Park, Jae Sung and Salehi, Mohammadreza and Muennighoff, Niklas and Lo, Kyle and Soldaini, Luca and others},
  booktitle={Proceedings of the Computer Vision and Pattern Recognition Conference},
  pages={91--104},
  year={2025}
}

@inproceedings{ramanathan2023paco,
  title={Paco: Parts and attributes of common objects},
  author={Ramanathan, Vignesh and Kalia, Anmol and Petrovic, Vladan and Wen, Yi and Zheng, Baixue and Guo, Baishan and Wang, Rui and Marquez, Aaron and Kovvuri, Rama and Kadian, Abhishek and others},
  booktitle={Proceedings of the IEEE/CVF Conference on Computer Vision and Pattern Recognition},
  pages={7141--7151},
  year={2023}
}

@inproceedings{chen2014detect,
  title={Detect what you can: Detecting and representing objects using holistic models and body parts},
  author={Chen, Xianjie and Mottaghi, Roozbeh and Liu, Xiaobai and Fidler, Sanja and Urtasun, Raquel and Yuille, Alan},
  booktitle={Proceedings of the IEEE conference on computer vision and pattern recognition},
  pages={1971--1978},
  year={2014}
}

@article{zhang2025latentsketchpad,
          title={Latent Sketchpad: Sketching Visual Thoughts to Elicit Multimodal Reasoning in MLLMs},
          author={Zhang, Huanyu and Wu, Wenshan and Li, Chengzu and Shang, Ning and Xia, Yan and Huang, Yangyu and Zhang, Yifan and Dong, Li and Zhang, Zhang and Wang, Liang and Tan, Tieniu and Wei, Furu},
          journal={arXiv preprint arXiv:2510.24514},
          year={2025}
        }

@misc{zhang2025deepsketcherinternalizingvisualmanipulation,
      title={DeepSketcher: Internalizing Visual Manipulation for Multimodal Reasoning}, 
      author={Chi Zhang and Haibo Qiu and Qiming Zhang and Zhixiong Zeng and Lin Ma and Jing Zhang},
      year={2025},
      eprint={2509.25866},
      archivePrefix={arXiv},
      primaryClass={cs.CV},
      url={https://arxiv.org/abs/2509.25866}, 
}

@misc{li2025imaginereasoningspacemultimodal,
      title={Imagine while Reasoning in Space: Multimodal Visualization-of-Thought}, 
      author={Chengzu Li and Wenshan Wu and Huanyu Zhang and Yan Xia and Shaoguang Mao and Li Dong and Ivan Vulić and Furu Wei},
      year={2025},
      eprint={2501.07542},
      archivePrefix={arXiv},
      primaryClass={cs.CL},
      url={https://arxiv.org/abs/2501.07542}, 
}

@article{gu2025thinkmorph,
  title={ThinkMorph: Emergent Properties in Multimodal Interleaved Chain-of-Thought Reasoning},
  author={Gu, Jiawei and Hao, Yunzhuo and Wang, Huichen Will and Li, Linjie and Shieh, Michael Qizhe and Choi, Yejin and Krishna, Ranjay and Cheng, Yu},
  journal={arXiv preprint arXiv:2510.27492},
  year={2025}
}

@article{su2025openthinkimg,
  title={OpenThinkIMG: Learning to Think with Images via Visual Tool Reinforcement Learning},
  author={Su, Zhaochen and Li, Linjie and Song, Mingyang and Hao, Yunzhuo and Yang, Zhengyuan and Zhang, Jun and Chen, Guanjie and Gu, Jiawei and Li, Juntao and Qu, Xiaoye and others},
  journal={arXiv preprint arXiv:2505.08617},
  year={2025}
}

@misc{ou2025bridgingdynamicperceptiongap,
      title={Bridging the Dynamic Perception Gap: Training-Free Draft Chain-of-Thought for Dynamic Multimodal Spatial Reasoning}, 
      author={Siqu Ou and Hongcheng Liu and Pingjie Wang and Yusheng Liao and Chuan Xuan and Yanfeng Wang and Yu Wang},
      year={2025},
      eprint={2505.16579},
      archivePrefix={arXiv},
      primaryClass={cs.AI},
      url={https://arxiv.org/abs/2505.16579}, 
}

@article{menon2024whiteboard,
    title={Whiteboard-of-Thought: Thinking Step-by-Step Across Modalities},
    author={Sachit Menon and Richard Zemel and Carl Vondrick},
    journal={arXiv},
    year={2024}
}

@techreport{generative_ai_outlook,
  title        = {Generative AI Outlook},
  author       = {{Bloomberg Intelligence}},
  year         = {2024},
  institution  = {Bloomberg},
  address      = {New York},
  url          = {https://assets.bbhub.io/professional/sites/41/Generative-AI-Outlook.pdf},
  note         = {Accessed: 2026-01-18}
}

@misc{vikhyat_moondream_github,
  title        = {Moondream: Tiny Vision Language Model},
  author       = {Vikhyat},
  howpublished = {\url{https://github.com/vikhyat/moondream}},
  year         = {2023},
  note         = {Open-source vision-language model with small-footprint multimodal capabilities}
}

@misc{allenai_molmo_2024,
  title        = {Molmo: An Open Vision-Language Model from Allen AI},
  author       = {{Allen Institute for AI}},
  year         = {2024},
  howpublished = {\url{https://github.com/allenai/molmo}},
  note         = {Open-source multimodal model family for vision-language tasks; accessed 2026-01-18}
}

@article{taesiri2025understanding_genai_image_editing,
  title   = {Understanding Generative AI Capabilities in Everyday Image Editing Tasks},
  author  = {Taesiri, Mohammad Reza and Collins, Brandon and Bolton, Logan and Lai, Viet Dac
             and Dernoncourt, Franck and Bui, Trung and Nguyen, Anh Totti},
  journal = {arXiv preprint arXiv:2505.16181},
  year    = {2025},
  url     = {https://arxiv.org/abs/2505.16181},
  doi     = {10.48550/arXiv.2505.16181},
  note    = {Version 2, submitted 26 May 2025}
}

@misc{wu2023vstar,
  title={V*: Guided Visual Search as a Core Mechanism in Multimodal LLMs},
  author={Penghao Wu and Saining Xie},
  year={2023},
  eprint={2312.14135},
  archivePrefix={arXiv},
  primaryClass={cs.CV},
  doi={10.48550/arXiv.2312.14135},
  url={https://arxiv.org/abs/2312.14135}
}

@misc{chatapps,
author = {Sarah Perez},
  title = {ChatGPT's user growth has slowed, report finds | TechCrunch},
  url = "https://techcrunch.com/2025/12/05/chatgpts-user-growth-has-slowed-report-finds/",
month = {12},
year = {2025},
  note = "[Online; accessed 2026-01-28]"
}

@inproceedings{
  zhang2025mllms,
  title={{MLLM}s Know Where to Look: Training-free Perception of Small Visual Details with Multimodal {LLM}s},
  author={Jiarui Zhang and Mahyar Khayatkhoei and Prateek Chhikara and Filip Ilievski},
  booktitle={The Thirteenth International Conference on Learning Representations},
  year={2025},
  url={https://arxiv.org/abs/2502.17422}
}

@article{izadi2025visual,
  title         = {Visual Structures Helps Visual Reasoning: Addressing the Binding Problem in VLMs},
  author        = {Izadi, Amirmohammad and Banayeeanzade, Mohammad Ali and Askari, Fatemeh and Rahimiakbar, Ali and Vahedi, Mohammad Mahdi and Hasani, Hosein and Soleymani Baghshah, Mahdieh},
  journal       = {arXiv preprint arXiv:2506.22146},
  year          = {2025},
  archivePrefix = {arXiv},
  eprint        = {2506.22146},
  primaryClass  = {cs.CV},
  doi           = {10.48550/arXiv.2506.22146}
}

@misc{doshi2025gemini3flash,
  title        = {Gemini 3 Flash: frontier intelligence built for speed},
  author       = {Google DeepMind},
  year         = {2025},
  month        = dec,
  day          = {17},
  howpublished = {The Keyword (Google Blog)},
  url          = {https://blog.google/products-and-platforms/products/gemini/gemini-3-flash/},
  urldate      = {2026-01-25}
}

@misc{singh2025openaigpt5card,
      title={OpenAI GPT-5 System Card}, 
      author={OpenAI},
      year={2025},
      eprint={2601.03267},
      archivePrefix={arXiv},
      primaryClass={cs.CL},
      url={https://arxiv.org/abs/2601.03267}, 
}

@misc{pichai2025gemini3,
  title        = {A new era of intelligence with Gemini 3},
  author       = {Sundar Pichai and Demis Hassabis and Koray Kavukcuoglu},
  year         = {2025},
  month        = nov,
  day          = {18},
  howpublished = {The Keyword (Google Blog)},
  url          = {https://blog.google/products-and-platforms/products/gemini/gemini-3/},
  urldate      = {2026-04-11}
}

@InProceedings{pmlr-v235-chen24h,
  title = 	 {{MLLM}-as-a-Judge: Assessing Multimodal {LLM}-as-a-Judge with Vision-Language Benchmark},
  author =       {Chen, Dongping and Chen, Ruoxi and Zhang, Shilin and Wang, Yaochen and Liu, Yinuo and Zhou, Huichi and Zhang, Qihui and Wan, Yao and Zhou, Pan and Sun, Lichao},
  booktitle = 	 {Proceedings of the 41st International Conference on Machine Learning},
  pages = 	 {6562--6595},
  year = 	 {2024},
  editor = 	 {Salakhutdinov, Ruslan and Kolter, Zico and Heller, Katherine and Weller, Adrian and Oliver, Nuria and Scarlett, Jonathan and Berkenkamp, Felix},
  volume = 	 {235},
  series = 	 {Proceedings of Machine Learning Research},
  month = 	 {21--27 Jul},
  publisher =    {PMLR},
  url = 	 {https://proceedings.mlr.press/v235/chen24h.html}
}

@misc{Google025NanoBananaPro,
  author = {Google DeepMind},
  title  = {Introducing Nano Banana Pro},
  year   = {2025},
  month  = nov,
  day    = {20},
  url    = {https://blog.google/innovation-and-ai/products/nano-banana-pro/},
  note   = {. Accessed: 2026-01-25}
}

@inproceedings{lei-etal-2025-scaffolding,
    title = "Scaffolding Coordinates to Promote Vision-Language Coordination in Large Multi-Modal Models",
    author = "Lei, Xuanyu  and
      Yang, Zonghan  and
      Chen, Xinrui  and
      Li, Peng  and
      Liu, Yang",
    editor = "Rambow, Owen  and
      Wanner, Leo  and
      Apidianaki, Marianna  and
      Al-Khalifa, Hend  and
      Eugenio, Barbara Di  and
      Schockaert, Steven",
    booktitle = "Proceedings of the 31st International Conference on Computational Linguistics",
    month = jan,
    year = "2025",
    address = "Abu Dhabi, UAE",
    publisher = "Association for Computational Linguistics",
    url = "https://aclanthology.org/2025.coling-main.195/",
    pages = "2886--2903"
}

@article{zou2025unimmmumassivemultidisciplinemultimodal,
      title={{Uni-MMMU}: A Massive Multi-discipline Multimodal Unified Benchmark},
      author = {Kai Zou and Ziqi Huang and Yuhao Dong and Shulin Tian and Dian Zheng and Hongbo Liu and Jingwen He and Bin Liu and Yu Qiao and Ziwei Liu},
      journal={arXiv preprint arXiv:2510.13759},
      year = {2025}
}

@article{latif2025sketchmind,
  title={SketchMind: A Multi-Agent Cognitive Framework for Assessing Student-Drawn Scientific Sketches},
  author={Latif, Ehsan and Khan, Zirak and Zhai, Xiaoming},
  journal={arXiv preprint arXiv:2507.22904},
  year={2025}
}

@article{li2026gebench,
  title={GEBench: Benchmarking Image Generation Models as GUI Environments},
  author={Li, Haodong and Wu, Jingwei and Sun, Quan and Li, Guopeng and Tian, Juanxi and Zhang, Huanyu and Lai, Yanlin and An, Ruichuan and Peng, Hongbo and Dai, Yuhong and others},
  journal={arXiv preprint arXiv:2602.09007},
  year={2026}
}

@misc{zhao2025pyvision,
  title         = {PyVision: Agentic Vision with Dynamic Tooling},
  author        = {Shitian Zhao and Haoquan Zhang and Shaoheng Lin and
                   Ming Li and Qilong Wu and Kaipeng Zhang and Chen Wei},
  year          = {2025},
  eprint        = {2507.07998},
  archivePrefix = {arXiv},
  primaryClass  = {cs.CL},
  url           = {https://arxiv.org/abs/2507.07998}
}

@article{carion2025sam,
  title={Sam 3: Segment anything with concepts},
  author={Carion, Nicolas and Gustafson, Laura and Hu, Yuan-Ting and Debnath, Shoubhik and Hu, Ronghang and Suris, Didac and Ryali, Chaitanya and Alwala, Kalyan Vasudev and Khedr, Haitham and Huang, Andrew and others},
  journal={arXiv preprint arXiv:2511.16719},
  year={2025}
}

@online{caradvertisementopenai,
  author  = {Kiri Masters},
  title   = {Why {OpenAI}'s ad announcement should worry retail media networks},
  year    = {2026},
  month   = jan,
  website = {The Drum},
  url     = {https://www.thedrum.com/opinion/why-openai-s-ad-announcement-should-worry-retail-media-networks},
  urldate = {2026-02-27}
}

@online{openai2026fixwithchatgpt,
  author       = {{OpenAI}},
  title        = {Fix With ChatGPT},
  year         = {2026},
  month        = feb,
  day          = {13},
  organization = {YouTube},
  url          = {https://www.youtube.com/watch?v=PHKpsVIdAcc},
  urldate      = {2026-02-27}
}

@article{deng2025bagel,
  title   = {Emerging Properties in Unified Multimodal Pretraining},
  author  = {Deng, Chaorui and Zhu, Deyao and Li, Kunchang and Gou, Chenhui and Li, Feng and Wang, Zeyu and Zhong, Shu and Yu, Weihao and Nie, Xiaonan and Song, Ziang and Shi, Guang and Fan, Haoqi},
  journal = {arXiv preprint arXiv:2505.14683},
  year    = {2025}
}

@misc{kimiteam2026kimik25visualagentic,
      title={Kimi K2.5: Visual Agentic Intelligence}, 
    author={{Kimi Team}},

      year={2026},
      eprint={2602.02276},
      archivePrefix={arXiv},
      primaryClass={cs.CL},
      url={https://arxiv.org/abs/2602.02276}, 
}

@misc{bai2025qwen25vltechnicalreport,
      title={Qwen2.5-VL Technical Report}, 
      author={Shuai Bai and Keqin Chen and Xuejing Liu and Jialin Wang and Wenbin Ge and Sibo Song and Kai Dang and Peng Wang and Shijie Wang and Jun Tang and Humen Zhong and Yuanzhi Zhu and Mingkun Yang and Zhaohai Li and Jianqiang Wan and Pengfei Wang and Wei Ding and Zheren Fu and Yiheng Xu and Jiabo Ye and Xi Zhang and Tianbao Xie and Zesen Cheng and Hang Zhang and Zhibo Yang and Haiyang Xu and Junyang Lin},
      year={2025},
      eprint={2502.13923},
      archivePrefix={arXiv},
      primaryClass={cs.CV},
      url={https://arxiv.org/abs/2502.13923}, 
}

@InProceedings{Cheng_2021_CVPR,
    author    = {Cheng, Bowen and Girshick, Ross and Dollar, Piotr and Berg, Alexander C. and Kirillov, Alexander},
    title     = {Boundary IoU: Improving Object-Centric Image Segmentation Evaluation},
    booktitle = {Proceedings of the IEEE/CVF Conference on Computer Vision and Pattern Recognition (CVPR)},
    month     = {June},
    year      = {2021},
    pages     = {15334-15342}
}

@article{yu2024visualprompting,
  title={Visual Prompting in Multimodal Large Language Models: A Survey},
  author={Yu, Tong and others},
  journal={arXiv preprint arXiv:2409.15310},
  year={2024},
  doi={10.48550/arXiv.2409.15310},
  url={https://arxiv.org/abs/2409.15310}
}

@misc{kimiteam2026kimik3,
  title         = {Kimi K3: Open Frontier Intelligence},
  author        = {{Kimi Team}},
  year          = {2026},
  eprint        = {2607.24653},
  archivePrefix = {arXiv},
  primaryClass  = {cs.CL},
  url           = {https://arxiv.org/abs/2607.24653}
}

@misc{gemmateam2026gemma4,
  title         = {Gemma 4 Technical Report},
  author        = {{Gemma Team}},
  year          = {2026},
  eprint        = {2607.02770},
  archivePrefix = {arXiv},
  primaryClass  = {cs.CL},
  url           = {https://arxiv.org/abs/2607.02770}
}

@misc{qwenteam2026qwen35,
  title  = {Qwen3.5: Towards Native Multimodal Agents},
  author = {{Qwen Team}},
  year   = {2026},
  url    = {https://qwen.ai/blog?id=qwen3.5},
  note   = {Accessed: 2026-08-27}
}

\clearpage
\appendix

\renewcommand{\thesection}{\Alph{section}}
\counterwithin{figure}{section}
\counterwithin{table}{section}
\renewcommand{\thefigure}{\thesection\arabic{figure}}
\renewcommand{\thetable}{\thesection\arabic{table}}

\makeatletter
\renewcommand{\theHsection}{appendix.\Alph{section}}
\renewcommand{\theHsubsection}{appendix.\Alph{section}.\arabic{subsection}}
\renewcommand{\theHsubsubsection}{appendix.\Alph{section}.\arabic{subsection}.\arabic{subsubsection}}
\makeatother

\ifincludeappendix

\setcounter{tocdepth}{2} 


\tableofcontents

\clearpage

\twocolumn

\section{Real World Applications}
\label{sec:real_world_applications}

\begin{figure}[H]
  \centering
  \begin{adjustbox}{width=0.99\textwidth}
    \setlength{\tabcolsep}{0.5pt}
    \renewcommand{\arraystretch}{0}
    \newcommand{\imgw}{0.247\textwidth}

    \begin{tabular}{@{}c@{\hspace{1pt}}c@{\hspace{1pt}}c@{\hspace{1pt}}c@{}}
      \includegraphics[width=\imgw]{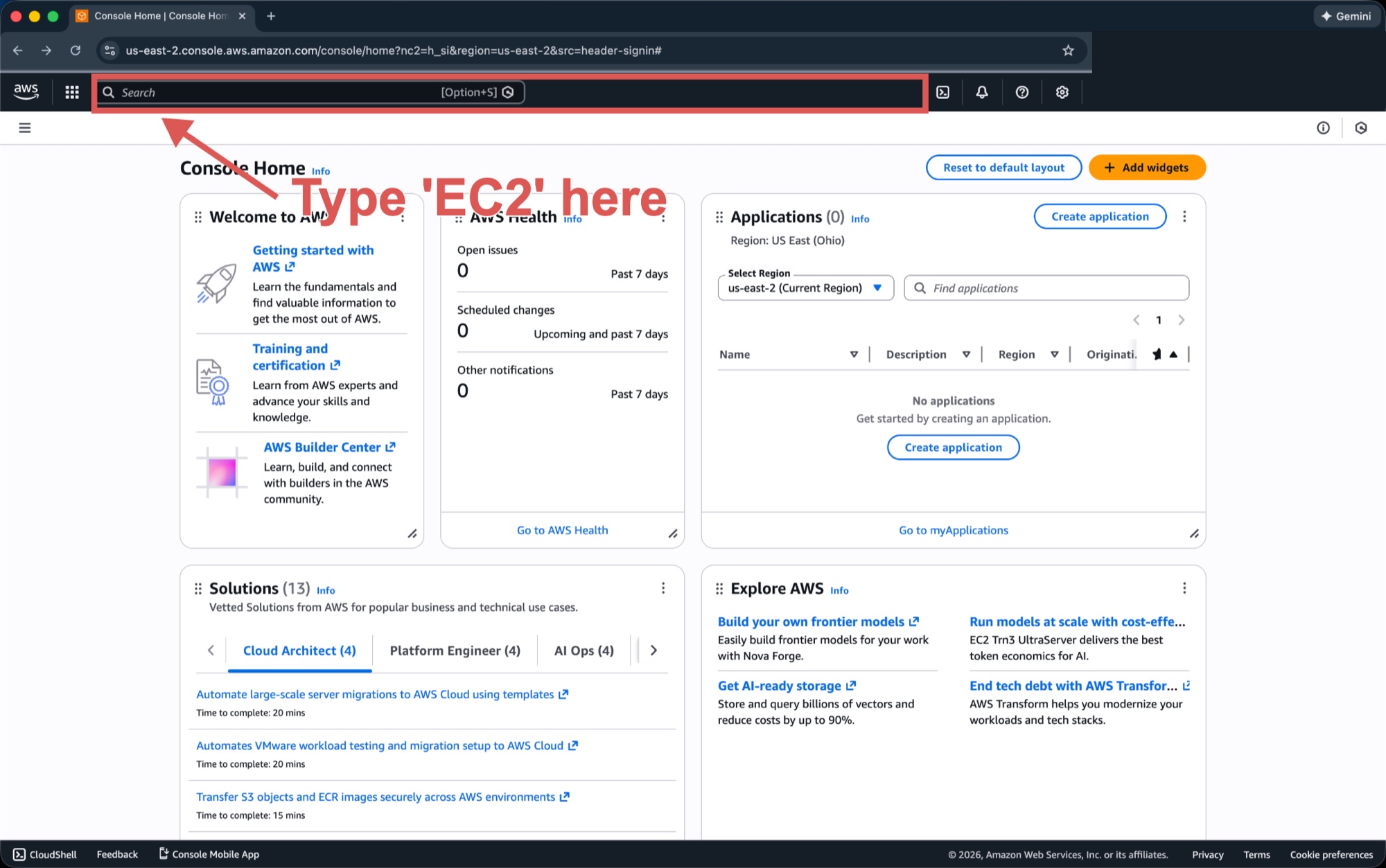} &
      \includegraphics[width=\imgw]{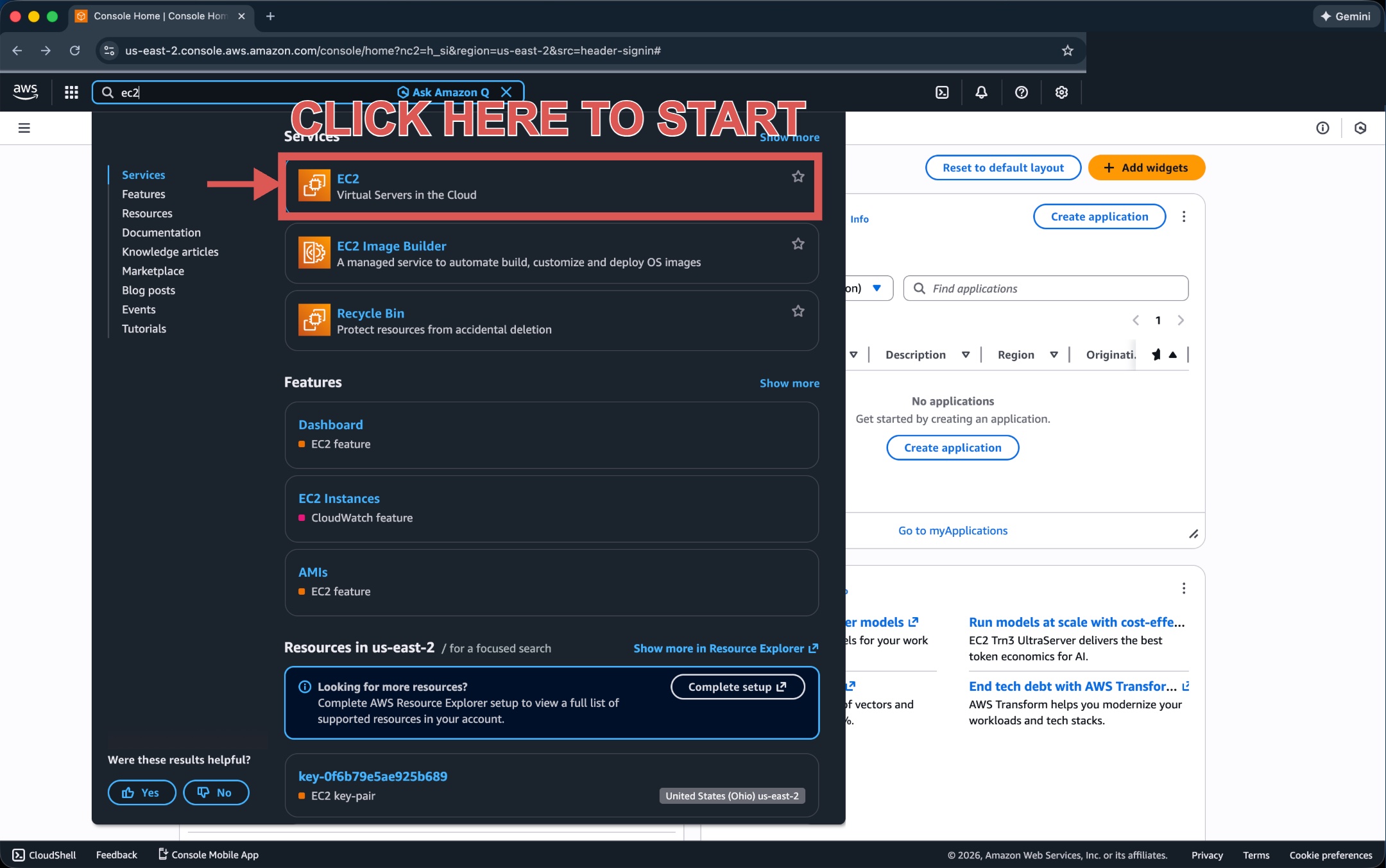} &
      \includegraphics[width=\imgw]{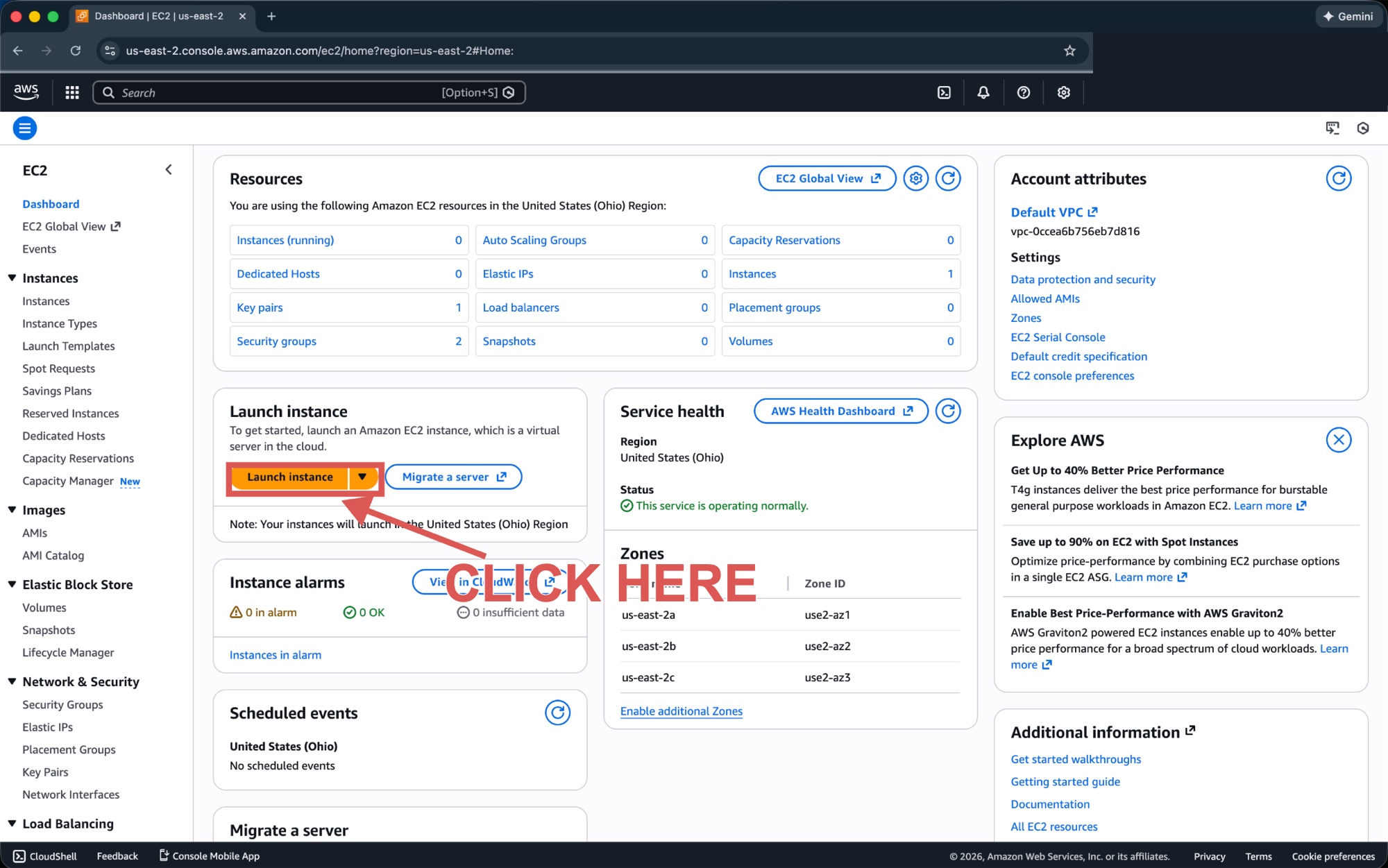} &
      \includegraphics[width=\imgw]{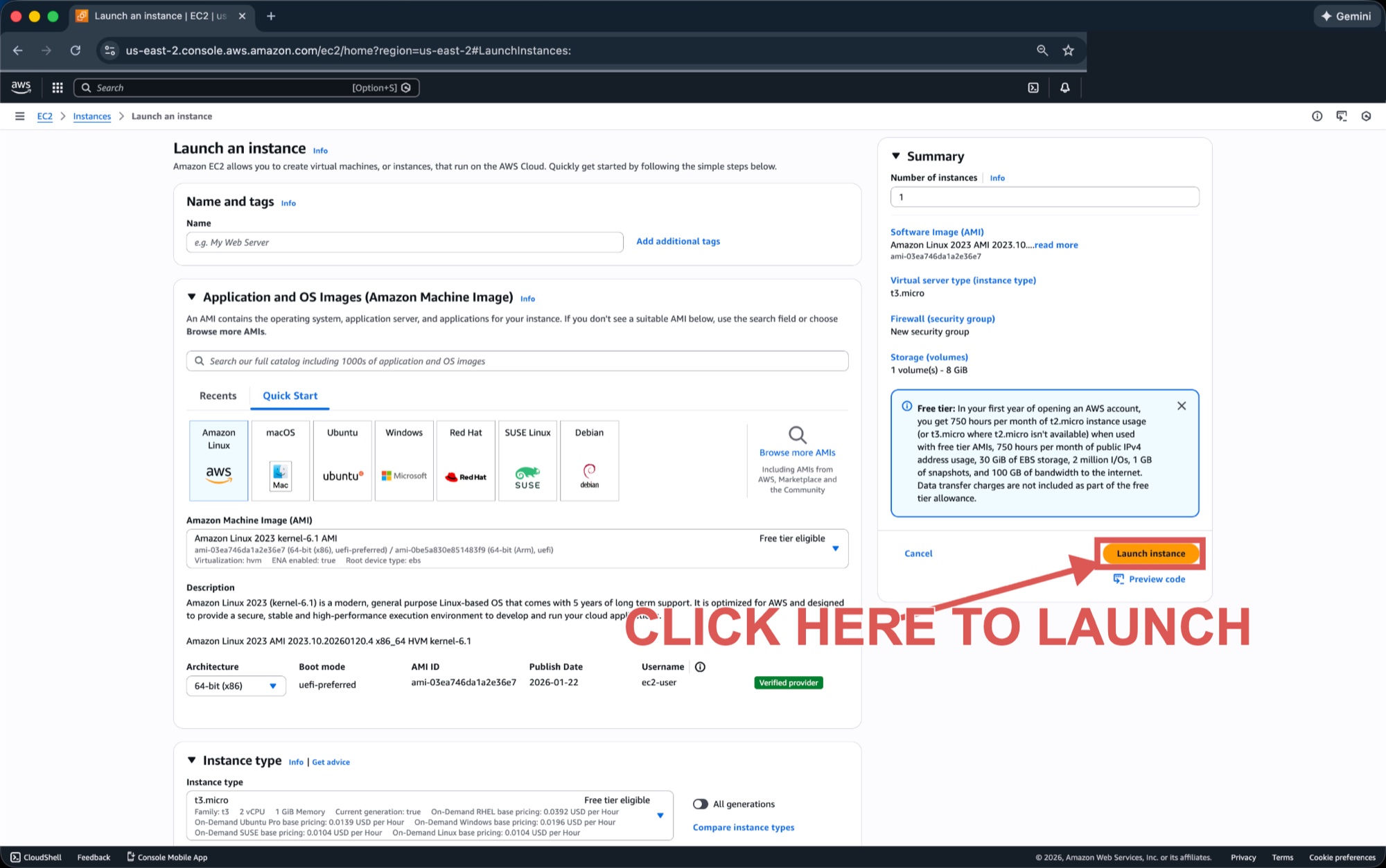} \\[-1pt]
      \includegraphics[width=\imgw]{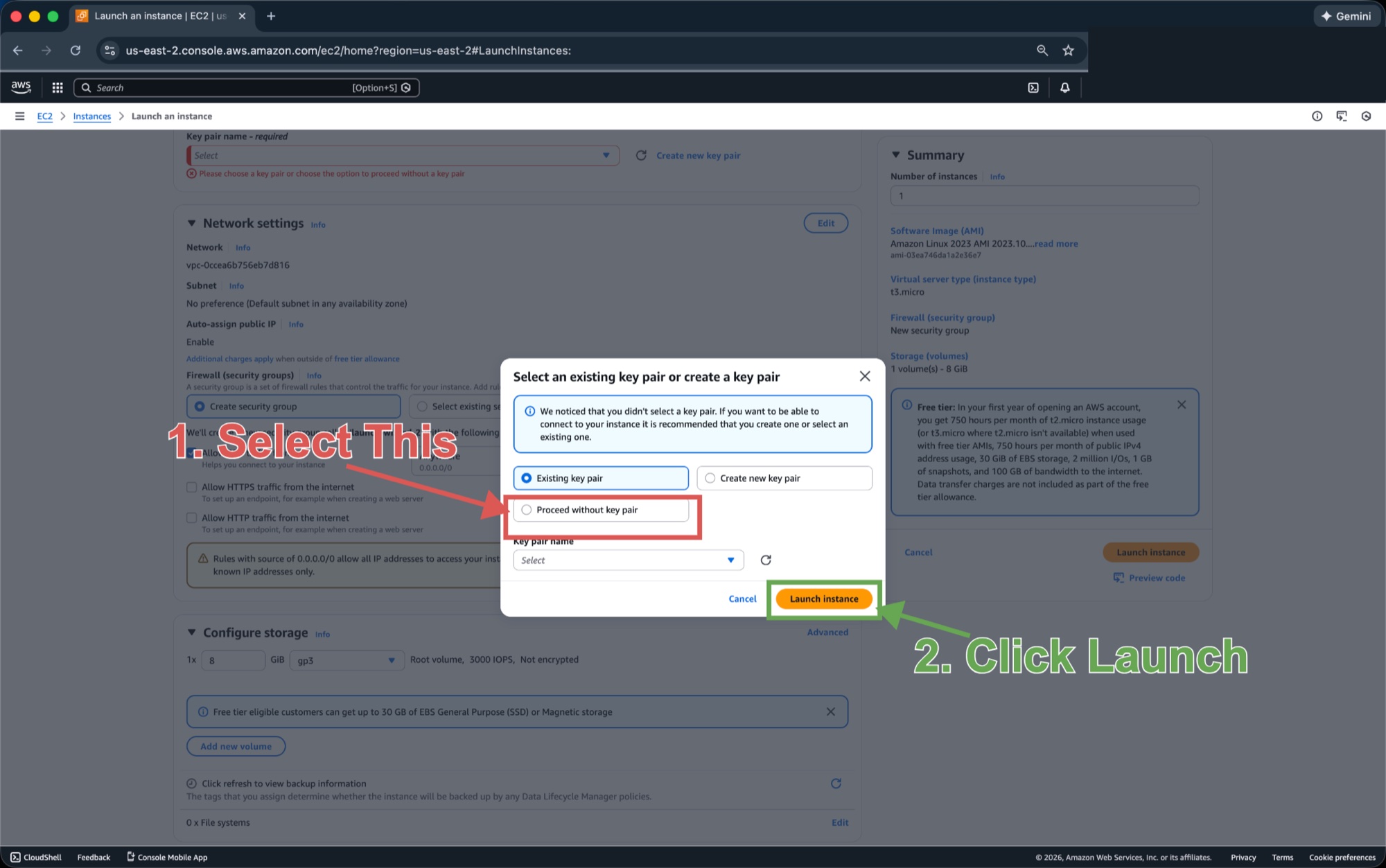} &
      \includegraphics[width=\imgw]{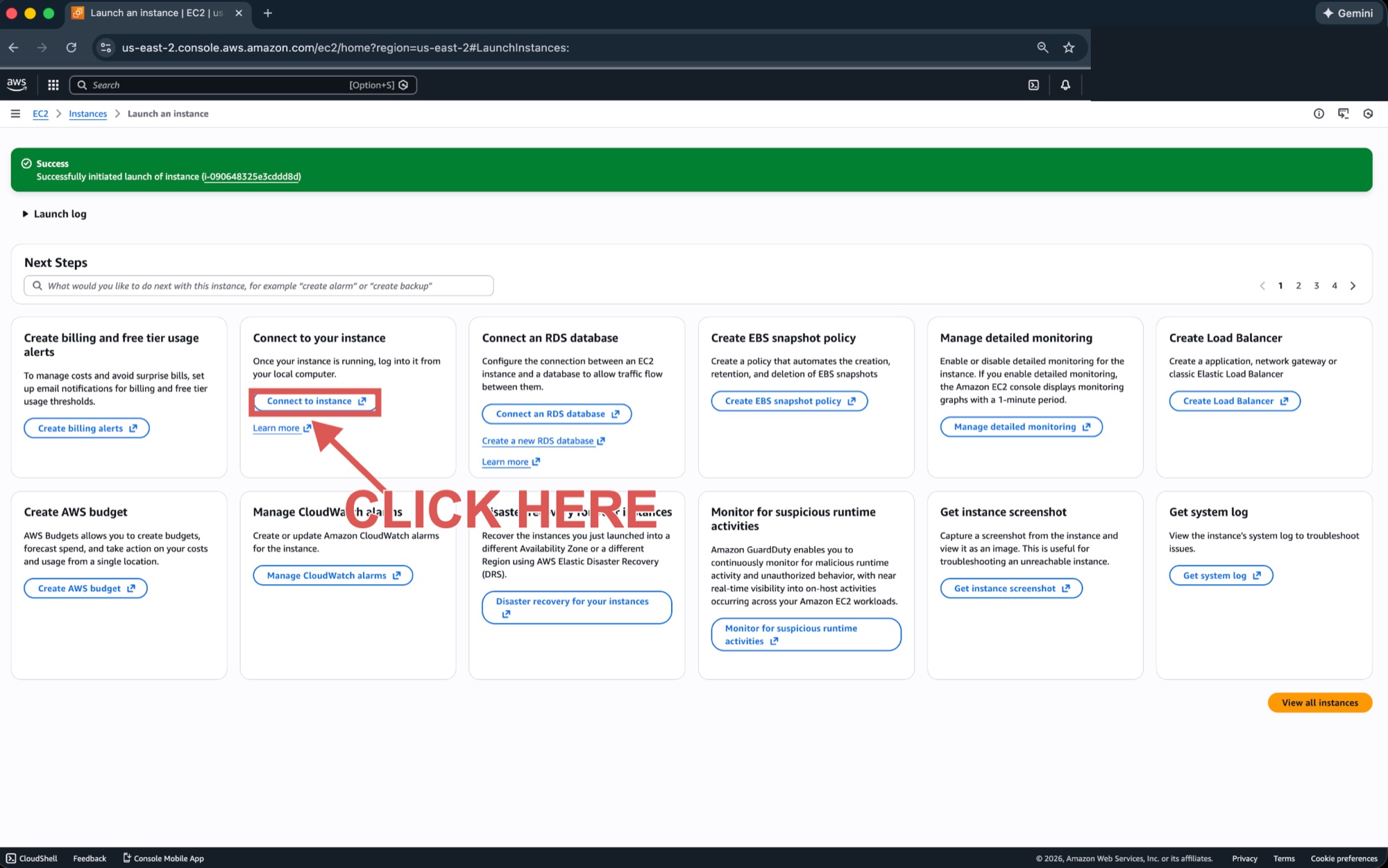} &
      \includegraphics[width=\imgw]{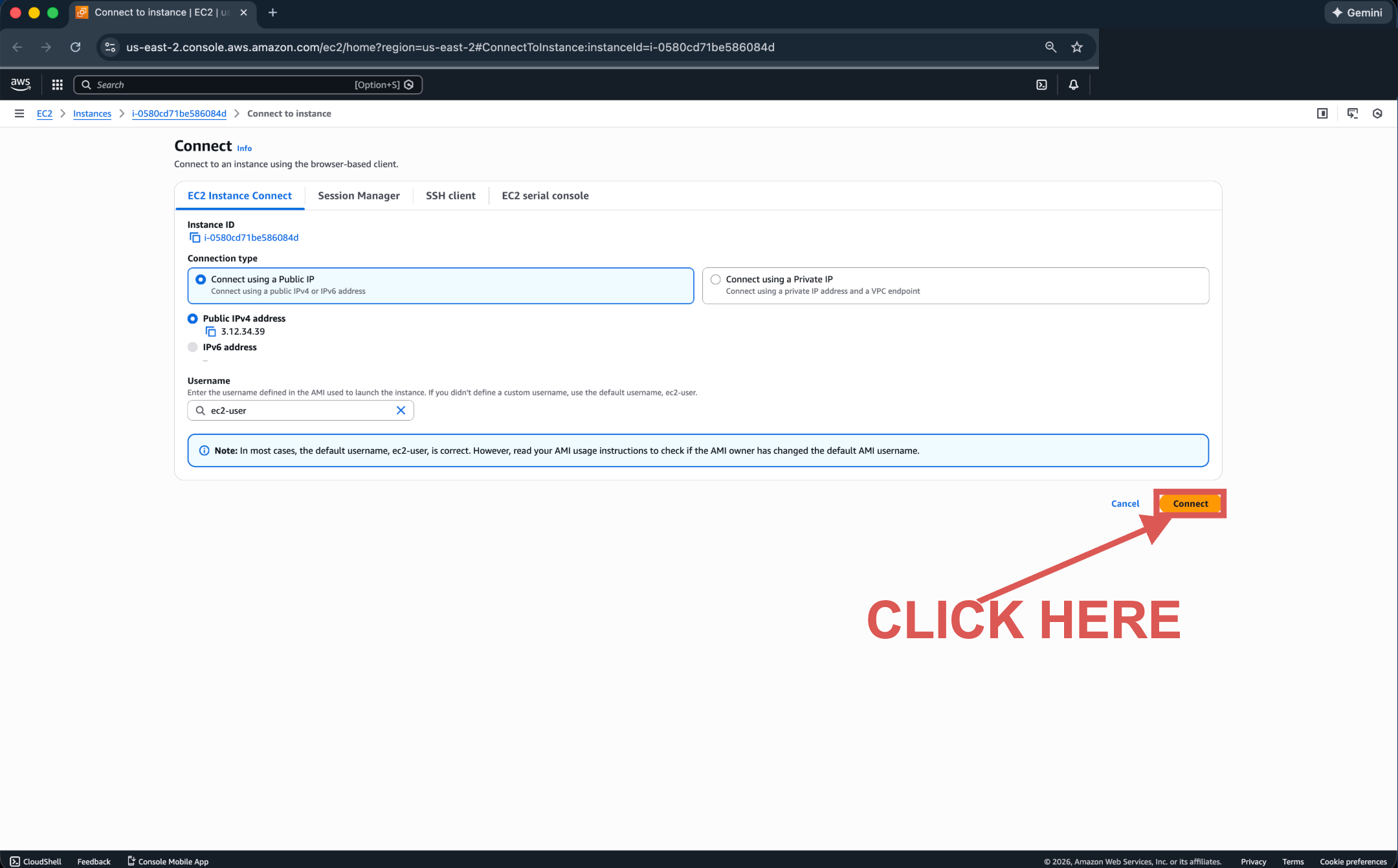} &
      \includegraphics[width=\imgw]{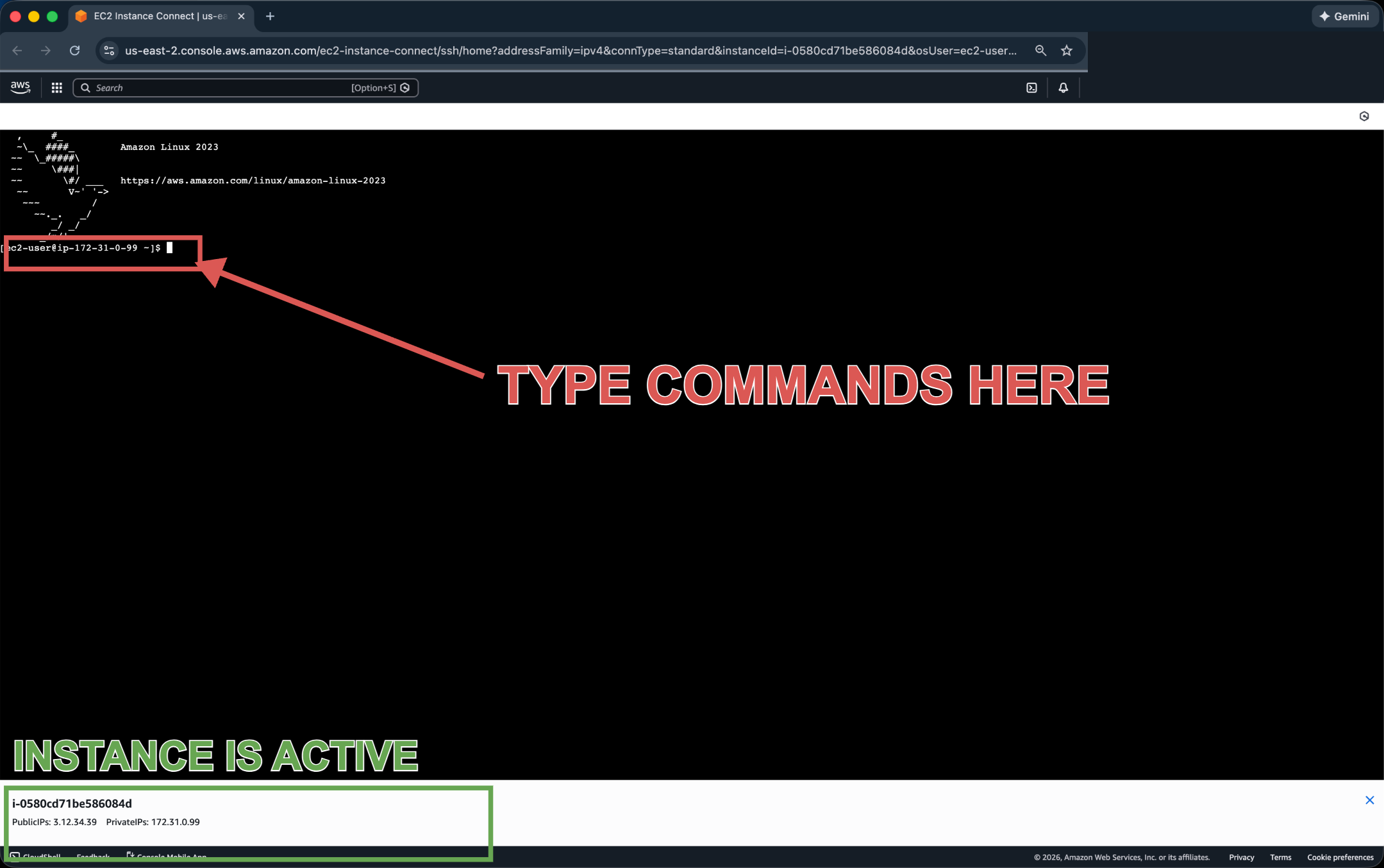}
    \end{tabular}
  \end{adjustbox}

  \vspace{-4pt}
  \caption{Beginning at the top left image, our \SketchVLM framework visually guides the user how to set up a free EC2 instance through the notoriously non-user-friendly Amazon Web Services web interface.}
  \label{fig:aws_demo}
\end{figure}

\begin{figure}[H]
  \centering
  \begin{adjustbox}{width=0.85\textwidth} 
    \setlength{\tabcolsep}{0pt}
    \renewcommand{\arraystretch}{0}
    \newcommand{\imgw}{0.24\textwidth}
    \newcommand{\imgwleft}{0.30\textwidth}

    \begin{tabular}{cc@{\hspace{0.02\textwidth}}cc}
      \includegraphics[width=\imgwleft]{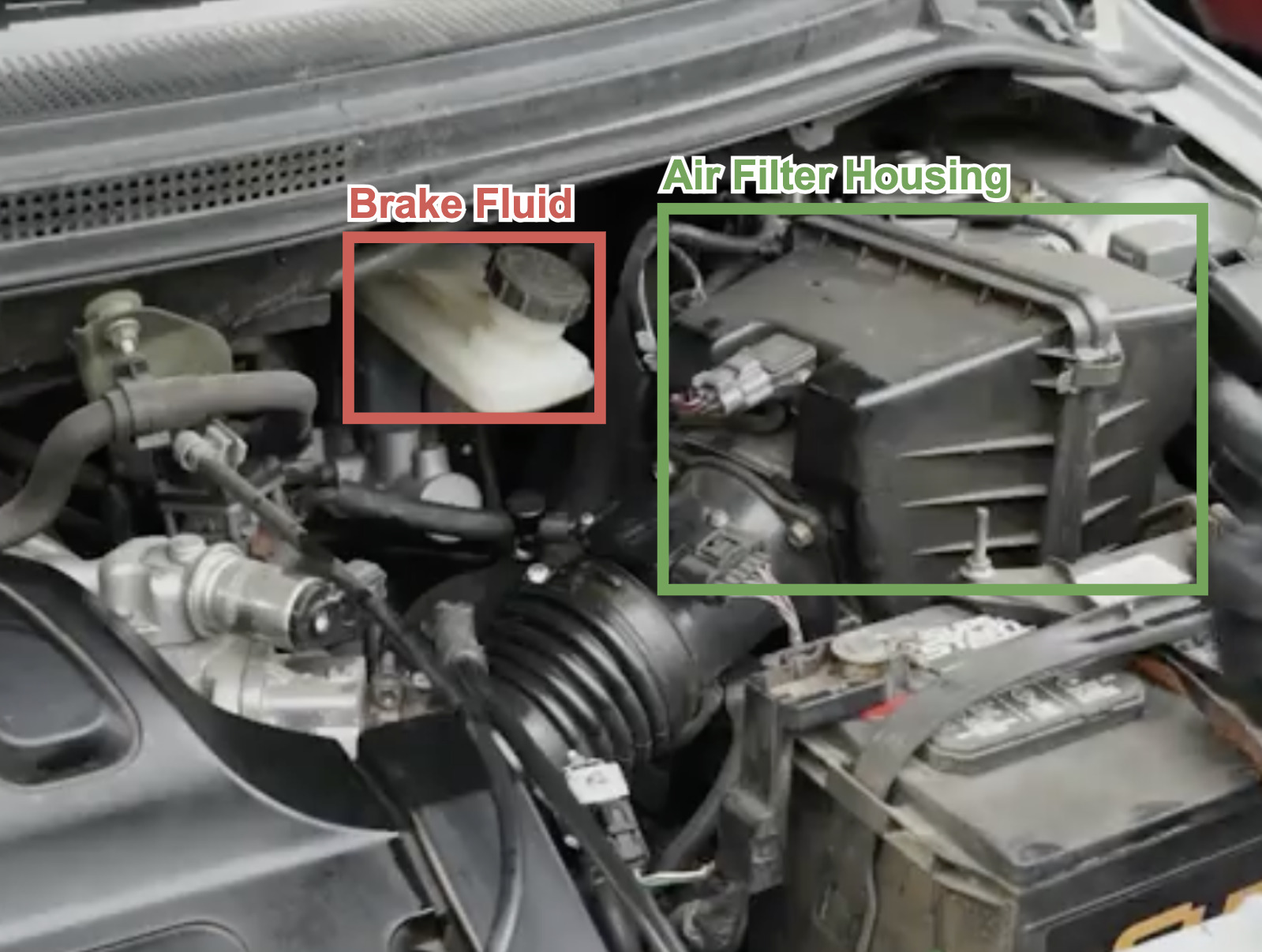} &
      \includegraphics[width=\imgw]{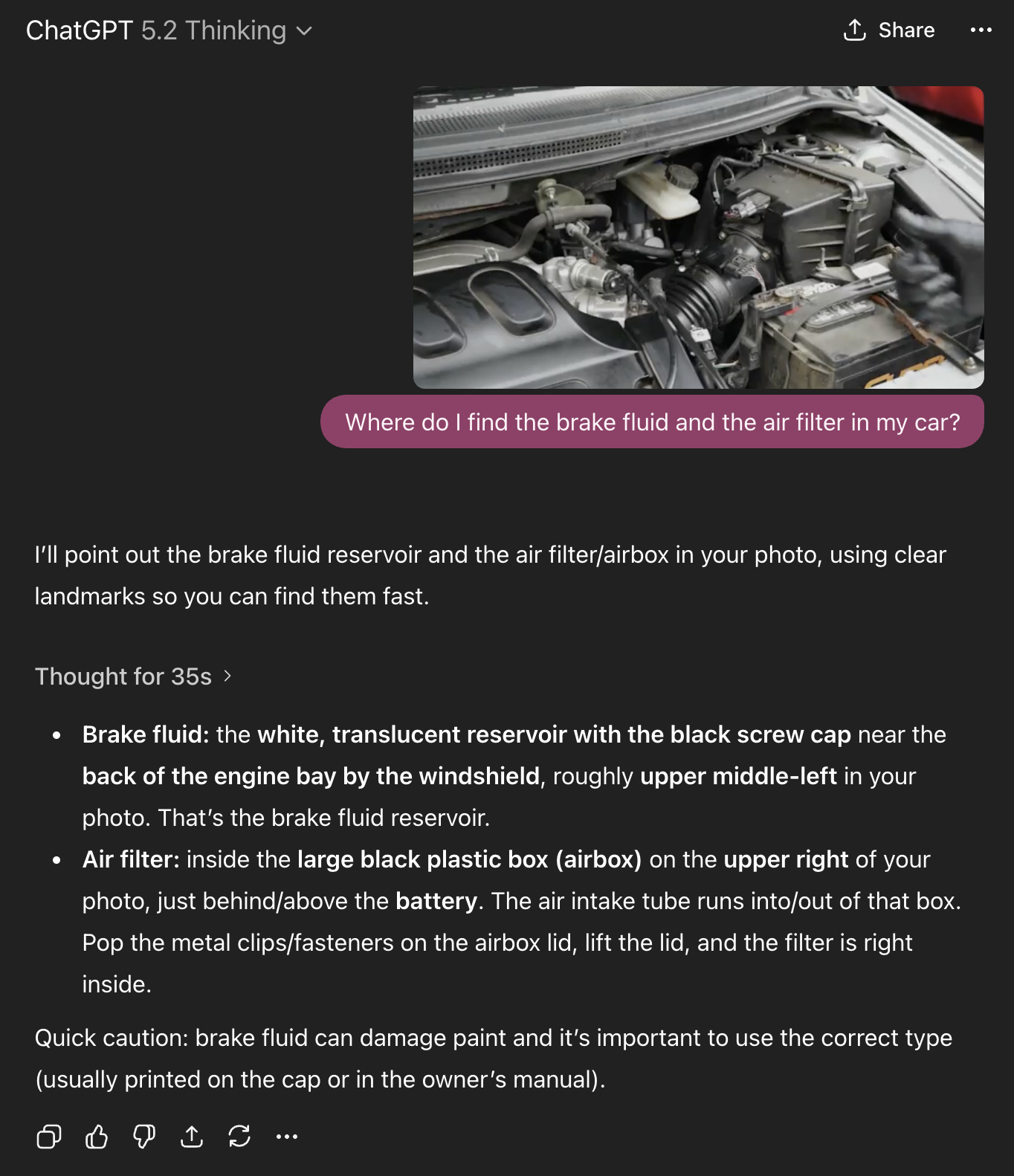} &
      
      \includegraphics[width=\imgwleft]{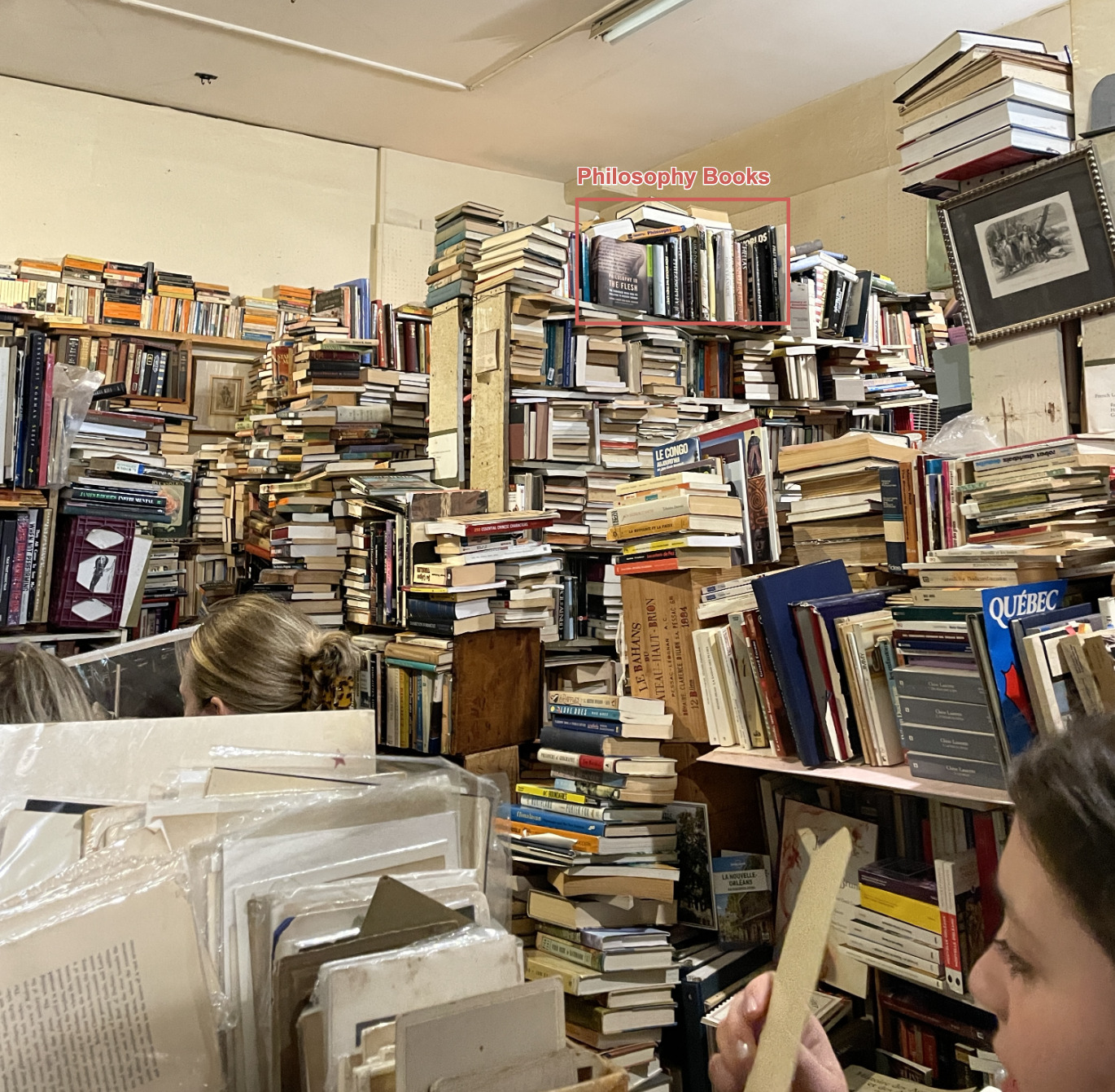} &
      \includegraphics[width=\imgw]{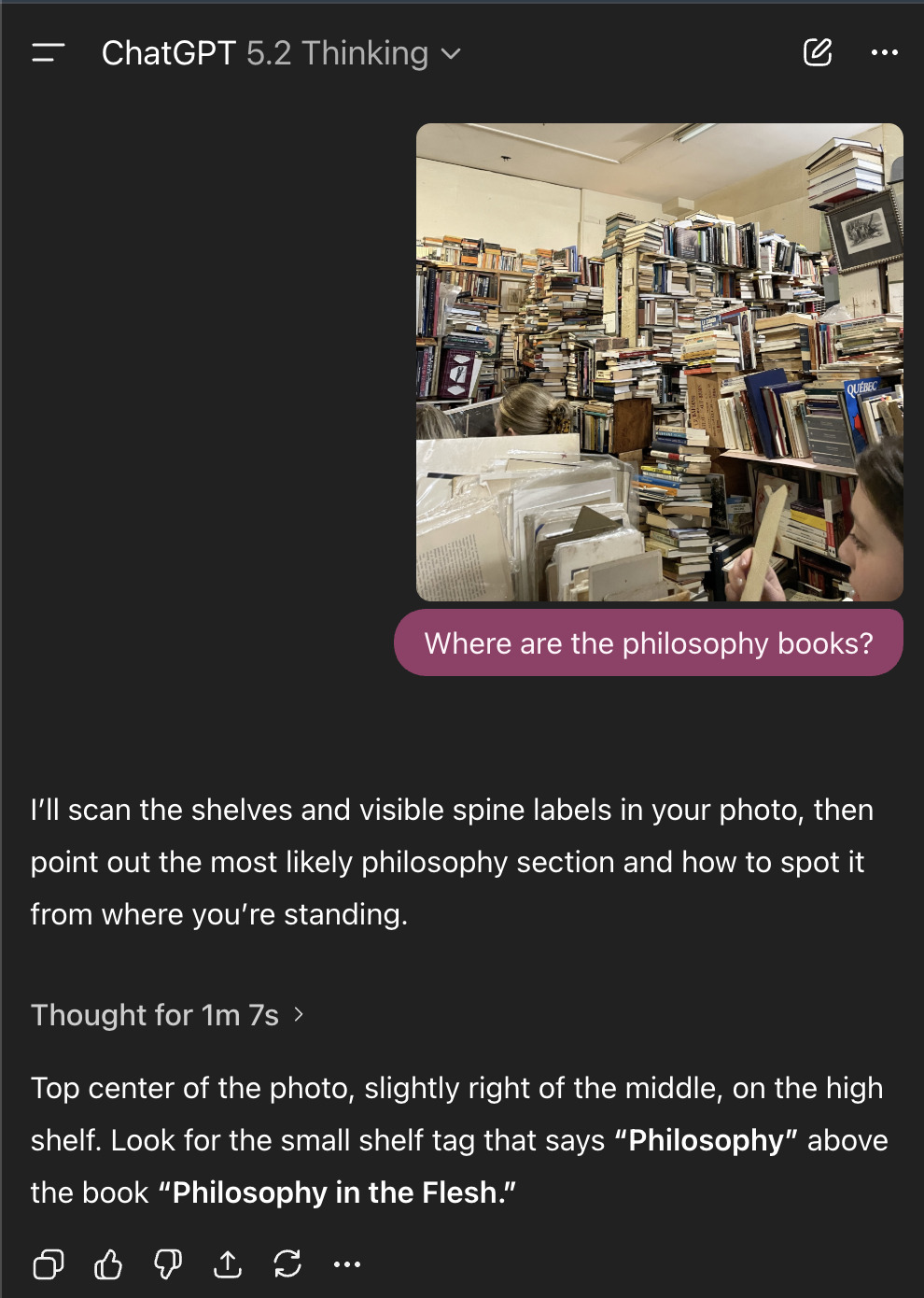} \\[-1pt]
      
      \includegraphics[width=\imgwleft]{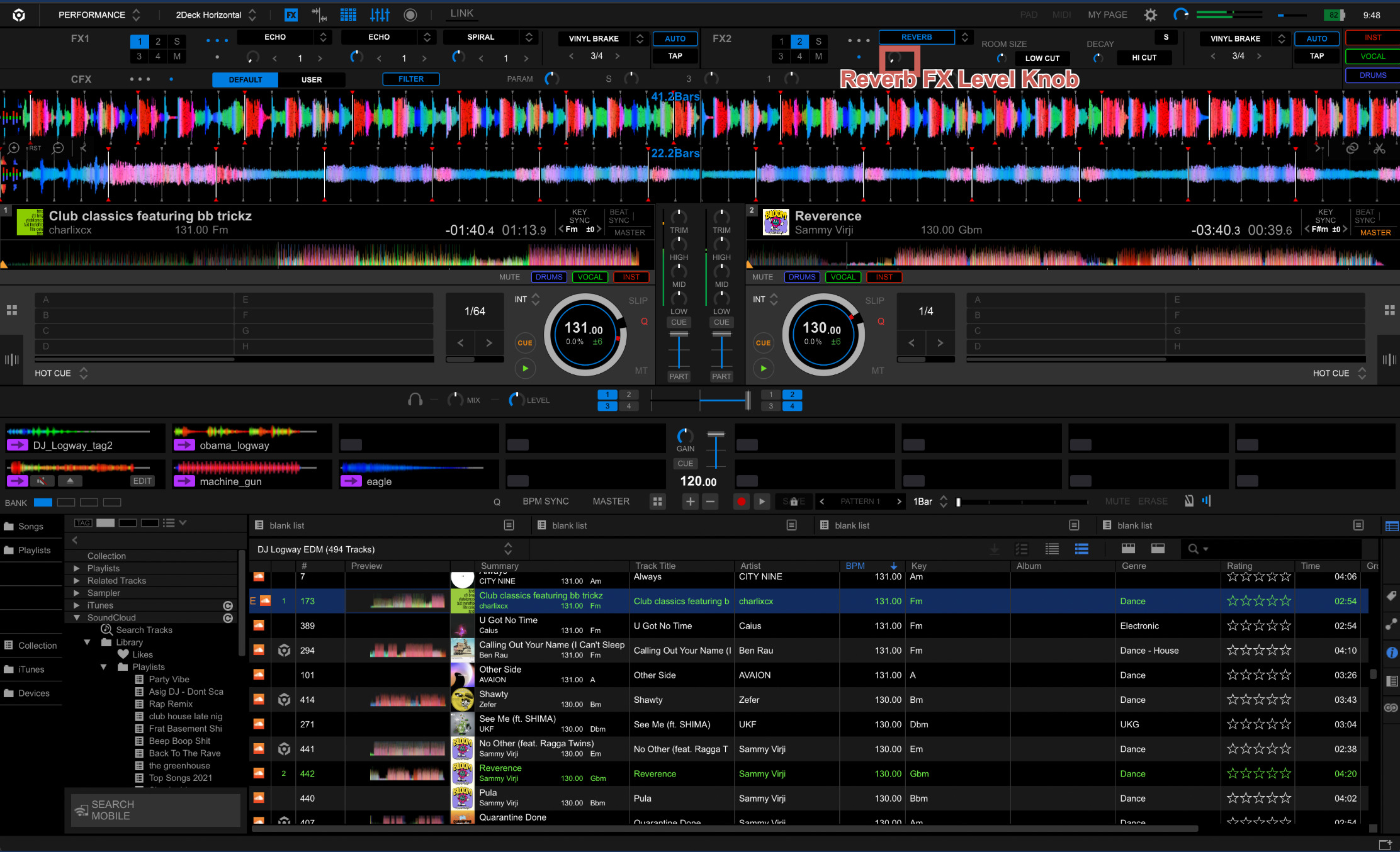} &
      \includegraphics[width=\imgw]{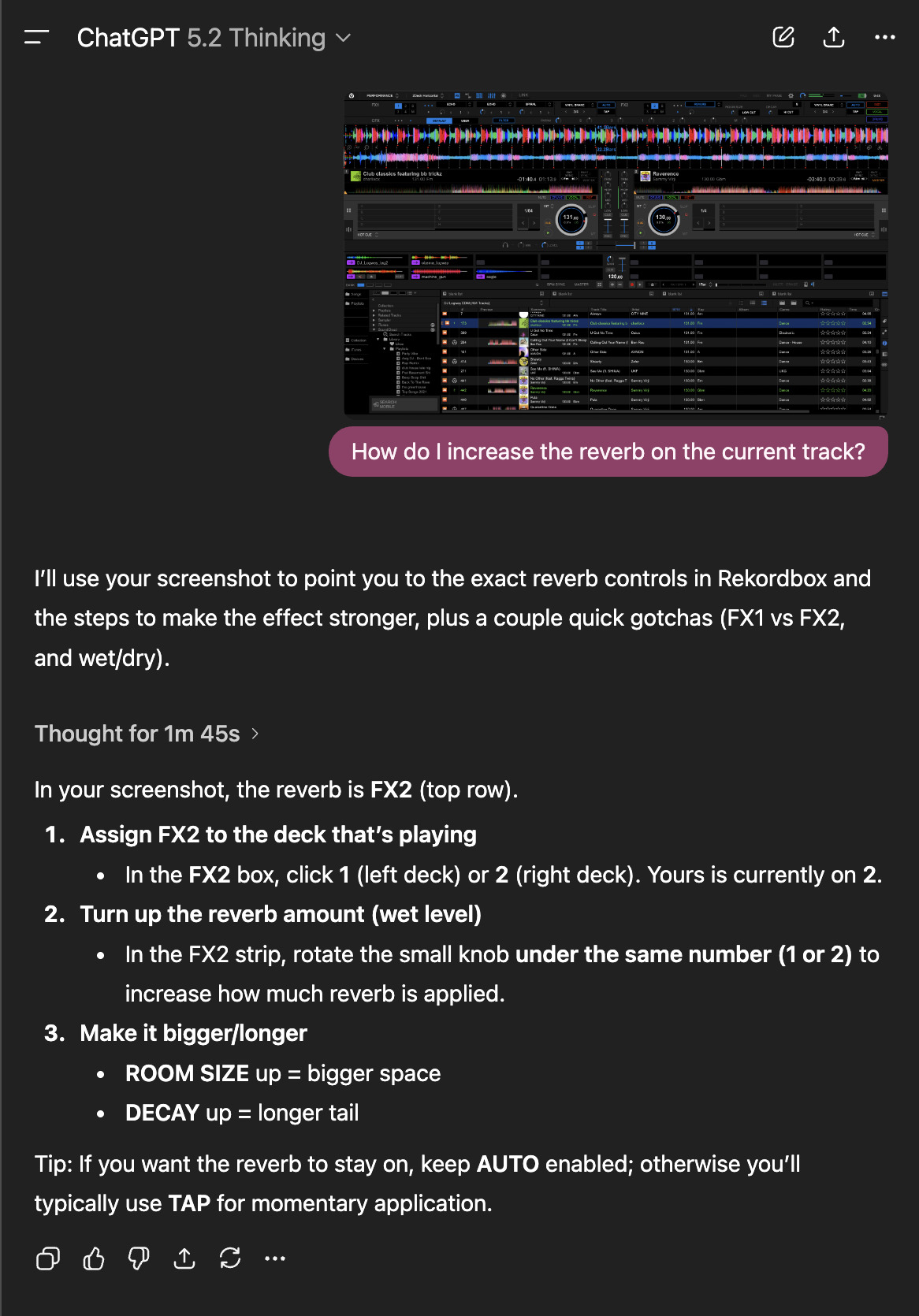} &
      
      \includegraphics[width=\imgwleft]{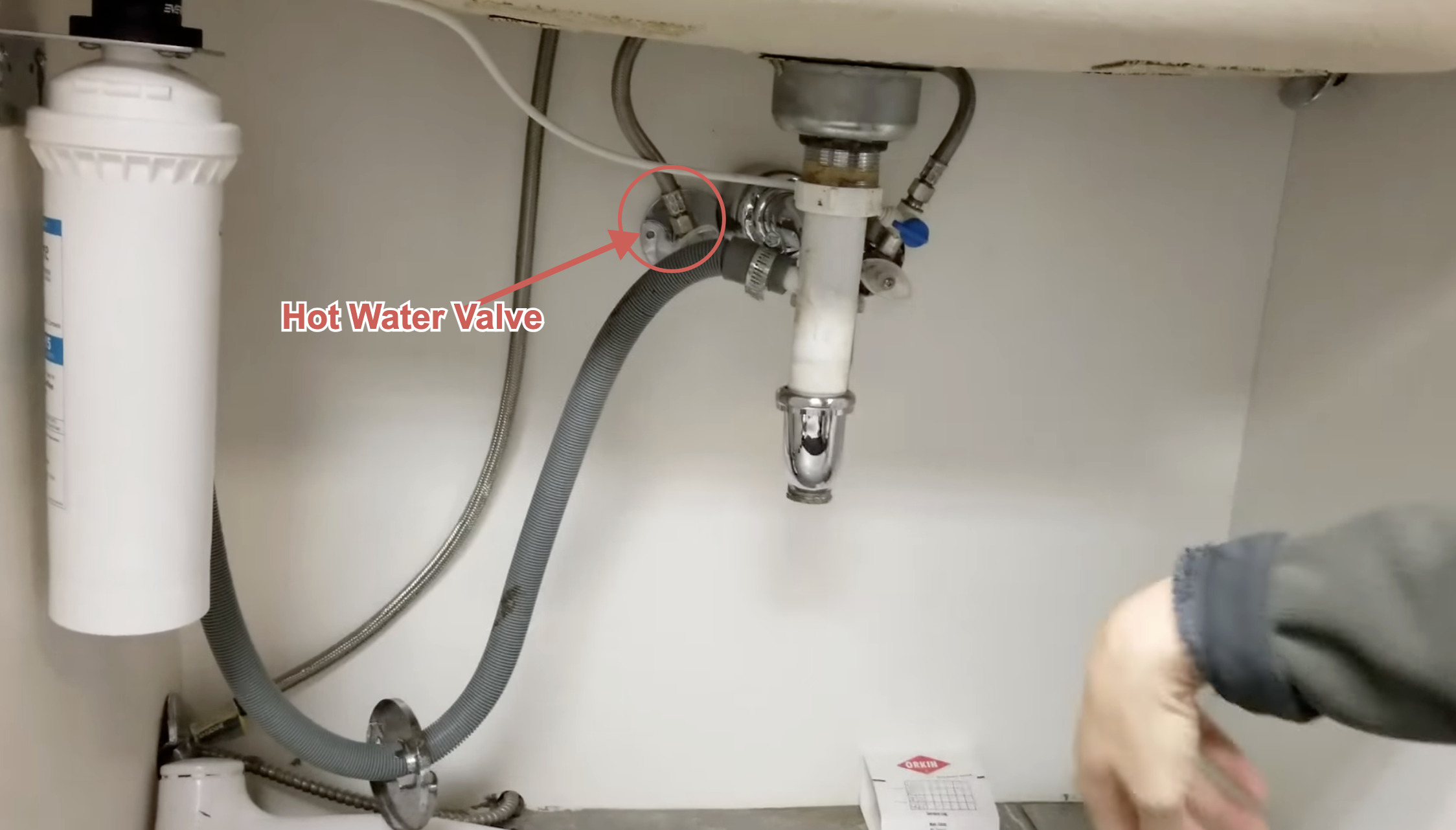} &
      \includegraphics[width=\imgw]{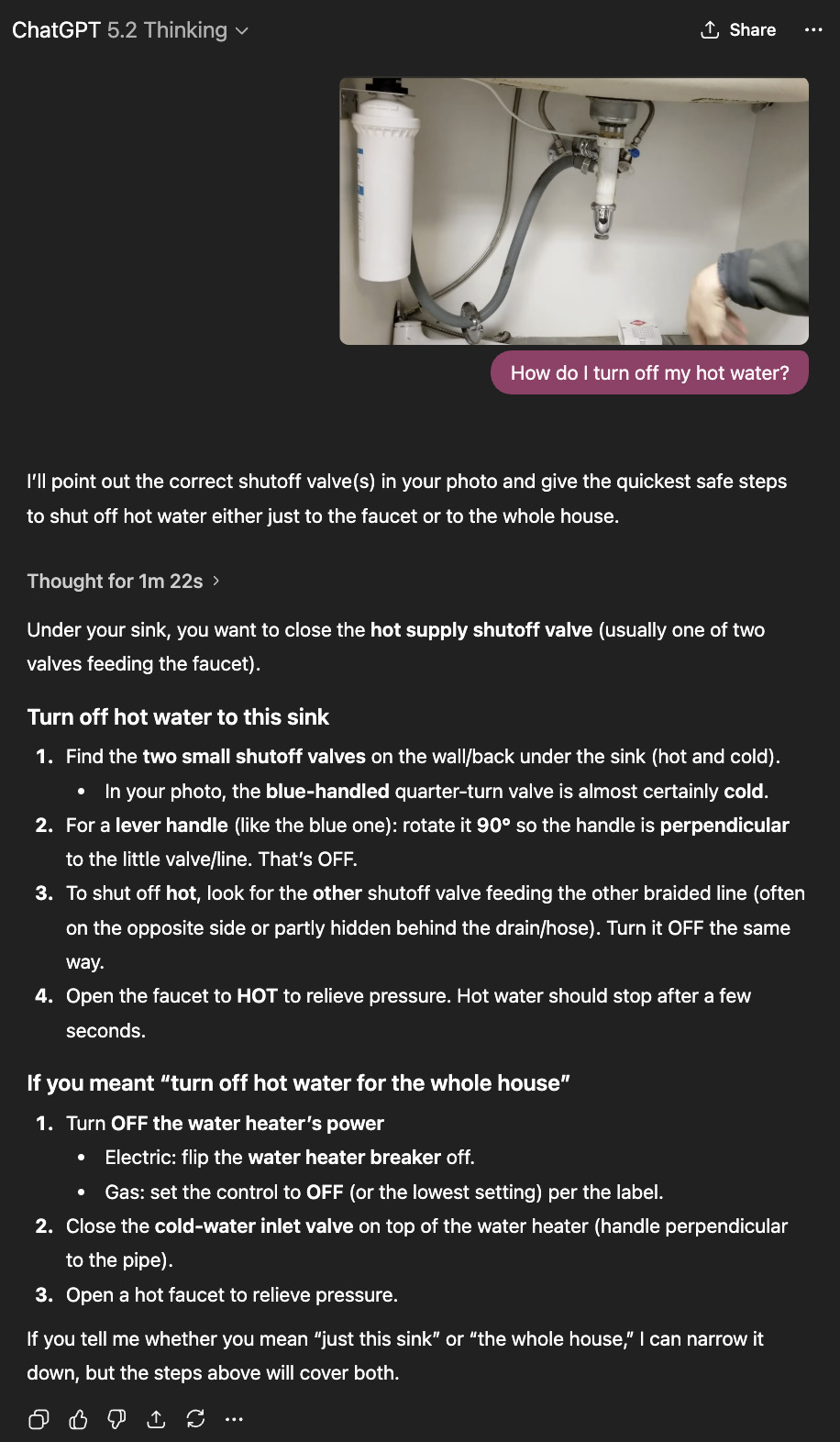} \\[-1pt]
      
      \includegraphics[width=\imgwleft]{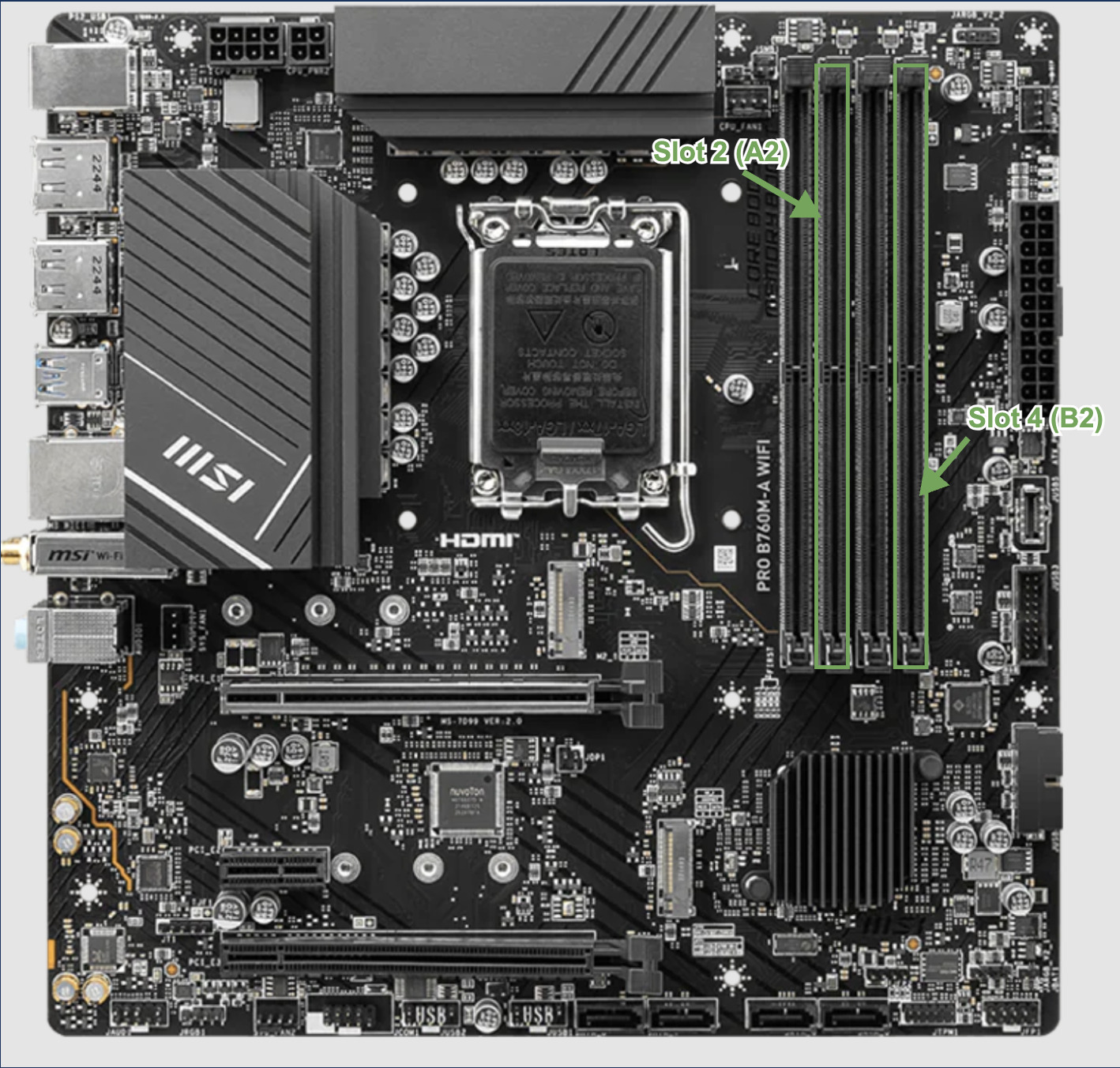} &
      \includegraphics[width=\imgw]{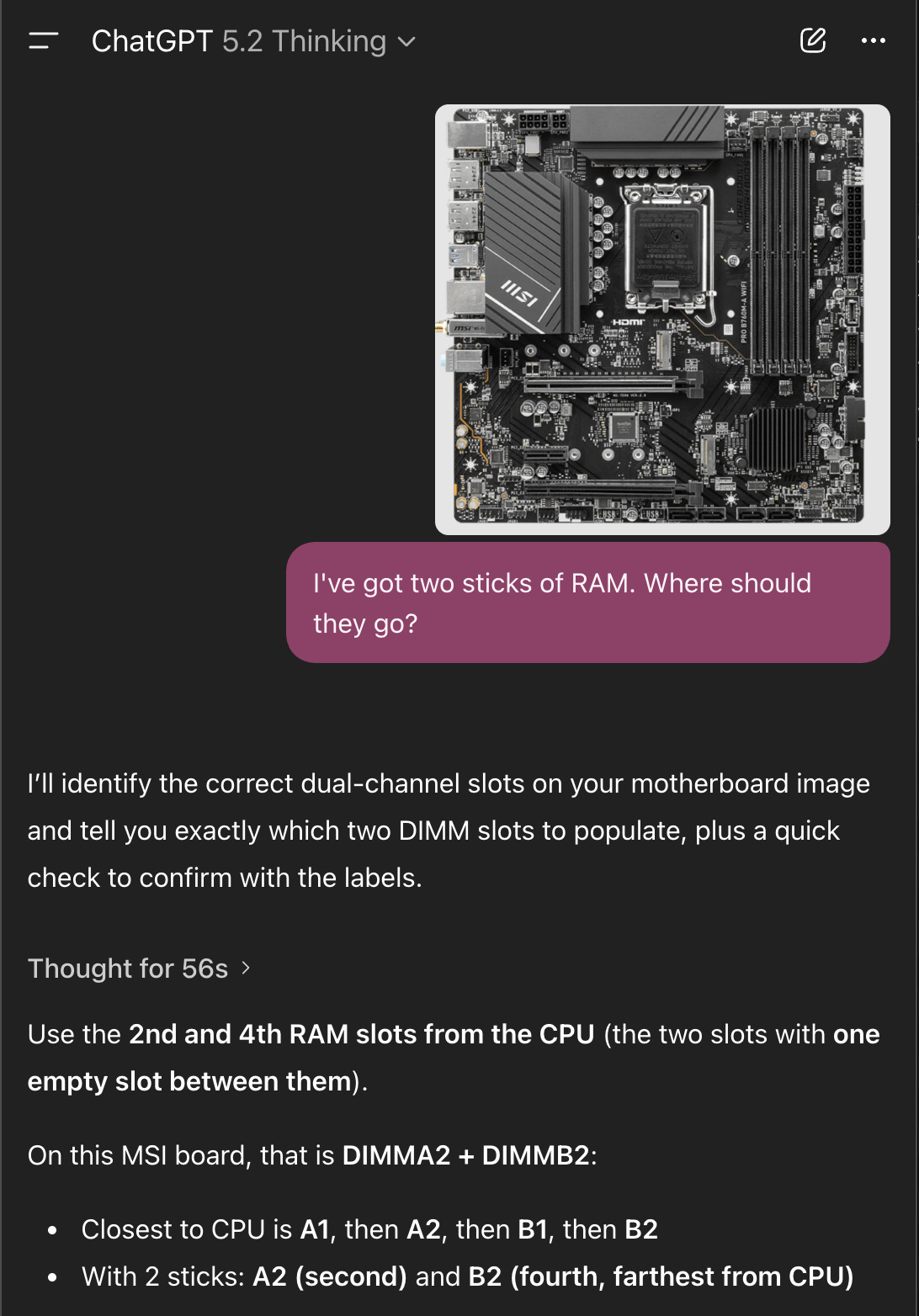} &
      
      \includegraphics[width=\imgwleft]{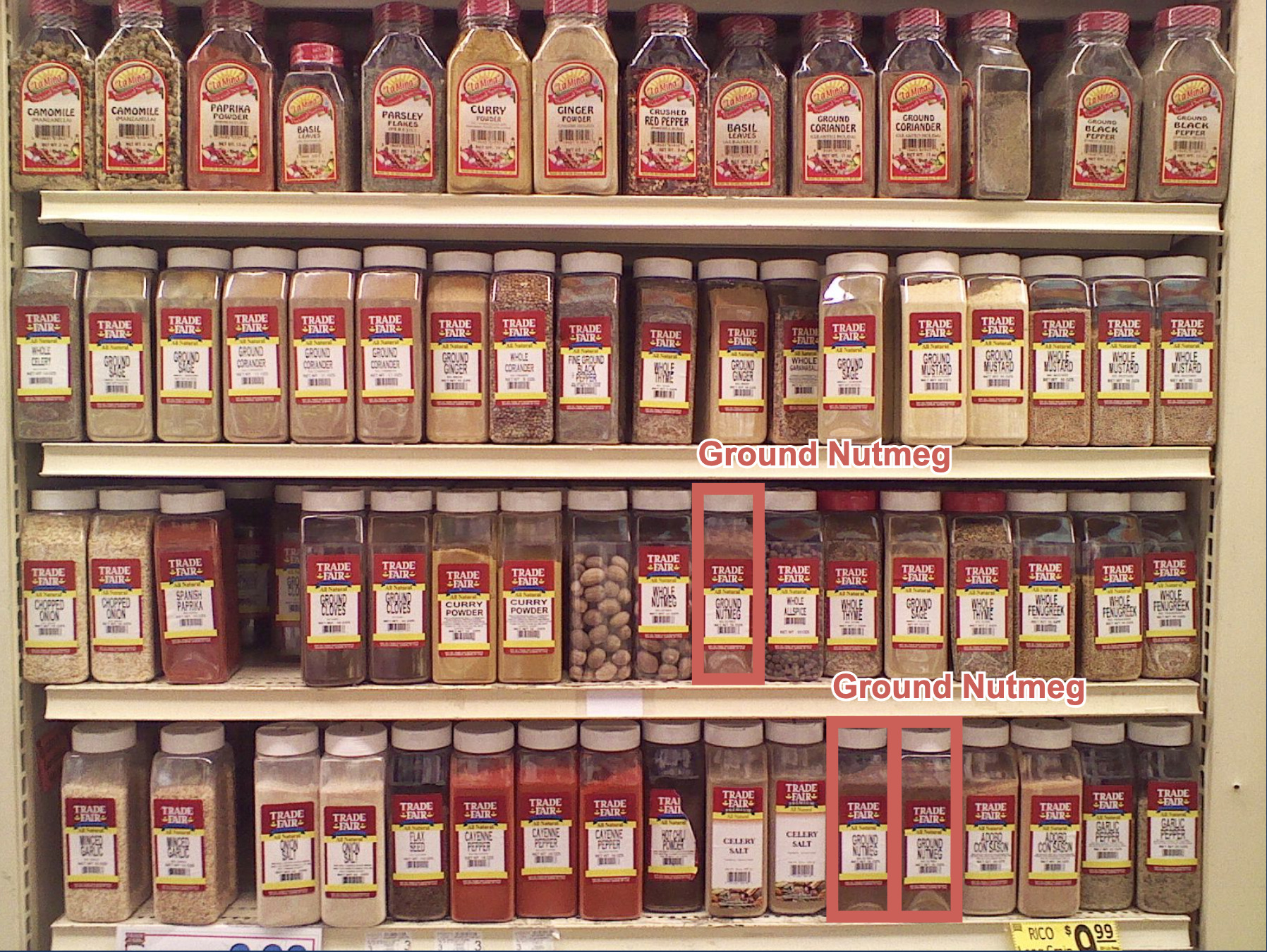} &
      \includegraphics[width=\imgw]{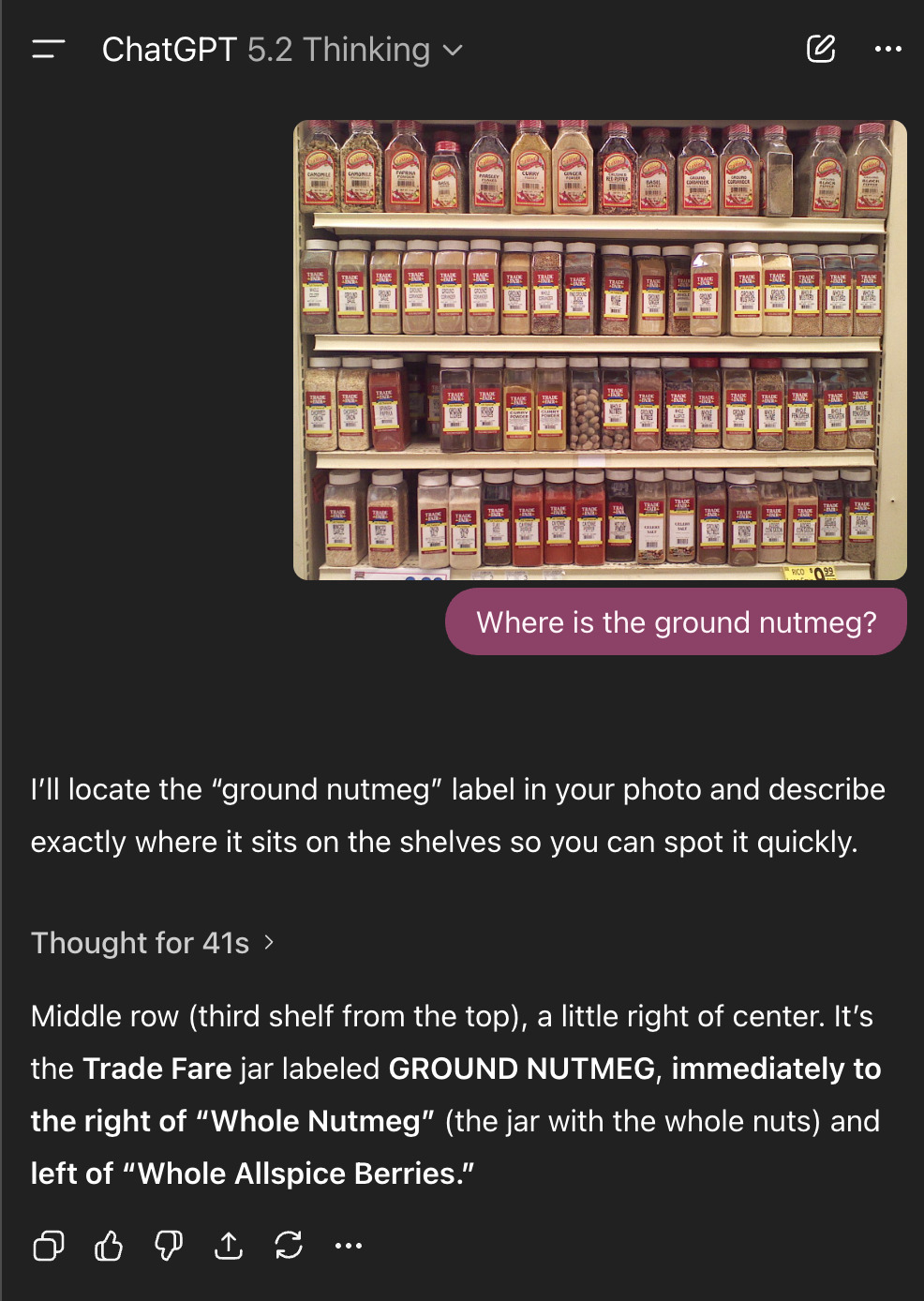}
    \end{tabular}
  \end{adjustbox}

  \vspace{-4pt}
  \caption{\SketchVLM responds via visual annotations while ChatGPT responds with text only.}
  \label{fig:image_chat_pairs}
\end{figure}

\clearpage

\begin{figure*}[t]
  \centering

  \begin{minipage}[t]{0.33\textwidth}
    \centering
    \vspace{0pt}%
    \includegraphics[width=\textwidth]{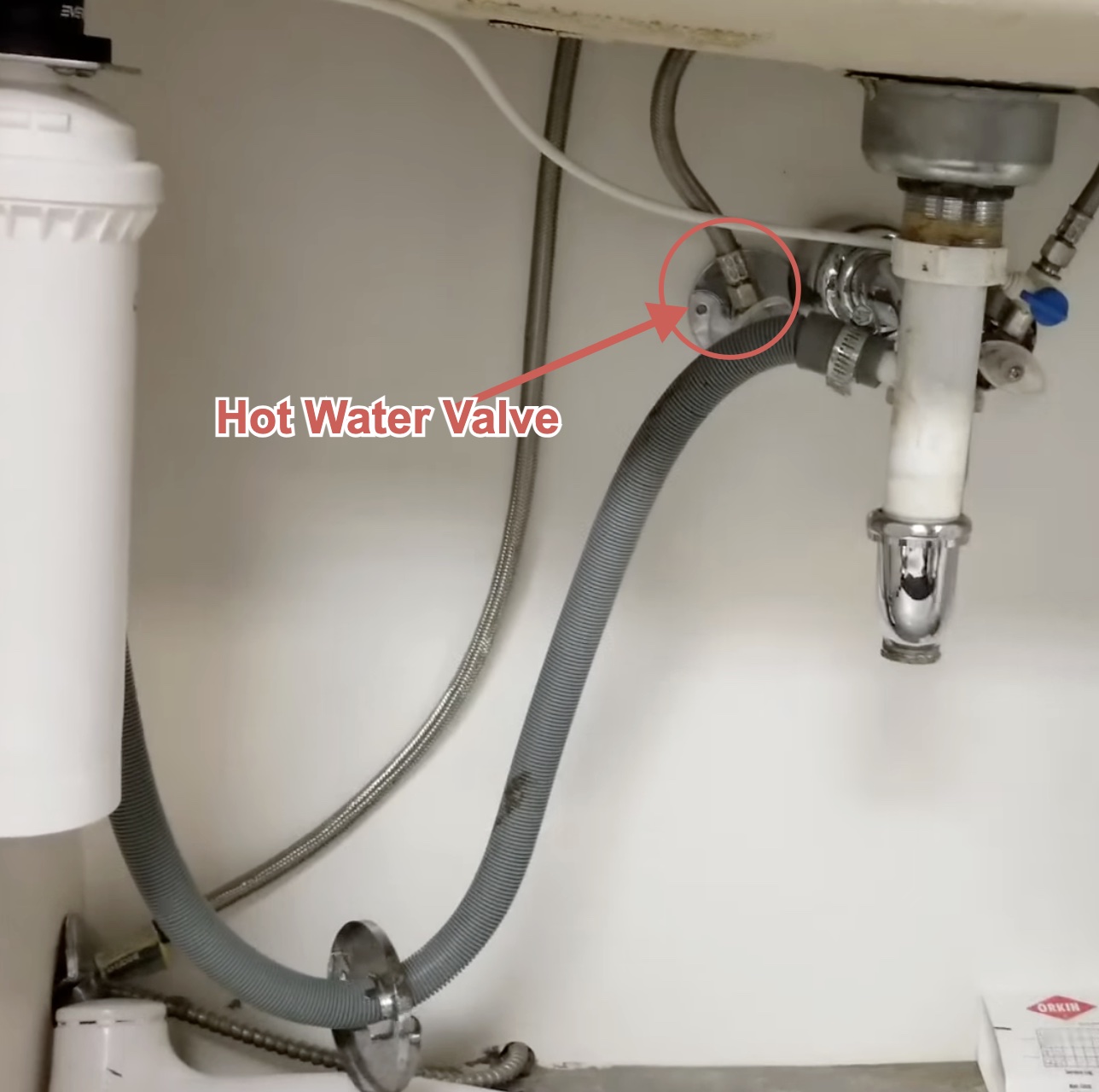}\par
    \vspace{0.5ex}
    \parbox{0.95\textwidth}{\centering\footnotesize How do I turn off my hot water?}
  \end{minipage}\hfill
  \begin{minipage}[t]{0.33\textwidth}
    \centering
    \vspace{0pt}%
    \includegraphics[width=\textwidth]{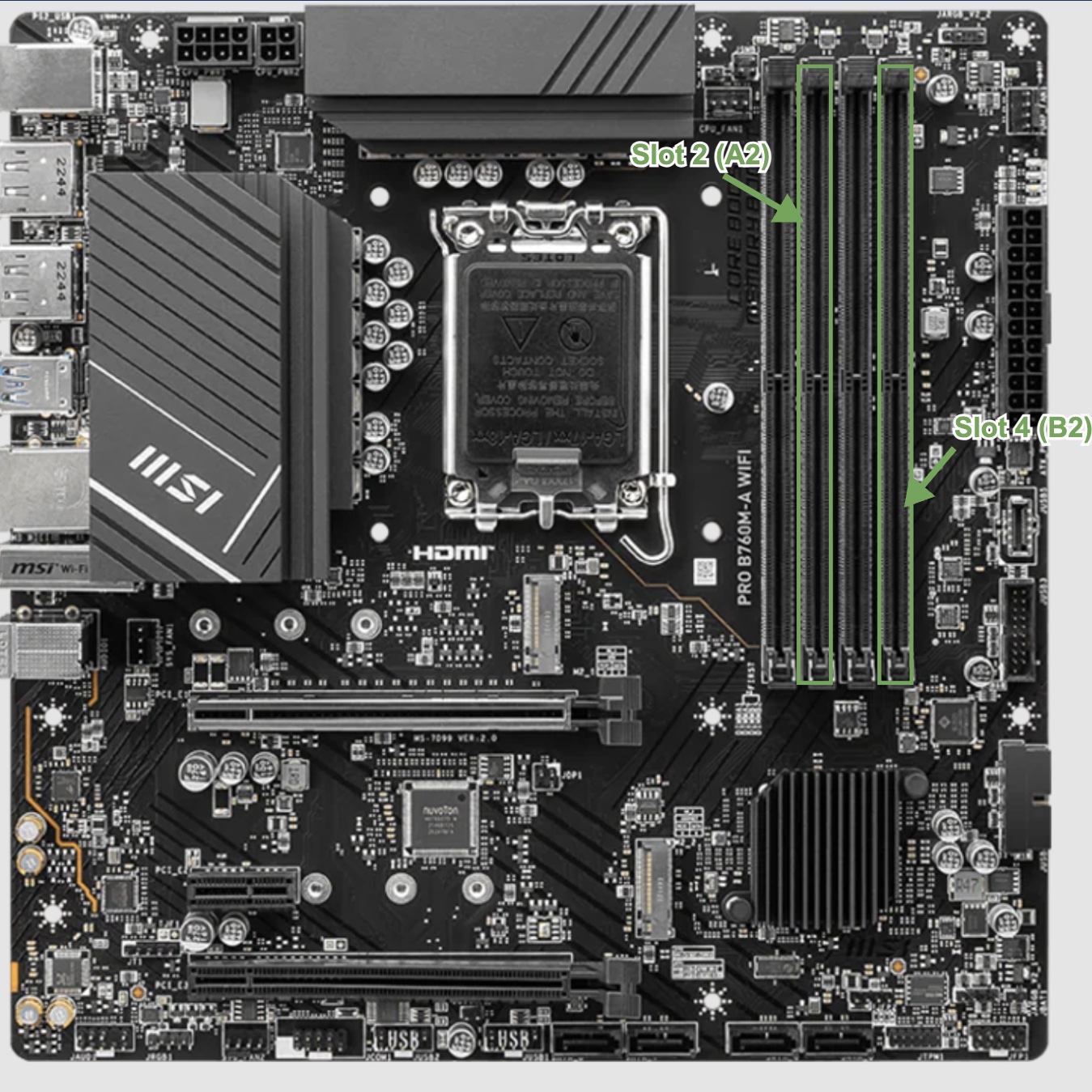}\par
    \vspace{0.5ex}
    \parbox{0.95\textwidth}{\centering\footnotesize I've got two sticks of RAM. Where do they go?}
  \end{minipage}\hfill
  \begin{minipage}[t]{0.33\textwidth}
    \centering
    \vspace{0pt}%
    \includegraphics[width=\textwidth]{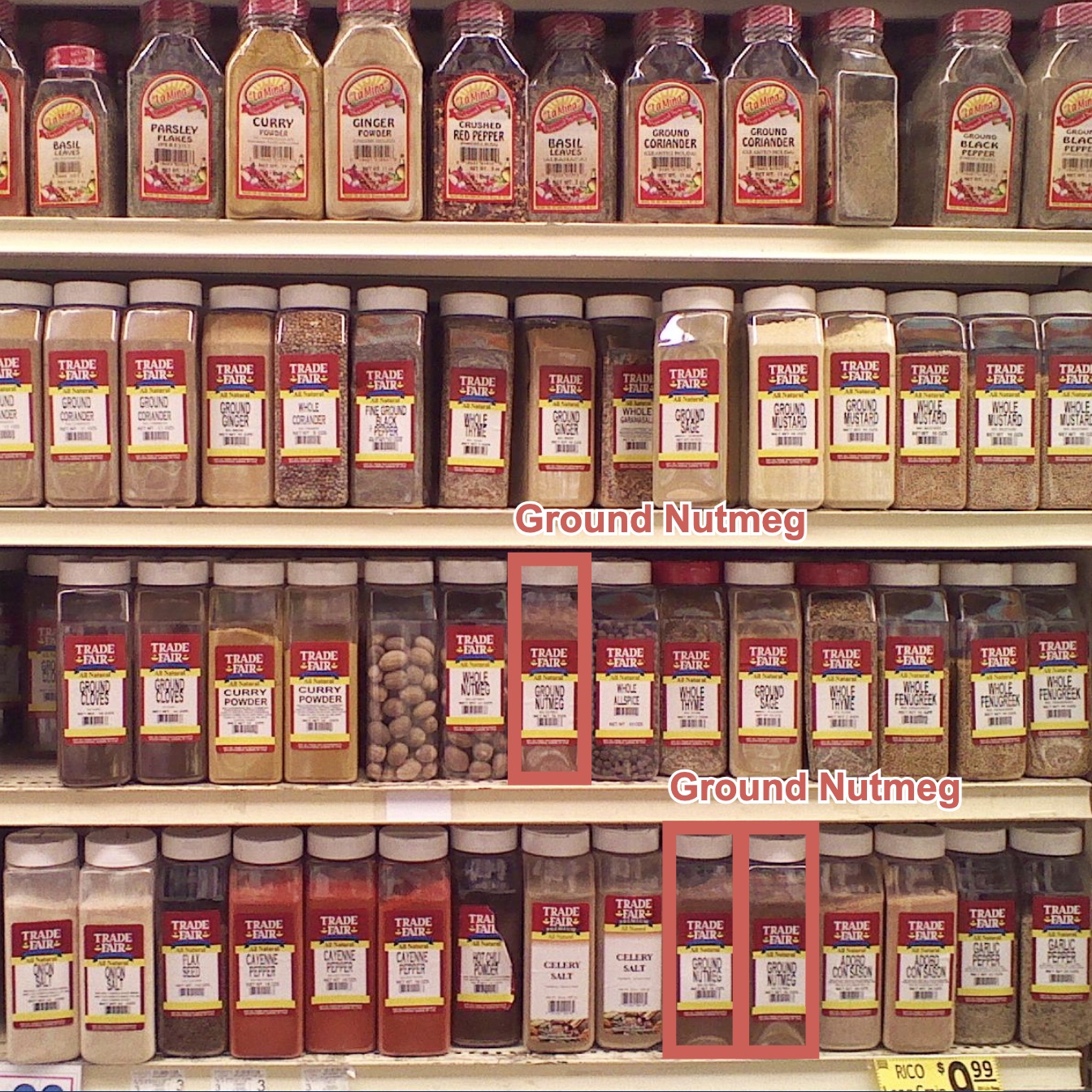}\par
    \vspace{0.5ex}
    \parbox{0.95\textwidth}{\centering\footnotesize Where is the ground nutmeg?}
  \end{minipage}

  \caption{\SketchVLM can be used in a variety of real-world use cases.}
  \label{fig:real_world_cases}
\end{figure*}

\begin{figure*}[t]
    \centering
    \includegraphics[width=\textwidth]{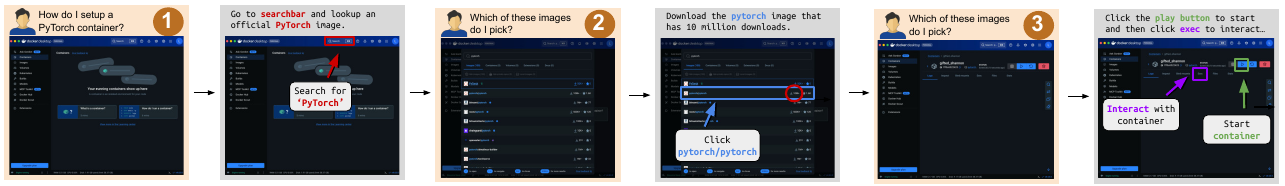}
    \caption{Multi-turn session where the model explains how to set up PyTorch in a Docker container.}
    \label{fig:docker_multiturn}
\end{figure*}

\begin{figure*}[t]
    \centering
    \includegraphics[width=\textwidth]{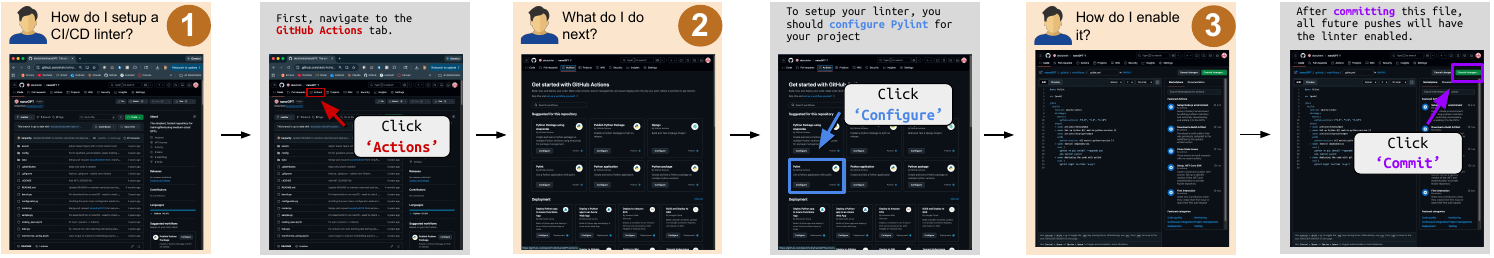}
    \caption{Multi-turn session where the model explains how to set up a CI/CD pipeline in GitHub.}
    \label{fig:github_multiturn}
\end{figure*}

\clearpage

\section{Dataset Creation}
\label{app:dataset_creation}

We use publicly available datasets, benchmarks, models, and APIs only for research evaluation. We cite the creators of the datasets, models, and tools used in our experiments in the corresponding dataset, baseline, and model-settings sections. We use these artifacts consistently with their intended research or API use. Our generated benchmarks, code, and demo are intended for research and should not be used for high-stakes decision making without additional validation. For real-world demo screenshots, we manually redact or crop user-specific information before inclusion.

\subsection{Connect-the-Dots}
\label{app:connect_dots_dataset_creation}

We collect 100 connect-the-dots images spanning three subsets with varying dot counts and background clutter.
\begin{enumerate}
    \item \textbf{Random dots:} We use a Python script with the Pillow image library to generate 3 images each with 4, 5...10 dots randomly placed for a total of 21 images. Each dot has a number that corresponds to the order it should be connected.
    \item \textbf{Outlines:} We convert 30 silhouette SVGs from Openclipart~\cite{openclipart} into connect-the-dots puzzles.
    Each SVG was flattened to polylines, simplified with the Douglas–Peucker algorithm \cite{douglas1973algorithms}, and normalized to a unit square.
    The main contour was then resampled to place 30 evenly spaced dots along the boundary, with number labels slightly offset.
    We manually filtered out cases with distorted or self-intersecting shapes, which often occurred for concave outlines.
    \item \textbf{Worksheets:} We gather 49 images of preexisting connect-the-dot worksheets from online sources. These images differ from the outlined images in how they often contain irrelevant information such as lines at the top of the image for students to mark their name. To obtain the ground truth strokes for these images, we manually annotate the coordinates of each dot.
\end{enumerate}

\subsection{Counting}
\label{app:counting_dataset_creation}
The counting dataset combines three sources: CountBench \cite{beyer2024paligemma}, TallyQA \cite{acharya2019tallyqa}, and Pixmo-Count \cite{deitke2025molmo}. The CountBench and TallyQA subsets together contain 746 samples, covering object counts ranging from 0 to 10. Pixmo-Count contributes 443 samples after removing unsuitable cases from the original 526-image test split and similarly includes object counts from 1 to 10.

\subsection{\drawingshapes}
\label{app:shapes_dataset_creation}
The testing dataset consists of 1,000 carefully selected images from the 5,000 COCO validation images, chosen to ensure a balanced distribution of object counts across classes and object sizes (small, medium, and large).

\subsection{Part Labeling}
\label{app:labeling}
We carefully selected images from two datasets, PACO \cite{ramanathan2023paco} and Pascal-Part \cite{chen2014detect}. 
The selected images satisfy the following criteria:
\begin{enumerate}
    \item Each image contains only one object corresponding to the target class name.
    \item The object's size occupies at least 10\% of the total image area.
    \item Each selected object has at least four part labels annotated.
    \item The dataset maintains a balanced distribution of objects across different classes.
\end{enumerate}

\noindent After selection, the final dataset used for the part labeling task consists of 985 images covering 52 class names.

\subsection{\maze}
\label{app:maze_dataset_creation}

Given a start point, an end point, and a set of direction commands (\eg, \texttt{Up, Down, Left, Right}), the model must determine if the path reaches the goal without crossing any border walls. We create 200 unique 3x3 grids where the shortest path length between the starting green square and ending red square varies from 3 to 8 steps. For each maze, we take the ground truth path and randomly change one of the direction steps in order to make an invalid path (for example, \texttt{\underline{Left}, Right, Down} could be changed to \texttt{\underline{Right}, Right, Down}) (\cref{fig:maze_valid_invalid_1row}).

\subsection{Ball Drop}
\label{app:ball_drop_dataset_creation}
We evaluate our framework using Visual Physics Comprehension Test (\vpct \cite{cbrowerVPCT2025}), which consists of 100 hand-crafted images where the model must determine which of the buckets the ball will fall into after it is dropped. While \vpct does provide a simple way to evaluate physics understanding of VLMs, it does not contain any ground truth data of the trajectory of the ball. Therefore, in order to evaluate how well \SketchVLM models can draw the true trajectory of the ball paths, we generate our own benchmark (\balldrop). We simulate the trajectory of the ball using PHYRE \cite{bakhtin2019phyre} to obtain ground truth ball trajectory data. In contrast to \vpct, our Ball Drop benchmark is synthetically generated. We generate 198 unique images, with an equal number of images containing 1, 2, and 3 randomly placed lines. We randomize the ball’s X position and fix its Y position near the top. There are four containers at the bottom of the image compared to \vpct's three containers, making it harder to guess answers correctly.

\clearpage
\twocolumn[{%
\centering
\section{Additional Task Results}
\subsection{Combined Results}
\vspace{-0.5em}
 
\captionof{table}{\SketchVLM improves visual reasoning task accuracy by +28.5 points over alternative sketching approaches. Averages are computed over VPCT, Ball Drop, Maze, and Counting tasks.}
\label{tab:avg_delta_accuracy}
\small
\begin{tabular}{l l | c c c c | c c}
\toprule
Category & Model & VPCT & \balllogo & \mazelogo & \countlogo & Avg & $\Delta$ \\
\midrule
\multirow{2}{*}{\SketchVLM}
  & \geminisketch & 96.0 & 79.7 & 98.0 & 94.5 & \multirow{2}{*}{\textbf{84.4}} & \multirow{2}{*}{\textcolor{green!50!black}{+28.5}} \\
  & \gptsketch    & 70.0 & 68.5 & 92.8 & 75.4 & & \\
\midrule
\multirow{3}{*}{Other Sketching}
  & \nanologo \nanobananapro & 63.0 & 62.6 & 93.3 & 91.7 & \multirow{3}{*}{55.9} & \multirow{3}{*}{} \\
  & \vilasrlogo \vilasr    & 37.0 & 35.9 & 50.8 & 48.6 & & \\
  & \thinkmorphlogo \thinkmorph & 27.0 & 30.3 & 62.5 & 68.1 & & \\
\bottomrule
\end{tabular}
 
\vspace{1.5em}
 
\captionof{table}{\SketchVLM produces higher quality annotations, scoring +48.3\% above alternative sketching approaches on a 1--5 VLM-judged drawing quality scale.}
\label{tab:avg_quality_delta}
\small
\begin{tabular}{l l | c c c | c c}
\toprule
Category & Model & VPCT & \balllogo & \mazelogo & Avg & $\Delta$ \\
\midrule
\multirow{2}{*}{\SketchVLM (Ours)}
  & \geminisketch & 3.12 & 4.28 & 3.69 & \multirow{2}{*}{\textbf{2.98}} & \multirow{2}{*}{\textcolor{green!50!black}{+48.3\%}} \\
  & \gptsketch    & 1.83 & 1.74 & 3.20 & & \\
\midrule
\multirow{3}{*}{Other Sketching}
  & \nanologo \nanobananapro & 1.56 & 2.56 & 3.68 & \multirow{3}{*}{2.01} & \multirow{3}{*}{} \\
  & \vilasrlogo \vilasr     & 1.36 & 1.28 & 2.78 & & \\
  & \thinkmorphlogo \thinkmorph & 1.62 & 2.11 & 1.17 & & \\
\bottomrule
\end{tabular}
 
\vspace{1.75em}
 
\subsection{Counting}
\vspace{-0.5em}
\captionof{table}{\geminisketch achieves high text-location accuracy, indicating that predicted counts are well aligned with target objects, whereas \gptsketch shows substantially weaker grounding.}
\label{tab:text_location_accuracy_counting}
\adjustbox{max width=0.30\linewidth}{%
\begin{tabular}{c|c}
\hline
Model & Accuracy (\%) \\
\hline
\geminisketch & \textbf{95.9} \\
\hline
\gptsketch & 51.0 \\
\hline
\vilasrlogo \vilasr & 59.9 \\
\hline
\kimisketch  & 70.4 \\
\hline
\end{tabular}%
}
\vspace{1em}

\makebox[\linewidth][c]{%
\begin{minipage}[t]{0.8\textwidth}
\centering
\vspace{0pt}

\captionof{table}{Counting accuracy for SketchVLM and Direct VQA across \gemmafour, Qwen3.5-397B, Kimi K2.5, and Kimi K3.}
\label{tab:counting_accuracy_comparison}

\setlength{\tabcolsep}{6pt}
\adjustbox{max width=0.6\linewidth}{%
  \begin{tabular}{l|cc}
  \toprule
  Model & SketchVLM & Direct VQA \\
  \midrule
  \gemmafour     & \textbf{55.11} & 36.33 \\
  Qwen3.5-397B   & 58.15 & \textbf{74.31} \\
  Kimi K2.5      & 63.26 & \textbf{66.85} \\
  Kimi K3      & \textbf{93.65} & 86.19 \\
  \bottomrule
  \end{tabular}%
}

\end{minipage}%
}

\vspace{1em}
}]

\paragraph{Kimi~K3 is the strongest open-source backbone for Counting.}
We compare SketchVLM and Direct VQA on Counting across four open-source
backbones (\Cref{tab:counting_accuracy_comparison}). Kimi~K3 is best under both
setups, reaching $93.65$ with SketchVLM, $19.34$ points above the best
configuration of any other model. Its lead also widens under SketchVLM, from
$11.88$ to $30.39$ points over the runner-up. The gain is not universal:
SketchVLM adds $18.78$ points on \gemmafour{} but costs $16.16$ on
Qwen3.5-397B and $3.59$ on Kimi~K2.5. The best open-source configuration is
therefore SketchVLM with Kimi~K3, while weaker backbones require per-model
validation.

\twocolumn[{%
\centering
\subsection{Connect-the-Dots}
\label{sec:nano_connect_dots_eval}
\vspace{1.5em}
 
\includegraphics[width=0.4\textwidth]{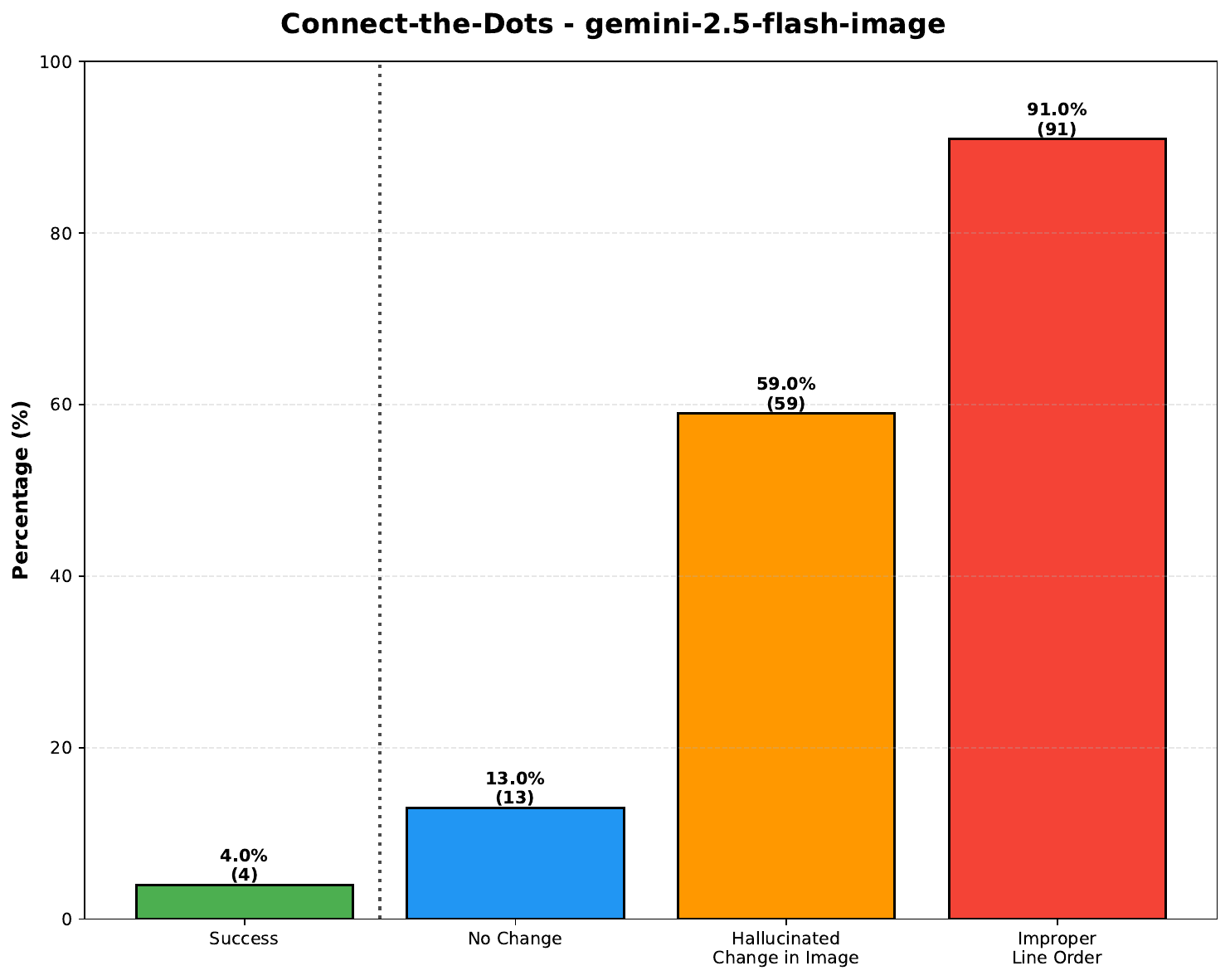}
\includegraphics[width=0.4\textwidth]{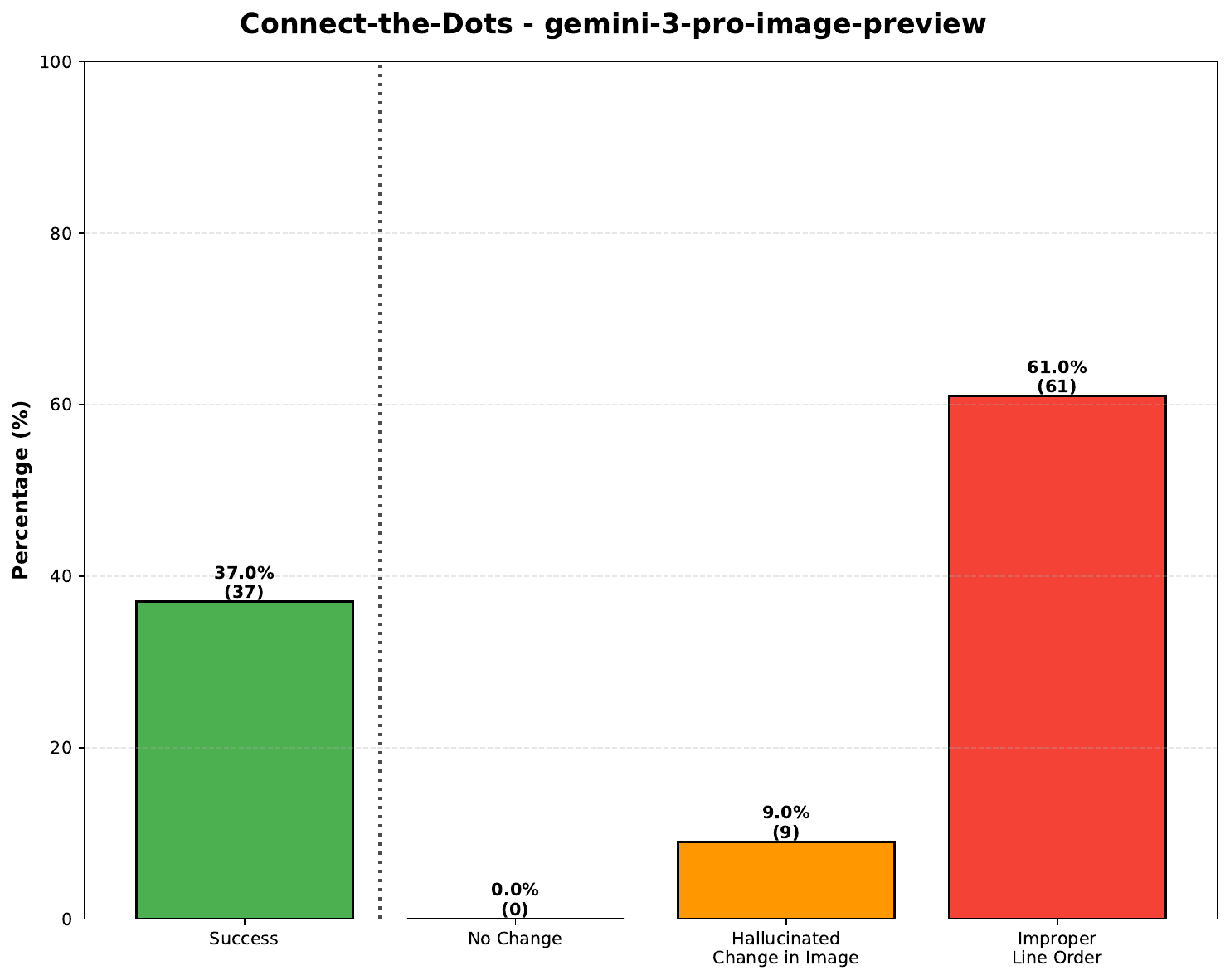}
\captionof{figure}{\nanobanana frequently adds non-existent points to the image or does not correctly follow the proper order of the connected dots. \nanobananapro is significantly better than \nanobanana, but still only completes the task without any errors 37\% of the time.}
\label{fig:nano_dots_failures}
 
\vspace{1.75em}
 
\scriptsize
\setlength{\tabcolsep}{4pt}
\renewcommand{\arraystretch}{1.2}

\adjustbox{max width=\textwidth}{%
\begin{tabular}{l
                *{3}{c}  
                *{3}{c}  
                *{3}{c}  
                *{3}{c}  
                }
\toprule
& \multicolumn{3}{c}{\textbf{Random Dots (n=21)}}
& \multicolumn{3}{c}{\textbf{Worksheets (n=49)}}
& \multicolumn{3}{c}{\textbf{Outlines (n=30)}}
& \multicolumn{3}{c}{\textbf{Total (n=100)}} \\

\cmidrule(lr){2-4}
\cmidrule(lr){5-7}
\cmidrule(lr){8-10}
\cmidrule(lr){11-13}

\textbf{Model}
& No Grid & With Grid & $\Delta$
& No Grid & With Grid & $\Delta$
& No Grid & With Grid & $\Delta$
& No Grid & With Grid & $\Delta$ \\

\midrule

\textbf{Gemini-2.5-Pro}
& 32.1 & \textbf{20.7} & \textcolor{ForestGreen}{-35.51\%}
& 64.5 & \textbf{30.9} & \textcolor{ForestGreen}{-52.09\%}
& 82.3 & \textbf{34.6} & \textcolor{ForestGreen}{-57.96\%}
& 61.1 & \textbf{29.5} & \textcolor{ForestGreen}{-51.72\%} \\

\textbf{Gemini-3-Pro}
& \textbf{2.4} & 23.2 & \textcolor{red}{+866.67\%}
& \textbf{5.0} & 31.9 & \textcolor{red}{+538.00\%}
& \textbf{4.7} & 65.1 & \textcolor{red}{+1285.11\%}
& \textbf{4.6} & 33.2 & \textcolor{red}{+621.74\%} \\

\textbf{GPT-5 (low)}
& 90.8 & \textbf{19.7} & \textcolor{ForestGreen}{-78.30\%}
& 109.4 & \textbf{37.7} & \textcolor{ForestGreen}{-65.54\%}
& 124.4 & \textbf{54.7} & \textcolor{ForestGreen}{-56.03\%}
& 110.5 & \textbf{40.1} & \textcolor{ForestGreen}{-63.71\%} \\

\textbf{Qwen3.5-397B-A17B}
& \textbf{18.4} & 20.7 & \textcolor{red}{+12.50\%}
& 44.7 & \textbf{27.8} & \textcolor{ForestGreen}{-37.81\%}
& 39.4 & \textbf{27.1} & \textcolor{ForestGreen}{-31.22\%}
& 34.2 & \textbf{26.5} & \textcolor{ForestGreen}{-22.51\%} \\

\textbf{Kimi K2.5}
& 22.1 & \textbf{18.3} & \textcolor{ForestGreen}{-17.19\%}
& 56.7 & \textbf{44.5} & \textcolor{ForestGreen}{-21.52\%}
& 111.9 & \textbf{45.7} & \textcolor{ForestGreen}{-59.16\%}
& 57.7 & \textbf{40.2} & \textcolor{ForestGreen}{-30.33\%} \\

\textbf{Kimi K3}
& 1.1 & -- & --
& 4.1 & -- & --
& 3.2 & -- & --
& 3.3 & -- & -- \\

\textbf{Gemma-4-31B}
& \textbf{5.8} & 20.5 & \textcolor{red}{+253.45\%}
& \textbf{24.7} & 34.1 & \textcolor{red}{+38.06\%}
& \textbf{27.7} & 39.7 & \textcolor{red}{+43.32\%}
& \textbf{23.7} & 33.9 & \textcolor{red}{+43.04\%} \\

\textbf{ViLaSR}
& 154.8 & -- & --
& 155.9 & -- & --
& 188.8 & -- & --
& 175.5 & -- & -- \\

\bottomrule
\end{tabular}
}

\captionof{figure}{\textbf{Connect-the-Dots median per-sample RMSE with
categories as columns.} Each entry shows the median per-sample RMSE without
and with the grid, and the percent change $\Delta$ (negative is better). \geminiproThree and \gemmafour perform better without the grid which we hypothesize about in \cref{fig:coord_1000_vs_2000}, while the models that are worse at localizing benefit from the grid, including \gptfive, \kimitwofive, and \qwenthreefive.}
\label{tab:connectdots_categories_mse}

\vspace{1em}
}]

\twocolumn[{%
\centering
\subsection{\drawingshapes}
\vspace{-0.5em}
 
\captionof{table}{Sketch-based localization improves accuracy for medium and large objects, while reducing performance on small objects, compared to coordinate-based bounding boxes (AP50).}
\label{tab:shape_ap50_gemini}
\begin{tabular}{lccc}
\toprule
& \textbf{\geminilogo} & \geminisketch & \geminisketch \\
\midrule
Output format & Bounding box & Rectangle & Oval \\
\midrule
AP50 (all)    & \textbf{63.1} & 58.8\,\decrease{4.3} & 55.4\,\decrease{7.7} \\
AP50 (small)  & \textbf{35.6} & 25.5\,\decrease{10.1} & 26.4\,\decrease{9.2} \\
AP50 (medium) & 62.6 & \textbf{63.2}\,\increase{0.6} & 55.4\,\decrease{7.2} \\
AP50 (large)  & 77.7 & \textbf{79.1}\,\increase{1.4} & 77.6\,\decrease{0.1} \\
\bottomrule
\end{tabular}
 
\vspace{1.5em}
 
\captionof{table}{\geminilogo and \geminilogo SketchVLM achieve the same precision; however, SketchVLM has lower recall than the baseline because it produces fewer true positive (TP) detections.}
\label{tab:total_detection_metrics}
\adjustbox{max width=\linewidth}{
\begin{tabular}{c|ccc|ccc|cc|cc|cc}
\hline
GT
& \multicolumn{3}{c|}{\geminilogo}
& \multicolumn{3}{c|}{\geminilogo SketchVLM}
& \multicolumn{2}{c|}{TP}
& \multicolumn{2}{c|}{FP}
& \multicolumn{2}{c}{FN} \\
& P & R & AP & P & R & AP
& \geminilogo & \geminisketch
& \geminilogo & \geminisketch
& \geminilogo & \geminisketch \\
\hline
1060
& 18.7 & 47.9 & 33.1
& 18.7 & 35.4 & 26.0
& \textbf{508} & 375
& 2204 & \textbf{1629}
& \textbf{552} & 685 \\
\hline
\end{tabular}
}
 
\vspace{1.5em}
 
\captionof{table}{Different prompt variations do not improve AP performance for the drawing shape task.}
\label{tab:drawing_shape_ablation}
\adjustbox{max width=\linewidth}{
\begin{tabular}{l|c|c|c|c|c}
\hline
 & \geminilogo & \geminilogo SketchVLM & \geminilogo SketchVLM & \geminilogo SketchVLM & \geminilogo SketchVLM \\
 &  & Draw rectangles & Draw all visible/part visible & Scan region before draw & Count object before draw \\
\hline
Output shape & Bounding box & Rectangles & Rectangles & Rectangles & Rectangles \\
\hline
AP50 (all)    & \textbf{63.1} & 58.8 & 57.1 & 59.0 & 56.2 \\
AP50 (small)  & \textbf{35.6} & 25.5 & 27.5 & 26.8 & 23.2 \\
AP50 (medium) & 62.6 & \textbf{63.2} & 60.7 & 59.9 & 58.9 \\
AP50 (large)  & 77.7 & 79.1 & 77.8 & \textbf{80.7} & 78.3 \\
\hline
\end{tabular}
}
\vspace{1em}



\captionof{table}{SketchVLM improves drawing-shape localization for \gemmafour, while Direct VQA performs better for Qwen3.5-397B and Kimi K2.5.}
\label{tab:drawing_shape_comparison}
\adjustbox{max width=\linewidth}{%
\begin{tabular}{l|c|c|c|c|c|c|c|c}
\hline
 & \gemmafour & \gemmafour 
 & Qwen3.5-397B & Qwen3.5-397B
 & Kimi K2.5 & Kimi K2.5
 & Kimi K3 & Kimi K3 \\
 & SketchVLM & Direct VQA
 & SketchVLM & Direct VQA
 & SketchVLM & Direct VQA
 & SketchVLM & Direct VQA \\
\hline
AP50 (all)    & 50.70 & 41.01 & 45.43 & 49.71 & 29.36 & 45.21 & 46.02 & \textbf{55.22} \\
AP50 (small)  & 12.09 & 12.64 & 11.68 & \textbf{19.31} & 4.52 & 10.44 & 8.96 & 15.55 \\
AP50 (medium) & \textbf{52.51} & 42.14 & 43.34 & 48.53 & 25.67 & 44.65 & 40.72 & 51.26 \\
AP50 (large)  & 75.78 & 53.35 & 73.03 & 67.40 & 51.99 & 65.71 & 71.05 & \textbf{76.84} \\
\hline
\end{tabular}%
}

\vspace{1em}
}]

\paragraph{Direct VQA is the stronger setup on Drawing Shape except for \gemmafour{}.}
We compare both setups on Drawing Shape across four open-source backbones
(\Cref{tab:drawing_shape_comparison}). SketchVLM helps only \gemmafour{},
raising AP50 by $9.69$ points overall and letting the smallest model outperform
Qwen3.5-397B and Kimi~K2.5 in both setups. For the larger backbones it is
harmful, costing $4.28$, $15.85$, and $9.20$ points on Qwen3.5-397B, Kimi~K2.5,
and Kimi~K3. The gain also depends on object scale: \gemmafour{} improves by
$22.43$ points on large objects but not on small ones ($-0.55$), and every
backbone loses accuracy on small objects under SketchVLM. Explicit intermediate
representations therefore substitute for capability the backbone lacks rather
than adding to strong backbones, and do not help at small object scales. The
best configuration remains Direct VQA with Kimi~K3 at $55.22$ AP50.

\twocolumn[{%
\centering
\subsection{\labeling}
\vspace{-0.5em}
 
\captionof{table}{Error type breakdown for labeling and part labeling with \geminilogo and \gptlogo original models and SketchVLM.}
\label{tab:appendix_labeling_error_breakdown}
\setlength{\tabcolsep}{5pt}
\begin{tabular}{l|cc|cc}
\hline
Wrong Type
& \geminilogo(\%) & \geminilogo SketchVLM(\%)
& \gptlogo(\%) & \gptlogo SketchVLM(\%) \\
\hline
Missing Label   & \textbf{10.8} & 3.6 & \textbf{0.69} & 0.32 \\
Wrong Position  & 25.1 & \textbf{36.1} & \textbf{80.17} & 79.32 \\
\hline
Total  & 35.9 & \textbf{39.7} & \textbf{80.86} & 79.64 \\
\hline
\end{tabular}
 
\vspace{1.75em}
 
\makebox[\linewidth][c]{%
  \begin{minipage}[t]{0.62\linewidth}
    \centering
    \vspace{0pt}
    \captionof{table}{Labels placed by SketchVLMs land very close to the correct location. \gptsketch is more accurate than the baseline at every tolerance level, and \geminisketch matches it within a few pixels.}
    \label{tab:dilation_accuracy}
    \setlength{\tabcolsep}{6pt}
    \adjustbox{max width=0.98\linewidth}{%
      \begin{tabular}{c|cc|cc}
      \toprule
      Dilation $r$ (px)
      & \geminilogo & \geminisketch
      & \gptlogo & \gptsketch  \\
      \midrule
      0  & \textbf{64.1} & 60.3 \decrease{3.8} & 19.1 & \textbf{20.4} \increase{1.3} \\
      3  & \textbf{69.0} & 66.5 \decrease{2.5} & 21.9 & \textbf{23.3} \increase{1.4} \\
      5  & \textbf{70.8} & 69.3 \decrease{1.5} & 23.9 & \textbf{25.3} \increase{1.4} \\
      7  & \textbf{71.8} & 71.2 \decrease{0.6} & 25.6 & \textbf{27.2} \increase{1.6} \\
      10 & 73.1 & \textbf{73.5} \increase{0.4} & 28.6 & \textbf{30.2} \increase{1.6} \\
      15 & 75.1 & \textbf{78.3} \increase{3.2} & 33.7 & \textbf{35.9} \increase{2.2} \\
      \bottomrule
      \end{tabular}%
    }
  \end{minipage}%
  \hspace{0.02\linewidth}%
  \begin{minipage}[t]{0.34\linewidth}
    \centering
    \vspace{0pt}
    \includegraphics[width=0.90\linewidth]{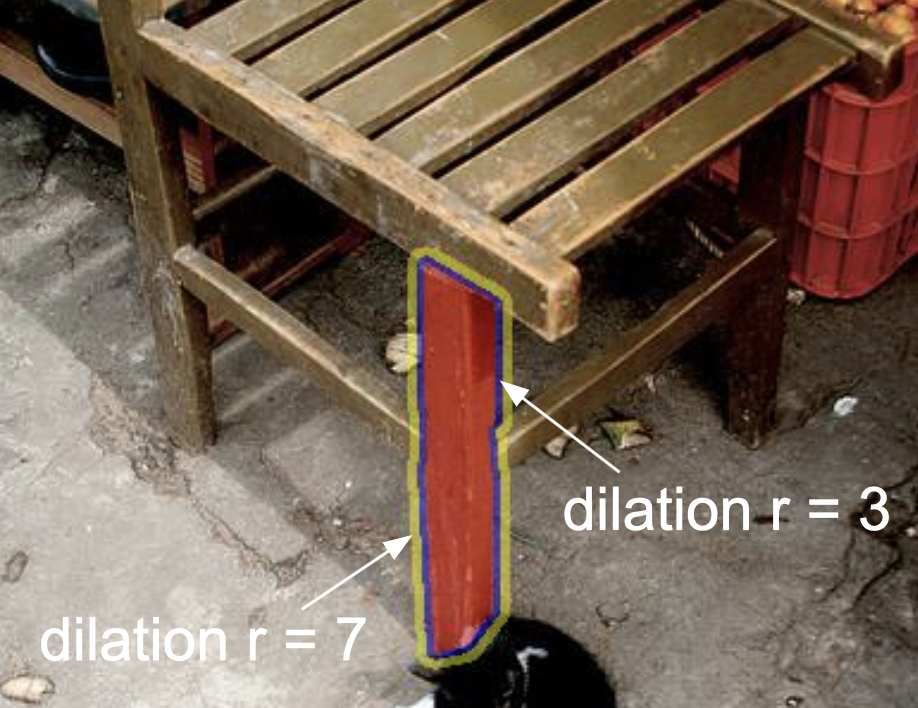}
    \vspace{2pt}
    \captionof{figure}{From $r=0$ (red) to modest dilation ($r=7$, yellow), small boundary offsets become visually negligible.}
    \label{fig:boundary_dilation}
  \end{minipage}%
}
\vspace{1em}

\makebox[\linewidth][c]{%
\begin{minipage}[t]{0.9\textwidth}
\centering
\vspace{0pt}





\captionof{table}{Part-labeling accuracy across different dilation radii for \gemmafour, Qwen3.5-397B, Kimi K2.5, and Kimi K3.}
\label{tab:part_labeling_comparison}

\setlength{\tabcolsep}{6pt}
\adjustbox{max width=0.98\linewidth}{%
  \begin{tabular}{c|cc|cc|cc|cc}
  \toprule
  Dilation $r$ (px)
  & \gemmafour & \gemmafour
  & Qwen3.5-397B & Qwen3.5-397B
  & Kimi K2.5 & Kimi K2.5
  & Kimi K3 & Kimi K3 \\
  & SketchVLM & Direct VQA
  & SketchVLM & Direct VQA
  & SketchVLM & Direct VQA
  & SketchVLM & Direct VQA \\
  \midrule
  0  & 48.31 & \textbf{50.12}
     & 38.53 & \textbf{46.44}
     & 23.31 & \textbf{42.94}
     & 45.46 & \textbf{51.34} \\

  3  & 54.19 & \textbf{55.30}
     & 44.43 & \textbf{50.85}
     & 26.84 & \textbf{47.64}
     & 51.13 & \textbf{56.99} \\

  5  & 57.74 & \textbf{57.97}
     & 48.49 & \textbf{52.83}
     & 29.44 & \textbf{50.20}
     & 54.47 & \textbf{60.32} \\

  7  & \textbf{60.44} & 59.91
     & 51.50 & \textbf{54.42}
     & 31.61 & \textbf{52.12}
     & 56.73 & \textbf{62.63} \\

  10 & \textbf{63.74} & 62.16
     & 56.10 & \textbf{56.29}
     & 35.10 & \textbf{54.42}
     & 60.07 & \textbf{66.10} \\

  15 & \textbf{67.90} & 64.70
     & \textbf{62.36} & 58.71
     & 40.79 & \textbf{57.61}
     & 63.80 & \textbf{70.13} \\
  \bottomrule
  \end{tabular}%
}

\end{minipage}%
}

\vspace{1em}

}]

\paragraph{Direct VQA is the stronger setup on Part Labeling.}
We compare both setups on Part Labeling across four open-source backbones at
dilation radii $r \in \{0,\dots,15\}$~px (\Cref{tab:part_labeling_comparison}).
Direct VQA is better in $20$ of $24$ settings, and Kimi~K3 with Direct VQA is
the best configuration at every radius, reaching $70.13$ at $r=15$. The deficit
of SketchVLM is largest for Kimi~K2.5 ($-19.6$ points at $r=0$) and smallest for
\gemmafour{} ($-1.8$), and it narrows as $r$ increases for \gemmafour{} and
Qwen3.5-397B while remaining constant for the two Kimi models. Unlike Counting
and Drawing Shape, Part Labeling shows no backbone for which SketchVLM is
clearly preferable.

\twocolumn[{%
\centering
\subsection{Additional Model Results}
\vspace{-0.5em}
 
\captionof{table}{Ablation across inputs in single-turn mode for additional
models. ``Sketch'' adds strokes/system prompt; ``Grid'' additionally overlays
the coordinate grid. Median per-sample RMSE is reported for \connectdots
while accuracy is reported for the other tasks. ``Order Accuracy'' is the
percentage of connect-the-dots samples with correct point ordering
(higher is better).}
\label{tab:ablation_other_models}

\setlength{\tabcolsep}{6pt}
\renewcommand{\arraystretch}{1.0}

\adjustbox{max width=0.95\linewidth}{%
\begin{tabular}{l l c c c c c}
\toprule
Model & Input & \vpct & \balllogo & \mazelogo
& \connectlogo & Order Accuracy \\
\midrule

\multirow{4}{*}{Kimi K2.5}
& Image
& 44.0 & 50.50 & \textbf{93.5} & -- & -- \\
& + Grid
& 55.0 & \textbf{58.08} & 97.0 & -- & -- \\
& + Sketch
& \textbf{56.0} & 48.48 & 88.00 & 57.7 & 59\% \\
& + Sketch + Grid
& \textbf{56.0} & 50.50 & 89.75 & \textbf{40.2} & 67\% \\

\midrule

\multirow{4}{*}{Gemma-4-31B}
& Image
& 38.0 & 38.90 & 90.5 & -- & -- \\
& + Grid
& 44.0 & \textbf{44.40} & 86.0 & -- & -- \\
& + Sketch
& 46.0 & 40.90 & 89.0 & \textbf{23.7} & 56\% \\
& + Sketch + Grid
& \textbf{51.0} & 39.40 & \textbf{93.8} & 33.9 & \textbf{62\%} \\

\midrule

\multirow{4}{*}{Qwen3.5-397B-A17B}
& Image
& \textbf{83.00} & \textbf{69.19} & 96.75 & -- & -- \\
& + Grid
& 76.00 & 67.67 & \textbf{97.50} & -- & -- \\
& + Sketch
& 81.00 & 61.11 & 96.00 & 34.2 & 64.00\% \\
& + Sketch + Grid
& 79.00 & 61.62 & 95.75 & \textbf{26.5} & \textbf{87.00\%} \\

\midrule

\multirow{2}{*}{Kimi K3}
& Image
& \textbf{95.0} & \textbf{85.35} & \textbf{100} & -- & -- \\
& + Sketch
& 92.0 & 79.29 & 99.75 & \textbf{3.3} & 97.62\% \\

\bottomrule
\end{tabular}%
}
 
\vspace{1.5em}
 
\captionof{table}{Average number of turns per task group in multi-turn evaluation.}
\label{tab:avg_turns}
\begin{tabular}{lcc}
\toprule
\textbf{Task} & \textbf{Gemini-3-Pro} & \textbf{GPT-5 (low)} \\
\midrule
VPCT       & 5.10  & 6.18  \\
\balllogo       & 5.00  & 2.58  \\
\connectlogo    & 16.61 & 19.00 \\
\mazelogo  & 3.42  & 4.29  \\
\midrule
\textbf{Overall} & \textbf{5.91} & \textbf{5.92} \\
\bottomrule
\end{tabular}
\vspace{1em}
}]

\clearpage

\twocolumn[{%
\centering

\captionof{table}{
Accuracy comparison between SketchVLM and Direct VQA across seven tasks
using \gemmafour, Qwen3.5-397B, Kimi K2.5, and Kimi K3. Drawing Shape
reports AP50 (all) for rectangle localization, Part Labeling reports
accuracy at dilation radius $r=0$, and Connecting Dots reports median
per-sample RMSE (lower is better).
}
\label{tab:seven_dataset_comparison}

\vspace{0.5em}

\small
\setlength{\tabcolsep}{5pt}

\adjustbox{max width=\textwidth}{%
\begin{tabular}{ll|ccccccc}
\toprule

Model & Setup
& VPCT
& \balllogo
& \mazelogo
& \countlogo
& \labellogo
& \shapeslogo
& \connectlogo \\

&
& \scriptsize Video Physics
& \scriptsize Ball Drop
& \scriptsize Maze
& \scriptsize Counting
& \scriptsize Part Labeling
& \scriptsize Drawing Shape
& \scriptsize Connecting Dots \\

&
& Acc. (\%)
& Acc. (\%)
& Acc. (\%)
& Acc. (\%)
& Acc. (\%)
& AP50 (\%)
& Median RMSE $\downarrow$ \\

\midrule

\multirow{2}{*}{Gemma4-31B}
& SketchVLM
& 46.00
& 40.90
& 89.00
& 55.11
& 48.31
& 50.70
& 23.7 \\

& Direct VQA
& 38.00
& 38.90
& 90.50
& 36.33
& 50.12
& 41.01
& -- \\

\midrule

\multirow{2}{*}{Qwen3.5-397B}
& SketchVLM
& 81.00
& 61.11
& 96.00
& 58.15
& 38.53
& 45.43
& 34.2 \\

& Direct VQA
& 83.00
& 69.19
& 96.75
& 74.31
& 46.44
& 49.71
& -- \\

\midrule

\multirow{2}{*}{Kimi K2.5}
& SketchVLM
& 56.00
& 48.48
& 88.00
& 63.26
& 23.31
& 29.36
& 57.7 \\

& Direct VQA
& 44.00
& 50.50
& 93.50
& 66.85
& 42.94
& 45.21
& -- \\

\midrule

\multirow{2}{*}{Kimi K3}
& SketchVLM
& 92.00
& 79.29
& 99.75
& 93.65
& 45.46
& 46.02
& 3.3 \\

& Direct VQA
& 95.00
& 85.35
& 100.00
& 86.19
& 51.34
& 55.22
& -- \\

\bottomrule
\end{tabular}%
}

\vspace{3pt}

\parbox{\textwidth}{\raggedright\scriptsize
\textit{Bold indicates the best result in each column. Higher is better for
all metrics except Connecting Dots median RMSE ($\downarrow$). Drawing Shape
uses AP50 (all) for rectangle localization, and Part Labeling uses dilation
radius $r=0$.}
}

\vspace{1em}
}]
\paragraph{Generalization to open-source models.}
Since our main results use closed-source models, we further evaluate
SketchVLM with four open-source models to assess whether its benefits
generalize across different backbones. Among them, Kimi K3 achieves the
strongest overall performance. The benefit of SketchVLM is task-dependent.
On Counting, SketchVLM improves over Direct VQA by 18.8 points with
Gemma4-31B and 7.5 points with Kimi K3. It also enables Connecting Dots,
which Direct VQA cannot address because it does not produce the required
coordinate-based outputs. In contrast, Direct VQA performs better on Part
Labeling across all four models and on Drawing Shape for all but
\gemmafour. Overall, these results show that SketchVLM generalizes to
open-source backbones, with the largest gains on tasks that benefit from
explicit intermediate visual representations.

\clearpage

\twocolumn[{%
\centering
\subsection{Latency and Cost}
\label{sec:latency_cost}
\vspace{-0.5em}

\captionof{table}{Average latency and cost per item for Direct VQA and Sketch + Grid across different models on VPCT and Maze Navigation.}
\label{tab:latency_cost_comparison}
\adjustbox{max width=\linewidth}{%
\begin{tabular}{l|l|c|c}
\toprule
Model & Method & Avg. Latency (s/item) & Avg. Cost (\$/item) \\
\midrule

GPT-5
& Direct VQA
& $24.87 \pm 20.52$
& $0.014225 \pm 0.005350$ \\
& Sketch + Grid
& $31.24 \pm 30.67$
& $0.021859 \pm 0.006280$ \\
\midrule

Gemini-3.1-Pro
& Direct VQA
& $27.13 \pm 16.94$
& $0.044237 \pm 0.038279$ \\
& Sketch + Grid
& $14.33 \pm 4.81$
& $0.029351 \pm 0.006184$ \\
\midrule

Gemma-4-31B
& Direct VQA
& $9.59 \pm 6.98$
& $0.000399 \pm 0.000069$ \\
& Sketch + Grid
& $10.11 \pm 6.24$
& $0.000397 \pm 0.000070$ \\
\midrule

Qwen3.5-397B
& Direct VQA
& $229.6 \pm 203.8$
& $0.03461 \pm 0.02075$ \\
& Sketch + Grid
& $212.1 \pm 110.4$
& $0.03685 \pm 0.02083$ \\

\bottomrule
\end{tabular}%
}

\vspace{1em}

\centering
\caption{Average latency and running cost across Counting, Drawing Shape, and Part Labeling tasks.}
\label{tab:latency_cost_comparison_2}
\adjustbox{max width=\textwidth}{%
\begin{tabular}{ll|l|cc}
\hline
\textbf{Task} & \textbf{Model} & \textbf{Setup} &
\textbf{Avg. Latency (s/item)} & \textbf{Avg. Cost (\$/item)} \\
\hline

\multirow{6}{*}{Counting}
& \multirow{2}{*}{Gemma4-31B}
& SketchVLM  & $1.874 \pm 1.18$ & $0.000730 \pm 0.00$ \\
& & Direct VQA & $0.616 \pm 0.52$ & $0.000064 \pm 0.00$ \\
\cline{2-5}

& \multirow{2}{*}{Qwen3.5-397B}
& SketchVLM  & $24.916 \pm 28.93$ & $0.008143 \pm 0.01$ \\
& & Direct VQA & $26.273 \pm 59.52$ & $0.006952 \pm 0.02$ \\
\cline{2-5}

& \multirow{2}{*}{Kimi K2.5}
& SketchVLM  & $14.659 \pm 19.08$ & $0.003915 \pm 0.00$ \\
& & Direct VQA & $5.611 \pm 13.17$ & $0.001291 \pm 0.00$ \\

\hline

\multirow{6}{*}{Drawing Shape}
& \multirow{2}{*}{Gemma4-31B}
& SketchVLM  & $3.434 \pm 13.74$ & $0.000936 \pm 0.00$ \\
& & Direct VQA & $0.885 \pm 1.00$ & $0.000198 \pm 0.00$ \\
\cline{2-5}

& \multirow{2}{*}{Qwen3.5-397B}
& SketchVLM  & $36.307 \pm 41.85$ & $0.011462 \pm 0.01$ \\
& & Direct VQA & $9.676 \pm 25.00$ & $0.002588 \pm 0.01$ \\
\cline{2-5}

& \multirow{2}{*}{Kimi K2.5}
& SketchVLM  & $34.220 \pm 26.93$ & $0.008392 \pm 0.01$ \\
& & Direct VQA & $36.106 \pm 31.54$ & $0.007568 \pm 0.01$ \\

\hline

\multirow{6}{*}{Part Labeling}
& \multirow{2}{*}{Gemma4-31B}
& SketchVLM  & $2.556 \pm 1.46$ & $0.000878 \pm 0.00$ \\
& & Direct VQA & $1.085 \pm 0.66$ & $0.000190 \pm 0.00$ \\
\cline{2-5}

& \multirow{2}{*}{Qwen3.5-397B}
& SketchVLM  & $55.898 \pm 32.03$ & $0.018339 \pm 0.01$ \\
& & Direct VQA & $18.406 \pm 33.30$ & $0.004773 \pm 0.01$ \\
\cline{2-5}

& \multirow{2}{*}{Kimi K2.5}
& SketchVLM  & $47.406 \pm 24.82$ & $0.011871 \pm 0.00$ \\
& & Direct VQA & $55.227 \pm 36.94$ & $0.011977 \pm 0.01$ \\

\hline
\end{tabular}%
}

\vspace{1em}

\centering
\captionof{table}{Average latency and cost of Kimi K3 with SketchVLM and Direct VQA across seven datasets. Lower is better for both metrics.}
\label{tab:kimi_k3_latency_cost}

\adjustbox{max width=\textwidth}{%
\begin{tabular}{llccccccc}
\toprule
\textbf{Metric}
& \textbf{Setup}
& \textbf{VPCT}
& \balllogo
& \mazelogo
& \countlogo
& \labellogo
& \shapeslogo
& \connectlogo \\
&
& \scriptsize Video Physics
& \scriptsize Ball Drop
& \scriptsize Maze
& \scriptsize Counting
& \scriptsize Part Labeling
& \scriptsize Drawing Shape
& \scriptsize Connecting Dots \\
\midrule

\multirow{2}{*}{\textbf{Latency (s)} $\downarrow$}
& SketchVLM
& $155.47 \pm 85.09$
& $\mathbf{112.26 \pm 87.93}$
& $60.25 \pm 31.07$
& $40.31 \pm 25.89$
& $54.00 \pm 36.25$
& $24.47 \pm 19.75$
& $40.32 \pm 25.48$ \\

& Direct VQA
& $\mathbf{101.33 \pm 47.19}$
& $124.55 \pm 73.83$
& $\mathbf{23.60 \pm 10.35}$
& $\mathbf{6.99 \pm 8.62}$
& $\mathbf{14.45 \pm 10.02}$
& $\mathbf{9.20 \pm 6.80}$
& -- \\
\midrule

\multirow{2}{*}{\textbf{Cost (\$)} $\downarrow$}
& SketchVLM
& $\mathbf{0.13 \pm 0.05}$
& $\mathbf{0.10 \pm 0.05}$
& $0.05 \pm 0.01$
& $0.03 \pm 0.01$
& $0.04 \pm 0.01$
& $0.02 \pm 0.02$
& $0.04 \pm 0.02$ \\

& Direct VQA
& $0.15 \pm 0.06$
& $0.15 \pm 0.08$
& $\mathbf{0.03 \pm 0.01}$
& $\mathbf{0.01 \pm 0.01}$
& $\mathbf{0.02 \pm 0.01}$
& $\mathbf{0.01 \pm 0.01}$
& -- \\

\bottomrule
\end{tabular}%
}
}]

\paragraph{Overhead of SketchVLM.}
SketchVLM issues additional generation calls, so we measure per-item latency and
cost against Direct VQA under the same model and task set. On tasks where a
Direct VQA baseline exists, SketchVLM is uniformly slower, by $1.5\times$ on
VPCT up to $5.8\times$ on Counting, since the sketch is produced before the
answer. Cost, however, does not follow latency. On the short tasks the sketching
call dominates and cost increases by $1.5$--$3\times$, whereas on the VPCT and Ball Drop
tasks SketchVLM is cheaper than Direct VQA despite comparable latency
($\$0.125$ vs.\ $\$0.154$ on VPCT and $\$0.098$ vs.\ $\$0.149$ on Ball Drop).
Connecting Dots is excluded from this comparison, as Direct VQA does not produce
the required coordinate outputs. The overhead is therefore concentrated on short
tasks with small input contexts, and is smallest, or negative in cost, on tasks
with large visual inputs.

\clearpage

\twocolumn[{%
\centering
\subsection{Ablation Study of Kimi K3 with Different Reasoning Levels}
\captionof{table}{
Comparison of Kimi K3 across different reasoning levels.
Bold indicates the best result for each metric.
$\uparrow$ indicates higher is better, while $\downarrow$ indicates lower is better.
Cost and latency are averaged per sample.
}
\label{tab:kimi_k3_reasoning_levels}

\vspace{0.5em}

\adjustbox{max width=\textwidth}{%
\begin{tabular}{l|l|cccc}
\toprule

\textbf{Metric}
& \textbf{Reasoning Level}
& VPCT
& \labellogo
& \shapeslogo
& \connectlogo \\

&
& \scriptsize Video Physics
& \scriptsize Part Labeling
& \scriptsize Drawing Shape
& \scriptsize Connecting Dots \\

&
& Acc. (\%) $\uparrow$
& Acc. (\%) $\uparrow$
& AP50 (\%) $\uparrow$
& RMSE $\downarrow$ \\

\midrule


\multirow{3}{*}{\textbf{Performance}}
& Low
& 86.00
& 39.29
& 55.51
& 6.6 \\

& High
& \textbf{94.00}
& 45.81
& 57.08
& \textbf{4.1} \\

& Max
& 91.80
& \textbf{46.93}
& \textbf{57.70}
& 4.2 \\

\midrule


\multirow{3}{*}{\textbf{Cost (\$)} $\downarrow$}
& Low
& \textbf{0.044}
& \textbf{0.017}
& \textbf{0.022}
& \textbf{0.025} \\

& High
& 0.123
& 0.030
& 0.030
& 0.042 \\

& Max
& 0.227
& 0.035
& 0.082
& 0.086 \\

\midrule


\multirow{3}{*}{\textbf{Latency (s)} $\downarrow$}
& Low
& \textbf{52.648}
& \textbf{13.190}
& \textbf{11.015}
& \textbf{19.570} \\

& High
& 140.813
& 19.070
& 16.185
& 39.600 \\

& Max
& 208.245
& 18.794
& 69.265
& 86.230 \\

\bottomrule
\end{tabular}%
}

\vspace{1em}
}]

\paragraph{Accuracy--cost trade-off across reasoning levels.}
Reasoning models expose an inference-time budget that increases accuracy at the
expense of cost and latency, so the practical question is not which level is
most accurate but which level offers the best trade-off. We therefore evaluate
Kimi~K3 at three reasoning levels (Low, High, and Max), keeping the model,
prompts, and evaluation protocol fixed, and report per-sample cost and
end-to-end latency alongside task performance. Low is clearly insufficient,
trailing High by 8.0 points on VPCT and 6.5 points on Part Labeling. High
matches or outperforms Max across all four tasks: it is within 1.1 points on
Part Labeling and 0.6 points on Drawing Shape, and is in fact 2.2 points
better on VPCT and achieves a lower median per-sample RMSE on Connecting Dots
(4.1 versus 4.2, worksheets subset only), while requiring up to $2.7\times$
less cost and $4.3\times$ less latency. Unlike the other tasks, additional
reasoning budget does not help fine-grained coordinate prediction: Max's
median RMSE is no better than High's, despite roughly $2\times$ the cost and
latency on Connecting Dots specifically. Overall, High offers the best
accuracy--cost trade-off across all four tasks, and we adopt it as our
default; none of the tasks in this ablation justify the added expense of Max.

\clearpage
\twocolumn[{%
\centering
\section{Qualitative Samples}
\label{sec:qual-samples}

\begin{minipage}[t]{0.24\linewidth}
  \centering
  \vspace{0pt}
  \parbox[t][1.0em][t]{\linewidth}{\centering Input}\\[0.01ex]
  \includegraphics[width=\linewidth]{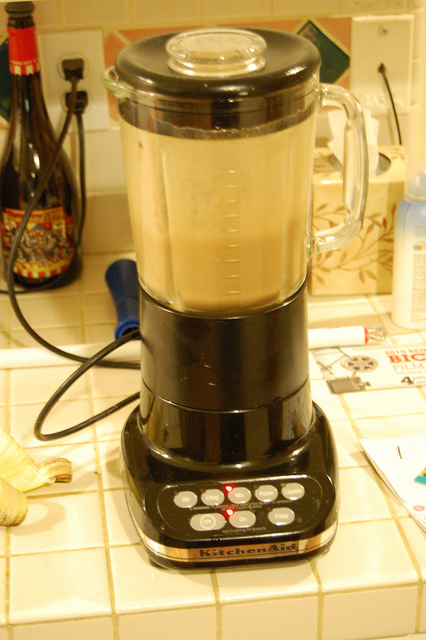}
\end{minipage}\hfill
\begin{minipage}[t]{0.24\linewidth}
  \centering
  \vspace{0pt}
  \parbox[t][1.0em][t]{\linewidth}{\centering (a) \nanologo \nanobananapro}\\[0.1ex]
  \includegraphics[width=\linewidth]{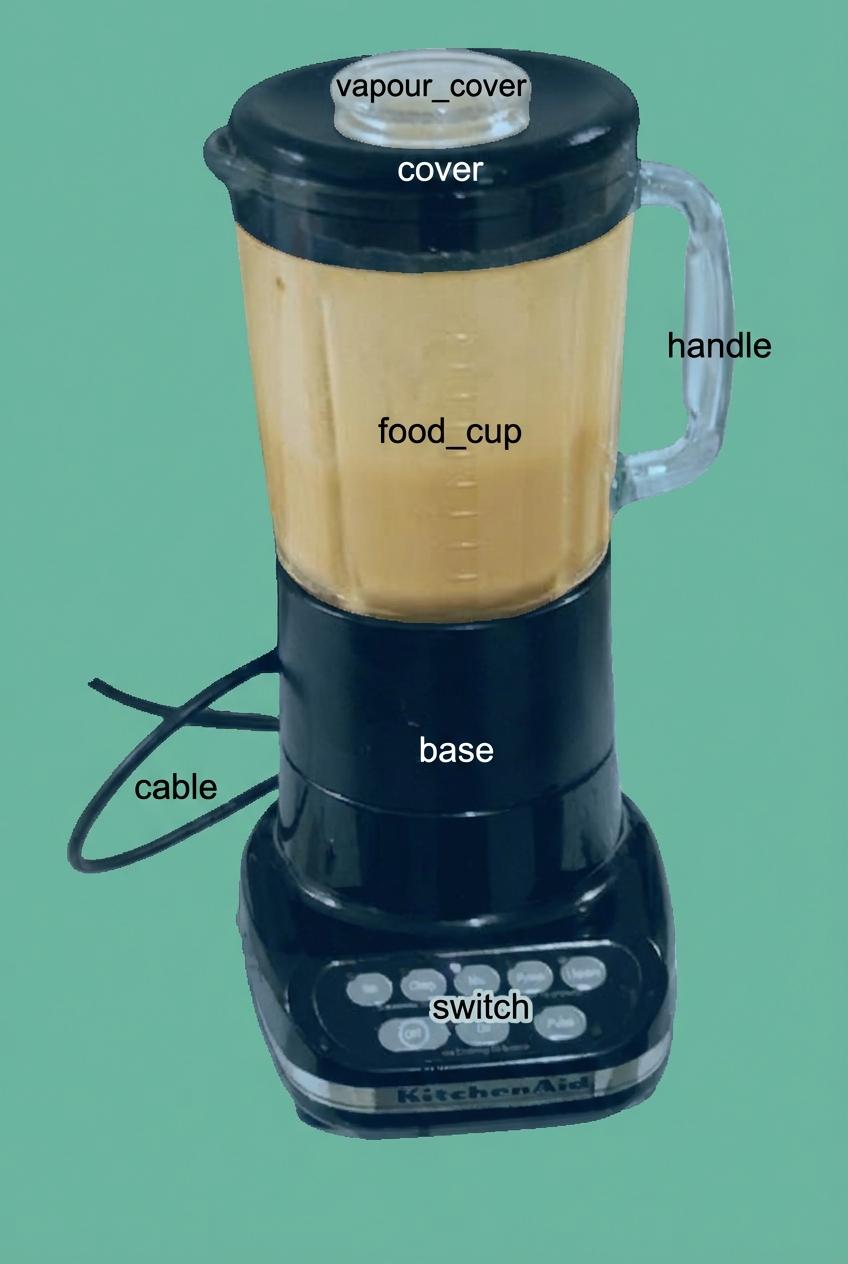}
\end{minipage}\hfill
\begin{minipage}[t]{0.24\linewidth}
  \centering
  \vspace{0pt}
  \parbox[t][1.0em][t]{\linewidth}{\centering (b) \geminisketch}\\[0.1ex]
  \includegraphics[width=\linewidth]{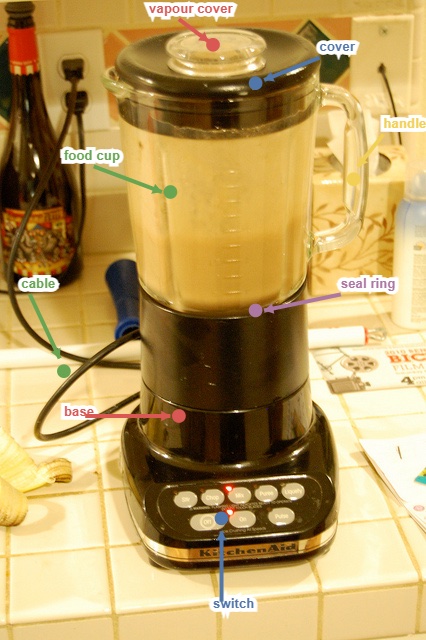}
\end{minipage}\hfill
\begin{minipage}[t]{0.24\linewidth}
  \centering
  \vspace{0pt}
  \parbox[t][1.0em][t]{\linewidth}{\centering (c) \geminilogo}\\[0.1ex]
  \includegraphics[width=\linewidth]{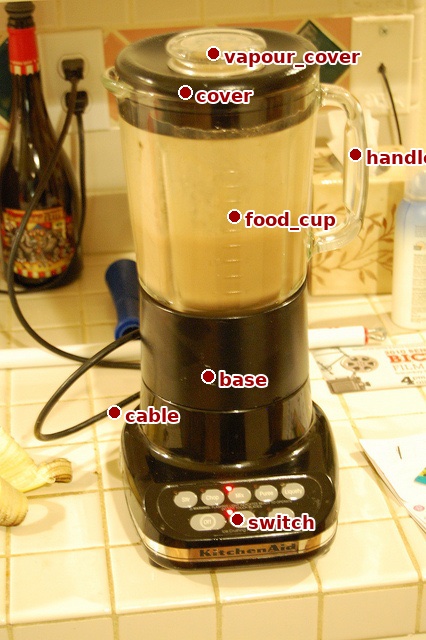}
\end{minipage}
 
\captionof{figure}{%
  Qualitative comparison on the part labeling task.
  (b) \geminisketch places each part label directly on its corresponding region while preserving the original image, producing more interpretable part annotations than (a) \nanologo \nanobananapro or (c) \geminilogo \geminiproThree.}
\label{fig:part_labeling_task_example_blender}
 
\vspace{0.5em}
 
\newlength{\imgheight}
\setlength{\imgheight}{2.6cm}
 
\begin{minipage}[t]{0.19\linewidth}
  \centering
  \vspace{0pt}%
  \raisebox{0pt}[\heightof{\geminisketch}][0pt]{\small Input}\\[1ex]
  \includegraphics[width=\linewidth,height=\imgheight,keepaspectratio]{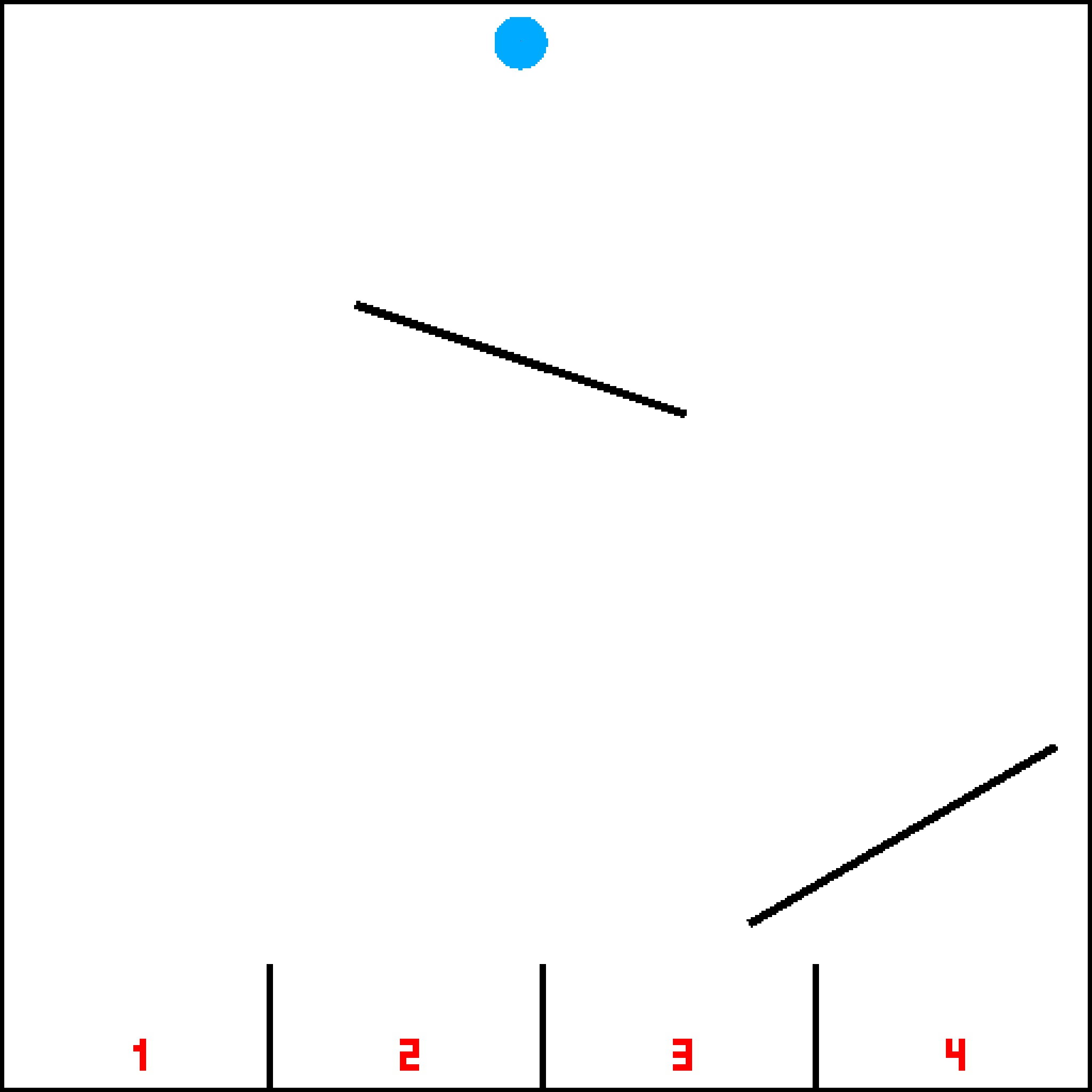}%
\end{minipage}\hfill%
\begin{minipage}[t]{0.19\linewidth}
  \centering
  \vspace{0pt}%
  \geminisketch \\[1ex]
  \includegraphics[width=\linewidth,height=\imgheight,keepaspectratio]{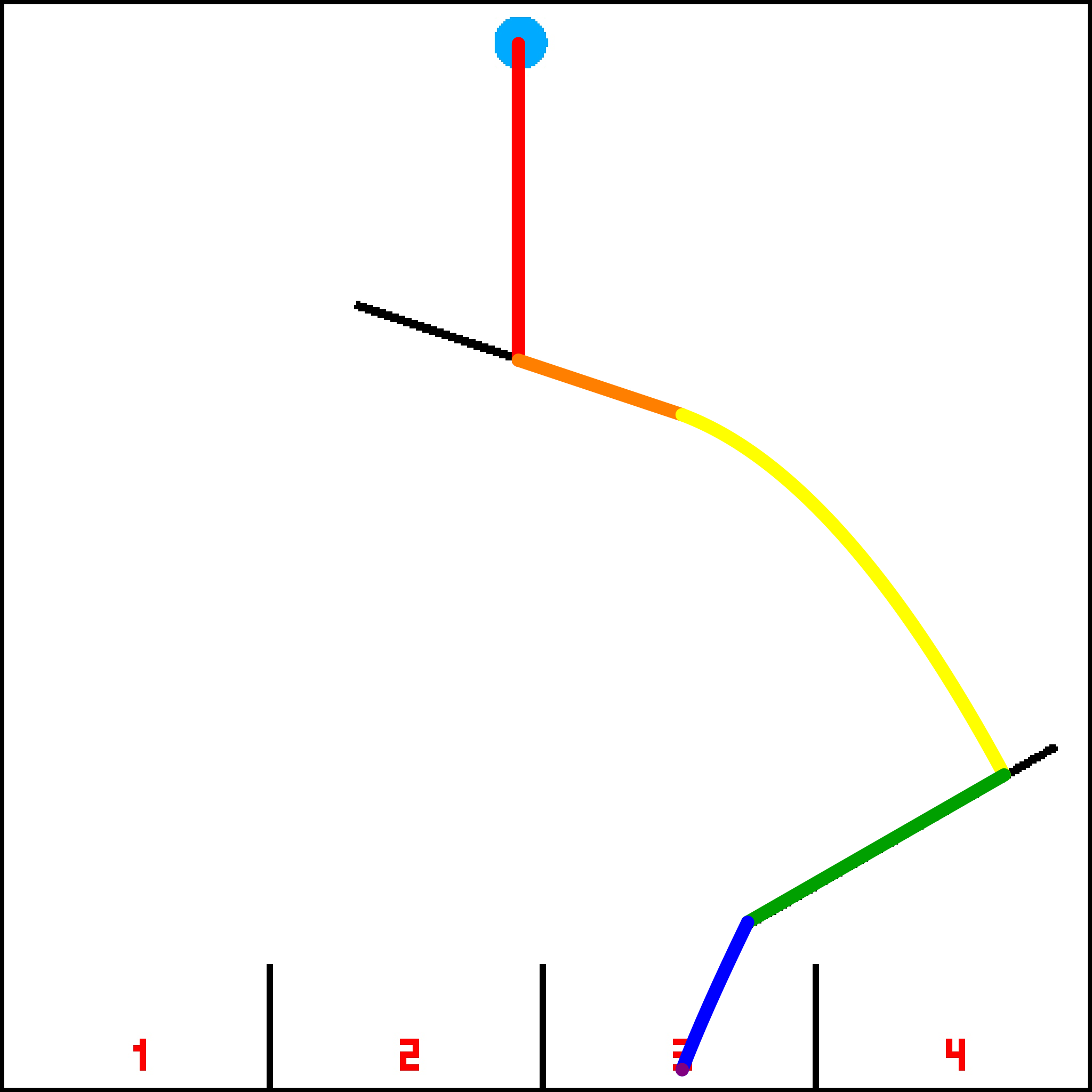}%
\end{minipage}\hfill%
\begin{minipage}[t]{0.19\linewidth}
  \centering
  \vspace{0pt}%
  \raisebox{0pt}[\heightof{\geminisketch}][0pt]{\nanologo \nanobananapro}\\[1ex]
  \includegraphics[width=\linewidth,height=\imgheight,keepaspectratio]{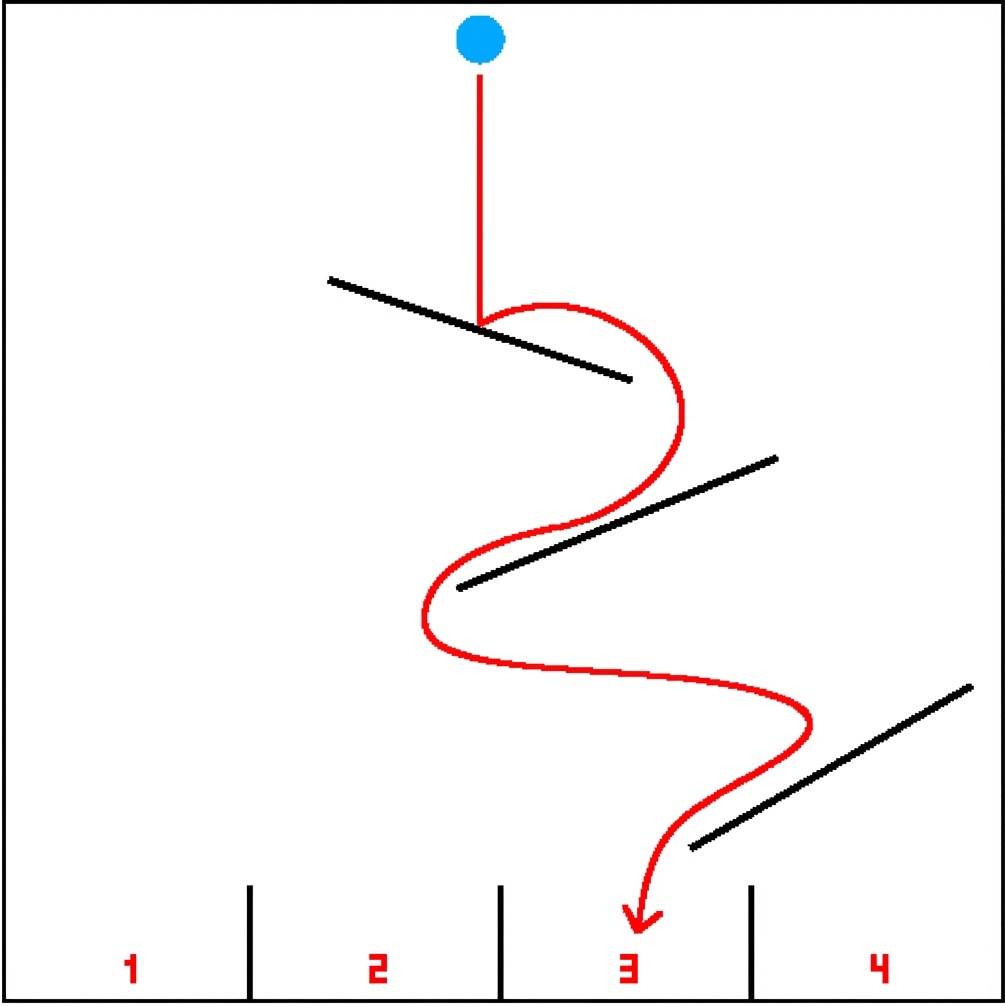}%
\end{minipage}\hfill%
\begin{minipage}[t]{0.19\linewidth}
  \centering
  \vspace{0pt}%
  \raisebox{0pt}[\heightof{\geminisketch}][0pt]{\thinkmorphlogo ThinkMorph}\\[1ex]
  \includegraphics[width=\linewidth,height=\imgheight,keepaspectratio]{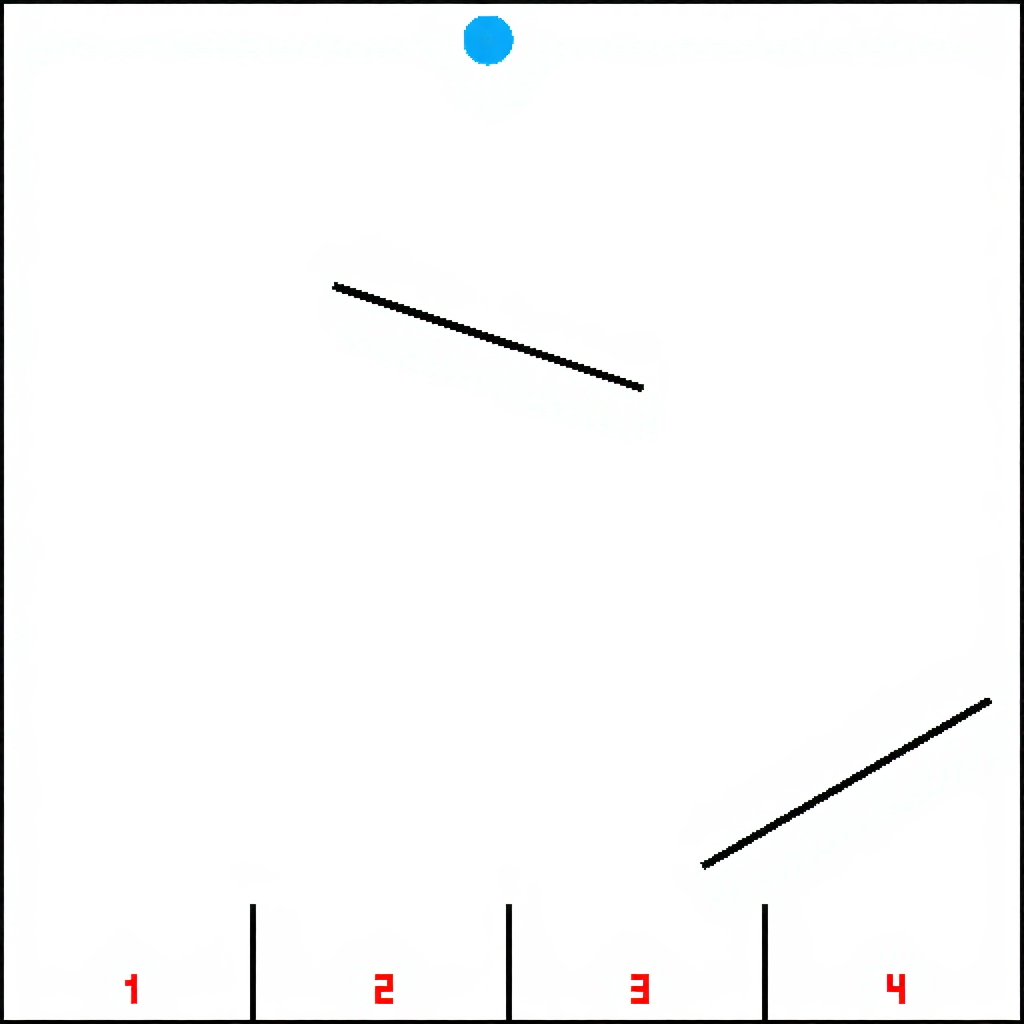}%
\end{minipage}\hfill%
\begin{minipage}[t]{0.19\linewidth}
  \centering
  \vspace{0pt}%
  \raisebox{0pt}[\heightof{\geminisketch}][0pt]{\vilasrlogo ViLaSR}\\[1ex]
  \includegraphics[width=\linewidth,height=\imgheight,keepaspectratio]{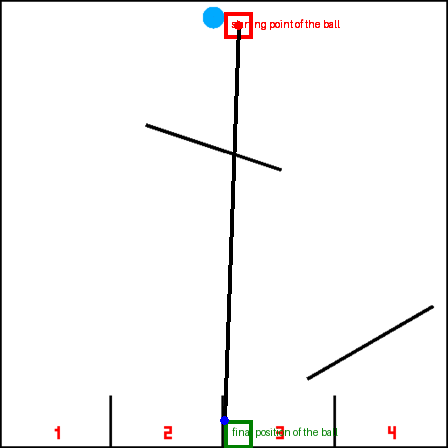}%
\end{minipage}
 
\captionof{figure}{\SketchVLM generates the most accurate \balldrop images compared to other baselines.}
\label{fig:ball_drop_examples}
}]

\twocolumn[{%
\centering
\subsection{Ball Drop}
\label{sec:qual-physics}
\vspace{0.3em}

\includegraphics[width=\textwidth]{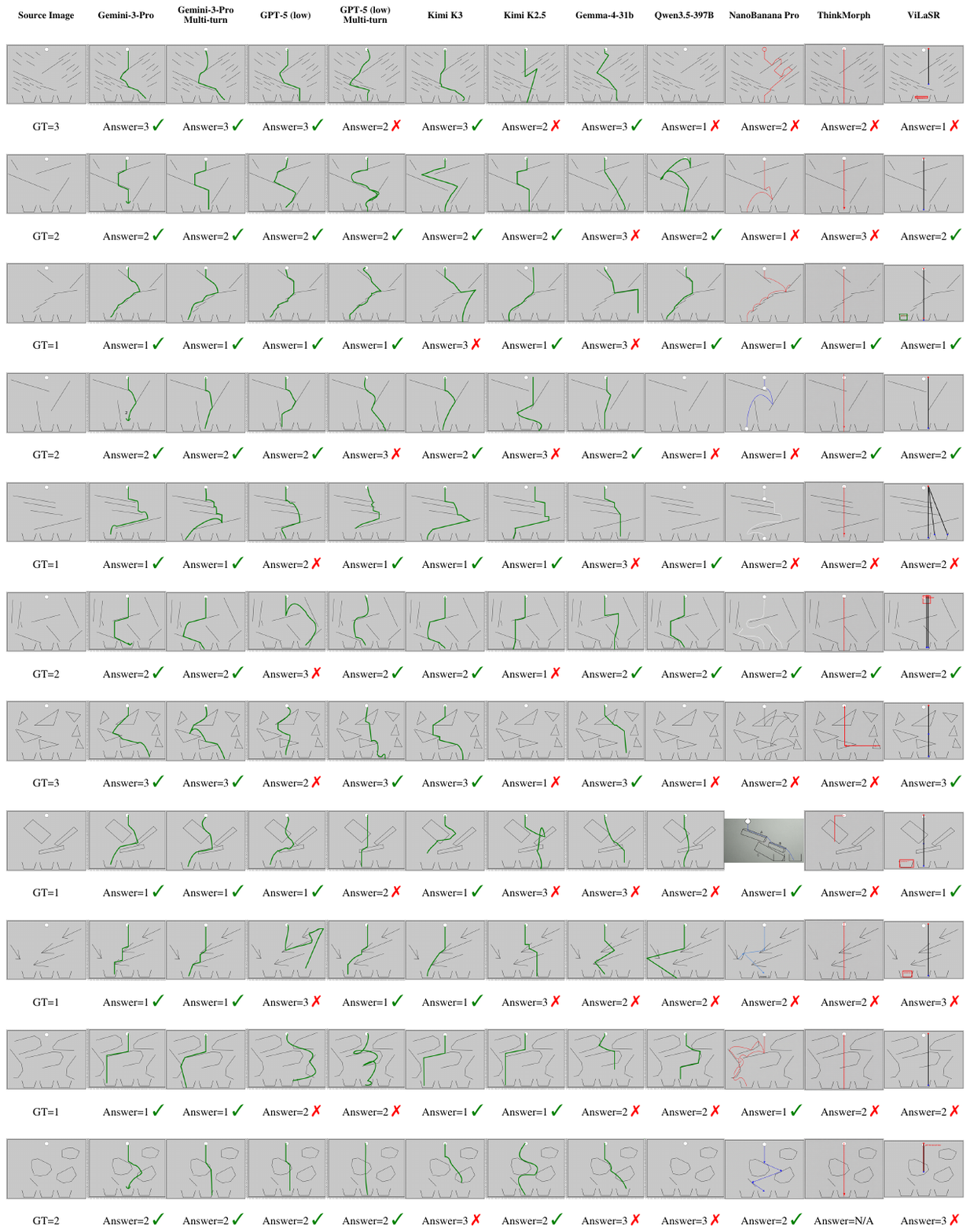}

\captionof{figure}{Qualitative examples of models on VPCT. Gemini-3-Pro-Preview in single-turn produces the most accurate annotations, while NanoBanana Pro, ThinkMorph and ViLaSR often draw paths that cross walls and provide the wrong answer.}
\label{fig:vpct_qual_ex}
}]

\begin{figure*}[t]
  \centering
  \includegraphics[width=\textwidth]{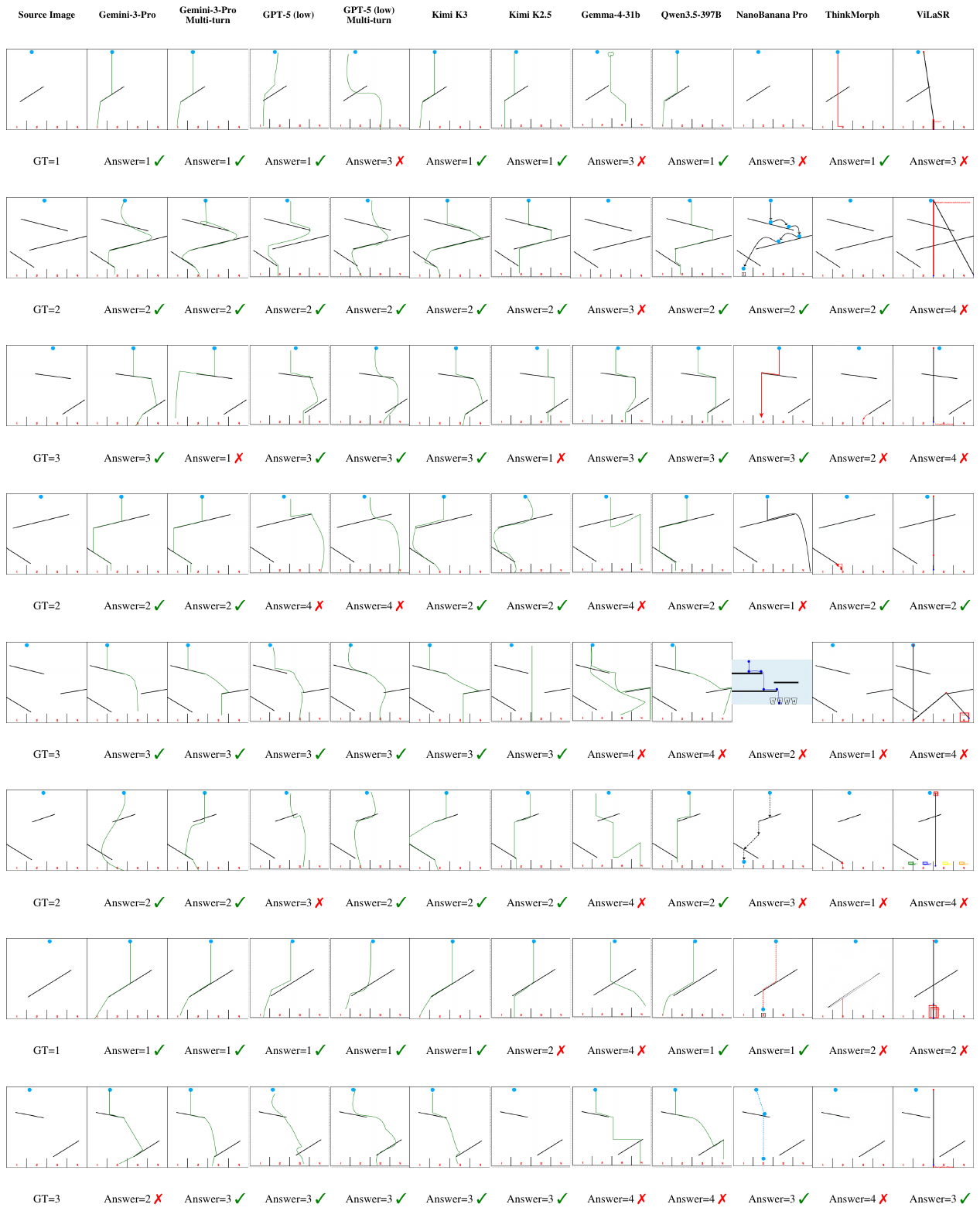}
  \caption{Qualitative examples of models on our Ball Drop dataset. Gemini-3-Pro-Preview in single-turn produces the most accurate annotations, while NanoBanana Pro, ThinkMorph and ViLaSR often draw paths that cross walls and provide the wrong answer.}
  \label{fig:ball_drop_qual_ex}
\end{figure*}

\begin{figure*}[t]
  \centering
  \includegraphics[width=0.3\linewidth]{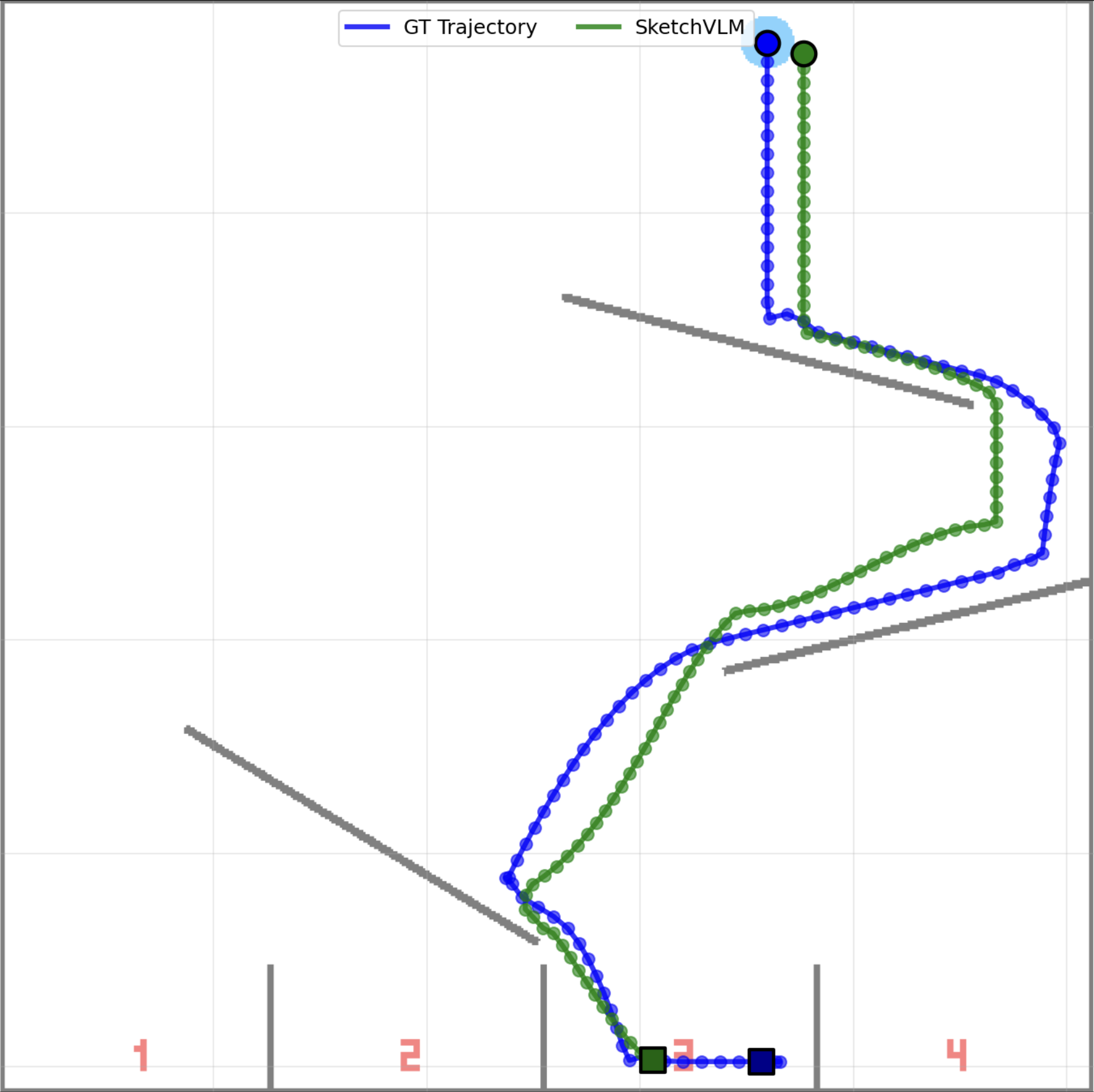}
  \caption{The \SketchVLM framework boosts VLM accuracy on the Ball Drop task while also letting them draw ball trajectory paths that closely simulate the ground truth data.}
  \label{fig:vpct_accuracy_mse}
\end{figure*}

\clearpage

\twocolumn[{%
\centering
\subsection{\maze}
\label{app:qual_maze}
\vspace{0.3em}

\includegraphics[width=\textwidth]{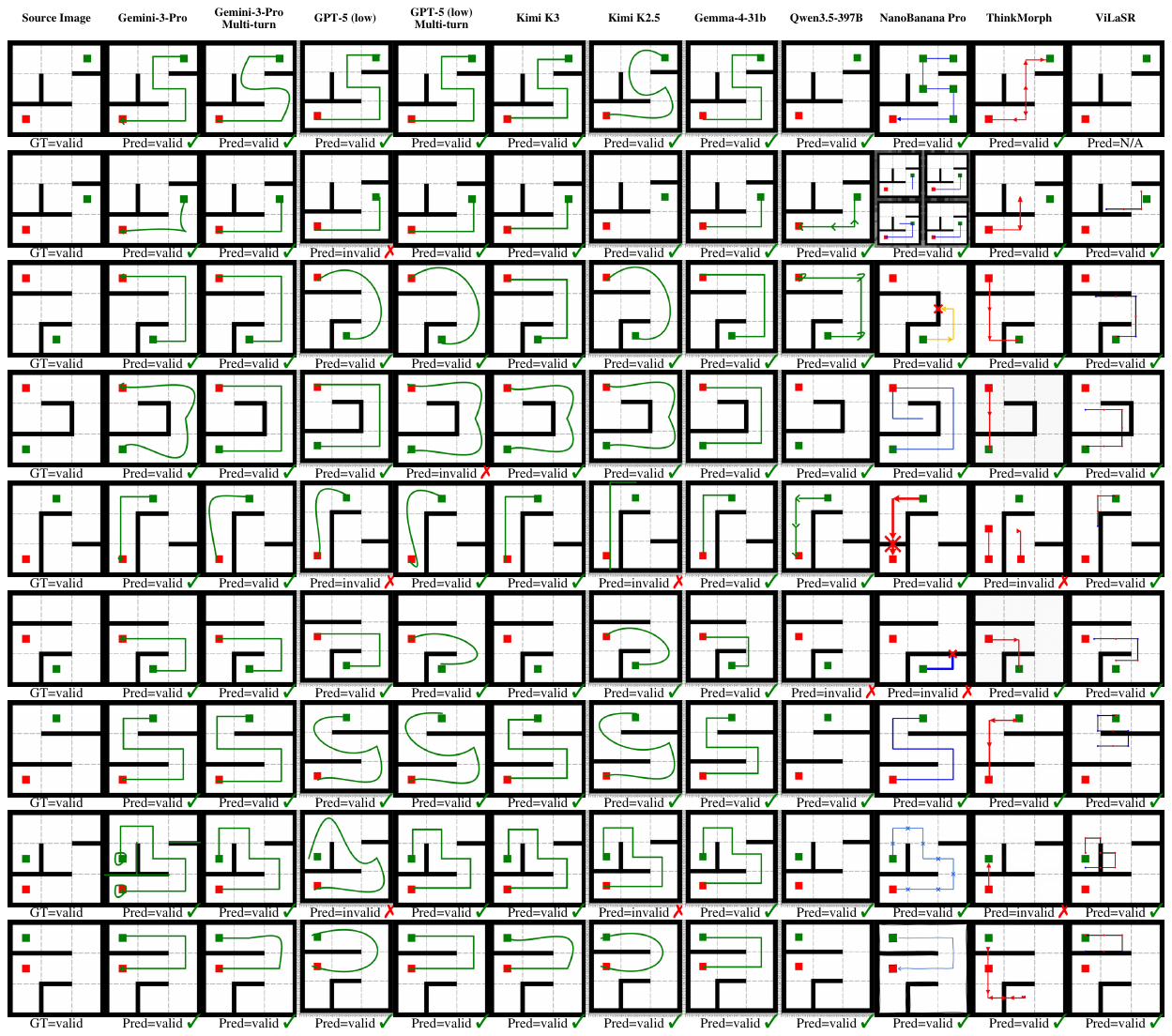}

\captionof{figure}{Qualitative examples of models on valid paths in \maze.}
\label{fig:maze_valid_qual_ex}
}]

\begin{figure*}[t]
  \centering
  \includegraphics[width=\textwidth]{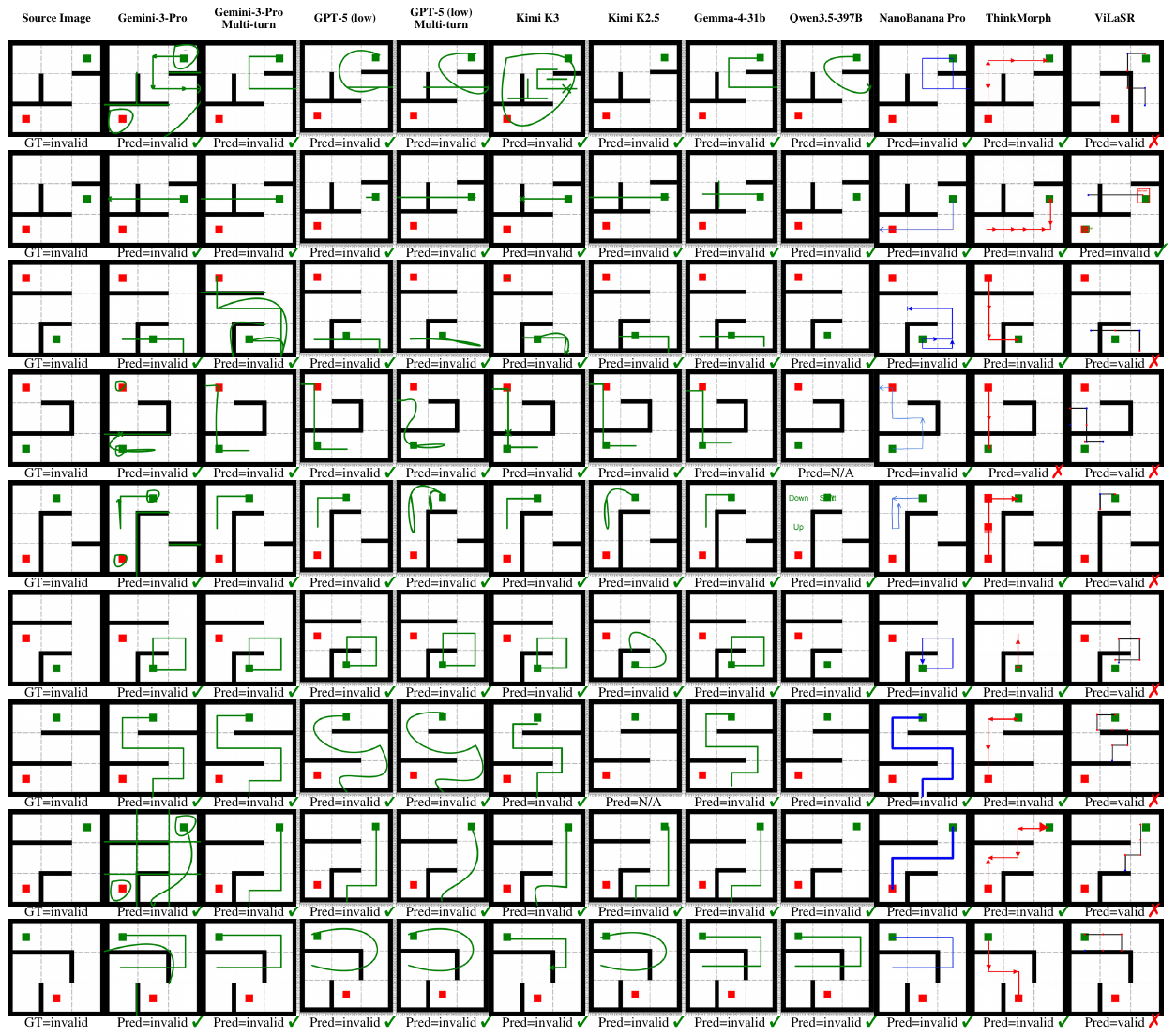}
  \caption{Qualitative examples of models on invalid paths in \maze.}
  \label{fig:maze_invalid_qual_ex}
\end{figure*}

\clearpage

\twocolumn[{%
  \centering
  \subsection{Connect Dots}
  \label{app:qual-connect-dots}
  \vspace{0.2em}

  \includegraphics[width=\textwidth,height=0.82\textheight,keepaspectratio]{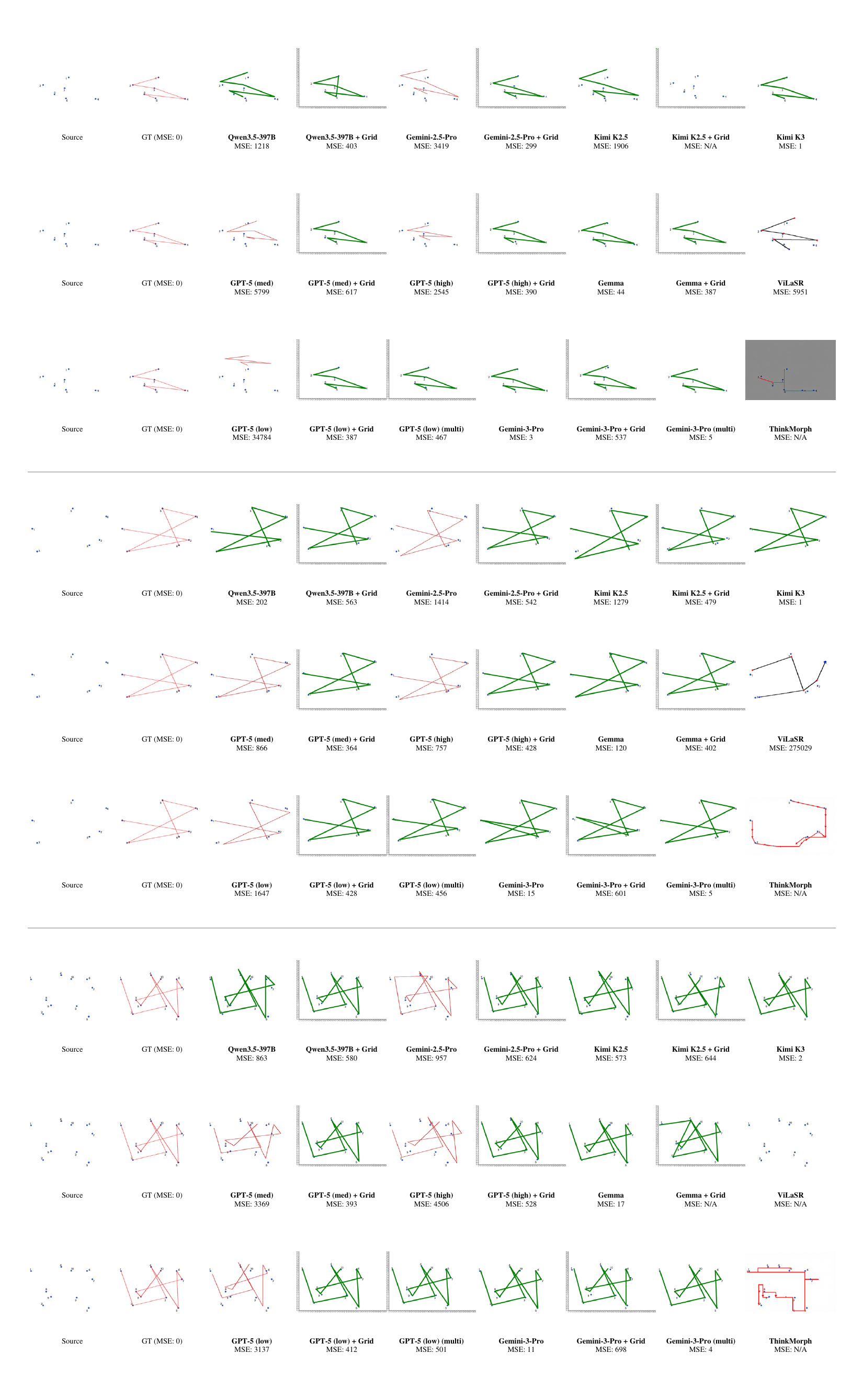}

  \captionof{figure}{Connect-the-Dots qualitative comparisons for the random dots subset. Each cell shows the overlay and MSE.}
  \label{fig:connectdots_random_pairs}
  \vspace{0.3em}
}]

\begin{figure*}[p]
  \centering
  \includegraphics[width=\textwidth,height=0.92\textheight,keepaspectratio]{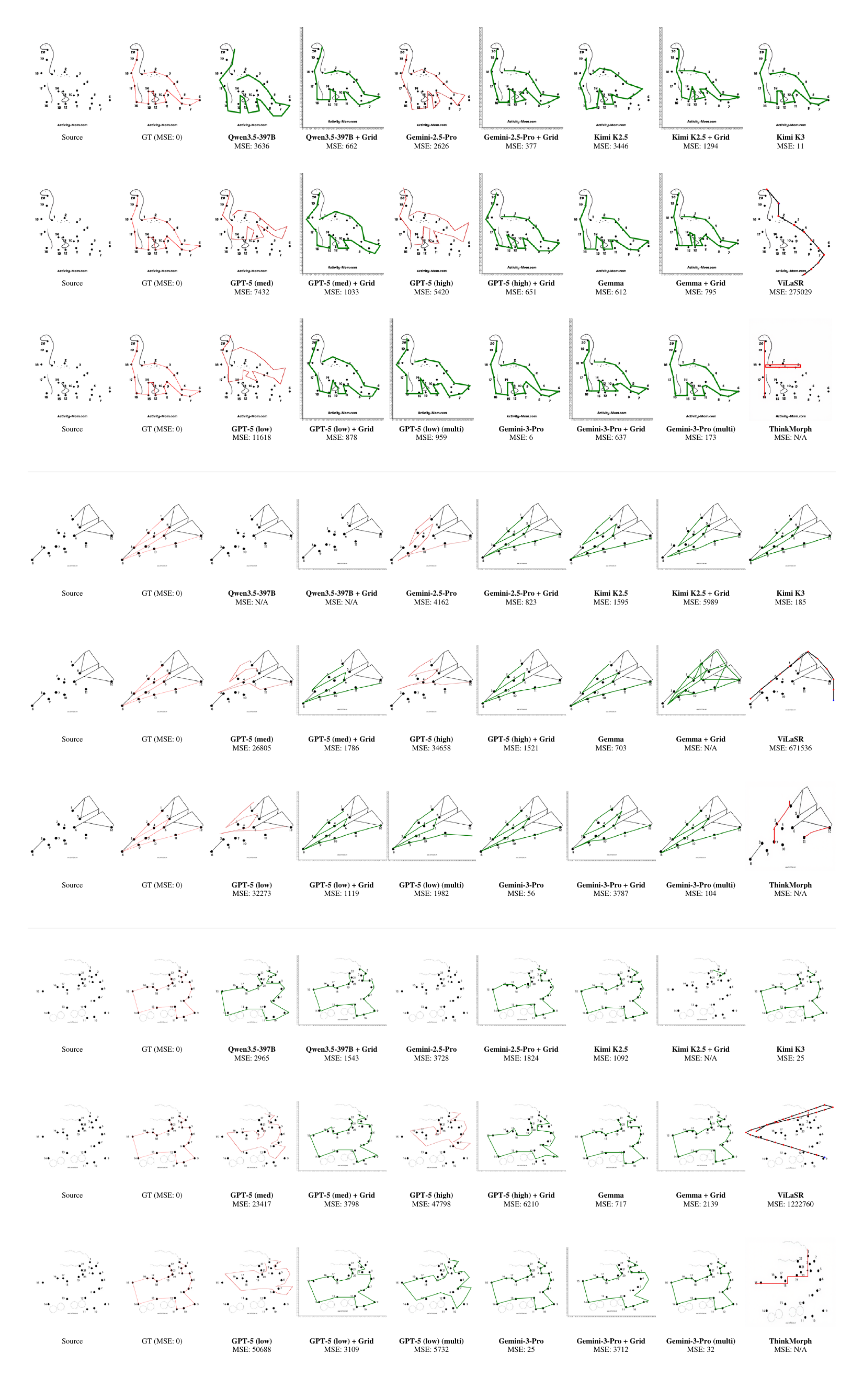}
  \caption{Connect-the-Dots qualitative comparisons for the worksheets subset.}
  \label{fig:connectdots_worksheets_19_36_44}
\end{figure*}

\clearpage

\begin{figure*}[p]
  \centering
  \includegraphics[width=\textwidth,height=0.92\textheight,keepaspectratio]{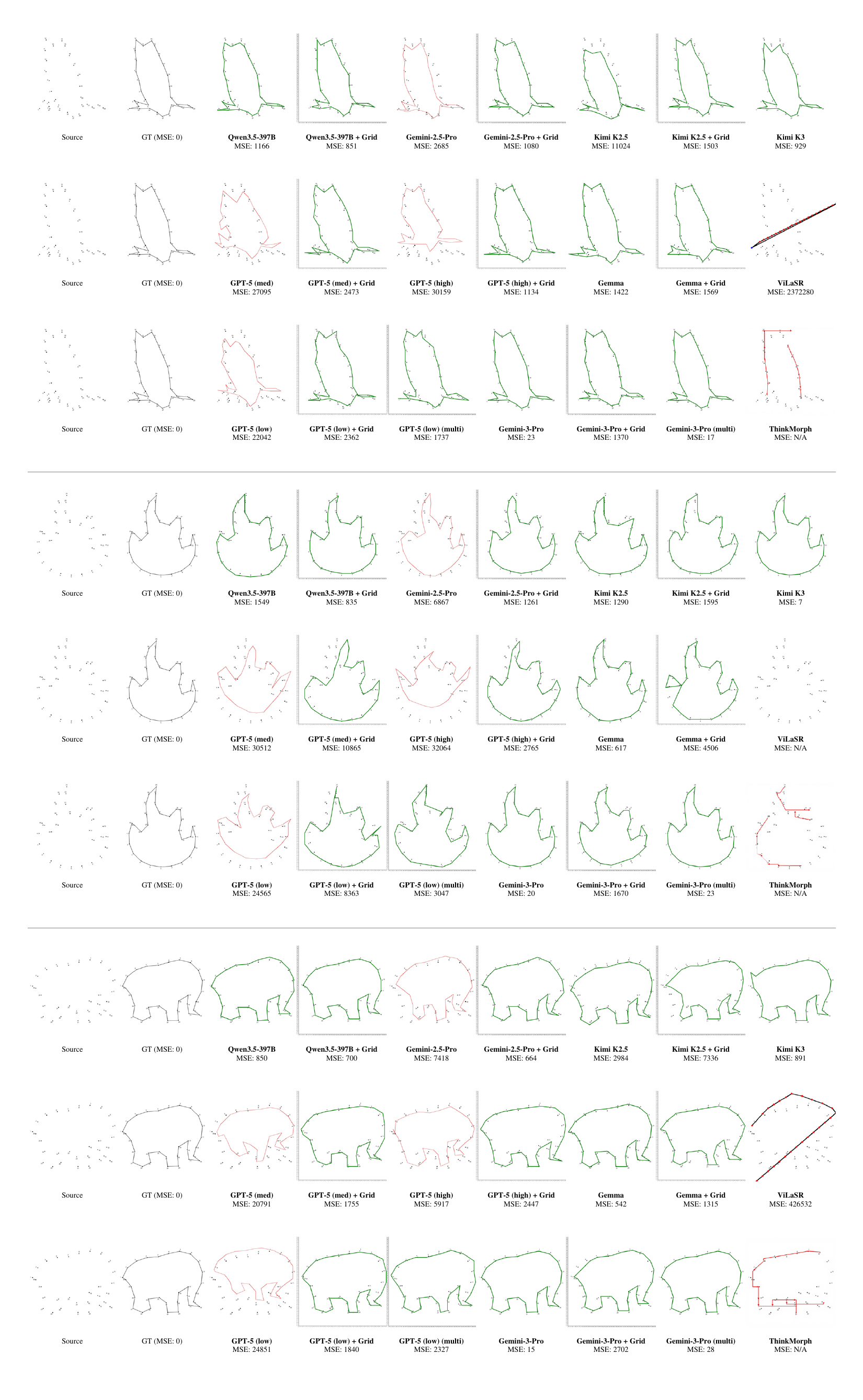}
  \caption{Connect-the-Dots qualitative comparisons for the outlines subset.}
  \label{fig:connectdots_outlines_4_8_21}
\end{figure*}

\clearpage
\subsection{Counting}
\label{app:qual_counting}
\begin{center}

\includegraphics[
  page=1,
  width=0.95\textwidth,
  trim=0mm 50mm 0mm 0mm,
  clip
]{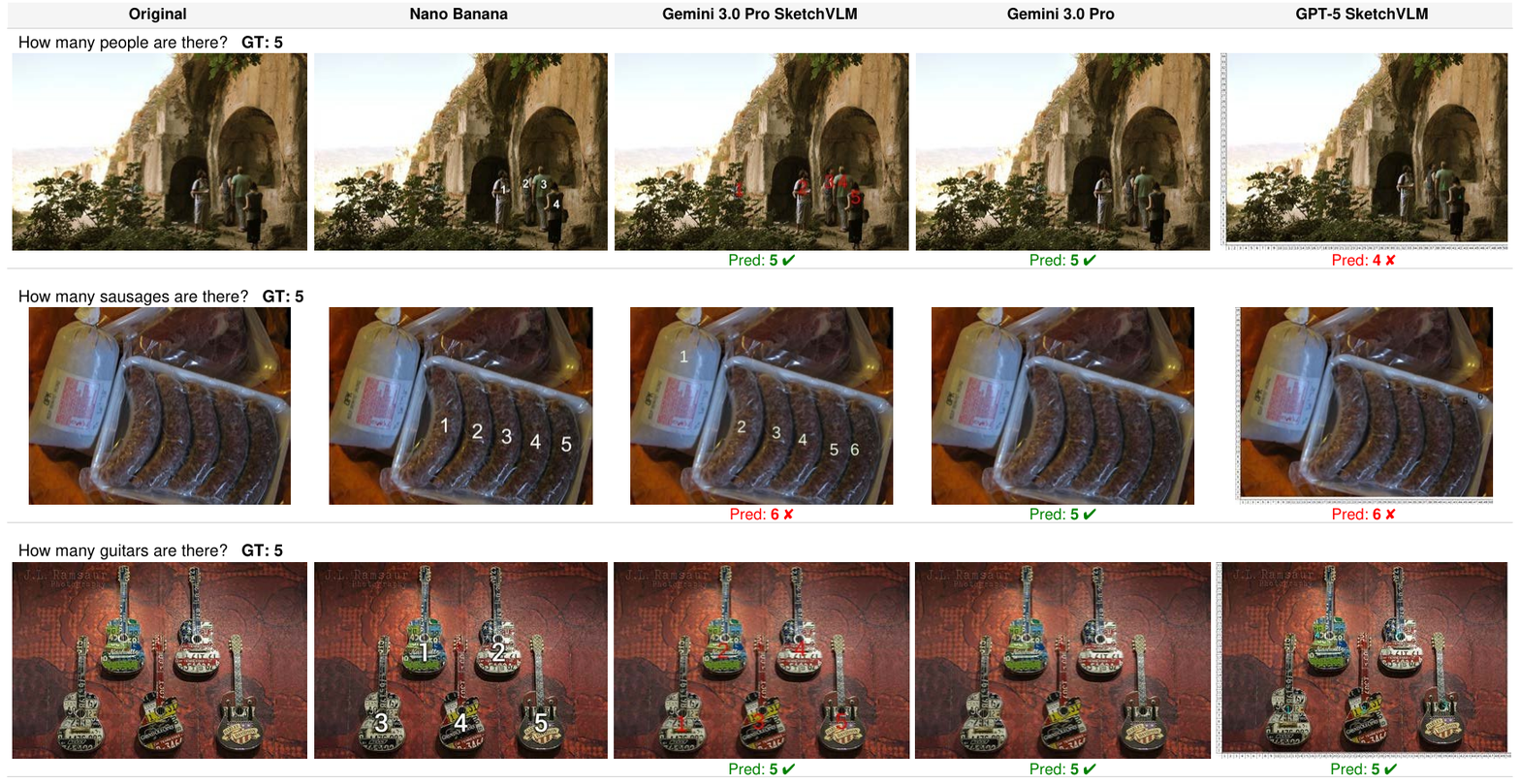}

\vspace{-6pt}
\includegraphics[
  page=2,
  width=0.95\textwidth,
  trim=0mm 50mm 0mm 0mm,
  clip
]{./figure/tasks/counting/qualitative_compare_counting_pixmo_all_models.pdf}

\vspace{-6pt}
\includegraphics[
  page=3,
  width=0.95\textwidth,
  trim=0mm 100mm 0mm 0mm,
  clip
]{./figure/tasks/counting/qualitative_compare_counting_pixmo_all_models.pdf}

\phantomsection
\parbox{0.95\textwidth}{
\captionof{figure}{
Qualitative comparison on counting task between \geminiproThree, \geminiplussketch, \gptplussketch, and \nanobananapro.
}
}
\label{fig:counting_qualitative_all_models}
\end{center}

\clearpage
\label{app:qual_counting_gemma4_qwen35}
\begin{center}

\includegraphics[
  page=1,
  width=0.95\textwidth
]{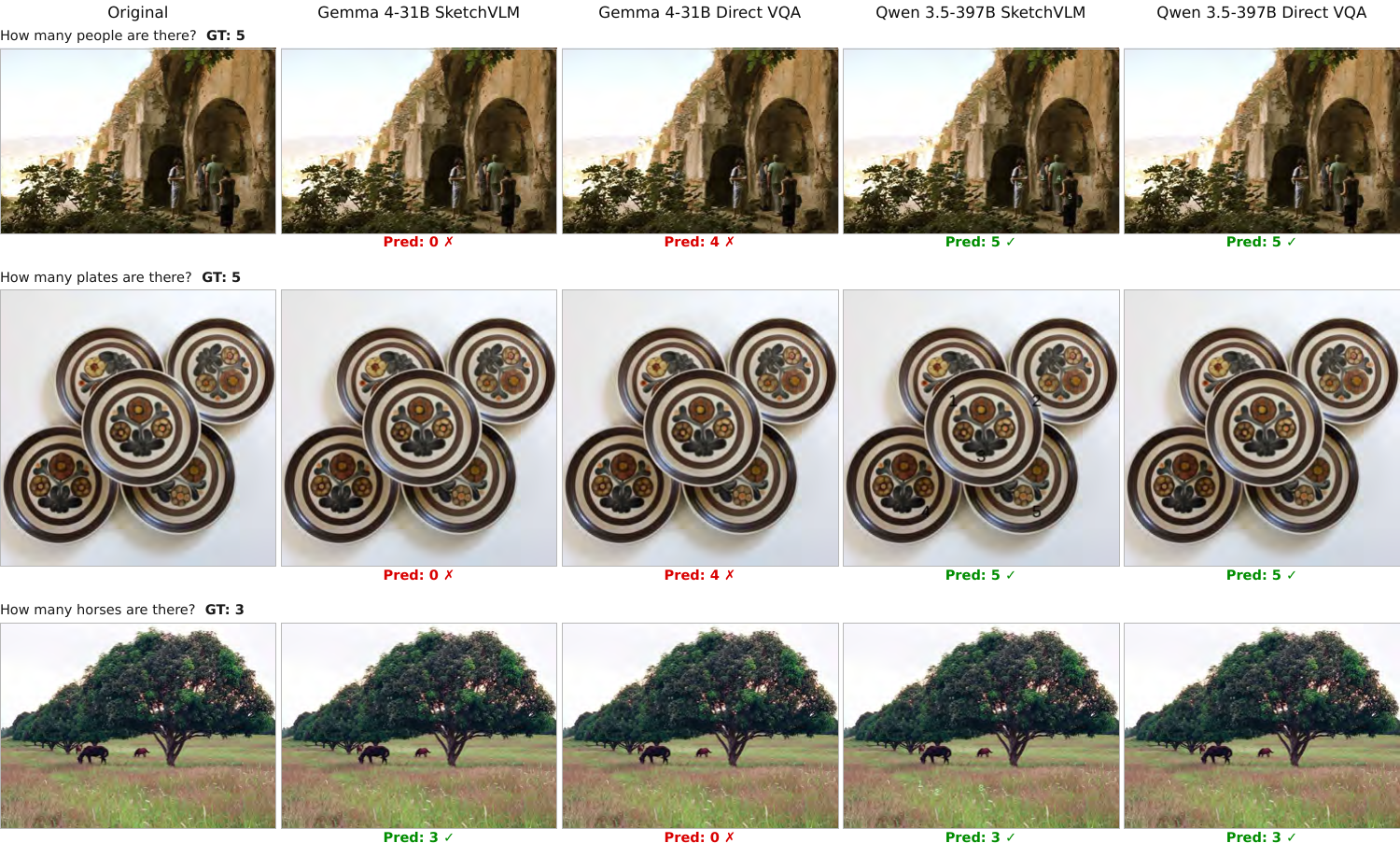}

\vspace{2pt}
\includegraphics[
  page=2,
  width=0.95\textwidth
]{./figure/tasks/counting/counting_gemma4_qwen35_sketchvlm_vs_vqa_100_samples.pdf}

\phantomsection
\parbox{0.95\textwidth}{
\captionof{figure}{
Qualitative comparison on the counting task between \gemmafour and Qwen3.5-397B using SketchVLM and Direct VQA.
}
}
\label{fig:counting_qualitative_gemma4_qwen35}
\end{center}

\clearpage
\begin{center}

\vspace{-10pt}
\includegraphics[
  page=11,
  width=0.92\textwidth
]{./figure/tasks/counting/counting_k25_k3_sketchvlm_vs_vqa_100_samples.pdf}

\vspace{-4pt}
\includegraphics[
  page=13,
  width=0.92\textwidth
]{./figure/tasks/counting/counting_k25_k3_sketchvlm_vs_vqa_100_samples.pdf}

\vspace{-4pt}
\includegraphics[
  page=19,
  width=0.92\textwidth,
  trim=0mm 48mm 0mm 0mm,
  clip
]{./figure/tasks/counting/counting_k25_k3_sketchvlm_vs_vqa_100_samples.pdf}

\vspace{6pt}
\phantomsection
\parbox{0.95\textwidth}{
\captionof{figure}{
Qualitative comparison on the counting task between Kimi K2.5 and Kimi K3 using SketchVLM and Direct VQA.
}
}
\label{fig:counting_qualitative_k25_k3}
\end{center}

\clearpage
\subsection{Drawing Shape}
\label{fig:drawing_more_samples}
\begin{center}
  \includegraphics[
    page=1,
    width=0.92\textwidth,
    trim=0mm 110mm 0mm 0mm,
    clip
  ]{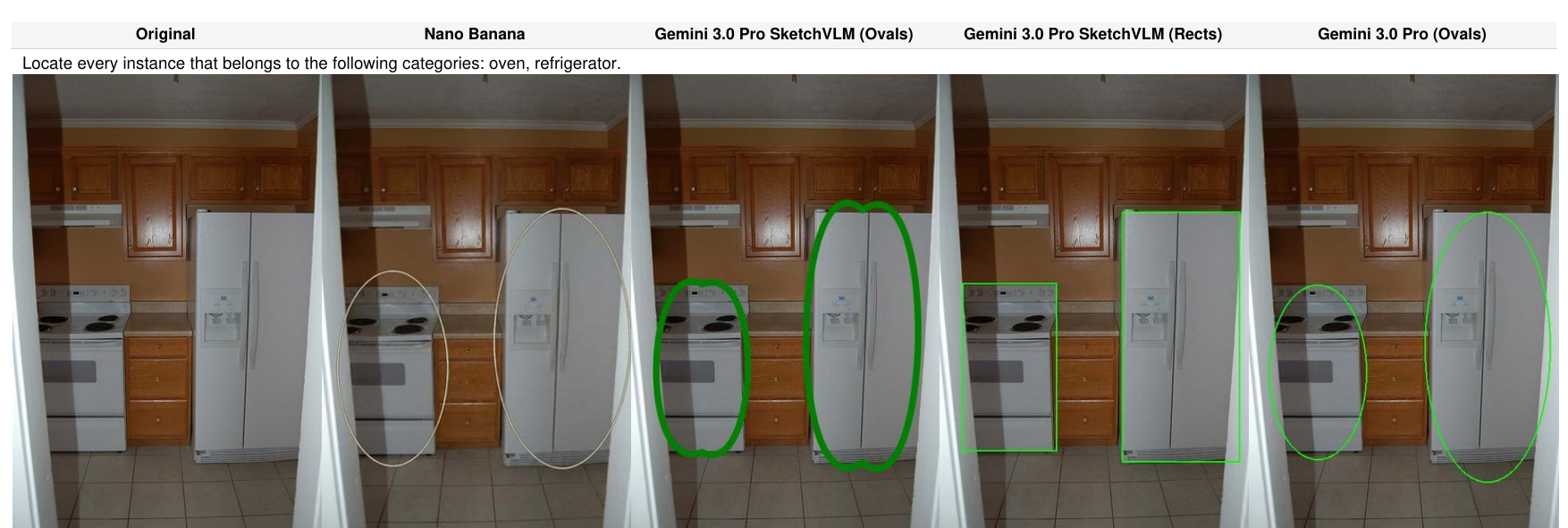}

  \vspace{-6pt}
  \includegraphics[
    page=2,
    width=0.92\textwidth,
    trim=0mm 152mm 0mm 0mm,
    clip
  ]{./figure/tasks/object_detection/qualitative_compare_shape_all_models_1_per_page.pdf}

  \vspace{-6pt}
  \includegraphics[
    page=3,
    width=0.92\textwidth,
    trim=0mm 120mm 0mm 0mm,
    clip
  ]{./figure/tasks/object_detection/qualitative_compare_shape_all_models_1_per_page.pdf}

  \vspace{-6pt}
  \includegraphics[
    page=4,
    width=0.92\textwidth,
    trim=0mm 160mm 0mm 0mm,
    clip
  ]{./figure/tasks/object_detection/qualitative_compare_shape_all_models_1_per_page.pdf}

  \vspace{-6pt}
  \includegraphics[
    page=5,
    width=0.92\textwidth,
    trim=0mm 156mm 0mm 0mm,
    clip
  ]{./figure/tasks/object_detection/qualitative_compare_shape_all_models_1_per_page.pdf}

  \vspace{-6pt}
  \includegraphics[
    page=6,
    width=0.92\textwidth,
    trim=0mm 105mm 0mm 0mm,
    clip
  ]{./figure/tasks/object_detection/qualitative_compare_shape_all_models_1_per_page.pdf}

    \parbox{0.92\textwidth}{
    \captionof{figure}{
    Qualitative comparison on drawing shape tasks between \geminiproThree, \geminiplussketch, and \nanobananapro.
    }
    }
  \label{fig:drawing_shape_qualitative_all_models}
\end{center}

\clearpage

\clearpage
\label{fig:drawing_more_samples_gemma4_qwen35}
\begin{center}
  \includegraphics[
    page=1,
    width=0.98\textwidth,
    trim=0mm 0mm 0mm 0mm,
    clip
  ]{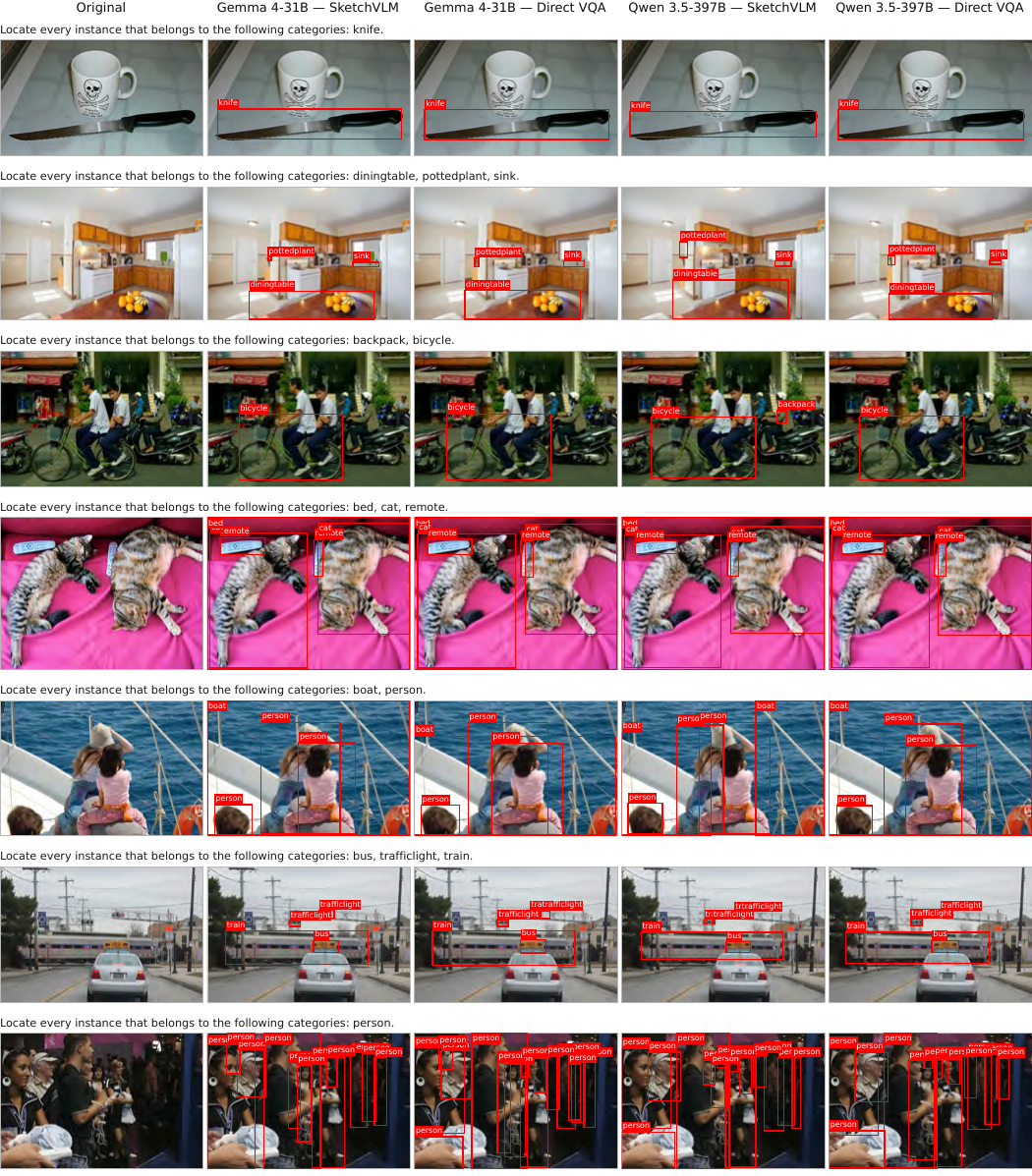}

  \parbox{0.98\textwidth}{
    \captionof{figure}{
    Qualitative comparison of SketchVLM and Direct VQA on the drawing shape task using Gemma 4 and Qwen3.5.
    }
  }
  \label{fig:drawing_shape_qualitative_gemma4_qwen35_sketchvlm_vqa}
\end{center}

\clearpage
\label{fig:drawing_more_samples_k25_k3}
\begin{center}
  \includegraphics[
    page=1,
    width=0.98\textwidth,
    trim=0mm 0mm 0mm 0mm,
    clip
  ]{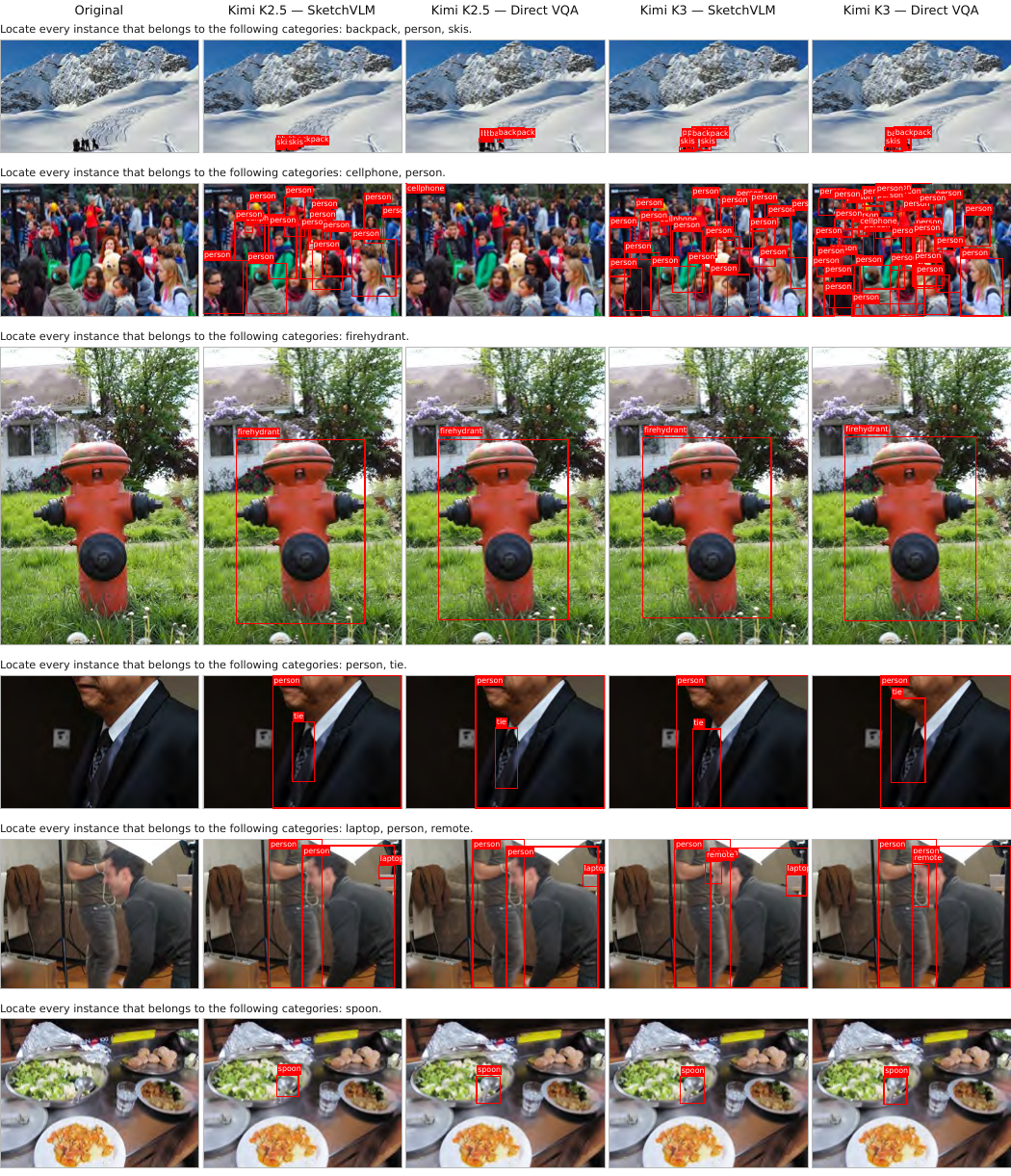}

    \parbox{0.98\textwidth}{
    \captionof{figure}{
    Qualitative comparison of SketchVLM and Direct VQA on the drawing shape task using Kimi K2.5 and Kimi K3.
    }
    }
  \label{fig:drawing_shape_qualitative_k25_k3_sketchvlm_vqa}
\end{center}

\clearpage
\subsection{Part Labeling}
\label{fig:part_labelling_more_samples}
\begin{center}
  \includegraphics[
    page=1,
    width=0.98\textwidth,
    trim=0mm 138mm 0mm 0mm,
    clip
  ]{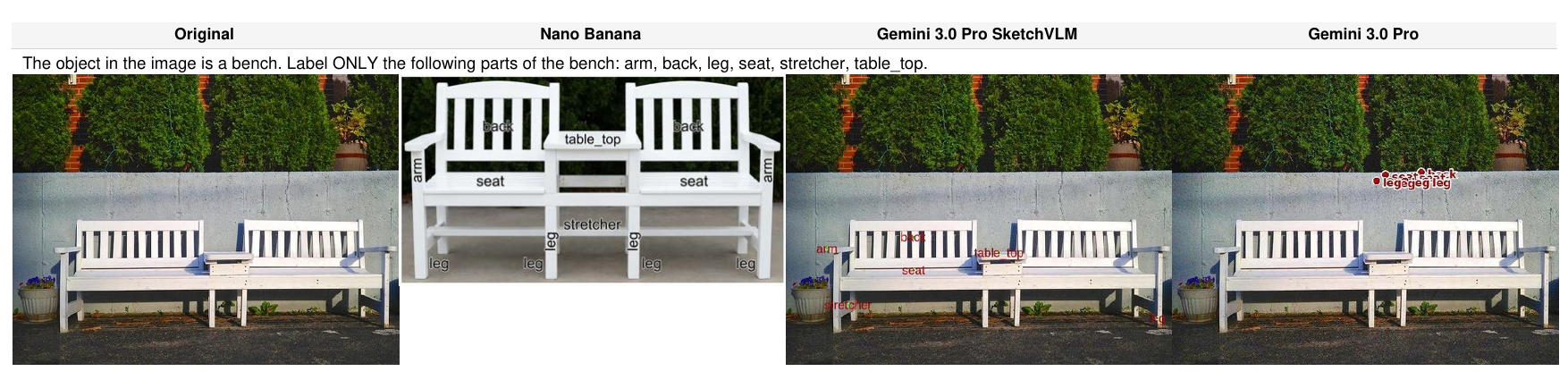}

  \vspace{-6pt}
  \includegraphics[
    page=2,
    width=0.95\textwidth,
    trim=0mm 145mm 0mm 0mm,
    clip
  ]{./figure/tasks/labelling/qualitative_compare_partlabelling_paco_all_models_1_per_page.pdf}

  \vspace{-6pt}
  \includegraphics[
    page=3,
    width=0.95\textwidth,
    trim=0mm 135mm 0mm 0mm,
    clip
  ]{./figure/tasks/labelling/qualitative_compare_partlabelling_paco_all_models_1_per_page.pdf}

  \vspace{-6pt}
  \includegraphics[
    page=4,
    width=0.95\textwidth,
    trim=0mm 140mm 0mm 0mm,
    clip
  ]{./figure/tasks/labelling/qualitative_compare_partlabelling_paco_all_models_1_per_page.pdf}

  \vspace{-6pt}
  \includegraphics[
    page=5,
    width=0.95\textwidth,
    trim=0mm 143mm 0mm 0mm,
    clip
  ]{./figure/tasks/labelling/qualitative_compare_partlabelling_paco_all_models_1_per_page.pdf}

  \vspace{-6pt}
  \includegraphics[
    page=6,
    width=0.95\textwidth,
    trim=0mm 110mm 0mm 0mm,
    clip
  ]{./figure/tasks/labelling/qualitative_compare_partlabelling_paco_all_models_1_per_page.pdf}

    \parbox{0.98\textwidth}{
    \captionof{figure}{
    Qualitative comparison on part labeling tasks between \geminiproThree, \geminiplussketch, and \nanobananapro.
    }
    }
  \label{fig:partlabelling_qualitative_all_models}
\end{center}

\clearpage
\label{app:part_labelling_gemma4_qwen35}
\begin{center}

  \includegraphics[
    page=1,
    width=0.98\textwidth
  ]{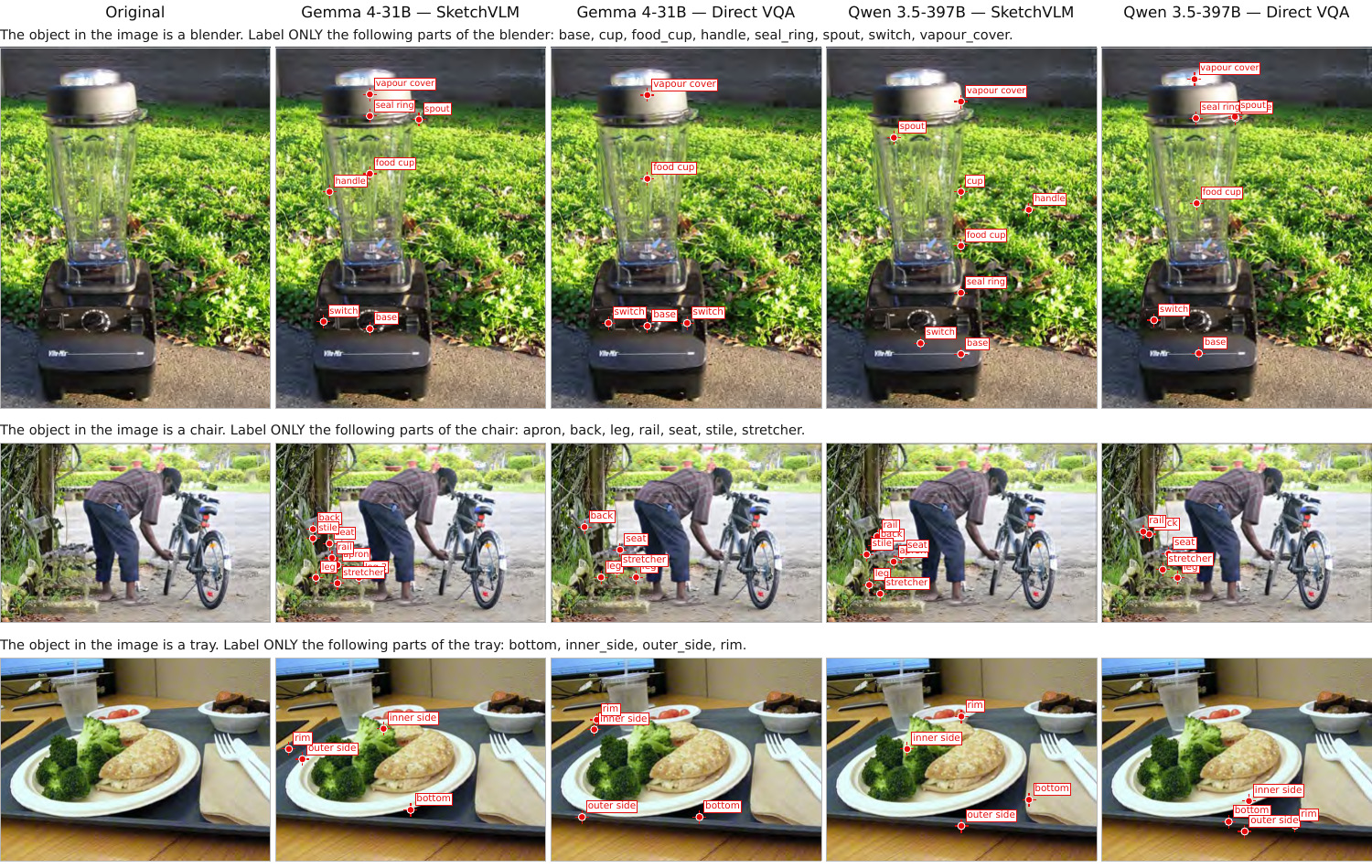}

  \vspace{-6pt}
  \includegraphics[
    page=2,
    width=0.98\textwidth
  ]{./figure/tasks/labelling/part_labelling_gemma4_qwen35_sketchvlm_vs_vqa_100_samples.pdf}

  \vspace{-6pt}
  \includegraphics[
    page=3,
    width=0.98\textwidth,
    trim=0mm 75mm 0mm 0mm,
    clip
  ]{./figure/tasks/labelling/part_labelling_gemma4_qwen35_sketchvlm_vs_vqa_100_samples.pdf}

  \parbox{0.98\textwidth}{
  \captionof{figure}{
  Qualitative comparison of SketchVLM and Direct VQA on the part labeling task using Gemma 4 and Qwen3.5-397B.
  }
  }
  \label{fig:partlabelling_qualitative_gemma4_qwen35_sketchvlm_vqa}

\end{center}

\clearpage
\label{app:part_labelling_k25_k3}
\begin{center}

  \includegraphics[
    page=1,
    width=0.98\textwidth
  ]{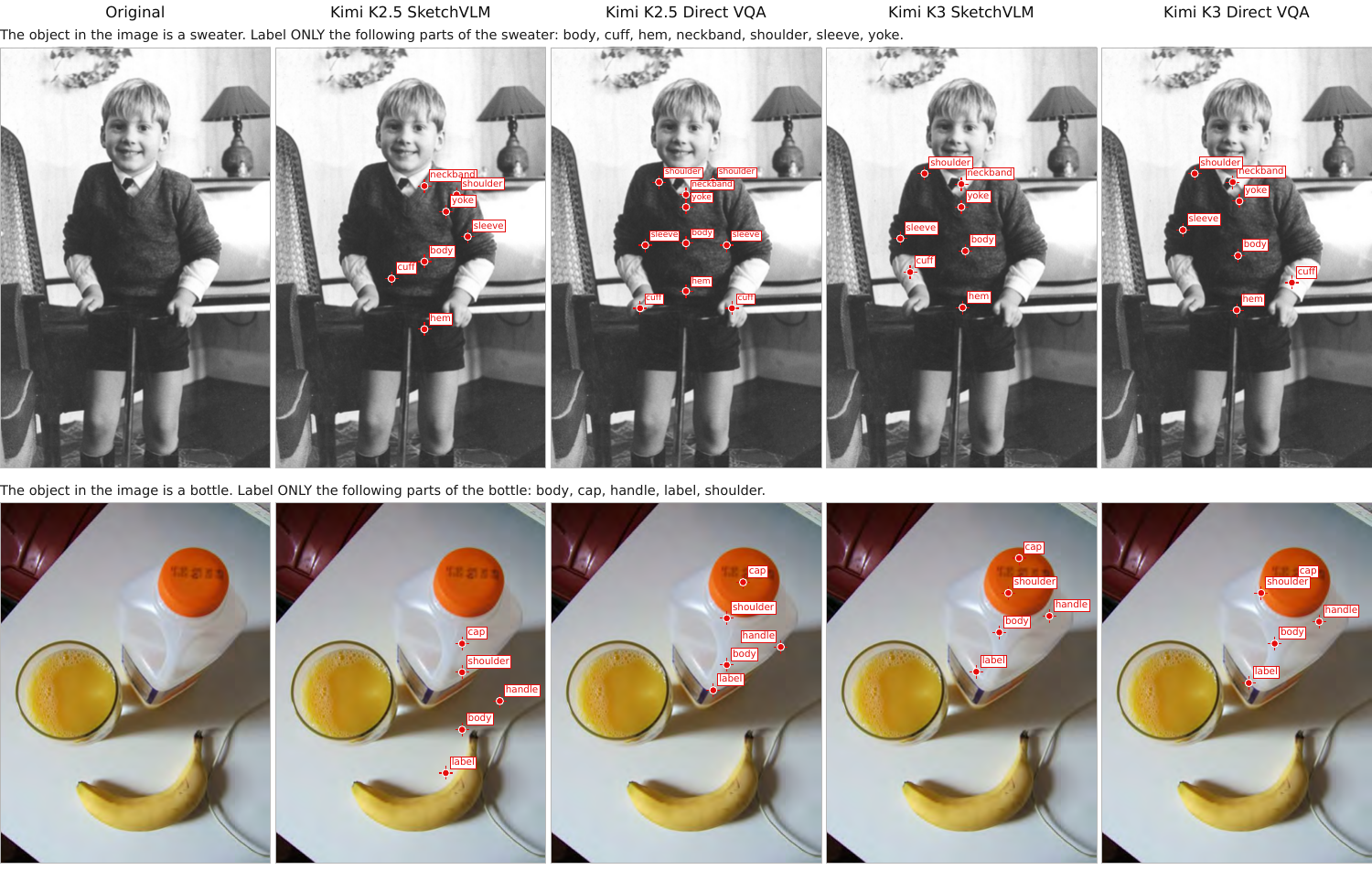}

  \vspace{-6pt}
  \includegraphics[
    page=2,
    width=0.98\textwidth
  ]{./figure/tasks/labelling/part_labelling_k25_k3_sketchvlm_vs_vqa_50_samples.pdf}

  \vspace{-6pt}
  \includegraphics[
    page=3,
    width=0.98\textwidth,
    trim=0mm 100mm 0mm 0mm,
    clip
  ]{./figure/tasks/labelling/part_labelling_k25_k3_sketchvlm_vs_vqa_50_samples.pdf}

  \parbox{0.98\textwidth}{
  \captionof{figure}{
  Qualitative comparison of SketchVLM and Direct VQA on the part labeling task using Kimi K2.5 and Kimi K3.
  }
  }
  \label{fig:partlabelling_qualitative_k25_k3_sketchvlm_vqa}

\end{center}

\clearpage

\subsection{Error Analysis}
\label{sec:qualitative_error_analysis}

As reported in \cref{sec:results_sketch_text_align}, Gemini-3-Flash infers the \SketchVLM{} text answer from the annotation alone in 95.5\% of \gptfive{} cases and 94.2\% of \geminiproThree{} cases. To better understand the remaining failures, we manually inspected 101 failed \maze{} examples and categorized them into three failure modes:

\begin{enumerate}
    \item \textit{Confusing sketch}: The sketch contains unnecessary strokes or exaggerated curves that make it difficult to interpret.
    \item \textit{Text contradiction}: The sketch directly contradicts the model's own text answer.
    \item \textit{Judge classification error}: The sketch is readily interpretable to a human, but the VLM judge makes a reasoning error and misclassifies it.
\end{enumerate}

The failures of \geminiproThree{} primarily consist of confusing sketches (72.4\%), whereas those of \gptfive{} primarily involve contradictions between the sketch and the model's own text answer (51.2\%) (\cref{tab:maze_alignment_failures}). Notably, 19.0\% and 34.9\% of the respective failures are attributable to judge classification errors rather than annotation errors. Thus, the reported alignment scores are conservative estimates of true annotation faithfulness.

We also randomly selected 100 \vpct{} examples and manually inspected the responses from both \geminiproThree{} and \gptfive{}. The most common annotation error is crossing walls in the image. The second most common error is incorrect physical reasoning, such as predicting that a ball will change direction in midair (\cref{tab:vpct_annotation_errors,fig:vpct_bad_physics,fig:vpct_bad_physics_wall}). Other examples of errors can be found throughout the qualitative samples \cref{app:qual_maze,app:qual_counting,app:qual-connect-dots,fig:maze_valid_qual_ex,fig:maze_invalid_qual_ex,fig:vpct_qual_ex,fig:ball_drop_qual_ex,fig:drawing_shape_qualitative_all_models,fig:partlabelling_qualitative_all_models}.

\begin{figure}[t]
    \centering
    \includegraphics[width=\columnwidth]{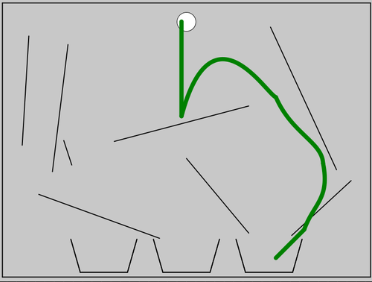}
    \caption{Example of sketching with bad physics understanding on \vpct{}. The predicted ball trajectory changes direction in midair.}
    \label{fig:vpct_bad_physics}
\end{figure}

\begin{figure}[t]
    \centering
    \includegraphics[width=\columnwidth]{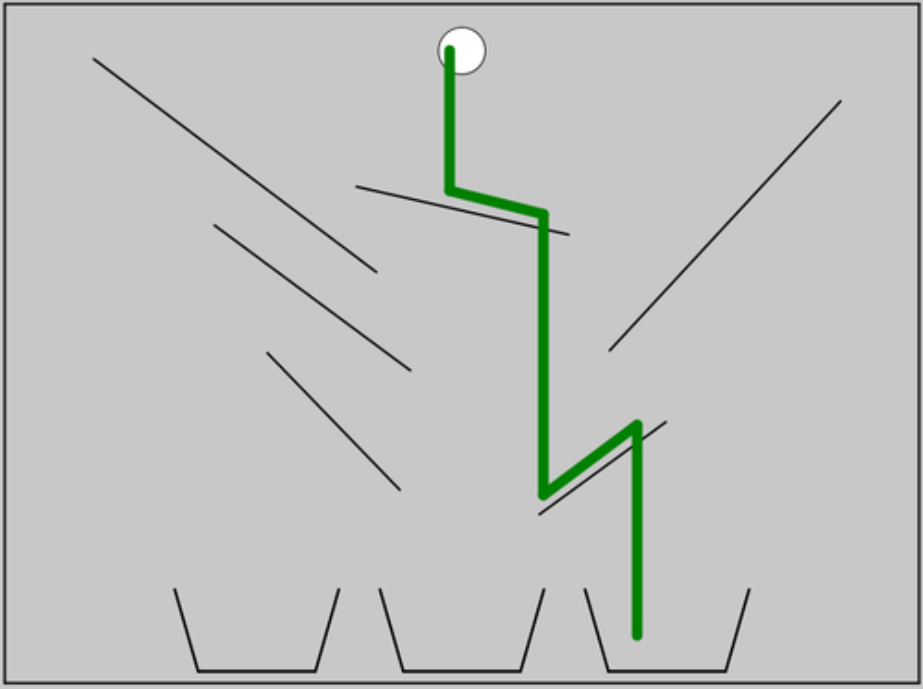}
    \caption{Example of sketching with both bad physics understanding and crossing walls on \vpct{}. The predicted ball trajectory is physically implausible and crosses a wall.}
    \label{fig:vpct_bad_physics_wall}
\end{figure}

\begin{table}[h]
\centering
\caption{Distribution of annotation--text alignment failures on \maze{}. Confusing sketches account for most \geminiproThree{} failures, whereas text contradictions account for most \gptfive{} failures.}
\label{tab:maze_alignment_failures}
\setlength{\tabcolsep}{4pt}
\renewcommand{\arraystretch}{1.12}

\adjustbox{max width=\columnwidth}{%
\begin{tabular}{lccc}
\toprule
\textbf{Model}
& \shortstack{\textbf{Confusing}\\\textbf{sketch}}
& \shortstack{\textbf{Text}\\\textbf{contradiction}}
& \shortstack{\textbf{Judge classification}\\\textbf{error}} \\
\midrule
\geminiproThree & 72.4\% & 8.6\%  & 19.0\% \\
\gptfive        & 14.0\% & 51.2\% & 34.9\% \\
\bottomrule
\end{tabular}%
}
\end{table}

\begin{table}[h]
\centering
\caption{Distribution of annotation errors on 100 randomly selected \vpct{} examples. Wall crossings are the most common annotation error for both models.}
\label{tab:vpct_annotation_errors}
\setlength{\tabcolsep}{5pt}
\renewcommand{\arraystretch}{1.12}

\adjustbox{max width=\columnwidth}{%
\begin{tabular}{lcccc}
\toprule
\textbf{Model}
& \textbf{No errors}
& \shortstack{\textbf{Crosses}\\\textbf{wall}}
& \shortstack{\textbf{Incorrect}\\\textbf{physics}}
& \textbf{Both} \\
\midrule
\geminiproThree & 91.0\% & 7.0\%  & 2.0\%  & 0.0\% \\
\gptfive        & 56.0\% & 28.0\% & 10.0\% & 6.0\% \\
\bottomrule
\end{tabular}%
}
\end{table}

\clearpage

\subsection{\connectdots: Grid versus No Grid}

\begin{figure}[H]
  \centering

    \begin{minipage}{\linewidth}
      \centering
      \small Source \\[1ex]
      \includegraphics[width=0.6\linewidth]{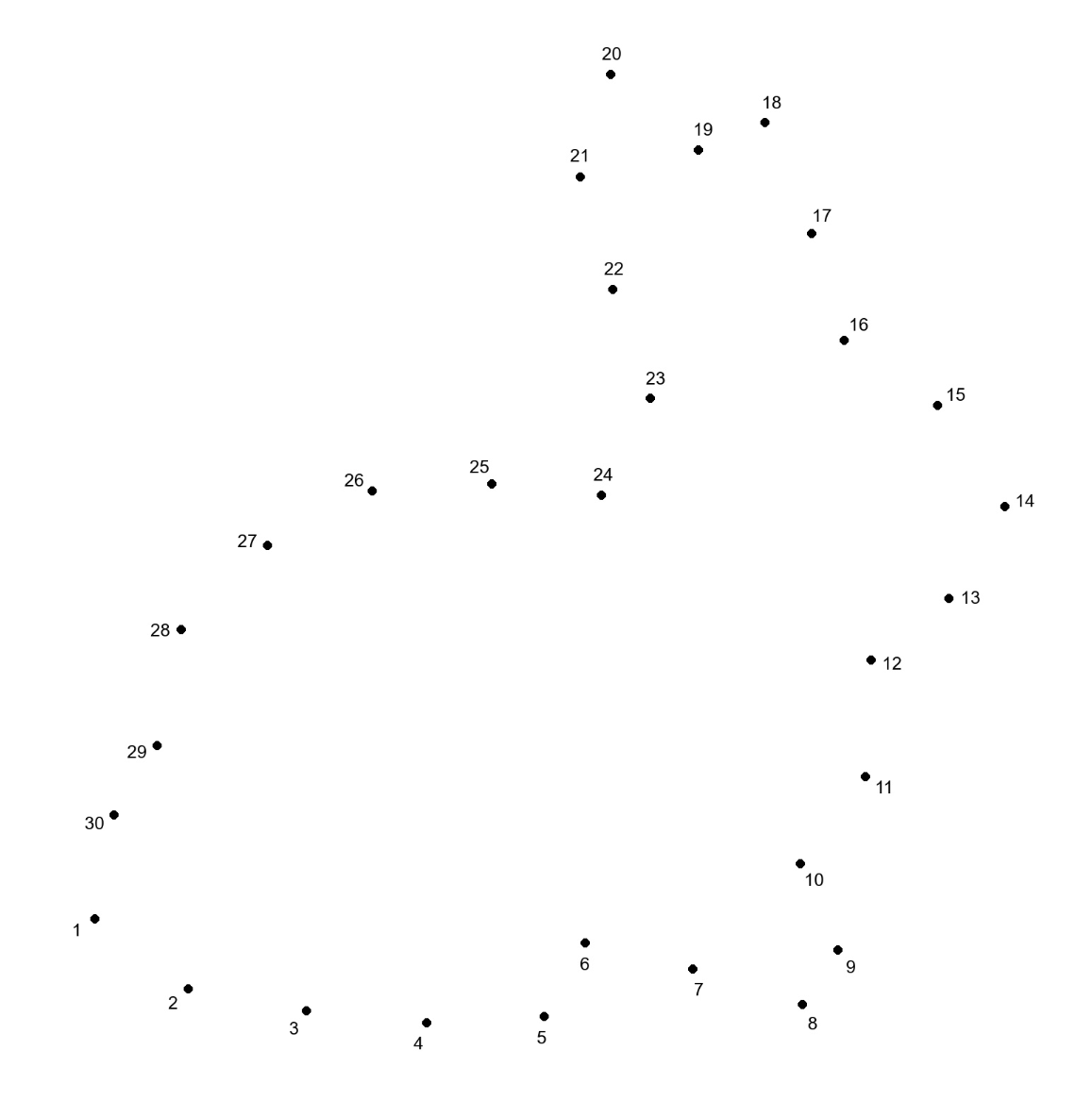}
    \end{minipage}\hfill
    \begin{minipage}{\linewidth}
      \centering
      \small \gptlogo\ \gptfive (without Grid) \\[1ex]
      \includegraphics[width=0.6\linewidth]{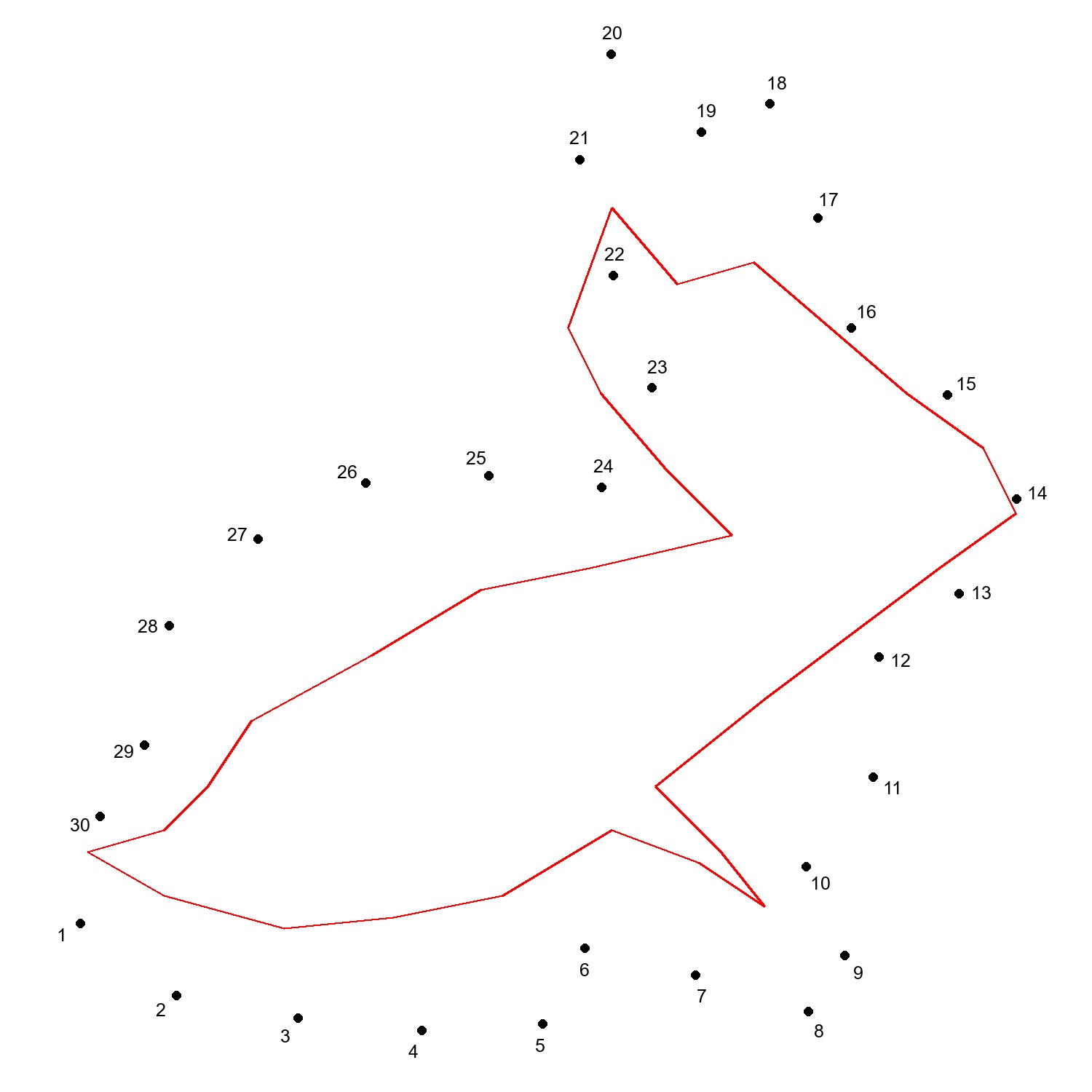}
    \end{minipage}\hfill
    \begin{minipage}{\linewidth}
      \centering
      \small \gptlogo\ \gptfive (with Grid) \\[1ex]
      \includegraphics[width=0.6\linewidth]{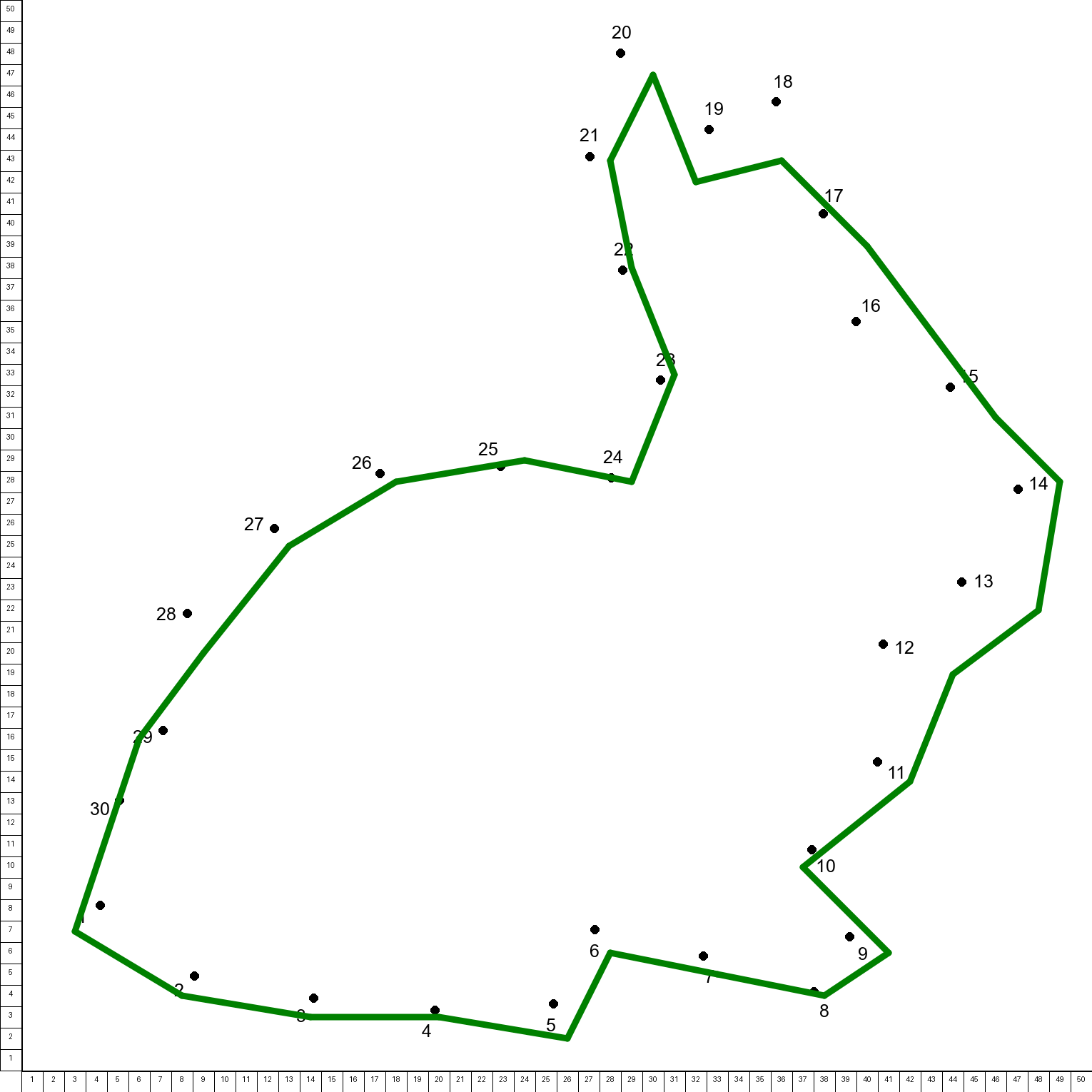}
    \end{minipage}

  \caption{Appending a reference coordinate grid to the edge of the input image allows \gptlogo \gptfive to be more precise, but does not help \geminilogo \geminiproThree.}
  \label{fig:worksheet_example}
\end{figure}

\clearpage

\subsection{Connect-the-Dots: Bézier Curves versus Lines}
\label{app:curves_vs_lines}

\begin{figure}[H]
  \centering

  \newlength{\imgH}\setlength{\imgH}{0.14\textheight} 
  \newcommand{\rowsep}{\vspace{3pt}}                   
  \newcommand{\captsize}{\small}                        


  \begin{minipage}[t]{0.485\linewidth}
      \centering
      \includegraphics[width=\linewidth,height=\imgH,keepaspectratio]{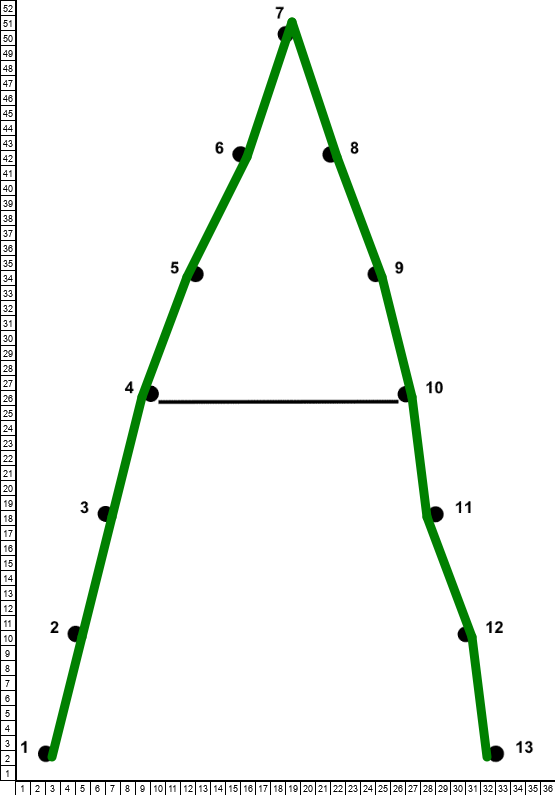}\vspace{1pt}\\[-1pt]
      {\captsize Straight Lines: 12 strokes}
    \end{minipage}\hfill
    \begin{minipage}[t]{0.485\linewidth}
      \centering
      \includegraphics[width=\linewidth,height=\imgH,keepaspectratio]{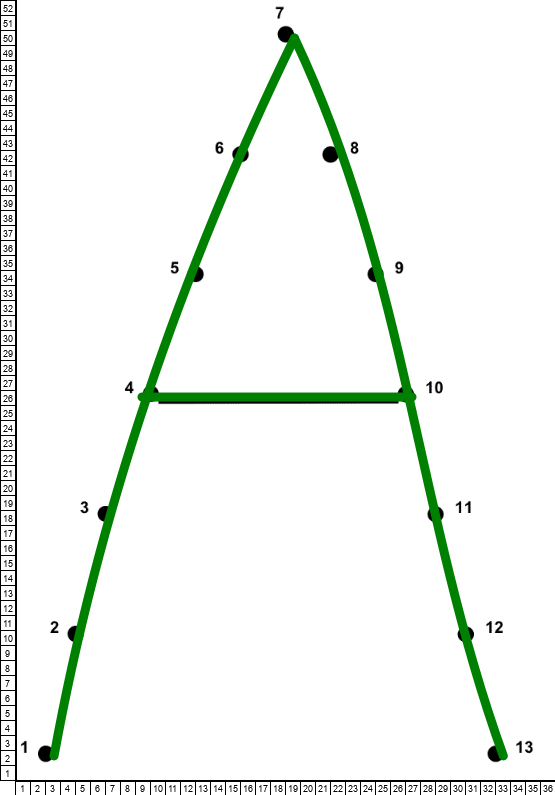}\vspace{1pt}\\[-1pt]
      {\captsize Bézier Curves: 2 strokes}
    \end{minipage}%

  \vspace{6pt}

 \begin{minipage}[t]{0.485\linewidth}
      \centering
      \includegraphics[width=\linewidth,height=\imgH,keepaspectratio]{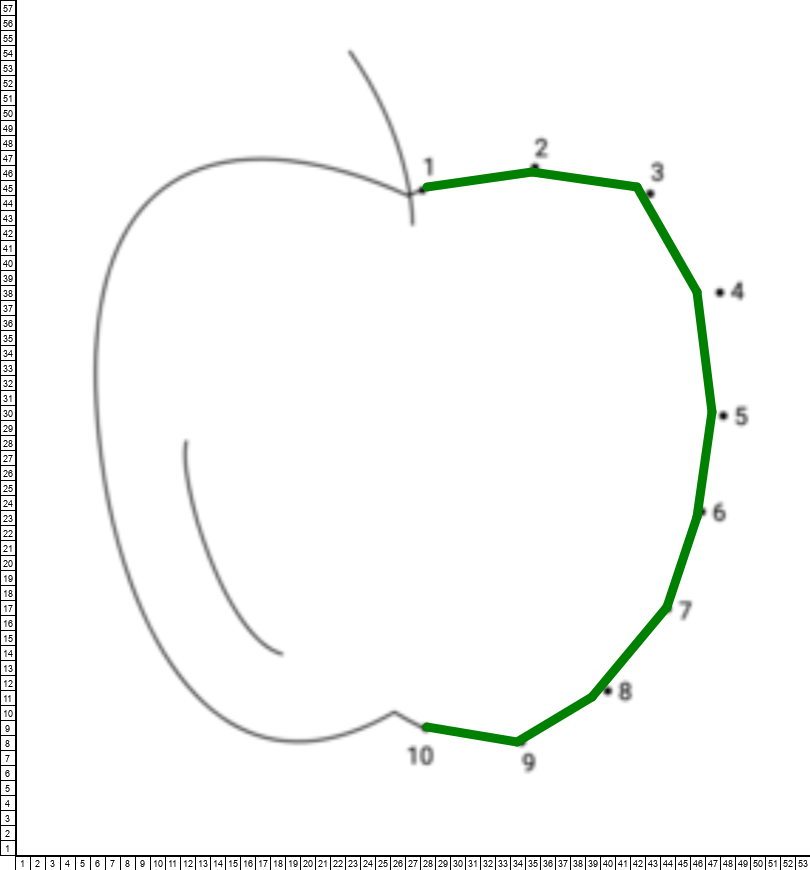}\vspace{1pt}\\[-1pt]
      {\captsize Straight Lines: 9 strokes}
    \end{minipage}\hfill
    \begin{minipage}[t]{0.485\linewidth}
      \centering
      \includegraphics[width=\linewidth,height=\imgH,keepaspectratio]{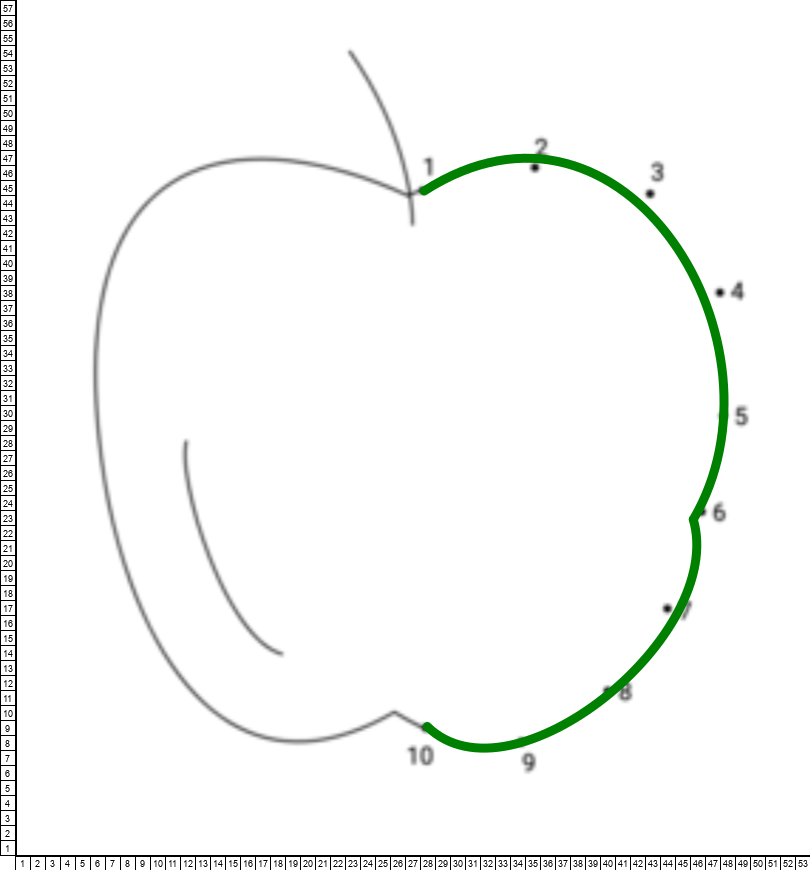}\vspace{1pt}\\[-1pt]
      {\captsize Bézier Curves: 1 stroke}
    \end{minipage}%

    \begin{minipage}[t]{0.485\linewidth}
      \centering
      \includegraphics[width=\linewidth,height=\imgH,keepaspectratio]{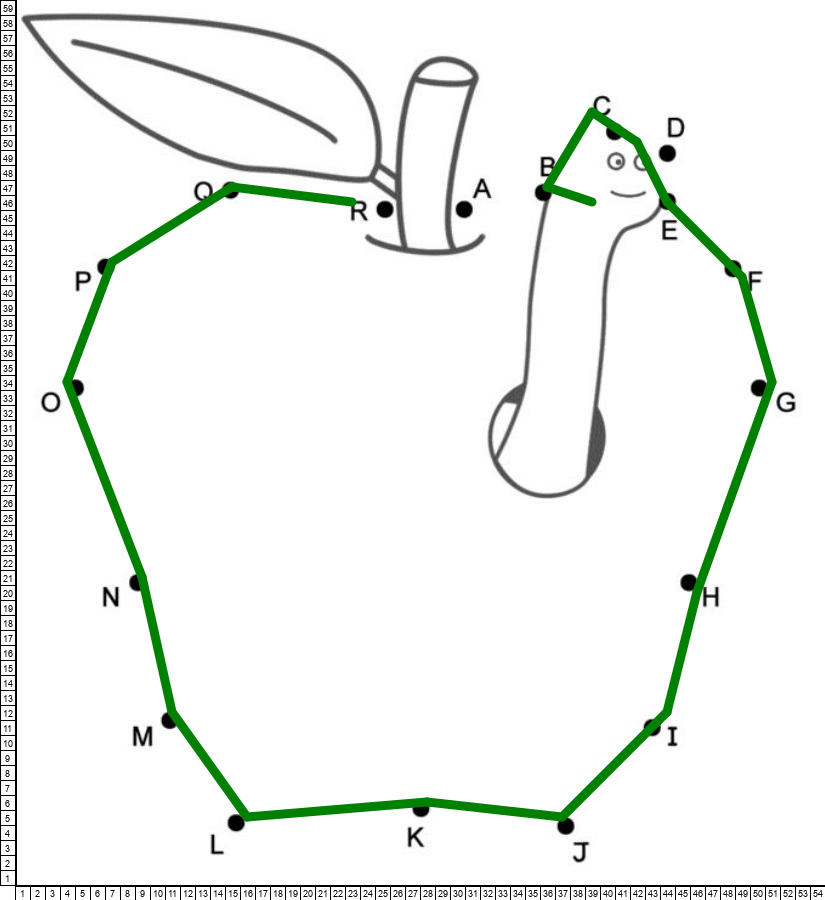}\vspace{1pt}\\[-1pt]
      {\captsize Straight Lines: 17 strokes}
    \end{minipage}\hfill
    \begin{minipage}[t]{0.485\linewidth}
      \centering
      \includegraphics[width=\linewidth,height=\imgH,keepaspectratio]{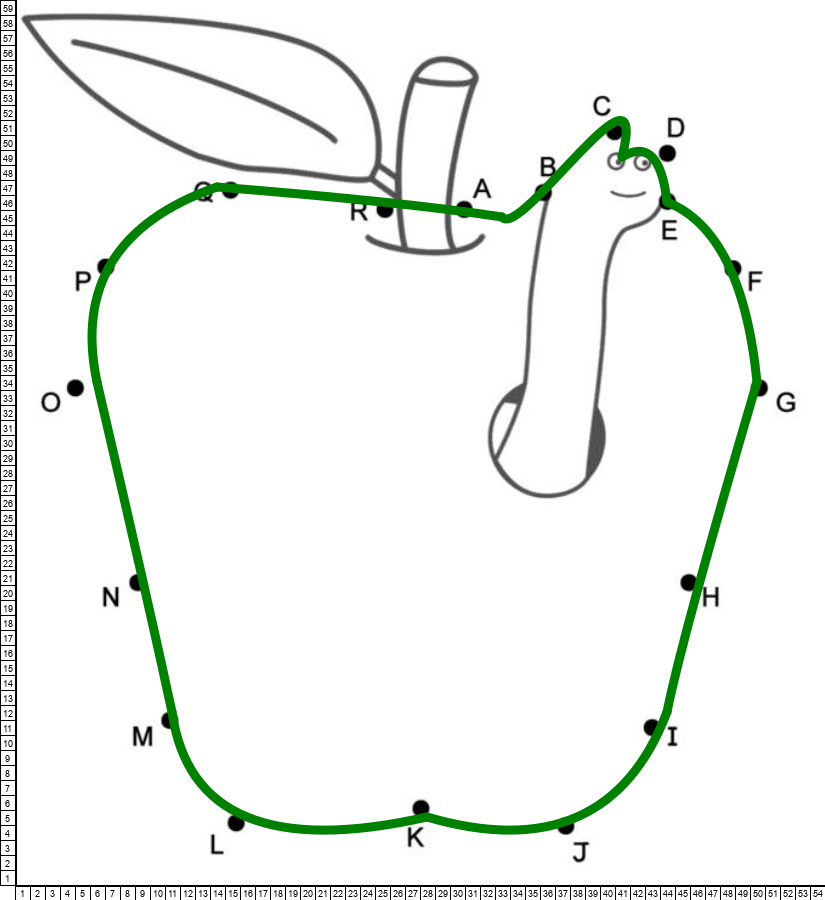}\vspace{1pt}\\[-1pt]
      {\captsize Bézier Curves: 9 strokes}
    \end{minipage}%

    \begin{minipage}[t]{0.485\linewidth}
      \centering
      \includegraphics[width=\linewidth,height=\imgH,keepaspectratio]{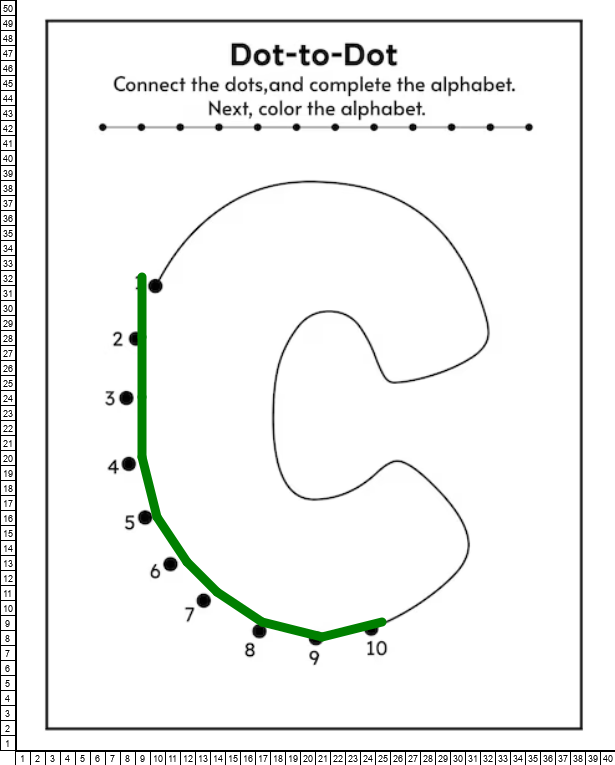}\vspace{1pt}\\[-1pt]
      {\captsize Straight Lines: 9 strokes}
    \end{minipage}\hfill
    \begin{minipage}[t]{0.485\linewidth}
      \centering
      \includegraphics[width=\linewidth,height=\imgH,keepaspectratio]{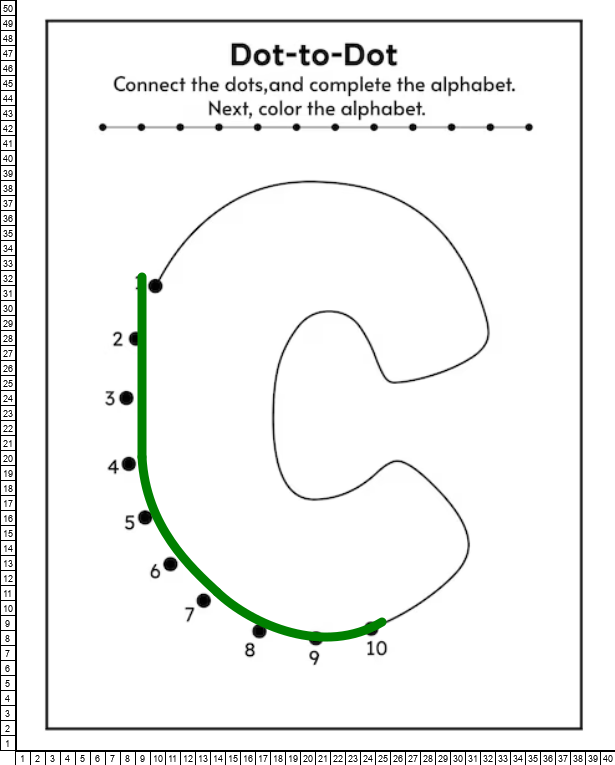}\vspace{1pt}\\[-1pt]
      {\captsize Bézier Curves: 3 strokes}
    \end{minipage}%

    \begin{minipage}[t]{0.485\linewidth}
      \centering
      \includegraphics[width=\linewidth,height=\imgH,keepaspectratio]{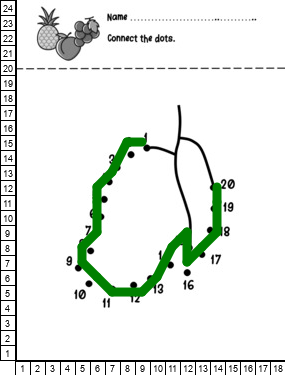}\vspace{1pt}\\[-1pt]
      {\captsize Straight Lines: 19 strokes}
    \end{minipage}\hfill
    \begin{minipage}[t]{0.485\linewidth}
      \centering
      \includegraphics[width=\linewidth,height=\imgH,keepaspectratio]{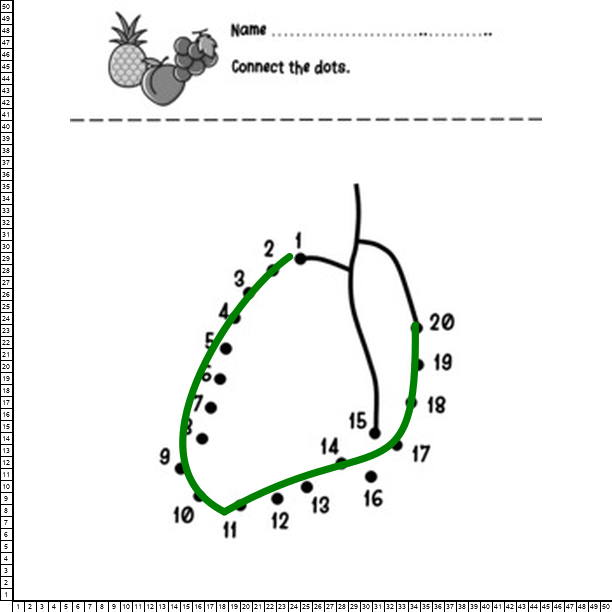}\vspace{1pt}\\[-1pt]
      {\captsize Bézier Curves: 1 stroke}
    \end{minipage}%

  \caption{Using Bézier curves instead of straight lines allows the model to connect the dots in fewer strokes, and it can draw less jagged shapes, leading to a more aesthetically pleasing result.}
  \label{fig:curves_vs_lines}
\end{figure}

\clearpage

\twocolumn[{%
\centering
\subsection{\geminiproThree Coordinate Systems}
\label{app:gempro3_coord_sys}
\vspace{0.3em}

\includegraphics[width=0.95\linewidth]{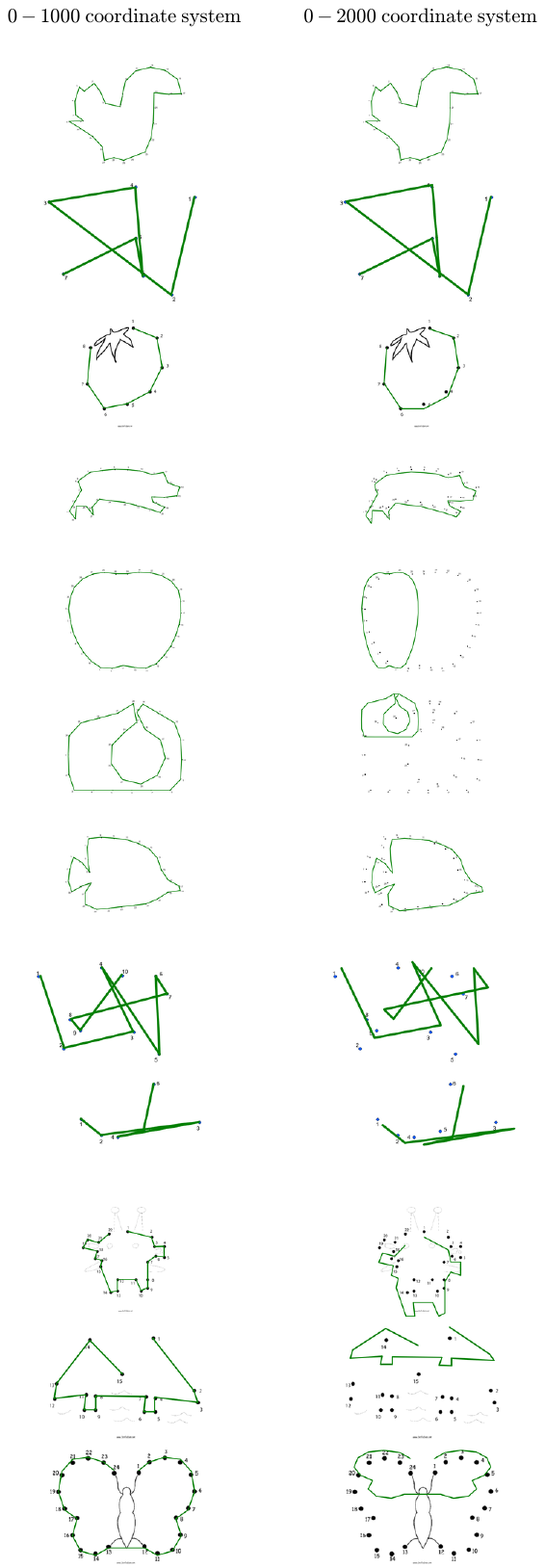}

\captionof{figure}{Comparing \geminiproThree annotations under two coordinate systems. The 0--2000 system (vs.\ the 0--1000 system the model is typically used to) often preserves the overall shape but produces annotations that are compressed or shifted. In most cases the model adapts to the new coordinate system, but these failure cases lead to a noticeable decrease in performance. We hypothesize this change is what causes \geminiproThree and \gemmafour to also perform worse with the grid (which has a different coordinate system).}
\label{fig:coord_1000_vs_2000}
}]

\clearpage

\twocolumn[{%
\centering
\subsection{Multi-turn Ablation}
\label{app:multi_without_text_rep}
\vspace{0.3em}

\begin{minipage}[t]{0.35\linewidth}
    \centering
    \small{Multiturn: Prompts + Image + Text Annotations sent back each turn}

    \includegraphics[width=0.48\linewidth]{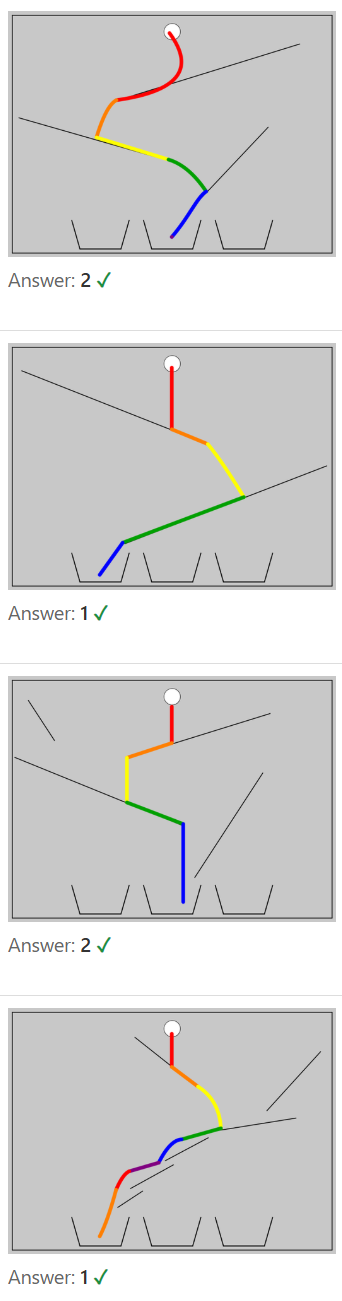}

    \includegraphics[width=0.48\linewidth]{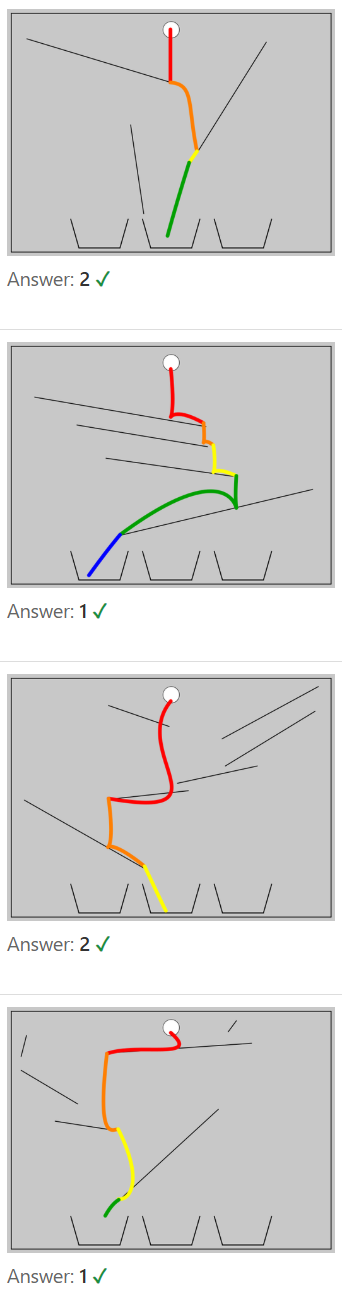}
\end{minipage}
\hfill
\begin{minipage}[t]{0.35\linewidth}
    \centering
    \small{Multiturn: Prompts + Image sent back each turn}

    \includegraphics[width=0.48\linewidth]{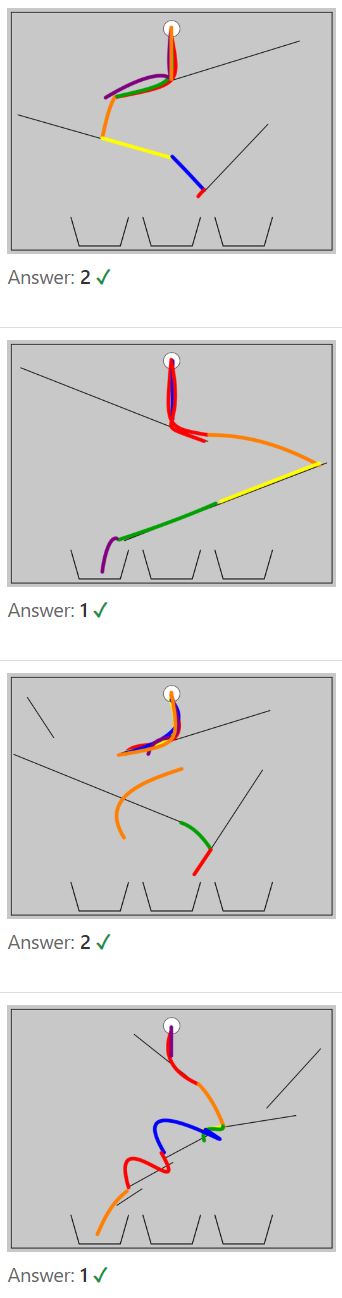}

    \includegraphics[width=0.48\linewidth]{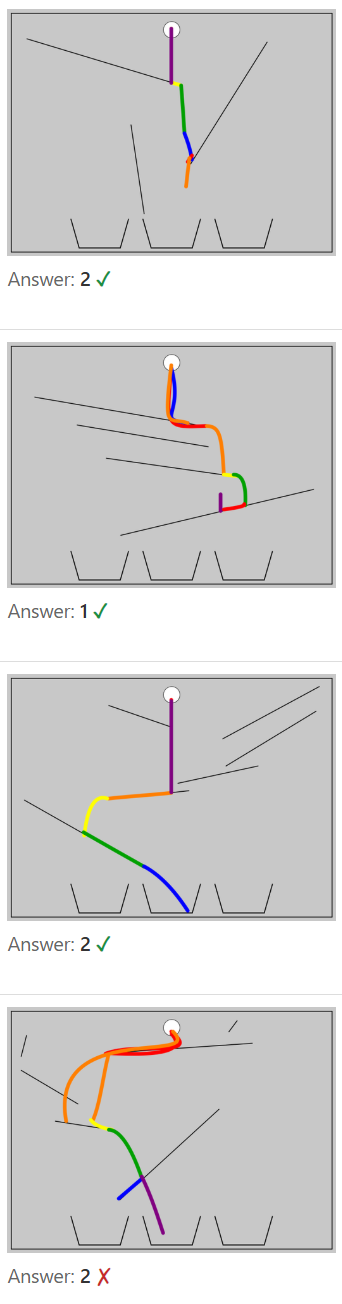}
\end{minipage}

\vspace{-0.5em}
\captionof{figure}{When testing \geminiproThree in the multi-turn setting, not sending back the text representation of prior annotations degrades drawing quality: the model often attempts to redraw earlier strokes and fails to properly connect new strokes to the existing trajectory.}
\label{fig:gemini3pro_multiturn_text_vs_no_text}
}]

\clearpage

\section{VLM-Judge Details}
\label{vlm-judge-details}

\noindent We use a VLM judge (Gemini-3-Flash-Preview \cite{doshi2025gemini3flash}) to evaluate annotation quality and annotation--text alignment. To verify that this automated metric is meaningful, we measure how well VLM judge ratings agree with human annotators.
The human annotators were recruited internally and were not asked to provide personal information. They were volunteers and not paid. Annotators were informed that their ratings would be used for research evaluation and reported anonymously.

\subsec{\textbf{Experiment}} 
Three independent human annotators each rate 50 annotations across the \balldrop, \vpct, and \maze tasks from all 5 models (\geminilogo \geminiproThree, \gptlogo \gptfive, \nanologo \nanobananapro, \vilasrlogo \vilasr, and \thinkmorphlogo \thinkmorph), totaling 2,250 labeled images. Both human and VLM judges use the same 1--5 grading rubric (\cref{vlm-rubrics}). We report agreement using quadratically weighted Cohen's $\kappa$, which penalizes larger disagreements more heavily on ordinal scales \cite{latif2025sketchmind}, and Pearson correlation, following recent VLM-judge benchmarks \cite{li2026gebench}.

\subsec{\textbf{Results}} 
Human--human agreement is high, with $\kappa = 0.84 \pm 0.04$ and Pearson $r = 0.85 \pm 0.04$ (\cref{tab:agreement_summary}), indicating strong consistency between annotators. Detailed per-dataset agreement statistics are reported in \cref{tab:human_human_agreement}. 

Human--VLM agreement is moderate, with $\kappa = 0.51 \pm 0.02$ and Pearson $r = 0.52 \pm 0.01$, with task-level results shown in \cref{tab:vlm_judge_agreement}. The VLM judge's ratings are positively correlated with human judgments across all tasks and models, making it a useful proxy for large-scale evaluation. However, the gap between human--human and human--VLM agreement indicates that the VLM judge can miss subtle errors, particularly logical violations such as trajectories clipping through walls (\cref{vlm-judge-examples}).

Finally, we test whether using Gemini-3-Flash-Preview as the VLM judge favors \geminiproThree. We compare its scores against those of an independent judge, Kimi K2.6. Both judges assign \geminiproThree lower scores relative to the human ratings than they do on average for the other four models, providing no evidence of bias toward \geminiproThree (\cref{tab:judge_bias}).

We therefore use the VLM judge for cost-effective large-scale comparison across models, while acknowledging that human annotation remains more reliable for fine-grained evaluation.

\begin{table}[h]
\centering
\caption{Agreement is measured by quadratic $\kappa$ and Pearson correlation (mean $\pm$ std). Human annotators are highly consistent, while the VLM judge shows moderate alignment with human judgment.}
\label{tab:agreement_summary}

\scalebox{0.8}{%
\begin{tabular}{lcc}
\toprule
Comparison & Kappa (Quadratic) & Pearson \\
\midrule
Human--Human
& $0.84 \pm 0.04$
& $0.85 \pm 0.04$ \\

Human--VLM Judge
& $0.51 \pm 0.02$
& $0.52 \pm 0.01$ \\
\bottomrule
\end{tabular}
}

\end{table}

\begin{table}[h]
\centering
\caption{Human--VLM agreement measured by quadratically weighted Cohen's $\kappa$ and Pearson correlation. The VLM judge achieves consistent moderate agreement with all annotators, with higher reliability on valid trajectories and lower agreement when annotations violate maze constraints.}
\label{tab:vlm_judge_agreement}
\setlength{\tabcolsep}{5pt}

\scalebox{0.7}{%
\begin{tabular}{lcc|cc|cc}
\toprule
& \multicolumn{2}{c}{Annotator 1} 
& \multicolumn{2}{c}{Annotator 2} 
& \multicolumn{2}{c}{Annotator 3} \\
\cmidrule(r){2-3} \cmidrule(r){4-5} \cmidrule(r){6-7}
Dataset & $\kappa_{quad}$ & Pearson 
        & $\kappa_{quad}$ & Pearson 
        & $\kappa_{quad}$ & Pearson \\
\midrule
ball-path    & 0.53 & 0.53 & 0.51 & 0.51 & 0.51 & 0.52 \\
maze-invalid & 0.40 & 0.41 & 0.44 & 0.46 & 0.42 & 0.44 \\
maze-valid   & 0.53 & 0.58 & 0.60 & 0.64 & 0.63 & 0.65 \\
vpct         & 0.42 & 0.49 & 0.41 & 0.44 & 0.46 & 0.50 \\
\bottomrule
\end{tabular}
}

\end{table}

\begin{table}[t]
\centering
\caption{Human--human agreement across datasets. Annotators show consistently high agreement ($\kappa \approx 0.85$--$0.92$), indicating reliable human evaluation, with slightly lower agreement on \textit{maze-invalid} cases.}
\label{tab:human_human_agreement}
\setlength{\tabcolsep}{5pt}

\adjustbox{max width=\linewidth}{%
\begin{tabular}{lcc|cc|cc}
\toprule
& \multicolumn{2}{c}{Annotator 1 vs Annotator 2}
& \multicolumn{2}{c}{Annotator 1 vs Annotator 3}
& \multicolumn{2}{c}{Annotator 2 vs Annotator 3} \\
\cmidrule(r){2-3} \cmidrule(r){4-5} \cmidrule(r){6-7}
Dataset & $\kappa_{quad}$ & Pearson
        & $\kappa_{quad}$ & Pearson
        & $\kappa_{quad}$ & Pearson \\
\midrule
ball-path    & 0.90 & 0.90 & 0.90 & 0.90 & 0.85 & 0.85 \\
maze-invalid & 0.80 & 0.82 & 0.50 & 0.52 & 0.53 & 0.53 \\
maze-valid   & 0.91 & 0.92 & 0.91 & 0.92 & 0.92 & 0.93 \\
vpct         & 0.89 & 0.90 & 0.88 & 0.89 & 0.86 & 0.87 \\
\bottomrule
\end{tabular}
}

\end{table}

\begin{table}[h]
\centering
\caption{Human evaluation score (1--5, higher is better). We report mean $\pm$ standard deviation for each task and the overall mean across all tasks. Best mean in each column is bolded. \SketchVLMs have higher mean annotation quality than other annotation models.}
\label{tab:human_score_by_task}
\setlength{\tabcolsep}{4.5pt}
\renewcommand{\arraystretch}{1.12}

\adjustbox{max width=\columnwidth}{%
\begin{tabular}{l | c c c c c}
\toprule
\textbf{Model}
& \multicolumn{5}{c}{Human Score (1--5)} \\
\cmidrule(lr){2-6}
& VPCT & Ball Drop & \maze{} (Invalid) & \maze{} (Valid) & Mean \\
\midrule
\gptsketch
& 3.18 $\pm$ 1.27
& 3.24 $\pm$ 1.08
& \textbf{4.13 $\pm$ 1.24}
& \textbf{4.45 $\pm$ 1.22}
& 3.70 $\pm$ 1.32 \\

\geminisketch
& \textbf{4.56 $\pm$ 0.85}
& \textbf{3.79 $\pm$ 1.30}
& 3.92 $\pm$ 1.27
& 4.36 $\pm$ 1.07
& \textbf{4.14 $\pm$ 1.18} \\
\midrule
\nanologo \nanobananapro
& 2.44 $\pm$ 1.30
& 2.94 $\pm$ 1.56
& 2.77 $\pm$ 1.41
& 4.29 $\pm$ 1.36
& 3.08 $\pm$ 1.57 \\

\vilasrlogo \vilasr
& 1.26 $\pm$ 0.88
& 1.01 $\pm$ 0.12
& 3.02 $\pm$ 1.34
& 1.96 $\pm$ 1.05
& 1.74 $\pm$ 1.20 \\

\hspace{1.6pt}\thinkmorphlogo \thinkmorph
& 1.44 $\pm$ 1.01
& 1.10 $\pm$ 0.38
& 1.17 $\pm$ 0.45
& 1.25 $\pm$ 0.79
& 1.24 $\pm$ 0.72 \\
\bottomrule
\end{tabular}%
}
\end{table}

\begin{table}[h]
\centering
\caption{Comparison of human and VLM-judge scores. Here, $\Delta$ is the judge score minus the human score. Relative bias is $\Delta$ for \geminiproThree minus the mean $\Delta$ across the other four models; negative values indicate that the judge rates \geminiproThree less favorably relative to the other models.}
\label{tab:judge_bias}
\setlength{\tabcolsep}{4pt}
\renewcommand{\arraystretch}{1.12}

\adjustbox{max width=\columnwidth}{%
\begin{tabular}{lccc}
\toprule
\textbf{Sketch model}
& \textbf{Human}
& \shortstack{\textbf{Kimi K2.6}\\\textbf{judge ($\Delta$)}}
& \shortstack{\textbf{Gemini-3-Flash}\\\textbf{judge ($\Delta$)}} \\
\midrule
\geminisketch
& \textbf{4.14}
& 3.93 ($-0.21$)
& 3.70 ($-0.44$) \\

\gptsketch
& 3.70
& 3.26 ($-0.44$)
& 2.26 ($-1.44$) \\

\nanologo \nanobananapro
& 3.08
& 3.32 ($+0.24$)
& 2.60 ($-0.48$) \\

\vilasrlogo \vilasr
& 1.74
& 1.81 ($+0.07$)
& 1.81 ($+0.07$) \\

\hspace{1.6pt}\thinkmorphlogo \thinkmorph
& 1.24
& 1.87 ($+0.63$)
& 1.63 ($+0.39$) \\
\midrule
\multicolumn{2}{l}{\textbf{Relative bias toward \geminiproThree}}
& \textit{$-0.34$}
& \textit{$-0.08$} \\
\bottomrule
\end{tabular}%
}
\end{table}

\clearpage

\newcommand{\judgeicon}{%
    \raisebox{-0.08em}{\includegraphics[height=0.9em]{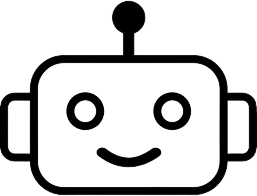}}%
}

\twocolumn[{%
\centering
\subsection{Annotation Score Explanation Examples}
\vspace{0.3em}

\begin{tabular*}{\textwidth}{@{}c@{\extracolsep{\fill}}cc@{}}

\begin{minipage}[t]{0.30\textwidth}
\centering
{\scriptsize \textbf{Input}\par}
\vspace{1pt}
\includegraphics[width=0.92\linewidth]{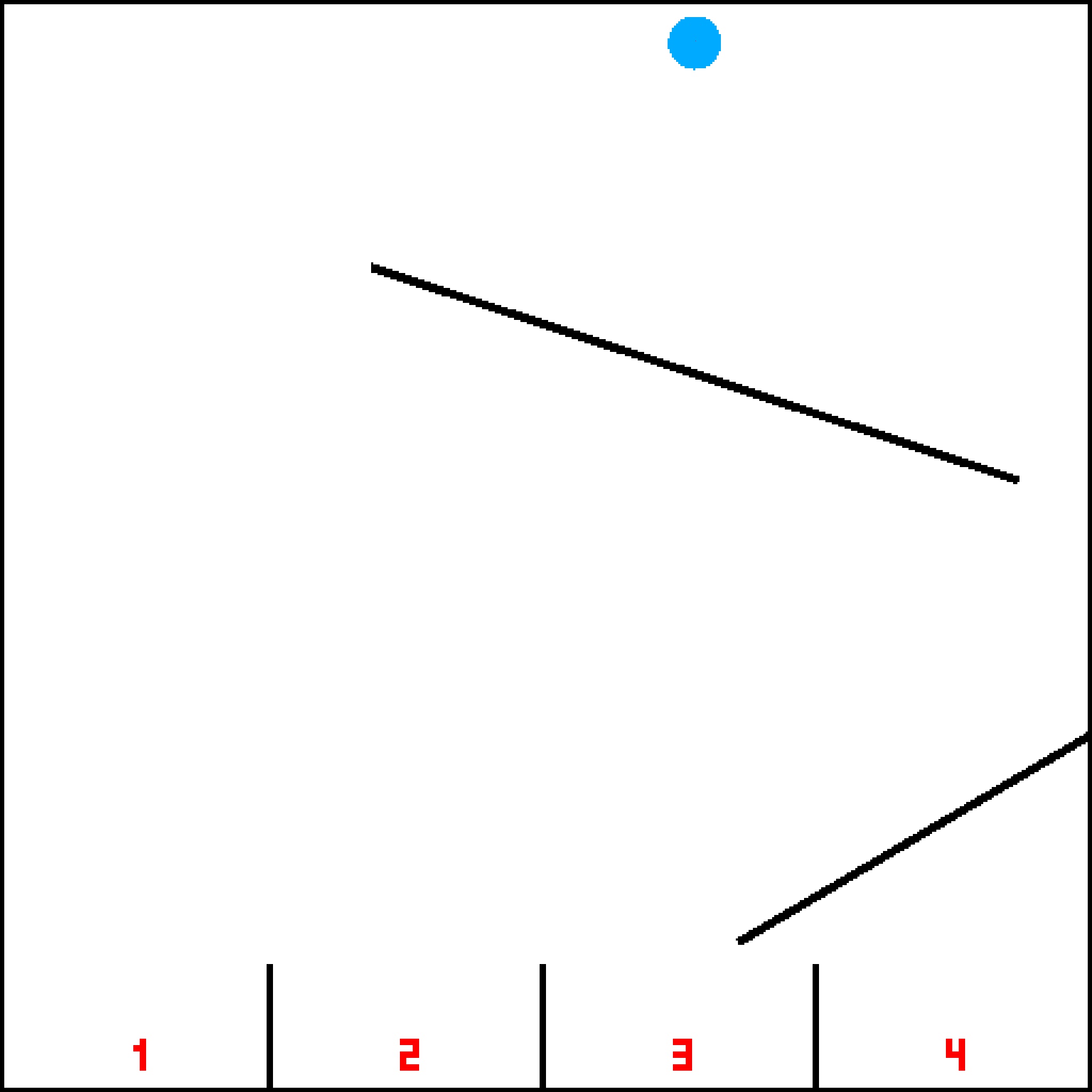}\par
\vspace{2pt}
\parbox[t]{0.92\linewidth}{\centering
    \scriptsize \textbf{Prompt:} ``Draw the path of the ball.''}
\end{minipage}
&
\begin{minipage}[t]{0.30\textwidth}
\centering
{\scriptsize \vilasrlogo~ViLaSR\par}
\vspace{1pt}
\includegraphics[width=0.92\linewidth]{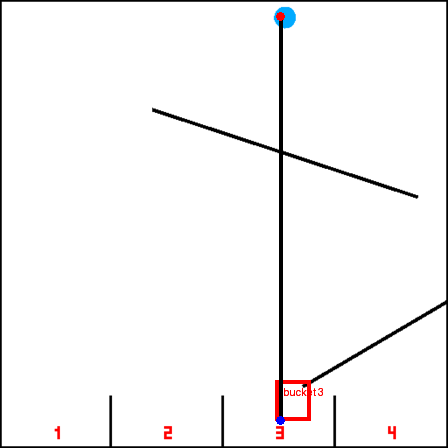}\par
\vspace{1pt}
{\scriptsize Annotation Quality Score: \textcolor{red}{1}\par}
\vspace{1pt}
\parbox[t]{0.92\linewidth}{\centering
    \scriptsize \judgeicon\hspace{0.25em}``The physics are completely \textcolor{red}{unrealistic}.''}
\end{minipage}
&
\begin{minipage}[t]{0.30\textwidth}
\centering
{\scriptsize \geminiplussketch\par}
\vspace{1pt}
\includegraphics[width=0.92\linewidth]{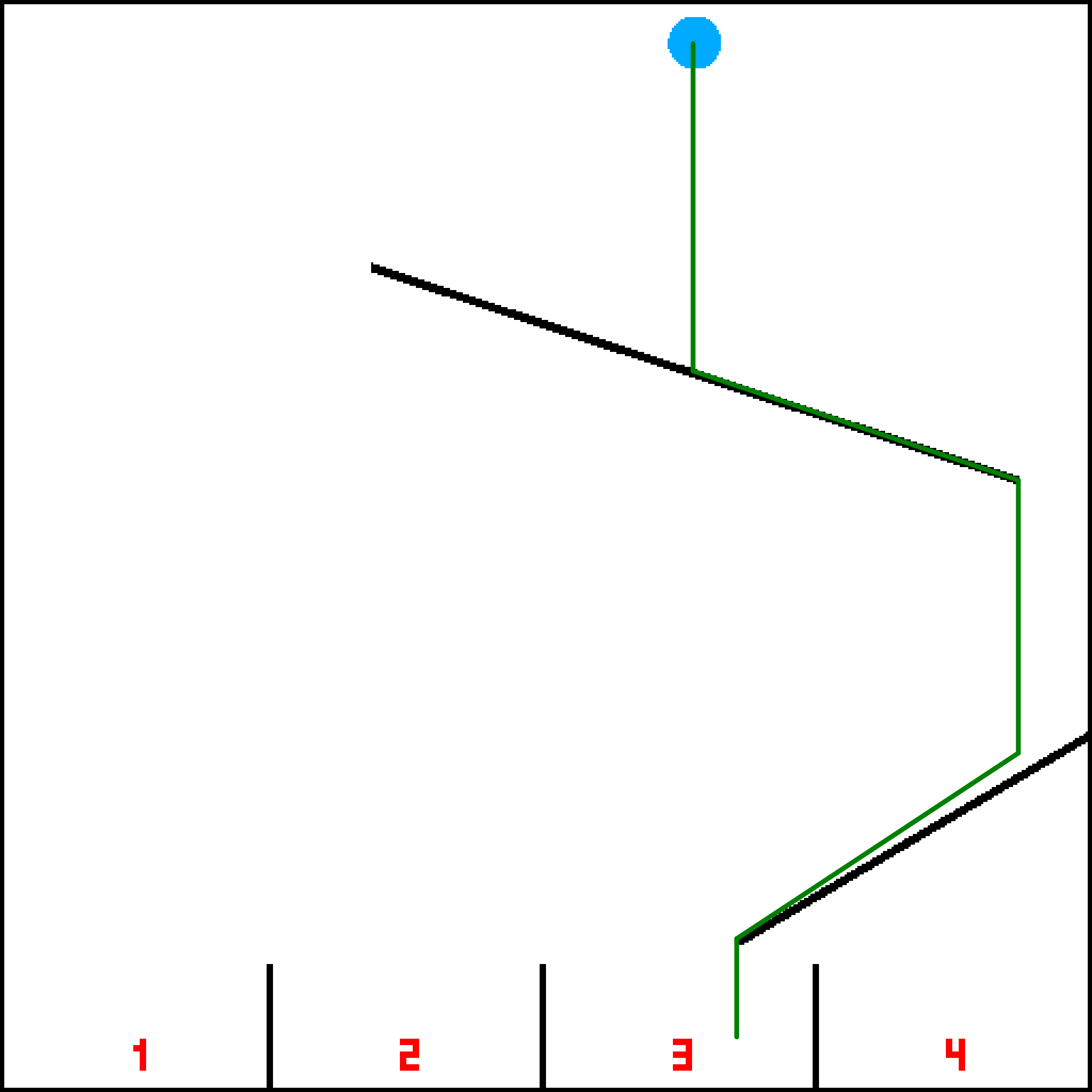}\par
\vspace{1pt}
{\scriptsize Annotation Quality Score: \textcolor{ForestGreen}{5}\par}
\vspace{1pt}
\parbox[t]{0.92\linewidth}{\centering
    \scriptsize \judgeicon\hspace{0.25em}``The path is \textcolor{ForestGreen}{logical}, follows gravity, and does not clip through walls.''}
\end{minipage}
\\[0.6em]

\begin{minipage}[t]{0.30\textwidth}
\centering
{\scriptsize \textbf{Input}\par}
\vspace{1pt}
\includegraphics[width=0.92\linewidth]{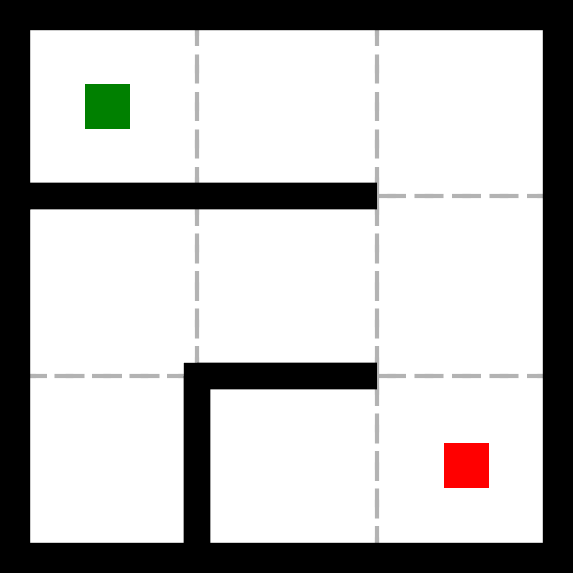}\par
\vspace{2pt}
\parbox[t]{0.92\linewidth}{\centering
    \scriptsize \textbf{Prompt:} ``Draw the proposed path: Right, Right, Down, Down.''}
\end{minipage}
&
\begin{minipage}[t]{0.30\textwidth}
\centering
{\scriptsize \thinkmorphlogo~ThinkMorph\par}
\vspace{1pt}
\includegraphics[width=0.92\linewidth]{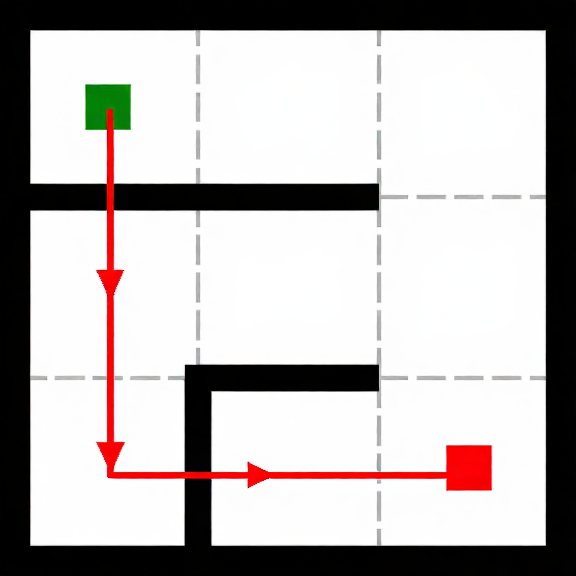}\par
\vspace{1pt}
{\scriptsize Annotation Quality Score: \textcolor{red}{1}\par}
\vspace{1pt}
\parbox[t]{0.92\linewidth}{\centering
    \scriptsize \judgeicon\hspace{0.25em}``The drawn path completely \textcolor{red}{contradicts} the given text path...clips through two solid black walls.''}
\end{minipage}
&
\begin{minipage}[t]{0.30\textwidth}
\centering
{\scriptsize \gptplussketch\par}
\vspace{1pt}
\includegraphics[width=0.92\linewidth]{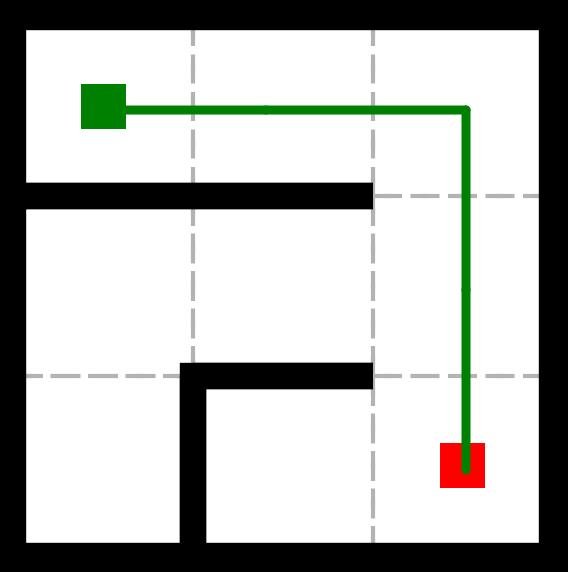}\par
\vspace{1pt}
{\scriptsize Annotation Quality Score: \textcolor{ForestGreen}{5}\par}
\vspace{1pt}
\parbox[t]{0.92\linewidth}{\centering
    \scriptsize \judgeicon\hspace{0.25em}``The sketch follows the proposed path...avoids all solid black walls...\textcolor{ForestGreen}{no logical errors}.''}
\end{minipage}

\end{tabular*}

\captionof{figure}{Low-quality annotations from \thinkmorphlogo ThinkMorph and \vilasrlogo ViLaSR may still lead to the correct final answer, but contain logical errors that are harder for users to verify than the high-quality annotations from SketchVLMs.}
\label{fig:annotation_quality_examples}
\vspace{-1mm}
}]

\clearpage

\twocolumn[{%
\centering
\subsection{Qualitative Examples}
\label{vlm-judge-examples}
\vspace{0.3em}

\includegraphics[width=0.6\textwidth]{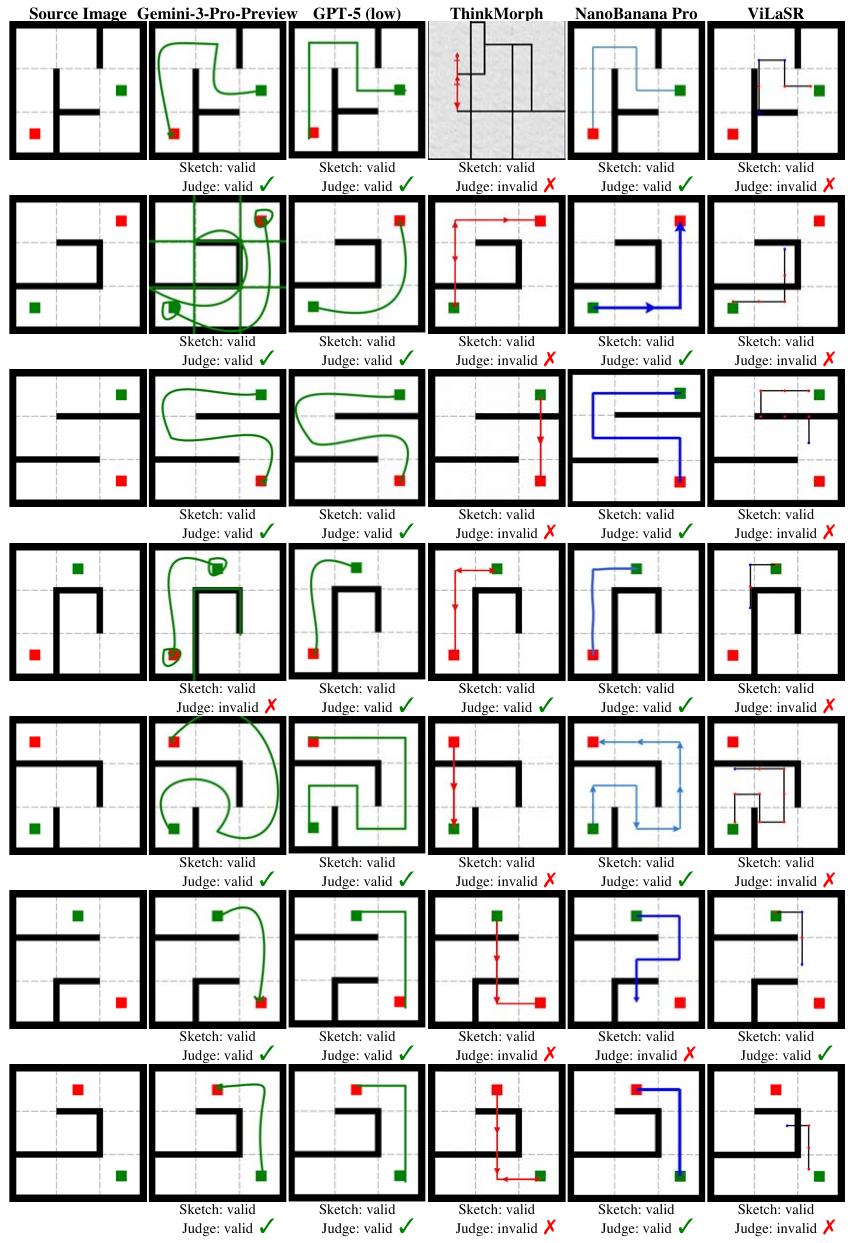}

\captionof{figure}{Comparison between the original sketching model's answer and the judge's answer for valid paths.}
\label{fig:consistency_maze_valid}
}]

\begin{figure*}[h]
  \centering
  \includegraphics[width=0.6\textwidth]{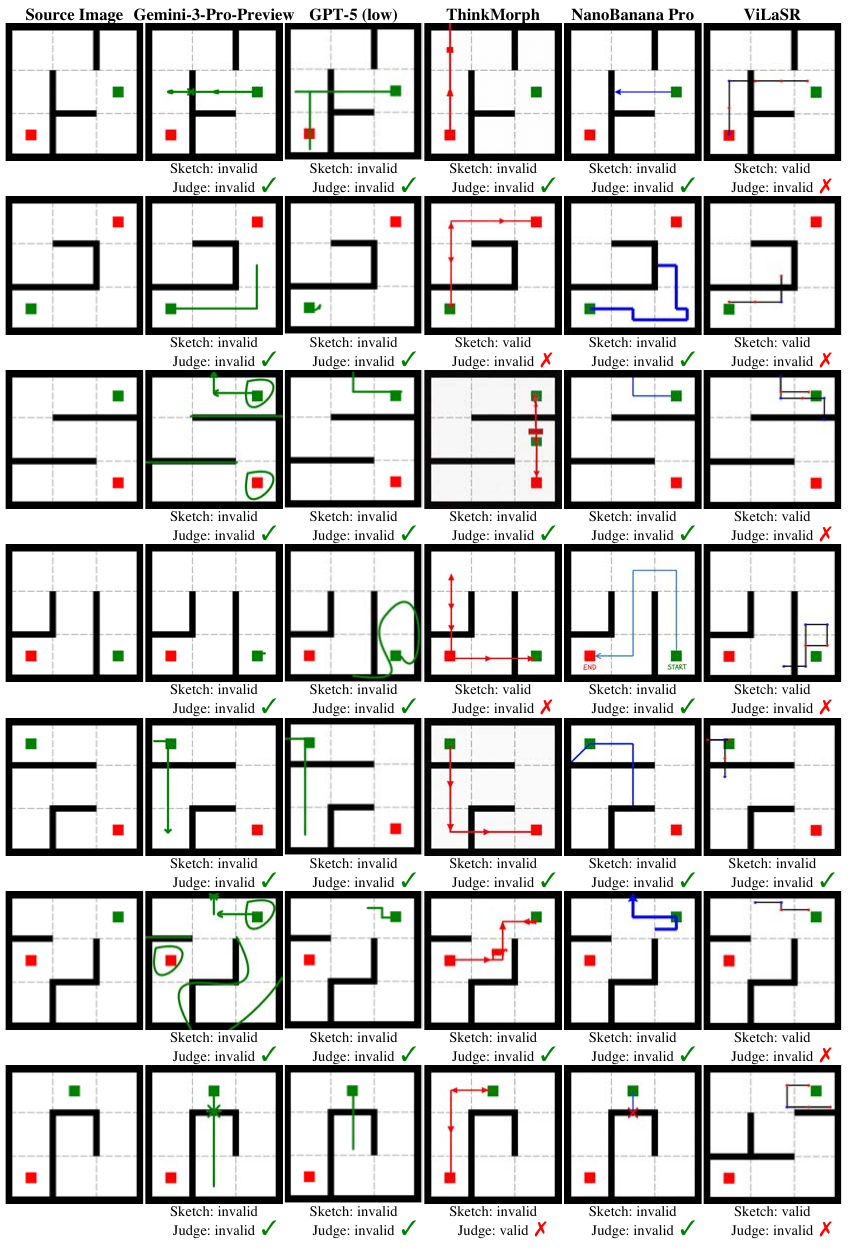}
  \caption{Comparison between the original sketching model's answer and the judge's answer for invalid paths.}
  \label{fig:consistency_maze_invalid}
\end{figure*}

\begin{figure*}[h]
  \centering
  \includegraphics[width=0.7\textwidth]{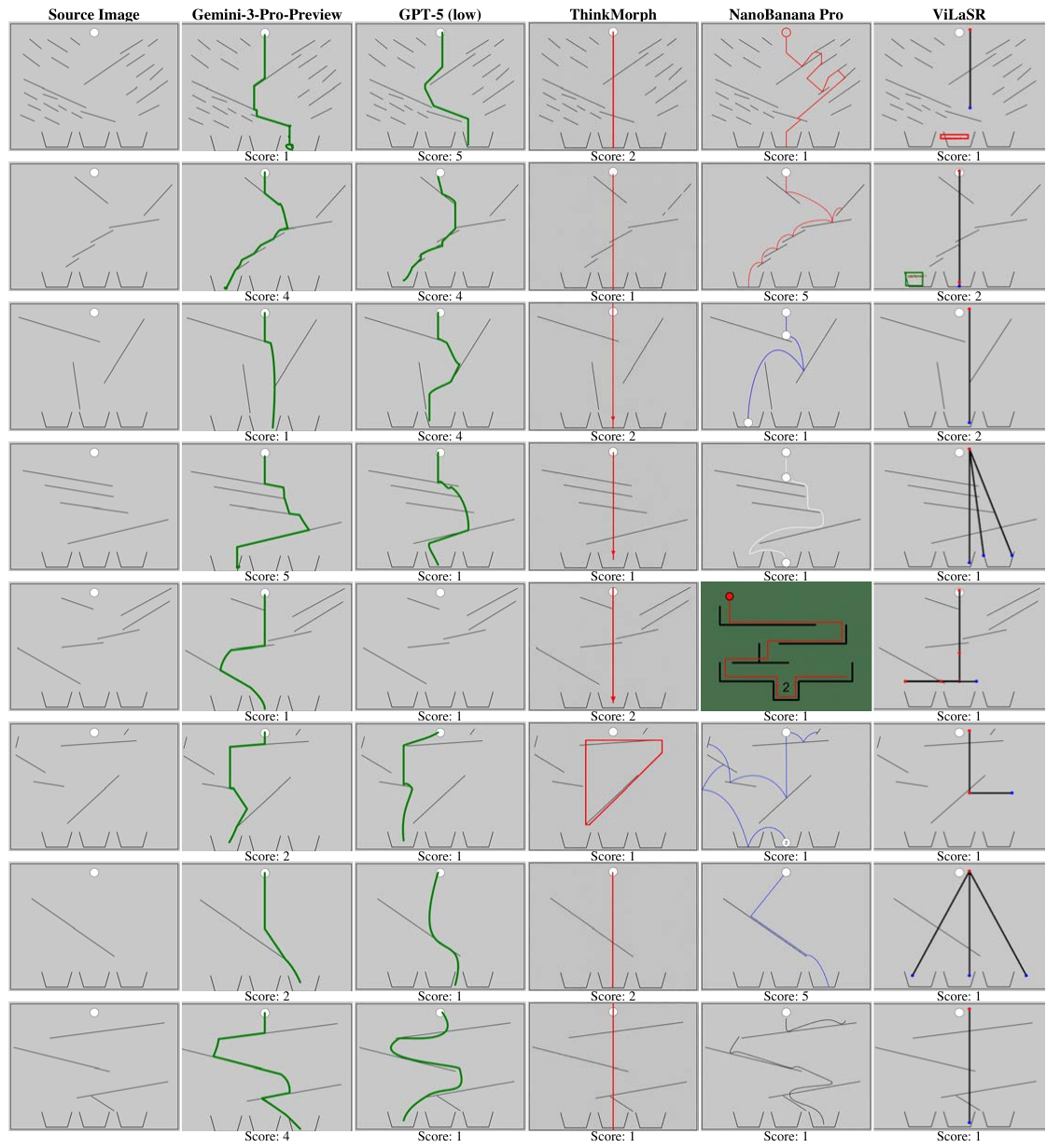}
  \caption{VLM-Judge quality scores for VPCT.}
  \label{fig:fig_vpct_quality_compare}
\end{figure*}

\begin{figure*}[h]
  \centering
  \includegraphics[width=0.7\textwidth]{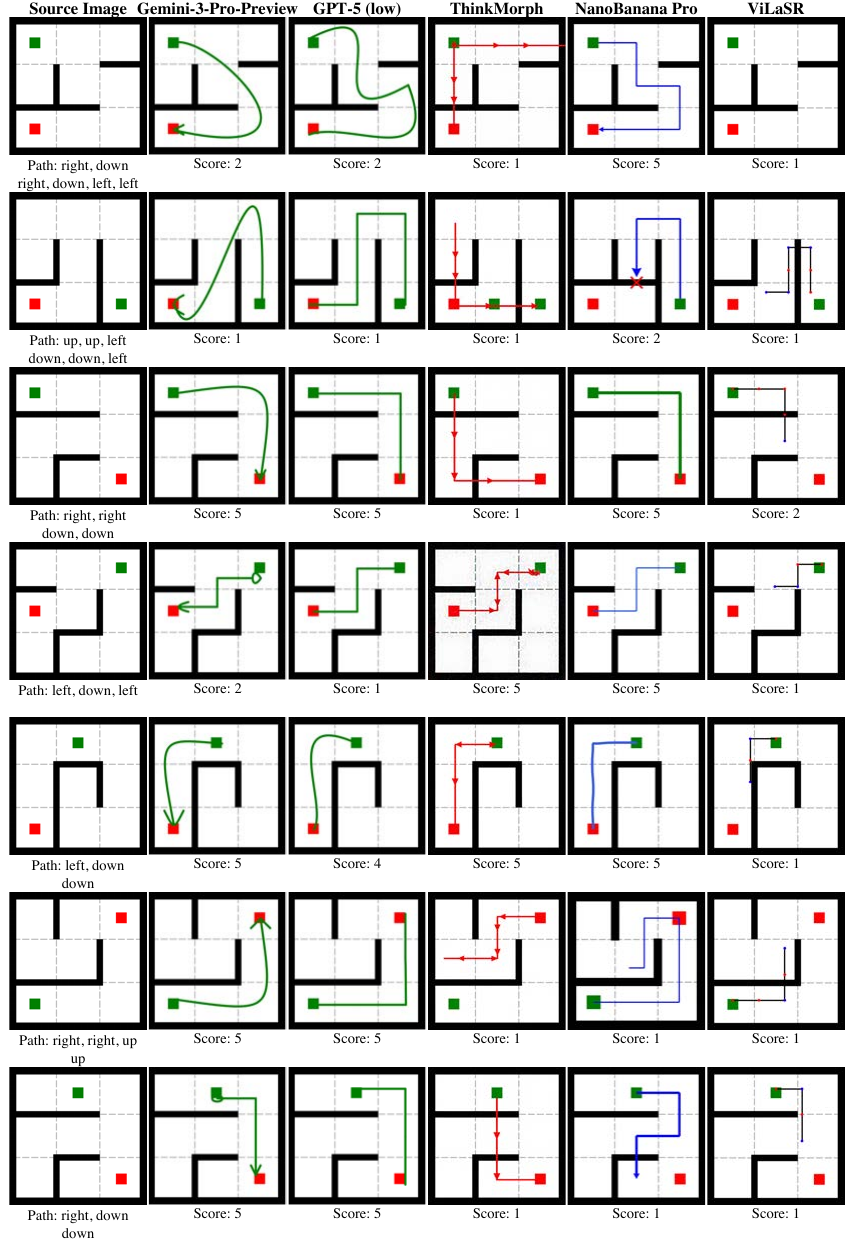}
  \caption{VLM-Judge quality scores for valid paths in the \maze task.}
  \label{fig:fig_maze_quality_valid_compare}
\end{figure*}

\begin{figure*}[h]
  \centering
  \includegraphics[width=0.7\textwidth]{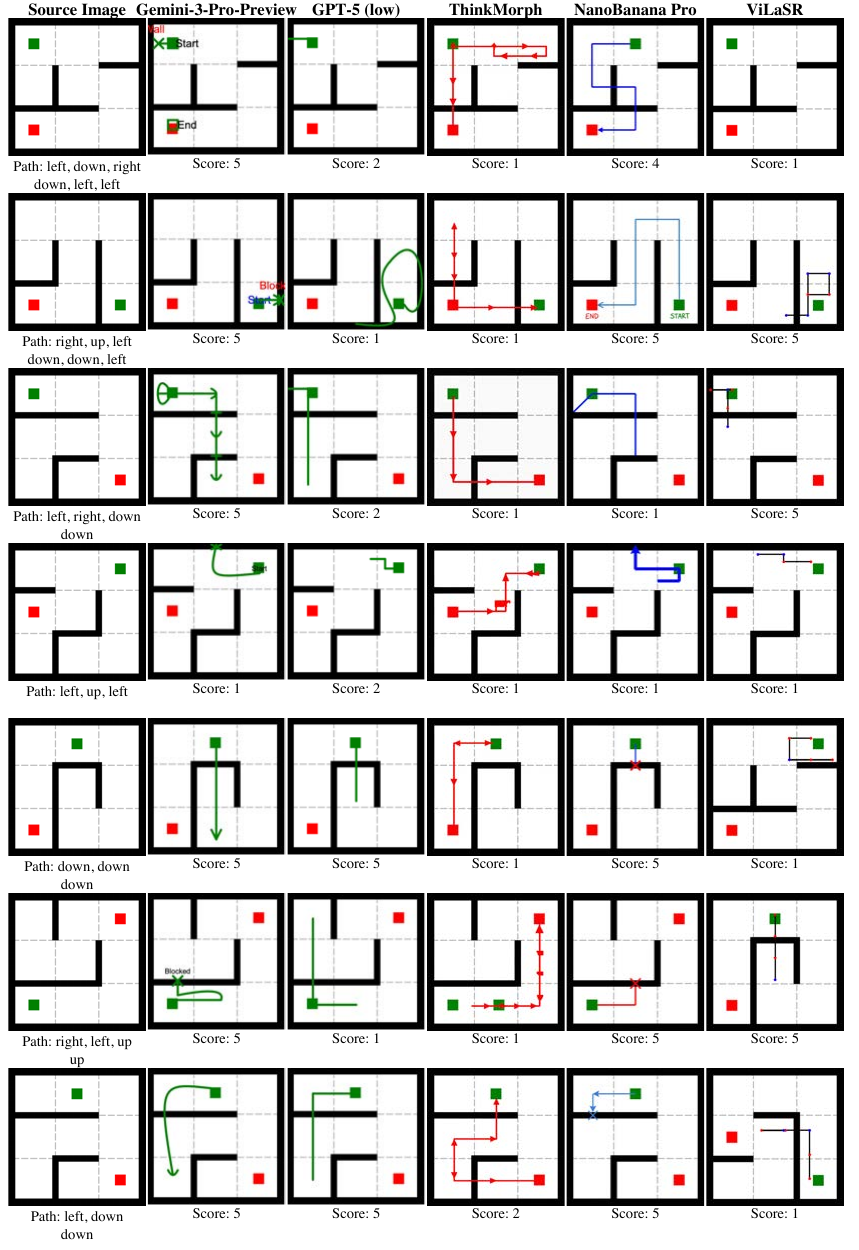}
  \caption{VLM-Judge quality scores for invalid paths in the \maze task.}
  \label{fig:fig_maze_quality_invalid_compare}
\end{figure*}

\clearpage

\subsection{Rubric Prompts}
\label{vlm-rubrics}
\begin{figure*}[t]
  \centering
\begin{tcblisting}{
  title=Rubric for VPCT and Ball Drop Quality,
  colback=black!5!white,
  colframe=black,
  fonttitle=\bfseries\color{white},
  coltitle=black,
  listing only,
  listing options={
    language=python,
    basicstyle=\ttfamily\tiny,
    keywordstyle=\color{orange},
    stringstyle=\color{black},
    identifierstyle=\color{black},
    showstringspaces=false,
    tabsize=1,
    breaklines=true 
  }
}
# Overall Instructions
You will be shown two images: the original image and an AI-annotated version. The annotation is supposed to show how the path the ball will take when it is dropped and where it will eventually end up. Your job is to grade the quality of the sketch using a rubric. You should grade only based on the following attributes that are given.

A minor issue is a violation that has little effect on the final answer of what bucket the ball will fall into. For example, if the ball path slightly clips through the walls, but the trajectory of the path still makes sense, then this is only a minor issue and not a major issue. A major issue is a violation that has a significant effect on the final answer of what bucket the ball will fall into. For example, if the ball path clips through the walls and the trajectory of the path is not logical, then this is a major issue. If there is no sketch provided, then the score should be a 1. The score should be based on the scale of 1 - 5

# Quality Checks

Below are more details:

## Things that are bad
1) The drawn path clips through any of the black walls.
2) The drawn path has very unrealistic physics. For example, the ball path momentarily disobeys gravity by moving upwards or moving in a direction that is not logical for gravity.
3) The drawn path substantially alters the original image (like adding or removing walls)
4) The drawn path contains multiple different paths instead of a single path.

## Things that are good
1) The drawn path is a single path that clearly shows where the final resting position of the ball will be.
2) The drawn path does not cut through any of the black walls.

## Things that are okay
1) If the path slightly clips through the walls, but the trajectory of the path still makes sense, then this is only a minor issue and not a major issue. We want to heavily penalize the ball going through walls and following a totally illogical path.
2) It's okay if the drawn path does not start exactly at the origin of the ball. As long as it is close, then this is not a minor or major issue.



# Scoring breakdown

1) The sketch has several critical flaws
2) The sketch has a critical flaw
3) The sketch contains some errors, overall direction of the path is valid and makes sense.
4) The sketch contains one minor logical error.
5) The sketch contains zero errors.

# Output Format
You should follow this output format EXACTLY with no other output:

{reasoning for logical consistency score}
Quality Score: {integer from 1 - 5}

# Example Output

<example_1>
The drawing contains multiple errors. The ball path barely clips through one of the platforms, but it is a minor issue as the path of the ball still makes sense. Additionally, the ball path defies gravity and begins to momentarily float straight upwards for no logical reason. The minor error combined with the critical error results in a logical score of 2/5.
Quality Score: 2
</example_1>

<example_2>
Overall, the path the ball takes is logical. It properly reflects off the first wall and then rolls to the right. Then, it falls down to the left and lands in bucket 2. The ball path slightly clips through one of the black edges, but this is a minor issue because the big picture trajectory of the ball is still logical. Given the overall path makes sense and there was just one minor issue, the quality score is 4/5.
Quality Score: 4
</example_2>
\end{tcblisting}
  \caption{Rubric for VPCT and Ball Drop Quality}
  \label{fig:supp-creativity_prompt_ball_physics}
\end{figure*}

\begin{figure*}[htbp]
  \centering
\begin{tcblisting}{
  title=Rubric for Path Navigation Quality,
  colback=black!5!white,
  colframe=black,
  fonttitle=\bfseries\color{white},
  coltitle=black,
  listing only,
  listing options={
    language=python,
    basicstyle=\ttfamily\tiny,
    keywordstyle=\color{black},
    stringstyle=\color{black},
    identifierstyle=\color{black},
    showstringspaces=false,
    tabsize=1,
    breaklines=true 
  }
}
# Overall Instructions
You will be shown two images: the original image and an AI-annotated version. Your job is to grade the quality of the sketch using a rubric. You should grade only based on the following attributes that are given.

# Sketch Instructions
The original sketch was drawn according to the following prompt:

"You are given an image of a maze where the green square marks the START cell and the red square marks the END cell of the maze. The walls of the maze are solid black lines. Dashed gray lines mark cell boundaries that can be crossed. You are given a proposed sequence of moves to reach the end of the maze starting from the green square and ending at the red square. Each move will move exactly one cell length in that direction. For example, "right" means move one cell in the maze to the right. A valid path must NOT cross any solid black walls and must end up in the red square cell. A valid path can also move through any of the dashed gray cell lines. The grid on the outside of the image is only there to help provide a reference for you. Moving one right means go one big cell in the maze right."

# Quality Checks

The main things that you should be looking for are:
1) Clipping through walls when it is not required to.
2) The drawn path does not go to the CENTER of each cell that it goes through.
3) The drawn path contradicts the given text path.

Below are more details:

## Things that are bad
1) The drawn path clips through any of the black walls when it is not required to. For example, even if the directions of the drawn path are correct, if the path touches or goes through a wall, then it is a bad sketch. That means that if the path goes through a wall even when it is not absolutely required to, then it is a bad sketch.
2) Each move in the drawn path should go to the **center** of the next cell in the path. If the drawn path is a curved path, then this does not apply. This is important! Look at each step in the path and make sure that the drawn path goes to the center of the next cell.
3) The sketch contains additional moves that are not in the path
4) The drawn sketch contradicts the given text path.
5) Even if the directions of the drawn path are correct, the end of the path does not end up touching the red square.
6) The drawn path does not start by touching the green square

## Things that are not an issue
1) If the proposed path is not valid, the drawn sketch shows exactly what the path should look like (even if it has to clip through walls or double back on itself).
2) If the proposed path is not valid, and the drawing ends as soon as there is an invalid move that is taken (such as requiring to go through a wall), then that is not an issue. It's okay for the drawing to not show all the steps of the path here since the sketch is emphasizing that the path is invalid.

## Score breakdown

1) The sketch makes absolutely no logical sense.
2) The sketch has some critical flaws that breaks the logic of the sketch.
3) The sketch contains multiple logical errors.
4) The sketch contains a minor error.
5) The sketch contains zero errors.

# Output Format
You should follow this output format EXACTLY with no other output:

{reasoning for quality score}
Quality Score: {integer from 1 - 5}

# Example Output
<example_1>
The drawing contains multiple errors. The drawn path goes up, up, left instead of up, up, right. This contradicts the given text path. Additionally, the end of the drawn path slightly clips through the solid black wall. Additionally, the draw path does NOT go to the center of each cell that it goes through. Instead, it only moves about 60 percent of the way across each cell and doesn't therefore doesn't follow the grid logic of moving from one center of each cell to the center of the next cell. The minor error combined with the multiple critical errors results in a score of 1/5.

Quality Score: 1
</example_1>

The original proposed path that the model should have followed is: {INSERT_ORIGINAL_PROMPT_PATH}
\end{tcblisting}
  \caption{Rubric for \maze Quality}
  \label{fig:supp-creativity_prompt_pathnav}
\end{figure*}

\clearpage

\clearpage

\twocolumn[{%
\section{Model Settings and Prompts}

\subsection{API Settings}

\begin{minipage}{\textwidth}
\centering

\captionof{table}{Model inference settings across providers. Temperature
uses the provider default for all models. Reasoning uses the provider
default unless otherwise specified.}
\label{tab:model_details}

\vspace{0.5em}

\scriptsize
\setlength{\tabcolsep}{5pt}
\renewcommand{\arraystretch}{1.12}

\begin{tabularx}{\textwidth}{
    @{}
    >{\raggedright\arraybackslash}p{0.17\textwidth}
    >{\raggedright\arraybackslash}p{0.31\textwidth}
    >{\raggedright\arraybackslash}p{0.19\textwidth}
    >{\raggedright\arraybackslash}X
    @{}
}
\toprule
\textbf{Model}
& \textbf{Model Identifier}
& \textbf{Reasoning}
& \textbf{API / Provider} \\
\midrule

GPT-5
& \path{gpt-5}
& Low, Medium, High
& OpenAI Responses API \\

Gemini 2.5 Flash
& \path{gemini-2.5-flash}
& Default
& Google AI Studio API \\

Gemini 2.5 Pro
& \path{gemini-2.5-pro}
& Default
& Google AI Studio API \\

Gemini 3 Pro
& \path{google/gemini-3-pro-preview}
& Default
& OpenRouter (Google AI Studio) \\

Gemini 3.1 Pro
& \path{google/gemini-3.1-pro-preview-20260219}
& Default
& OpenRouter; latency/cost evaluation only \\

Gemini 3 Flash
& \path{google/gemini-3-flash-preview}
& Default
& OpenRouter (Google AI Studio) \\

Nano Banana Pro
& \path{google/gemini-3-pro-image-preview}
& Default; image-only output
& OpenRouter (Google AI Studio) \\

Gemma-4-31B
& \path{google/gemma-4-31b-it}
& Default
& OpenRouter \\

Qwen3.5-397B-A17B
& \path{qwen/qwen3.5-397b-a17b}
& Default
& OpenRouter \\

Kimi K2.5
& \path{moonshotai/kimi-k2.5}
& Default
& OpenRouter \\

Kimi K3
& \path{moonshotai/kimi-k3}
& Low, High, Max
& OpenRouter \\

Qwen 2.5 VL 7B
& \path{qwen/qwen-2.5-vl-7b-instruct}
& Default
& OpenRouter (Hyperbolic) \\

\midrule

ThinkMorph
& \path{ThinkMorph/ThinkMorph-7B}
& Default
& Local; \href{https://github.com/ThinkMorph/ThinkMorph}{official code} \\

ViLaSR
& \path{inclusionAI/ViLaSR}
& Default
& Local; \href{https://github.com/AntResearchNLP/ViLaSR/}{official code} \\

\bottomrule
\end{tabularx}

\vspace{0.3em}

\end{minipage}

\vspace{1em}
}]

\clearpage

\subsection{SketchVLM System Prompt}
\label{sec:sketchvlm_system_prompt}

\newtcblisting{promptbox}[2][]{%
  enhanced,
  breakable,
  colback=black!1,
  colframe=black!25,
  boxrule=0.4pt,
  arc=1pt,
  boxsep=1pt,
  left=2pt,right=2pt,top=2pt,bottom=2pt,
  listing only,
  listing options={
    basicstyle=\ttfamily\fontsize{4}{5}\selectfont,
    breaklines=true,
    breakatwhitespace=false,
    columns=fullflexible,
    keepspaces=true,
    showstringspaces=false,
    tabsize=2,
  },
  title={#2},
  #1
}

\begin{figure}[H]
\centering
\captionsetup{font=footnotesize}

\begin{promptbox}{System prompt (1/3)}
# prompts.py

SYSTEM_PROMPT_BASE = """You are an expert artist specializing in drawing sketches that are visually appealing, expressive, and professional.
You will be provided with a blank grid. Your task is to specify where to place strokes on the grid to create a visually appealing sketch to complete the request.
The grid uses numbers (0 to {res_x}) along the bottom (x axis) and numbers (0 to {res_y}) along the left edge (y axis) to reference specific locations within the grid. The {origin_corner} is the origin. Each cell is uniquely identified by a combination of the corresponding x axis numbers and y axis number (\eg, the bottom-left cell is '{example_bottom_left}', the cell to its right is '{example_right_of_bottom_left}').
You can draw on this grid by specifying where to draw strokes. You can draw multiple strokes to depict the whole object, where different strokes compose different parts of the object. 
To draw a stroke on the grid, you need to specify the following:
Starting Point: Specify the starting point by giving the grid location (\eg, 'x0y0' for column 0, row 0).
Ending Point: Specify the ending point in the same way (\eg, 'x{res_x}y{res_y}' for column {res_x}, row {res_y}).
Intermediate Points: Specify at least two intermediate points that the stroke should pass through. List these in the order the stroke should follow, using the same grid location format (\eg, 'x6y5', 'x13y10' for points at column 6 row 5 and column 13 row 10).
Parameter Values (t): For each point (including the start and end points), specify a t value between 0 and 1 that defines the position along the stroke's path. t=0 for the starting point. t=1 for the ending point.
Intermediate points should have t values between 0 and 1 (\eg, "0.3 for x6y5, 0.7 for x13y10").

Examples:
To draw a smooth curve that starts at x8y6, passes through x6y7 and x6y10, ending at x8y11:
Points = ['x8y6', 'x6y7', 'x6y10', 'x8y11']
t_values = [0.00,0.30,0.80,1.00]
To close this curve into an ellipse shape, you can add another curve:
Points = ['x8y11', 'x11y10', 'x11y7', 'x8y6']
t_values = [0.00,0.30,0.70,1.00]
To draw a large circle that starts at x25y44 and ends at x25y44, passing through the cells x32y41, x35y35, x31y29, x25y27, x19y29, x15y35, x18y41:
Points = ['x25y44', 'x32y41', 'x35y35', 'x31y29', 'x25y27', 'x19y29', 'x15y35', 'x18y41', 'x25y44']
t_values = [0.00, 0.125, 0.25, 0.375, 0.50, 0.625, 0.75, 0.875, 1.00]
To draw non-smooth shapes (with corners) like triangles or rectangles, you need to specify the corner points twice with adjacent corresponding t values.
For example, to draw an upside-down "V" shape that starts at x13y27, ends at x24y27, with a pick (corner) at x18y37:
Points = ['x13y27', 'x18y37','x18y37', 'x24y27']
t_values = [0.00,0.55,0.5,1.00]
To draw a triangle with corners at x10y29, x15y33, and x9y35, start with drawing a "V" shape that starts at x10y29, ends at x9y35, with a pick (corner) at x15y33:
Points = ['x10y29', 'x15y33', 'x15y33', 'x9y35']
t_values = [0.00,0.55,0.5,1.00]
and then close it with a straight line from x13y27 to x24y27 to form a triangle:
Points = ['x13y27', 'x24y27']
t_values = [0.00,1.00]
Note that for a triangle, the start and end points should be different from each other.
To draw a rectangle with four corners at x13y27, x24y27, x24y11, x13y11:
Points = ['x13y27', 'x24y27', 'x24y27', 'x24y11', 'x24y11', 'x13y11', 'x13y11', 'x13y27']
t_values = [0.00,0.3,0.25,0.5,0.5,0.75,0.75,1.00]
To draw a small square with four corners at x26y25, x29y25, x29y21, x26y21:
Points = ['x26y25', 'x29y25', 'x29y25', 'x29y21', 'x29y21', 'x26y21', 'x26y21', 'x26y25']
t_values = [0.00,0.3,0.25,0.5,0.5,0.75,0.75,1.00]
To draw a single dot at x15y31 use:
Points = ['x15y31']
t_values = [0.00]
To draw a straight linear line that starts at x18y31 and ends at x35y14 use:
Points = ['x18y31', 'x35y14']
t_values = [0.00, 1.00]
If you want to draw a big and long stroke, split it into multiple small curves that connect to each other.
"""
\end{promptbox}

\vspace{-4pt}
\caption{System Prompt (1/3). Used for all SketchVLM models in single-turn and multi-turn.}
\label{fig:system_prompt_1of3}
\end{figure}

\begin{figure*}[t]
\centering
\captionsetup{font=footnotesize}

\begin{promptbox}{System prompt (2/3)}
# Sketch Methods

Below are the different sketching methods you can use for your task.

## FREEHAND SKETCH
- Emit one or more stroke blocks with points on the grid, no <text>.
- Use multiple strokes to compose shapes; curves/lines are both fine.
- the <id> tag should describe the part being drawn.

<s1>
  <points>'x12y20','x13y20','x14y21','x15y22'</points>
  <t_values>0.00,0.33,0.66,1.00</t_values>
  <id>part_1</id>
</s1>
<s2>
  <points>'x20y18','x20y14','x24y14','x24y18','x20y18'</points>
  <t_values>0.00,0.25,0.50,0.75,1.00</t_values>
  <id>part_2</id>
</s2>

## STRAIGHT LINE
<sN>
  <points>'x10y19','x40y19'</points>
  <t_values>0.00,1.00</t_values>
  <id>line_1</id>
</sN>

## Arrow (draw the shaft, and the arrowhead as two separate parts)

<s1>
  <points>'x12y32','x6y32'</points>
  <t_values>0.00,1.00</t_values>
  <id>arrow_shaft</id>
</s1>
<s2>
  <points>'x7y33','x6y32'</points>
  <t_values>0.00,1.00</t_values>
  <id>arrowhead_top</id>
</s2>
<s3>
  <points>'x7y31','x6y32'</points>
  <t_values>0.00,1.00</t_values>
  <id>arrowhead_bottom</id>
</s3>

## BOX / RECTANGLE (list the 4 corners in order)
<sN>
  <points>'x12y12','x20y12','x20y18','x12y18','x12y12'</points>
  <t_values>0.00,0.25,0.50,0.75,1.00</t_values>
  <id>box_1</id>
</sN>

## COUNTING (place numerals near each instance; one stroke per number; change text size based on object size and image resolution, so can be text size="1.0" or "2.0" up to "32.0" etc)
<sN>
  <points>'x08y22'</points>
  <t_values>0.00</t_values>
  <text size="4.0" color="black">'1'</text>
  <id>count_1</id>
</sN>

## LABELING (anchor a text label to the cell closest to center of the object/part; change text size based on object/part size and image resolution, so can be text size="1.0" or "2.0" up to "32.0" etc)
<sN>
  <points>'x26y17'</points>
  <t_values>0.00</t_values>
  <text size="3.2" color="black">'handlebar'</text>
  <id>label_handlebar</id>
</sN>

# Rules
- Output only <answer>...</answer> with a single <strokes>...</strokes> section.
- For counting/labeling tasks, prefer <text> with short values ('1','2',... or 'wheel','seat',...).
- Use <points> with exactly one anchor cell for each text label/number (one per item/part).
- Do not mix patterns: if the user asks to label, do not draw boxes; if the user asks to count, do not label names.
- Keep each stroke in its own <sN>...</sN> block; increment N in order without gaps.
- If the question requires an answer (\eg, "How many?"), include it at the end of your response, after the </strokes> tag, in a new <final_answer> tag.
"""

\end{promptbox}

\vspace{-4pt}
\caption{System Prompt (2/3), continued from \cref{fig:system_prompt_1of3}.}
\label{fig:system_prompt_2of3}
\end{figure*}

\begin{figure*}[t]
\centering
\captionsetup{font=footnotesize}

\begin{promptbox}{System prompt (3/3)}
def build_system_prompt(res_x: int, res_y: int, origin: str = "bottom_left") -> str:
    if origin == "top_left":
        origin_corner = "top left"
        example_bottom_left = f"x0y{res_y}"
        example_right_of_bottom_left = f"x1y{res_y}"
    else:
        origin_corner = "bottom left"
        example_bottom_left = "x0y0"
        example_right_of_bottom_left = "x1y0"

    return SYSTEM_PROMPT_BASE.format(
        res_x=res_x,
        res_y=res_y,
        origin_corner=origin_corner,
        example_bottom_left=example_bottom_left,
        example_right_of_bottom_left=example_right_of_bottom_left,
    )

COUNTING_PROMPT = """
Task: 
- COUNT all the {object} by placing numbered SVG text strokes on them (no curves).

Output example could be:
<answer>
<concept>Numbering each {object}</concept>
<strokes>
<s1>
    <text size="1.6" color="#ff0066">'1'</text>
    <points>'xAyB'</points>
    <t_values>0.00</t_values>
    <id>marker_{object}1</id>
</s1>
<!-- s2, s3, ... one per object -->
</strokes>

Rules:
- Use ONLY text strokes (no curves).
- Exactly one point per stroke ('xAyB') at the object's center-ish cell.
- You MAY style numbers: <text size="1.8" color="#0057ff"> or <style><font_size>...</font_size><color>...</color></style>.
- size is cells (multiplier) unless you suffix 'px'
- choose bigger text size for larger objects, smaller for tiny objects. use bigger size for higher resolution images, smaller for lower resolution.
- choose readable colors that will contrast well with the object that you are numbering and the background.
- If 0 objects, still return the full wrapper with an empty <strokes> block.
- Do not write anything outside <answer>...</strokes>.
"""

GENERIC_LABEL_PROMPT = """
Task:
- The object in the image is a {concept}.
- Label ONLY the following parts of the {concept}: {labels_hint}.
- Do not invent or add any new part names beyond this list.
- Use SVG text strokes (no curves) to place each label.

Output EXACTLY this XML shape:
<answer>
<concept>Labeling: {concept}</concept>
<strokes>
  <!-- one <sN> per label -->
  <s1>
    <text size="1.6" color="#ff0066">'head'</text>   <!-- size: cells or 'px' -->
    <points>'xAyB'</points>
    <t_values>0.00</t_values>
    <id>label_head</id>
  </s1>
</strokes>

Rules:
- Use ONLY text strokes (no curves).
- Anchor each label at the center of the corresponding part ('xAyB').
- You MAY style the text labels: <text size="1.8" color="#0057ff"> or <style><font_size>...</font_size><color>...</color></style>
- Use one <sN> per label name in the list above.
- Choose bigger text size for larger objects, smaller for tiny objects. use bigger size for higher resolution images, smaller for lower resolution.
- Choose readable colors that will contrast well with the object/part that you are labeling and the background.
- Do not write anything outside <answer>...</strokes>.
"""
\end{promptbox}

\vspace{-4pt}
\caption{System Prompt (3/3), continued from \cref{fig:system_prompt_2of3}. Parameters are res\_x, res\_y, and origin. res\_x controls how many x coordinates there are. res\_y controls how many y coordinates there are. origin controls whether the origin is at the top left or bottom left.}
\label{fig:system_prompt_3of3}
\end{figure*}

\clearpage

\lstdefinestyle{promptbox}{
  basicstyle=\ttfamily\scriptsize,
  columns=fullflexible,
  breaklines=true,
  breakatwhitespace=true,
  frame=single,
  framerule=0.3pt,
  rulecolor=\color{black},
  xleftmargin=0.5em,
  xrightmargin=0.5em,
  aboveskip=0.5em,
  belowskip=0.0em,
  showstringspaces=false
}

\begin{figure*}[t]
\vspace{-0.5em}
\centering

\begin{minipage}[t]{0.485\columnwidth}
\centering
\textbf{One-stroke guard (multiturn)}\\[-0.25em]
\begin{lstlisting}[style=promptbox]
[Mode: stepwise]
You are in stepwise mode. On this turn you output EXACTLY ONE stroke block:
<answer>
  <strokes>
    <sN>...</sN>
  </strokes>
</answer>
Do NOT output any other <sM> blocks, no <final_answer>, no explanations.
If the drawing is already complete and no further stroke is needed, output an empty <answer> with NO <strokes> block.
Stop immediately after </answer>.
\end{lstlisting}
\end{minipage}
\hfill
\begin{minipage}[t]{0.485\columnwidth}
\centering
\textbf{Final-answer guard (multiturn)}\\[-0.25em]
\begin{lstlisting}[style=promptbox]
[Mode: stepwise]
All strokes have already been provided. On this turn output ONLY:
<final_answer> ... </final_answer>
Do not output the previous strokes again. Stop immediately after </final_answer>.
\end{lstlisting}
\end{minipage}

\vspace{-0.5em}
\caption{During multi-turn, when the model must draw a stroke, we append the "One-stroke guard" to clarify to the model that it should only draw one stroke with no final answer. After the model has decided that no more strokes are needed (it does this by outputting no strokes on a turn), we run one more turn where we prompt the model with "Final-answer guard."}
\label{fig:system_prompt_multiturn}
\vspace{-0.75em}
\end{figure*}

\clearpage

\subsection{SketchVLM Output Example}

\newtcolorbox{annobox}[1][]{
  enhanced,
  colback=black!2,
  colframe=black!20,
  boxrule=0.6pt,
  arc=2pt,
  left=6pt,right=6pt,top=6pt,bottom=6pt,
  breakable,
  #1
}

\begin{figure}[H]
\centering

\begin{minipage}{0.45\textwidth}
  \centering
  \includegraphics[width=\linewidth]{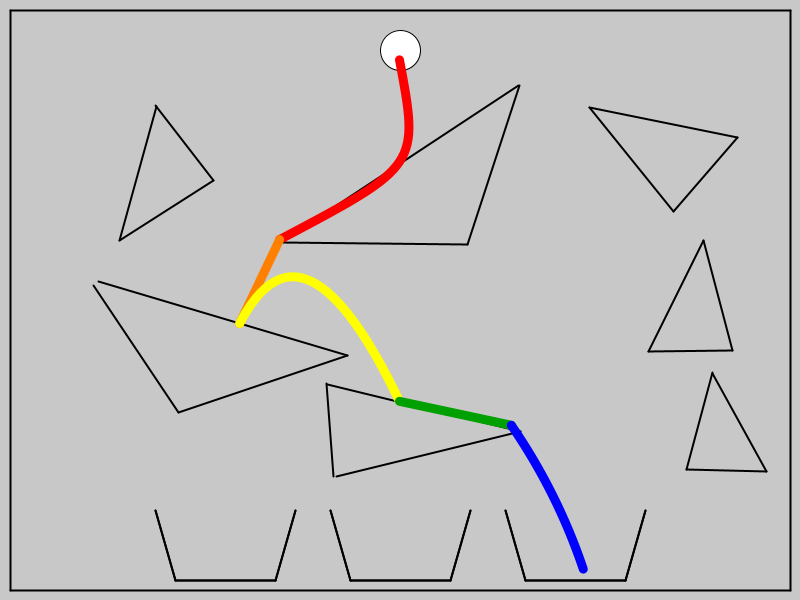}
  \vspace{2pt}

\end{minipage}

\vspace{6pt} 

\begin{minipage}{0.90\textwidth}
\begin{annobox}
\textbf{\footnotesize Stroke annotations (XML)}\vspace{4pt}

\begin{lstlisting}[
  basicstyle=\ttfamily\scriptsize,
  columns=fullflexible,
  breaklines=true,
  breakatwhitespace=true,
  keepspaces=true,
  showstringspaces=false
]
<s1>
  <points>'x500y100','x500y270','x500y270','x350y400'</points>
  <t_values>0.00,0.50,0.50,1.00</t_values>
  <id>path_1</id>
</s1>
<s2>
  <points>'x350y400','x325y470','x300y540'</points>
  <t_values>0.00,0.50,1.00</t_values>
  <id>drop_1</id>
</s2>
<s3>
  <points>'x300y540','x390y470','x500y670'</points>
  <t_values>0.00,0.50,1.00</t_values>
  <id>path_2</id>
</s3>
<s4>
  <points>'x500y670','x570y690','x640y710'</points>
  <t_values>0.00,0.50,1.00</t_values>
  <id>path_bounce</id>
</s4>
<s5>
  <points>'x640y710','x700y850','x730y950'</points>
  <t_values>0.00,0.60,1.00</t_values>
  <id>drop_2</id>
</s5>
<final_answer>3</final_answer>
\end{lstlisting}
\end{annobox}
\end{minipage}

\vspace{2pt}
\caption{Example sketch output and annotations for prompting \geminiplussketch on VPCT.
Each stroke is colored differently for viewing purposes.}
\label{fig:appendix_annotation_example}
\end{figure}

\clearpage

\section{Other Baselines}

\subsection{Baselines}\label{app:baselines}

\paragraph{Image-Editing Models.}\label{app:baselines:image_editing}
To benchmark the image-editing model \nanobananapro\, we first prompt it to generate a sketch for the input image. We then provide the resulting edited image to \geminiproThree\ to produce the final task answer. For Connect-the-Dots, we additionally report a manual evaluation of \nanobananapro, since its outputs do not always map cleanly to our structured sketch representation.

\paragraph{Fine-tuned Sketching Models.}\label{app:baselines:fine-tuned}
For fine-tuned sketching models such as \vilasr\ and \thinkmorph, we use the same task prompts as for SketchVLM. Because these models are trained to always output a sketch, they do not have a meaningful ``no-sketch'' baseline mode; therefore, we omit baseline VQA accuracy for these models in the main results and report only their sketch-conditioned performance.

\clearpage

\section{AI Use}
We utilized AI assistants for grammar editing, LaTeX formatting, and code debugging. All code, experiments, scientific claims, analyses, and final writing decisions were reviewed and verified by the authors.

\clearpage

\fi

\end{document}